\RequirePackage[english=usenglishmax]{hyphsubst}
\documentclass[twoside,openright,titlepage,numbers=noenddot,headinclude,footinclude=true,cleardoublepage=empty,listof=totoc,paper=a4,fontsize=11pt,english,twoside=semi,DIV=calc]{scrreprt} 
\newcommand{\myTitle}{Towards Reliable Simulation-based Inference}
\newcommand{\mySubmissionYear}{2024}
\newcommand{\mySubmissionMonth}{December}

\newcommand{\myFirstName}{Arnaud}
\newcommand{\myLastName}{Delaunoy}

\newcommand{\myDepartment}{Department of Electrical Engineering and Computer Science}
\newcommand{\myFaculty}{Faculty of Applied Sciences}
\newcommand{\myUni}{University of Liège}
\PassOptionsToPackage{utf8}{inputenc}
	\usepackage{inputenc}

\PassOptionsToPackage{eulerchapternumbers,listings,pdfspacing,beramono,dottedtoc}{classicthesis}

\newcounter{dummy} 
\providecommand{\mLyX}{L\kern-.1667em\lower.25em\hbox{Y}\kern-.125emX\@}

\PassOptionsToPackage{english}{babel}
  \usepackage{babel}

\usepackage{csquotes}
\PassOptionsToPackage{%
  backend=biber,
  bibencoding=ascii,
  language=auto,
  bibstyle=apa,
  citestyle=authoryear,
  sorting=none,
  minbibnames=1,
  maxbibnames=10,
  natbib=true
}{biblatex}
  \usepackage{biblatex}

  \usepackage{amsmath}

\usepackage{amssymb}
\usepackage{amsthm}
\usepackage{gensymb}
\usepackage{amsfonts}
\usepackage{dsfont}
\usepackage{lipsum}
\usepackage{mathtools}
\usepackage{float}
\PassOptionsToPackage{T1}{fontenc} 
    \usepackage{fontenc}
\usepackage{textcomp} 
\usepackage{scrhack} 
\usepackage{xspace} 
\usepackage{mparhack} 
\PassOptionsToPackage{printonlyused,smaller}{acronym}
  \usepackage{acronym} 

\usepackage{afterpage}
\usepackage[version=4]{mhchem}

\usepackage{tabularx} 
\usepackage{makecell}
\usepackage{multirow}
\usepackage{caption}
\usepackage{subcaption}
\usepackage{notoccite} 
\usepackage{longtable}%

\usepackage{comment}
\usepackage{pdflscape}
\usepackage[svgnames]{xcolor} 
\usepackage{colortbl}%
  
\usepackage{booktabs}

\usepackage{etoolbox}
\makeatletter
\newif\if@in@acrolist

\AtBeginEnvironment{acronym}{\@in@acrolisttrue}
\newrobustcmd{\LU}[2]{\if@in@acrolist#1\else#2\fi}

\newcommand{\ACF}[1]{{\@in@acrolisttrue\acf{#1}}}
\makeatother

\usepackage{listings}
\PassOptionsToPackage{pdftex,hyperfootnotes=false,pdfpagelabels}{hyperref}
\usepackage{hyperref}  
\PassOptionsToPackage{pdftex}{graphicx}
    \usepackage{graphicx}

\hypersetup{breaklinks=true}
\hypersetup{linktocpage=true}
\hypersetup{colorlinks=true}
\hypersetup{urlcolor=webbrown}
\hypersetup{linkcolor=RoyalBlue}
\hypersetup{citecolor=webgreen}
\hypersetup{pageanchor=true}
\hypersetup{plainpages=false}
\hypersetup{pdfstartpage=3}
\hypersetup{pdfstartview=FitV}
\hypersetup{pdfpagemode=UseNone}
\hypersetup{pageanchor=true}
\hypersetup{pdfpagemode=UseOutlines}
\hypersetup{pdftitle={\myTitle}}
\hypersetup{pdfauthor={\textcopyright\ \myFirstName\ \myLastName, \myUni, \myFaculty}}
\hypersetup{pdfsubject={}}
\hypersetup{pdfkeywords={}}
\hypersetup{pdfhighlight=/O}

\theoremstyle{plain}
\newtheorem{theorem}{Theorem}[section]
\newtheorem{proposition}[theorem]{Proposition}

\theoremstyle{definition}
\newtheorem{definition}[theorem]{Definition}

\theoremstyle{remark}

\makeatletter
\@ifpackageloaded{babel}%
{%
  \addto\extrasaustralian{%
  }%
  \addto\extrasngerman{%
  }%
}{\relax}
\makeatother

\usepackage{classicthesis}

\usepackage{lmodern} 
\usepackage{bm}
\renewcommand*{\mathbf}[1]{\ifmmode\bm{#1}\else\textbf{#1}\fi}

\usepackage{colortbl}
\usepackage{setspace}
\usepackage{fp}
\usepackage{soul}

\usepackage{textcomp}

\usepackage{pgfplots}
\usepackage{pgfplotstable}
\pgfplotsset{compat=newest}
\usepackage[separate-uncertainty = true,multi-part-units=single]{siunitx}

\usepackage[shortcuts]{extdash}

\usepackage{tikz-timing}[2011/01/09]
\usetikzlibrary{arrows, backgrounds, patterns, shapes}
\usepackage{arydshln}

\usepackage{algorithm}
\usepackage{algorithmic}

\renewcommand{\lstlistlistingname}{List of Listings}

\newcolumntype{L}[1]{>{\raggedright\arraybackslash}m{#1}}
\newcolumntype{C}[1]{>{\centering\arraybackslash}m{#1}}
\newcolumntype{R}[1]{>{\raggedleft\arraybackslash}m{#1}}

\numberwithin{equation}{chapter}
\numberwithin{figure}{chapter}
\numberwithin{table}{chapter}
\numberwithin{algorithm}{chapter}
\newcommand{\phat}{\hat{p}}
\newcommand{\qhat}{\hat{q}}
\newcommand{\rhat}{\hat{r}}

\newcommand{\alphahat}{\hat{\alpha}}

\newcommand{\varpitilde}{\Tilde{\varpi}}
\newcommand{\pitilde}{\Tilde{\pi}}

\newcommand{\btheta}{\boldsymbol{\theta}}
\newcommand{\bTheta}{\boldsymbol{\Theta}}

\newcommand{\bphi}{\boldsymbol{\phi}}

\newcommand{\bD}{\boldsymbol{D}}

\newcommand{\bw}{\mathbf{w}}
\newcommand{\bx}{\boldsymbol{x}}
\newcommand{\bX}{\boldsymbol{X}}

\newcommand{\by}{\boldsymbol{y}}
\newcommand{\bz}{\boldsymbol{z}}
\newcommand{\boldf}{\boldsymbol{f}}

\newcommand{\indicator}{\mathds{1}}

\newcommand{\DKL}{\text{KL}}
\newcommand{\Dchi}{\chi^2}
\newcommand{\Mid}{\, \Vert \,}
\renewcommand{\mid}{\, | \,}
\newcommand{\dmid}{\, || \,}

\newcommand{\E}{\mathbb{E}}

\DeclareMathOperator*{\argmax}{arg\,max}
\DeclareMathOperator*{\argmin}{arg\,min}

\newcommand{\sigmoid}{\sigma}

\usepackage{tcolorbox}
\newtcolorbox{custombox}[1][]
{
  colframe=black!40,
  colback=black!5,
  left=2pt,
  right=2pt,
  top=2pt,
  bottom=2pt,
  #1
}

\newtcolorbox{prologuebox}[1][]
{
  title=Prologue,
  colframe=black!70,
  colback=black!5,
  left=2pt,
  right=2pt,
  top=2pt,
  bottom=2pt,
  #1
}

\newtcolorbox{epiloguebox}[1][]
{
  title=Epilogue,
  colframe=black!70,
  colback=black!5,
  left=2pt,
  right=2pt,
  top=2pt,
  bottom=2pt,
  #1
}

\usepackage[skip=10pt plus1pt, indent=0pt]{parskip}
\usepackage{placeins}
\usepackage[a4paper,top=25.4mm,bottom=25.4mm,left=25mm, right=25mm,bindingoffset=6mm]{geometry} 

\begin{document}
  \frenchspacing
  \raggedbottom
  \selectlanguage{english}
  
  \pagestyle{plain}
  \pagenumbering{roman}

  \singlespacing
  \begin{titlepage}
  \doublespacing
  \large
  \hfill
  \vspace*{0.5cm}
  \begin{center}
    \doublespacing
    \textcolor{Maroon}{\huge\textbf{\myTitle}}
  \end{center}
  \vspace{1.25cm}
  \hrule
  \vspace{1.5cm}
  \onehalfspacing
  \begin{center}

    by\\
    \vspace{5mm}
    {\myFirstName} \textsc{\myLastName}

    \vspace{1cm}
    \myDepartment\\
    \myFaculty

    \hfill
    \vfill

    This dissertation has been submitted in partial fulfillment of the requirements for the Degree of Doctor of Applied Sciences.

    \hfill
    \vfill

    \emph{Jury members}\\
    \vspace{4mm}
    Prof. Gilles Louppe, University of Liège (Advisor)\\
    Prof. Vân Anh Huynh-Thu, University of Liège (President)\\
    Prof. Pierre Geurts, University of Liège \\
    Prof. Anne-Françoise Donneau, University of Liège \\
    Prof. Gaël Varoquaux, INRIA \\
    Prof. Pedro L. C. Rodrigues, INRIA \\
    \hfill
    \vfill

    \begin{minipage}[c]{0.5\textwidth}
        \includegraphics[width=5cm]{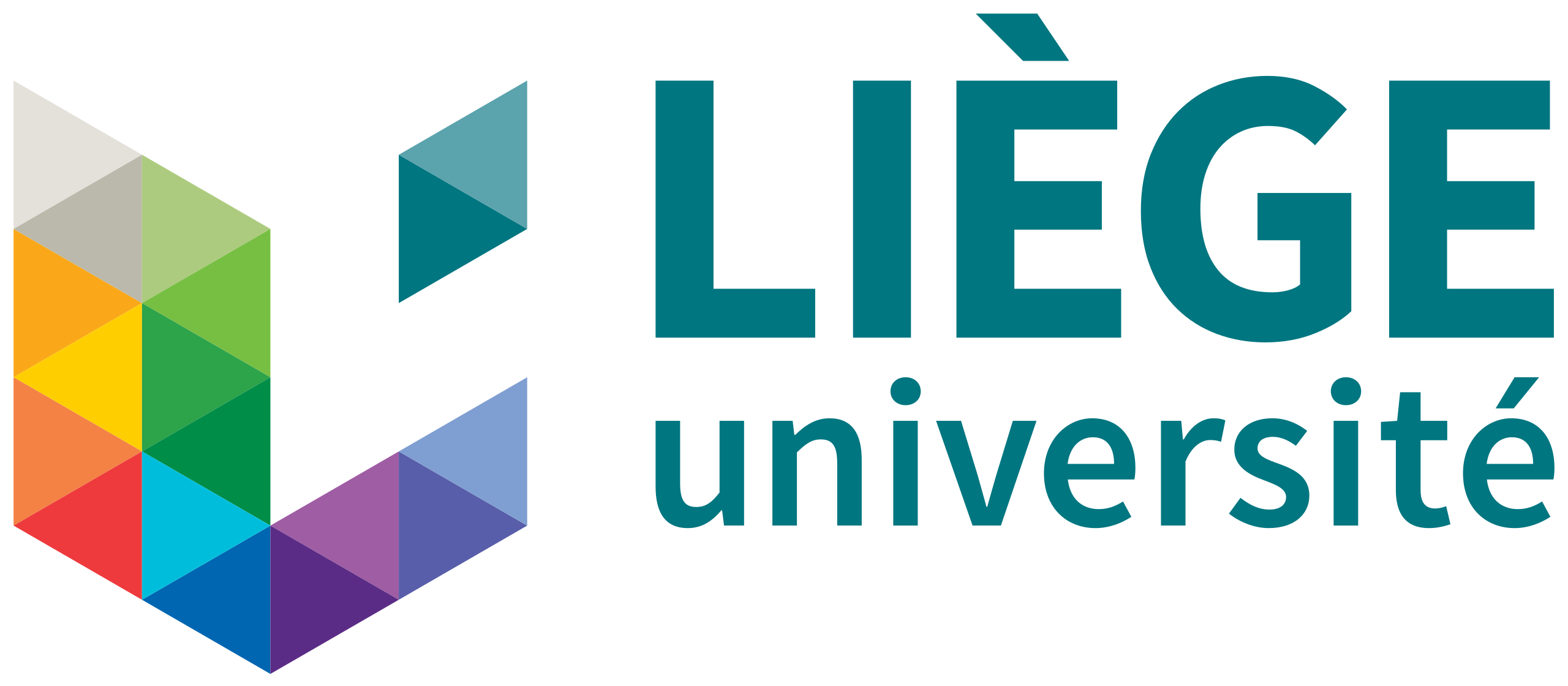}
    \end{minipage}
    \begin{minipage}[c]{0.4\textwidth}
    \begin{flushright}
        {\mySubmissionMonth} {\mySubmissionYear}
      \end{flushright}
    \end{minipage}\\[1.5cm]

  \end{center}

\end{titlepage}

  \newpage 
  
  \onehalfspacing
\refstepcounter{dummy}
\pdfbookmark[1]{\contentsname}{tableofcontents} 
\setcounter{tocdepth}{2}
\setcounter{secnumdepth}{3}
\manualmark
\markboth{\spacedlowsmallcaps{\contentsname}}{\spacedlowsmallcaps{\contentsname}}
\tableofcontents
\automark[section]{chapter}
\renewcommand{\chaptermark}[1]{\markboth{\spacedlowsmallcaps{#1}}{\spacedlowsmallcaps{#1}}}
\renewcommand{\sectionmark}[1]{\markright{\thesection\enspace\spacedlowsmallcaps{#1}}}

\clearpage






  
  \chapter*{Abstract}
\addcontentsline{toc}{chapter}{Abstract}

Scientific knowledge expands by observing the world, hypothesizing some theories about it, and testing them against collected data. When those theories take the form of statistical models, statistical analyses are involved in the process of testing and refining scientific hypotheses. Those analyses can become challenging with the increased complexity of proposed theories. In this thesis, we focus on statistical models that take the form of scientific simulators and provide background about how machine learning can be used for statistical analyses in this context.

The first part of this thesis is about showing empirically that performing statistical analyses with machine learning involves a degree of approximation. Specifically, all statistical analyses involve a level of uncertainty in the conclusions drawn, and we show that approximations can lead to overconfident conclusions. We draw caution regarding such overconfident conclusions and introduce a criterion to diagnose overconfident approximations.

In the second part, we introduce balancing, a way to regularize machine learning models to reduce overconfidence and favor calibrated or underconfident approximations. Balancing is first introduced for neural ratio estimation algorithms and then extended to other algorithms. Intuition about why balancing leads to less overconfident solutions is provided, and it is shown empirically that balanced algorithms are often either close to calibrated or underconfident. 

The third part shows that Bayesian neural networks can also be used to mitigate the overconfidence of approximations. Unlike balancing, no regularization is required, and this solution can then work with few training samples and, hence, computationally expensive simulators. To that end, a new Bayesian neural network prior tailored for simulation-based inference is developed, and empirical results show a reduction in overconfidence compared to similar solutions without Bayesian neural networks.
  
  \chapter*{Acknowledgments}
\addcontentsline{toc}{chapter}{Acknowledgments}

First of all, I would like to sincerely thank the members of my jury for accepting to evaluate this work and for the time and attention they devoted to reading it.

I would then like to warmly thank Professor Gilles Louppe. I believe I can say that I was fortunate to have a PhD experience that was both enjoyable and enriching. Many factors contributed to this, but you were certainly one of the main ones. You were always present and deeply committed to both research and the supervision of your PhD students. I never felt left alone when facing a difficulty without being able to come and discuss it with you. Beyond your support, I always felt your genuine desire to help me grow as a researcher — by encouraging me to explore my own ideas, even when some of them did not lead very far; by challenging those ideas; and by taking the time to provide thoughtful and sometimes lengthy feedback, from which I always emerged having learned something valuable. Outside of your role as supervisor, I will also always remember the many enjoyable moments shared during our coffee breaks.

Of course, Gilles is not the only reason these PhD years were so great. I would like to thank Adrien for putting up with sharing an office with me for four years. Thank you for not throwing my keyboard out of the window — even though I am sure the temptation must have been there at times. More seriously, I already miss our daily discussions, sometimes serious, often not.

I would also like to thank the “General Tour” team — these years would certainly not have been as joyful without all of you. Thank you as well to the “Cards” team for the wonderful lunch breaks and their occasional extensions. More broadly, I would like to thank the entire department for fostering such a positive and supportive atmosphere.

I am also deeply grateful to my family, in the broadest sense of the word. Thank you for always being there, for checking in on me, and for so often offering your valuable advice.

Finally, Valentine, I am incredibly lucky to have you by my side. Thank you for the joy you bring into my daily life, softening moments that might otherwise feel much more difficult. You are always there to support me — thank you for being so understanding of my little stresses and for always helping me overcome or rationalize them. Thank you as well for helping whenever you can, especially during these past few intense months. And thank you for taking an interest in what I do — even if I sometimes have to speak what feels like another language.

  \chapter*{List of publications}
\addcontentsline{toc}{chapter}{Publications}

The core part of this thesis is based on the following publications. Each publication comes with a publicly available implementation that can be found on their associated github repositories.
\begin{itemize}
    \item Hermans, J.$^*$, Delaunoy, A.$^*$, Rozet, F., Wehenkel, A., \& Louppe, G. (2022). A crisis in simulation-based inference? beware, your posterior approximations can be unfaithful. Transactions on Machine Learning Research. \\
    This paper is discussed in Chapter \ref{c:crisis} and the implementation can be found at \url{https://github.com/montefiore-institute/trust-crisis-in-simulation-based-inference}
    \item Delaunoy, A.$^*$, Hermans, J.$^*$, Rozet, F., Wehenkel, A., \& Louppe, G. (2022). Towards reliable simulation-based inference with balanced neural ratio estimation. Advances in Neural Information Processing Systems, 35, 20025-20037. \\
    This paper is discussed in Chapter \ref{c:bnre} and the implementation can be found at \url{https://github.com/montefiore-institute/balanced-nre}
    \item Delaunoy, A.$^*$, Miller, B. K.$^*$, Forré, P., Weniger, C., \& Louppe, G. (2023). Balancing Simulation-based Inference for Conservative Posteriors. In Fifth Symposium on Advances in Approximate Bayesian Inference. \\
    This paper is discussed in Chapter \ref{c:balancing_sbi} and the implementation can be found at \url{https://github.com/ADelau/balancing_sbi}
    \item Delaunoy, A.$^*$, Bonardeaux, M. D. L. B.$^*$, Mishra-Sharma, S., \& Louppe, G. (2024). Low-Budget Simulation-Based Inference with Bayesian Neural Networks. arXiv preprint arXiv:2408.15136. \\
    This paper is discussed in Chapter \ref{c:bnn} and the implementation can be found at \url{https://github.com/ADelau/low_budget_sbi_with_bnn}
\end{itemize}

In addition, I authored the following publications during my thesis.
\begin{itemize}
    \item Delaunoy, A., Wehenkel, A., Hinderer, T., Nissanke, S., Weniger, C., Williamson, A., \& Louppe, G. (2020). Lightning-Fast Gravitational Wave Parameter Inference through Neural Amortization. In Machine Learning and the Physical Sciences, NeurIPS 2020 workshop.
    \item Delaunoy, A., \& Louppe, G. (2021). SAE: Sequential Anchored Ensembles. In Bayesian Deep Learning, NeurIPS 2021 workshop.
\end{itemize}
\newpage
I have also participated to the following publication.
\begin{itemize}
   \item Falkiewicz, M., Takeishi, N., Shekhzadeh, I., Wehenkel, A., Delaunoy, A., Louppe, G., \& Kalousis, A. (2024). Calibrating neural simulation-based inference with differentiable coverage probability. Advances in Neural Information Processing Systems, 36.
\end{itemize}
  
  \cleardoublepage
  \pagestyle{scrheadings}
  \pagenumbering{arabic}
  \onehalfspacing

  
  \chapter{Introduction}\label{c:introduction}
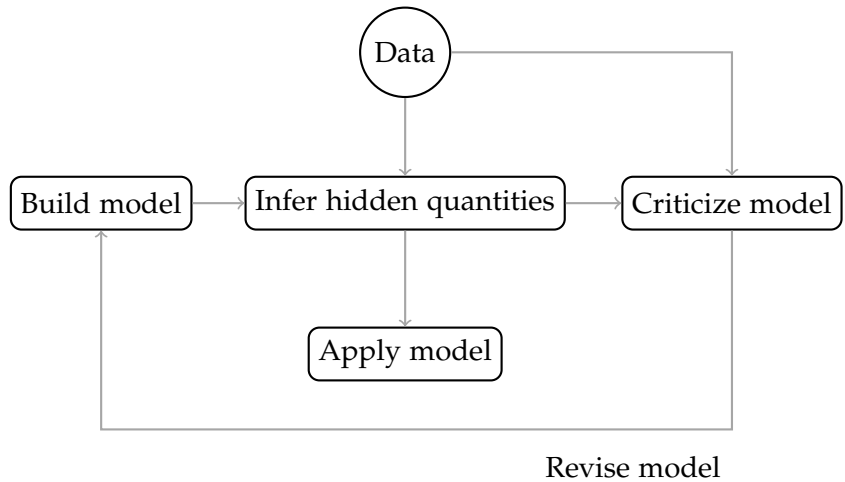
\begin{figure}
    \centering
    
    \begin{tikzpicture}[scale=1]
        \tikzstyle{box} = [draw=black, fill=white, thick, rectangle, anchor=west, rounded corners, minimum height=0.7cm]
        \tikzstyle{circlebox} = [draw=black, fill=white, thick, circle, minimum height=0.7cm]
        \tikzstyle{greyarrow} = [thick, ->, gray!70]
        \tikzstyle{blackarrow} = [thick, ->, black]
        \tikzstyle{bigarrow} = [thick, ->, black, double]
        \tikzstyle{line} = [thick, -, black]

        \node [box, anchor=center, align=center] at (0,0) (label_1) {Build model};
        \node [box, anchor=center, align=center] at (4,0) (label_2) {Infer hidden quantities};
        \node [box, anchor=center, align=center] at (8.3,0) (label_3) {Criticize model};
        \node [circlebox, anchor=center, align=center] at (4,2) (label_4) {Data};
        \node [box, anchor=center, align=center] at (4,-2) (label_5) {Apply model};
        
        \draw [greyarrow] (label_1.east) -- (label_2.west);
        \draw [greyarrow] (label_2.east) -- (label_3.west);
        \draw [greyarrow] (label_4.south) -- (label_2.north);
        \draw [greyarrow] (label_4.east) -- (8.3, 2) -- (label_3.north);
        \draw [greyarrow] (label_2.south) -- (label_5.north);
        \draw [greyarrow] (label_3.south) -- (8.3, -3) -- (0, -3) -- (label_1.south);

        \node [anchor=center] at (7, -3.5) {Revise model};
    \end{tikzpicture}

    \caption{Visualization of Box's loop \citep{box1976science, blei2014build}. The loop iterates over three main elements: model building, model's hidden quantities inference, and model criticism. Iterating over those three elements aims to build better scientific models, taking into account model criticism feedback.}
    \label{fig:box_loop}
    
\end{figure}

  Scientific discoveries are driven by data, either collected through controlled experiments or by observing the world. Those scientific theories can take the form of a mathematical model for some phenomenon in the world. In such cases, the model then represents our current knowledge about that phenomenon. For example, Newton's laws capture some knowledge about motion dynamics and Maxwell's equation knowledge about electromagnetism. Scientific theories could also take other forms when not suited for mathematical modeling. In this thesis, we keep the focus on mathematical modeling. \citet{box1976science, blei2014build} introduced the Box's loop, a scientific method for constructing models for phenomena from data. This method is represented in Figure \ref{fig:box_loop}. The loop contains three steps that are repeated to improve a model iteratively. The first step is to build a model. This involves some creativity and intuition to develop equations that could explain the observations made. The model might include parameters that are unknown. For example, if we consider motion under a gravitational force, it is often modeled as being proportional to object masses and the inverse of the squared distance between the objects. A multiplicative constant is then added to obtain the scale of this force. It would be hard to come up with a value for this multiplicative constant simply with intuition. The second step is then to infer hidden quantities of the model, such as this multiplicative constant, from data. Once a model has been built and its possible hidden quantities have been inferred, it can be used to make predictions about the world, and those predictions can be challenged against pieces of data. Discrepancies between predictions and data can be used to criticize the model, for example, by identifying some effects that are not taken into account by the model. This model criticism is then a feedback loop, allowing the building of more accurate models. By repeating this procedure, more and more accurate models can be built.

Box's loop provides a way to iteratively build a collection of models. \citet{popper1959logic} argue that a model cannot be verified. However, a model can be refuted empirically. This collection of models should then be pruned out by removing models that can be shown to not accurately reflect the world. This process is called falsification. Refuting models happens both on model structure and hidden quantities. If we take back the example of motion under gravitational force, one could refute that gravitational force is proportional to object masses and the inverse of the squared distance between the objects or refute some multiplicative constant values. 

Historically, scientists kept models simple enough such that inferring hidden quantities and testing the model could be done analytically. Nowadays, with the increase in computational power and advances in algorithmics for numerical approximations, it is possible to deal with models that do not allow analytical inference and testing. In addition, as we keep refining models to capture all aspects, such complex models are often needed. While those advances allow the construction of more accurate models, hidden quantity inference and model testing are now often done approximately. This raises questions as to how it impacts the Popperian falsification procedure. Can a model or hidden quantities be reliably refuted by an approximate inference or testing procedure? It could be that our procedure leads to refutation only because of approximations and that if that same procedure was applied without approximations, refutation could not be obtained. Incorrect refutation of models could then ultimately prevent scientists from exploring valid models and lead them in the wrong direction.

In this thesis, we aim to adapt modern scientific reasoning to take into account the approximations made. At the root of this adapted methodology is the inclusion of the additional source of uncertainty brought by those approximations. Statistical methods are modified to produce inflated uncertainty estimates based on how the approximations are likely to affect the end result. In particular, we argue that underestimating the uncertainty is more harmful than overestimating it for Popperian falsification. Indeed, underestimation of uncertainty can lead to incorrect refutation, while overestimation can only lead to failing to refute a model. With this in mind, we design new statistical methods to try to avoid underestimation of the uncertainty and, hence, overconfidence. These methods can then be used more reliably for scientific analyses, and more trust can be given to conclusions drawn. We focus on models that take the form of a simulator. Simulators are becoming heavily used in science as they can model complex phenomena. Examples include agent-based models that can be found in the fields of epidemiology or economics. Simulators are also used in many fields of physical sciences, such as cosmology or particle physics, as well as in biology. Stochastic simulators are inherently complex to deal with as they involve many stochastic steps to model the full phenomena. If we consider agent-based models, they involve the stochastic actions of all those agents or the stochastic consequences of those actions. In addition, we focus on the problem of inferring hidden quantities of those models and on machine learning based methods for approximate inference. 

The thesis is outlined as follows. In Chapter \ref{c:Statistical_modeling} and \ref{c:sbi}, we provide background in statistical modeling and simulation-based inference, respectively. In Chapter \ref{c:crisis}, we showcase how simulation-based inference methods can lead to overconfidence. In Chapter \ref{c:bnre}, we introduce balancing, the first improvement of simulation-based inference methods toward more reliability, and in Chapter \ref{c:balancing_sbi}, we extend balancing to a broader range of methods. We show in Chapter \ref{c:bnn} that Bayesian neural networks can be used to improve the reliability of simulation-based inference methods, even in low-data regimes. We conclude and provide additional discussions in Chapter \ref{c:discussion}.

  \chapter{Statistical modeling}\label{c:Statistical_modeling}
  Statistical modeling is about specifying relationships between one or more random variables and non-random variables. Those relationships usually come from a combination of prior knowledge and inference from data. In this chapter, we describe different ways of reasoning about those relationships. We will also illustrate the concept of model using the example of a feather falling due to gravity.

\section{Probability and statistical models}
A probability model is a mathematical representation, that includes randomness, of a phenomenon. It is defined in the following way.
\begin{definition}
    A probability model is defined by three components: a sample space $\Omega$, a space of possible events $\mathcal{A}$ within the sample space, and a mapping $P$ from an event $A \subset \Omega$ to a probability $P(A)$.
\end{definition}
Hence, it encapsulates all the information about the probabilities of experiment outcomes regarding that random phenomenon. In all generality, statistics is concerned with inferring from an outcome of a probabilistic experiment interesting things about the data generating process. For example, one might be interested in extracting, from a family of models, the probability model that is the most consistent with the experimental outcome (point estimation), a range of models that are consistent with this outcome could also be extracted (interval estimation). Another interesting piece of information is whether or not a hypothesis can be refuted from data with some probability (hypothesis testing). To answer such questions, we start with a statistical model.
\begin{definition}
A statistical model is a mathematical model that embodies a set of statistical assumptions concerning the generation of sample data.
\end{definition}
It often takes the form of a parametric family of probability models. In the remainder of this thesis, we will denote those parameters by $\btheta$. The statistical model with parameter values then jointly defines a probability model.

\begin{figure}[h!]
    \centering
    
    \begin{tikzpicture}[scale=1.]
        \tikzstyle{box} = [draw=black, fill=white, thick, rectangle, anchor=west, rounded corners, minimum height=0.7cm]
        \tikzstyle{greyarrow} = [thick, ->, gray!70]
        \tikzstyle{blackarrow} = [thick, ->, black]
        \tikzstyle{doubleblackarrow} = [thick, |-|, black]
        \tikzstyle{bigarrow} = [thick, ->, black, double]
        \tikzstyle{line} = [thick, -, black]

        \node [anchor=center, rotate=25] at (0,10) (label_1) {\includegraphics[height=2cm]{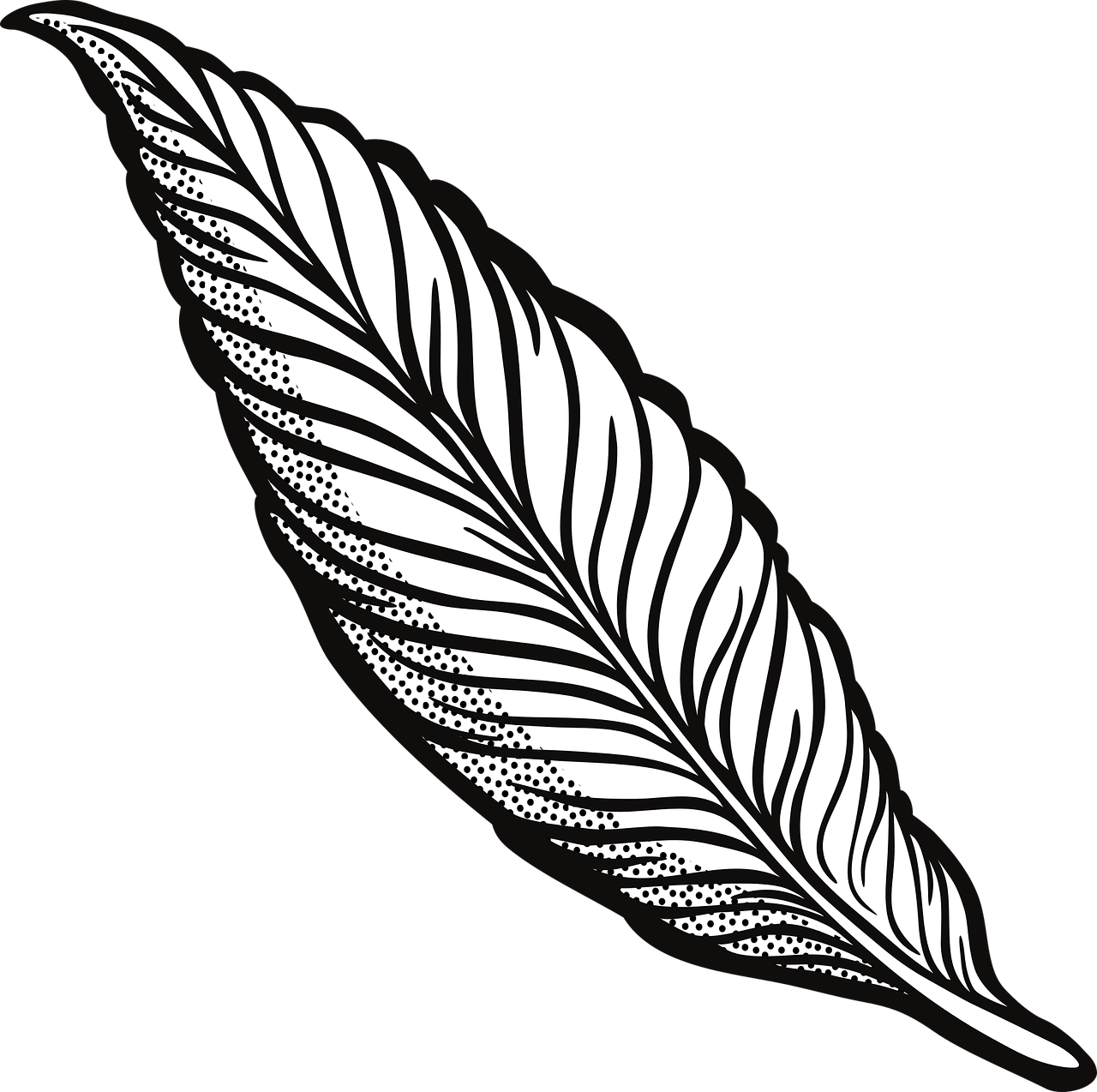}};
        \node [anchor=center, rotate=65] at (0,5) (label_2) {\includegraphics[height=2cm]{figures/stat_modeling/feather.png}};
        \node [anchor=center, rotate=45] at (0,0) (label_3) {\includegraphics[height=2cm]{figures/stat_modeling/feather.png}};

        \node [anchor=center] at (5,0) (label_4) {\includegraphics[height=2cm]{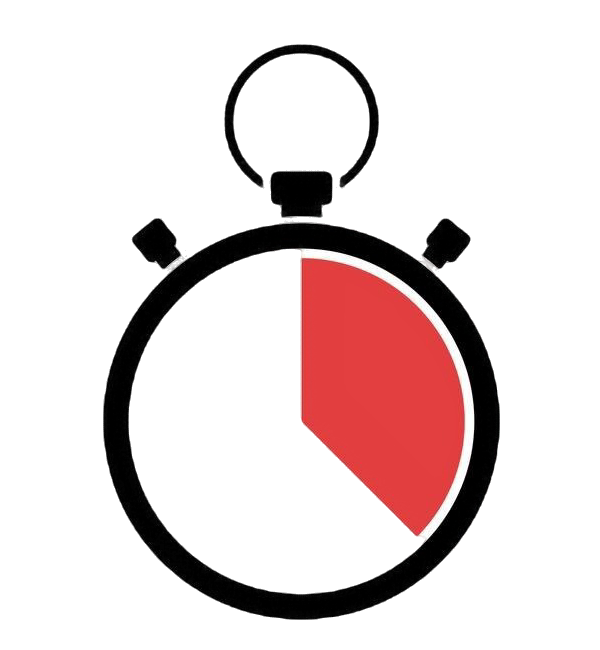}};
        \node [anchor=center] at (5,10) (label_5) {\includegraphics[height=2cm]{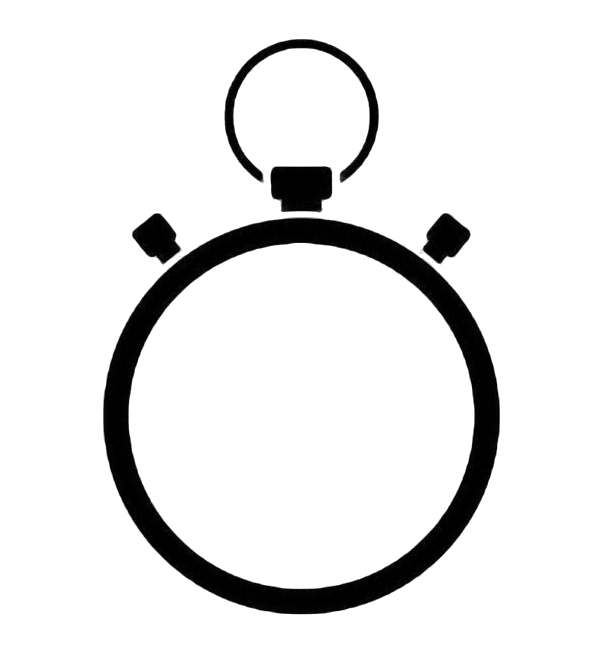}};
        \draw [line] (label_4.east) to[out=45,in=-45, distance=3cm] (label_5.east);
        \node [align=center] at (8.7, 5) {measured\\time};

        \draw [blackarrow] (0,9.2) -- (0, 8.5);
        \node [anchor=west] at (0.1,8.8) {$\mathbf{F} = m\ g\ \mathbf{e}$};

        \draw [doubleblackarrow] (-2,-0.5) -- (-2, 10.5);
        \node [anchor=east] at (-2.5, 5) {height};
        
    \end{tikzpicture}

    \caption{Illustration of the experiment of a feather falling. A gravitational force $\mathbf{F} = m\ g\ \mathbf{e}$ is applied on the feather and the fall time is measured.}
    \label{fig:feather_model}
    
\end{figure}

As an illustration, we will take the example of a feather falling. In this setting, we observe the time it takes for the feather to fall from a given height, and we want to model the fall. This is illustrated in Figure \ref{fig:feather_model}. We assume that the fall follows Newtonian dynamics and that the gravitational force applied to the feather is 
\begin{equation}
    \mathbf{F} = m\ g\ \mathbf{e},        
\end{equation}
where $m$ is the mass of the feather, $g$ is the gravity constant, and $\mathbf{e}$ is a unit vector pointing towards the earth. If no other forces are applied to the object, it takes $\sqrt{\frac{2h}{g}}$ seconds to fall from a height $h$. 

We will model the measurement error as a normal distribution centered at $0$:
\begin{equation}
    \text{measurement error} \sim \mathcal{N}(0, \sigma_{\text{measurement}}).
\end{equation}
If we combine the two, it gives the following statistical model for the measured fall time
\begin{equation}
    \text{measured time} \sim \mathcal{N}\left(\sqrt{\frac{2h}{g}}, \sigma_{\text{measurement}}\right).
\end{equation}
In this setting, if we assume that $\sigma_{\text{measurement}}$ and height $h$ are known, there is only one parameter $\btheta = g$.

\section{Frequentist and Bayesian statistics}
There are two main approaches to statistical inference: the frequentist and Bayesian paradigms. At the root of those two paradigms is the interpretation of probability. Frequentist statistics are based on the frequency interpretation of probability. The probability of an event is the observed frequency of that event in the limit of infinite observations. Bayesian statistics interpret probability as a degree of belief rather than observed frequencies. 

The most direct consequence of those two philosophical approaches to probability is that, in the frequentist philosophy, we cannot assign a probability to the parameters $\btheta$ of the statistical model as those are not random events that can be observed. In the example of a feather falling, it is not possible to design a random experiment that would yield several values for $g$. It is an unknown parameter that is not random, and it would then make no sense to assign a probability to $g$ in the frequentist philosophy. However, one can have a belief about what $g$ could be, based on observations, and a probability can then be assigned to parameters in the Bayesian philosophy.

In the remainder of this thesis, we will denote the observations by $\bx$ and by $p$ the probability density function or probability mass function for continuous and discrete random variables, respectively. The data-generating process that models the fact of performing an experiment will be denoted as $p^*(\bx)$. Performing an experiment is, hence, equivalent to sampling an observation from this data-generating process $\bx \sim p^*(\bx)$. A central component of statistical inference is the likelihood $p(\bx | \btheta)$. It defines the probability of observing $\bx$ according to the statistical model with parameters $\btheta$. In our example, it defines the probability of observing some fall time ($\bx$) given a gravity constant $g$ ($=\btheta$). The likelihood is used in both frequentist and Bayesian statistics. 

In the Bayesian philosophy, we can go a step further and define a belief over parameters $\btheta$ given the observations $\bx$. This belief is called the posterior, $p(\btheta|\bx)$, and is defined through the Bayes rule
\begin{equation}
    p(\btheta|\bx) = \frac{p(\bx | \btheta) p(\btheta)}{p(\bx)}.
\end{equation}
The term $p(\btheta)$ is called the prior and is a part of the statistical model that represents the belief about the parameters before observing $\bx$. The term $p(\bx)$ is called the evidence and is the marginal probability of observing $\bx$. It can be computed through the law of total probability
\begin{equation}
    p(\bx) = \int p(\bx|\btheta) p(\btheta) d\btheta .
\end{equation}
The posterior is a central component of Bayesian inference as it defines all the information we have about the parameters $\btheta$. It will be used to derive all the other quantities of interest.

\section{Point estimation}
Point estimation corresponds to the task of finding the parameters that lead to an appropriate probability model according to the data. In other words, we aim to find parameter values $\btheta$ such that $p(\bx | \btheta)$ is close to $p^*(\bx)$. If we measure how close those distributions are using the KL-divergence and assume we have access to $N$ samples $\bx_1, ..., \bx_N \sim p^*(\bx)$, we aim for
\begin{equation}
\begin{aligned}
    \argmin_{\btheta} \text{KL}\left[p^*(\bx) || p(\bx | \btheta) \right] 
    &= \argmin_{\btheta} \int \log\left(\frac{p^*(\bx)}{p(\bx | \btheta)}\right) p^*(\bx) d\bx\\
    &\simeq \argmin_{\btheta} \frac{1}{N}\sum_{i=1}^N \log\left(\frac{p^*(\bx_i)}{p(\bx_i | \btheta)}\right) \\
    &= \argmax_{\btheta} \frac{1}{N}\sum_{i=1}^N \log\left(p(\bx_i | \btheta)\right) - \frac{1}{N}\sum_{i=1}^N \log\left(p(\bx_i)\right) \\
    &= \argmax_{\btheta} \sum_{i=1}^N \log\left(p(\bx_i | \btheta)\right) \\
    &= \argmax_{\btheta} \prod_{i=1}^N p(\bx_i | \btheta). \\
\end{aligned}
\end{equation}

In the remainder, we will use $\bx$ to represent both single and multiple observations from a phenomenon. The model is then a model for a single or multiple observations from the phenomenon. A model for multiple observations can be obtained from a model for single observations. $p(\bx_1, ..., \bx_N | \btheta) = \prod_{i=1}^N p(\bx_i | \btheta)$. An appropriate probability model can then be the one that maximizes the likelihood. This leads to the maximum likelihood estimator (MLE) of those parameters.

\begin{definition}
   The maximum likelihood estimator $\btheta_{\text{MLE}}$ is defined as
    \begin{equation}
        \btheta_{\text{MLE}} = \argmax_{\btheta} p(\bx | \btheta).
    \end{equation}
\end{definition}

Let us assume that we make a feather falling experiments for heights $3\text{m}$, $5\text{m}$ and $7\text{m}$ and observe measured falling time of $4.68\text{s}$, $6.97\text{s}$ and $9.69\text{s}$ respectively. The probability of those observations given our model is then

\begin{equation}
    \begin{aligned}
    p(\bx | g) &= \frac{1}{\sigma_{\text{measurement}} \sqrt{2\pi}}\exp\left(-\frac{\left(4.68-\sqrt{\frac{6}{g}}\right)^2}{2\sigma_{\text{measurement}}^2}\right) \\
               &\times \frac{1}{\sigma_{\text{measurement}} \sqrt{2\pi}}\exp\left(-\frac{\left(6.97-\sqrt{\frac{10}{g}}\right)^2}{2\sigma_{\text{measurement}}^2}\right) \\
               &\times \frac{1}{\sigma_{\text{measurement}} \sqrt{2\pi}}\exp\left(-\frac{\left(9.69-\sqrt{\frac{14}{g}}\right)^2}{2\sigma_{\text{measurement}}^2}\right) \\
               &= \left(\frac{1}{\sigma_{\text{measurement}} \sqrt{2\pi}}\right)^3 \exp\left(\frac{-\left(4.68-\sqrt{\frac{6}{g}}\right)^2 - \left(6.97-\sqrt{\frac{10}{g}}\right)^2 - \left(9.69-\sqrt{\frac{14}{g}}\right)^2}{2\sigma_{\text{measurement}}^2}\right).
    \end{aligned}
\end{equation}
We have that
\begin{equation}
    \log p(\bx | g) \propto -\left(4.68-\sqrt{\frac{6}{g}}\right)^2 - \left(6.97-\sqrt{\frac{10}{g}}\right)^2 - \left(9.69-\sqrt{\frac{14}{g}}\right)^2.
\end{equation}

From that we can compute the maximum likelihood estimator

\begin{equation}
    \begin{aligned}
    g_{\text{MLE}}
    &= \argmax_{g} p(\bx | g) \\
    &= \argmax_{g} \log p(\bx | g) \\
    &= \argmax_{g} -\left(4.68-\sqrt{\frac{6}{g}}\right)^2 - \left(6.97-\sqrt{\frac{10}{g}}\right)^2 - \left(9.69-\sqrt{\frac{14}{g}}\right)^2 \\
    &\simeq 0.1849.
    \end{aligned}
\end{equation}

In Bayesian philosophy, it would make sense to consider instead the probability model corresponding to the parameters in which we have the highest belief. The belief over model parameters is the posterior. This leads to the maximum a posteriori estimator (MAP) of those parameters.
\begin{definition}
    The maximum a posteriori estimator $\btheta_{\text{MAP}}$ is defined as
    \begin{equation}
        \btheta_{\text{MAP}} = \argmax_{\btheta} p(\btheta | \bx) = \argmax_{\btheta} p(\bx | \btheta) p(\btheta).
    \end{equation}
\end{definition}
Let us assume that we have the a priori belief that $\log g$ should follow a uniform distribution between $\log(0.01)$ and $\log(1000)$. We have
\begin{equation}
    p(g) = \frac{1}{g(\log(1000) - \log(0.01))}.
\end{equation}
Consequently,
\begin{equation}
    \begin{aligned}
    \log p(g | \bx) &\propto \log p(\bx | g) + \log p(g) \\
    &= -3\log(\sigma_{\text{measurement}} \sqrt{2\pi}) + \frac{-\left(4.68-\sqrt{\frac{6}{g}}\right)^2 - \left(6.97-\sqrt{\frac{10}{g}}\right)^2 - \left(9.69-\sqrt{\frac{14}{g}}\right)^2}{2\sigma_{\text{measurement}}^2} \\
    &- \log(g) - \log(\log(1000) - \log(1)). \\
    \end{aligned}
\end{equation}
If we assume that $\sigma_{\text{measurement}} = 0.5\text{s}$, then the maximum a posteriori is
\begin{equation}
   g_{\text{MAP}} = \argmax_{g} p(g | \bx) \simeq 0.1838.
\end{equation}
The parameters $\btheta_{\text{MLE}}$ and $\btheta_{\text{MAP}}$ could then optionally be used to obtain a probability model. This probability model can then be used to make predictions on future observations. If we consider the maximum likelihood estimator $g_{\text{MLE}} = 0.1849$, we can predict that the measured fall time from a height $h=6\text{m}$ would follow the distribution $\mathcal{N}(\mu = \sqrt{\frac{2h}{g_{\text{MLE}}}} = 8.056, \sigma=\sigma_{\text{measurement}})$.

\section{Interval estimation}
Instead of a point estimation, we might often want to obtain a range of models that are consistent with the observation. This is done through interval or region estimation. In frequentist statistics, these are called confidence regions or intervals. 
\begin{definition}
    A level $\alpha$ confidence region generation function $C_\alpha(.)$ is a function such that
    \begin{equation}
        P(\btheta \in C_\alpha(\bX)) \geq \alpha, \quad \forall \btheta,
    \end{equation}
    where $\bX$ is the random variable associated with the observation generated according to the model with parameter $\btheta$.
\end{definition}
A level $\alpha$ confidence region generation function is then a function that, when given as input a random variable associated with the observation according to the model with some parameters $\btheta$, outputs a random region that contains those parameters with probability $\alpha$. The probability statement is then on the constructed region and not on the parameters $\btheta$ that are not random variables. When evaluated on an observation $\bx$, this yields a level $\alpha$ confidence region $C_\alpha(\bx)$. 

Such regions are sometimes hard to construct analytically and must be approximated. This leads to the definition of coverage of a region generation function. 
\begin{definition}
    The frequentist coverage probability of a region generator function is the probability that generated regions based on some observations generated from the model with some parameters contain those parameters,
    \begin{equation}
        \text{coverage}(C, \btheta) = P(\btheta \in C(\bX)).
    \end{equation}
\end{definition}
It follows that if a region generation function has coverage greater of equal to $\alpha$ for all $\btheta$, then this is a valid level $\alpha$ confidence region generation function. The coverage probability can then be used to measure the validity of a region generation function.

When the parameters $\btheta$ are one-dimensional and the generated regions are contiguous, those regions correspond to confidence intervals. Confidence intervals are defined by a lower function $L_\alpha$ and an upper function $U_\alpha$ such that
\begin{equation}
    P(L_\alpha(\bX) \leq \btheta \leq U_\alpha(\bX)) \geq \alpha, \quad \forall \btheta.
\end{equation}

In the Bayesian setting, we assign a belief probability on parameters, and credible regions are constructed.
\begin{definition}
    A level $\alpha$ credible region is defined as a region $C_\alpha$ such that
    \begin{equation}
        \int_{C_\alpha} p(\btheta|\bx) d \btheta = \alpha.
    \end{equation}
\end{definition}
It is a region in which we have a belief that the parameters have a probability $\alpha$ to be included in. This is then a probability statement about the unknown parameters themselves. Similarly to confidence intervals, when the parameters $\btheta$ are one-dimensional and we consider contiguous regions, those regions are defined by an upper and lower bound and are called credible intervals. There are many regions satisfying this property depending on which part of the parameter space is prioritized to be included. Although not the only valid choice, a common choice is to consider the highest posterior density credible region. This is defined as a credible region for which all included elements of the parameter space have a higher posterior density than the ones excluded. In other words, it prioritizes high posterior density elements to be included in the credible region.  

While the maximum a posteriori could be computed using only the likelihood $p(\bx | \btheta)$ and the prior $p(\btheta)$, computing a credible region requires the knowledge of the posterior $p(\btheta | \bx)$ explicitly. Computing the posterior requires computing the evidence term $p(\bx)$, which involves the computation of the integral $\int p(\bx|\btheta) p(\btheta) d\btheta$ that might be hard to solve analytically. Therefore, we often have to rely on approximations of the posterior to derive credible regions. Two widely used techniques to approximate the posterior are Markov Chain Monte-Carlo and Variational Inference techniques. 

\paragraph{Markov chain Monte-Carlo}
Markov Chain Monte-Carlo (MCMC) techniques consist in building a Markov chain whose stationary distribution is the posterior. A Markov chain defines a stochastic transition between states represented by a probability distribution over the next states, given the current state. In this setting, states are parameter values, and the transitions are also conditioned on the observation. The Markov chain then defines the transition $p(\btheta_{i+1}|\btheta_i, \bx)$. A stationary distribution of a Markov chain is a distribution over $\btheta$ such that applying the transition on elements from that distribution returns elements following the same distribution. In other words, $\pi$ is a stationary distribution if, for every parameter values $\tilde{\btheta}$,
\begin{equation}
    \int \pi(\btheta) p(\tilde{\btheta} | \btheta, \bx) d \btheta = \pi(\tilde{\btheta}).
\end{equation}
It follows that if this stationary distribution is the posterior and convergence to that stationary distribution is guaranteed, then parameters sampled from that Markov chain will asymptotically follow the posterior distribution. 

The Metropolis-Hastings algorithm \citep{metropolis1953equation, hastings1970monte} constructs a Markov chain that asymptotically follows the posterior distribution. It uses a proposal distribution $g(\tilde{\btheta}|\btheta_i, \bx)$ that is then corrected to get transitions leading to the posterior distribution. The transitions are defined as follows.
\begin{enumerate}
    \item Sample $\tilde{\btheta} \sim g(\tilde{\btheta} | \btheta_i, \bx)$.
    \item With probability $\min \left(1, \frac{p(\tilde{\btheta}|\bx) g(\btheta|\tilde{\btheta}, \bx)}{p(\btheta|\bx) g(\tilde{\btheta}|\btheta, \bx)} \right)$ set $\btheta_{i+1} = \tilde{\btheta}$, otherwise set $\btheta_{i+1} = \btheta_i$.
\end{enumerate}
To show that such transitions lead to the stationary distribution being the posterior, let us first show that a distribution $\pi$ is a stationary distribution if it satisfies the detailed balance equation
\begin{equation}
    \pi(\btheta) p(\tilde{\btheta} | \btheta, \bx) = \pi(\tilde{\btheta}) p(\btheta | \tilde{\btheta}, \bx).
\end{equation}
If that is the case, then 
\begin{equation}
    \int \pi(\btheta) p(\tilde{\btheta} | \btheta, \bx) d \btheta = \int \pi(\tilde{\btheta}) p(\btheta | \tilde{\btheta}, \bx) d \btheta = \pi(\tilde{\btheta}) \int p(\btheta | \tilde{\btheta}, \bx) d \btheta = \pi(\tilde{\btheta}).
\end{equation}
and $\pi$ is then a stationary distribution. If we assume that the transitions are built using a proposal distribution $g(\tilde{\btheta} | \btheta, \bx)$ and an acceptance probability $A(\tilde{\btheta}, \btheta, \bx)$, then
\begin{equation}
    p(\tilde{\btheta} | \btheta, \bx) \propto g(\tilde{\btheta}|\btheta, \bx) A(\tilde{\btheta}, \btheta, \bx).
\end{equation}
To satisfy the detailed balance equation, we need
\begin{equation}\label{eq:detailed_balance}
    \frac{A(\tilde{\btheta}, \btheta, \bx)}{A(\btheta, \tilde{\btheta}, \bx)} = \frac{\pi(\tilde{\btheta}) g(\btheta|\tilde{\btheta}, \bx) }{\pi(\btheta) g(\tilde{\btheta}|\btheta, \bx)}.
\end{equation}
This is satisfied by the following acceptance probability
\begin{equation}
    \label{eq:acceptance_1}
    A(\tilde{\btheta}, \btheta, \bx) = \min\left(1, \frac{\pi(\tilde{\btheta}) g(\btheta|\tilde{\btheta}, \bx) }{\pi(\btheta) g(\tilde{\btheta}|\btheta, \bx)}\right).
\end{equation}
as there is always one of $A(\tilde{\btheta}, \btheta, \bx)$ or $A(\btheta, \tilde{\btheta}, \bx)$ that is equal to one and the other is equal to the right-hand side of Equation \ref{eq:detailed_balance} or its inverse. We aim to build a Markov chain that has the posterior as a stationary distribution and set $\pi(\btheta) = p(\btheta|\bx)$. The posterior ratio can be easily computed as the evidence cancels out:
\begin{equation}
    \label{eq:acceptance_2}
    \frac{p(\tilde{\btheta}|\bx)}{p(\btheta|\bx)} = \frac{p(\bx | \tilde{\btheta})p(\tilde{\btheta})}{p(\bx | \btheta)p(\btheta)}.
\end{equation}
MCMC can, therefore, be used to draw samples from the posterior $p(\btheta | \bx)$ while only being able to evaluate the likelihood $p(\bx | \btheta)$ and the prior $p(\btheta)$. Those samples can then be used to construct approximate credible regions. Let us note that the samples drawn from the Markov chain are not independently and identically distributed (i.i.d.) due to the correlation between consecutive samples and that a sufficient number of steps must be performed before reaching the stationary distribution. Those two elements hence add error to the credible region approximations. With this in mind, it is a good practice to run an MCMC algorithm with multiple chains, i.e., multiple instances of the algorithm, to decrease the correlation between samples. It is also a good practice to discard the first samples produced by each chain when the stationary distribution has not yet been attained. This is called the burn-in period.

The algorithm is correct for any proposal distribution $g(\btheta|\tilde{\btheta}, \bx)$. However, the closer this distribution is to the posterior, the higher the acceptance rate. In addition, the less consecutive samples are correlated, the fewer steps will be required to visit many regions of the posterior space. Much of the research in MCMC has then been directed towards finding the best proposals. Some proposals do not use information about the posterior \citep{Goodman2010ensemble, nelson2013run} while others leverage quantities such as the gradient of the unnormalized posterior \citep{welling2011bayesian, neal2012mcmc}.

\begin{figure}[h!]
    \centering
    \includegraphics[width=0.5\textwidth]{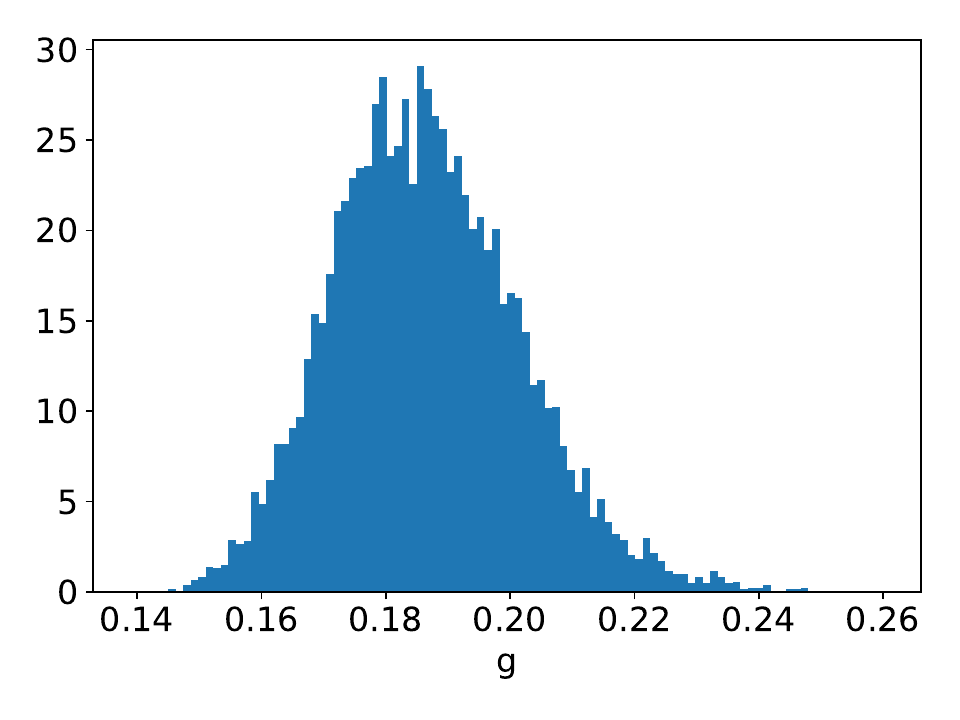}
    \caption{Histogram of samples obtained by running an MCMC algorithm on the feather falling model with measured time $4.68\text{s}$, $6.97\text{s}$ and $9.69\text{s}$ for heights of $3\text{m}$, $5\text{m}$ and $7\text{m}$ respectively.}
    \label{fig:mcmc_feather}
\end{figure}

As an illustration, Figure \ref{fig:mcmc_feather} shows a histogram of samples obtained by running an MCMC algorithm on the feather falling model with the same observations as for point prediction. This histogram then approximates the posterior density. We can observe that there is a single mode and that there are samples roughly between $g=0.14$ and $g=0.25$.

\paragraph{Variational inference}
While MCMC only allows drawing samples from the posterior and does not provide an analytical form of it, variational inference (VI) optimizes a family of distributions to find a member of that family that is close to the actual posterior. The idea is to start with a family of distributions over $\btheta$, also called variational family, which we will denote by $\pi_{\bphi}(\btheta)$. That family has parameters $\bphi$ that will be optimized to find a distribution close to the posterior. For example, if we assume that the parameters follow a multivariate Normal distribution, $\btheta \sim \mathcal{N}(\boldsymbol{\mu}, \boldsymbol{\Sigma})$, then the parameters are $\bphi = (\boldsymbol{\mu}, \boldsymbol{\Sigma})$. 

We optimize those parameters to minimize the KL divergence between the variational distribution and the posterior:
\begin{equation}
    \bphi^* = \argmin_{\bphi} \text{KL}\left[\pi_{\bphi}(\btheta) || p(\btheta | \bx) \right]. 
\end{equation}
Because the posterior $p(\btheta | \bx)$ is unknown, the KL divergence cannot be computed, and an alternative objective will be used. We first rewrite the objective as
\begin{align}
\begin{split}
    \text{KL}\left[\pi_{\bphi}(\btheta) || p(\btheta | \bx) \right] &= \int \pi_{\bphi}(\btheta) \log\frac{\pi_{\bphi}(\btheta)}{p(\btheta | \bx)} d \btheta \\
    &= \int \pi_{\bphi}(\btheta) \log\frac{\pi_{\bphi}(\btheta)}{p(\bx | \btheta)p(\btheta)} d \btheta + \log(p(\bx)).
\end{split}
\end{align}
By reorganizing the terms, we have that
\begin{equation}\label{eq:elbo_derivation}
    \log(p(\bx)) = \text{KL}\left[\pi_{\bphi}(\btheta) || p(\btheta | \bx) \right] + \int \pi_{\bphi}(\btheta) \log\frac{p(\bx | \btheta)p(\btheta)}{\pi_{\bphi}(\btheta)} d \btheta,
\end{equation}
and because $\text{KL}\left[\pi_{\bphi}(\btheta) || p(\btheta | \bx) \right] \geq 0$,
\begin{equation}
    \label{eq:elbo_approx}
    \log(p(\bx)) \geq \int \pi_{\bphi}(\btheta) \log\frac{p(\bx | \btheta)p(\btheta)}{\pi_{\bphi}(\btheta)} d \btheta = \mathbb{E}_{\pi_{\bphi}(\btheta)}\left[\log\frac{p(\bx | \btheta)p(\btheta)}{\pi_{\bphi}(\btheta)}\right].
\end{equation}
For that reason, the term $\mathbb{E}_{\pi_{\bphi}(\btheta)}\left[\log\frac{p(\bx | \btheta)p(\btheta)}{\pi_{\bphi}(\btheta)}\right]$ is often called the evidence lower bound (ELBO). From Equation \ref{eq:elbo_derivation}, we observe that, by maximizing the ELBO and hence making it closer to $\log(p(\bx))$, we push the term $\text{KL}\left[\pi_{\bphi}(\btheta) || p(\btheta | \bx) \right]$ towards zero and hence minimize it. Maximizing the ELBO is hence equivalent to minimizing $\text{KL}\left[\pi_{\bphi}(\btheta) || p(\btheta | \bx) \right]$. The ELBO can be approximated using Monte-Carlo
\begin{equation}
    \mathbb{E}_{\pi_{\bphi}(\btheta)}\left[\log\frac{p(\bx | \btheta)p(\btheta)}{\pi_{\bphi}(\btheta)}\right] \simeq \frac{1}{N} \sum_{i=1}^N \log\frac{p(\bx | \btheta_i)p(\btheta_i)}{\pi_{\bphi}(\btheta_i)}, \quad \btheta_i \sim \pi_{\bphi}(\btheta),
\end{equation}
and is, hence, a valid training objective.

\begin{figure}[h!]
    \centering
    \includegraphics[width=0.5\textwidth]{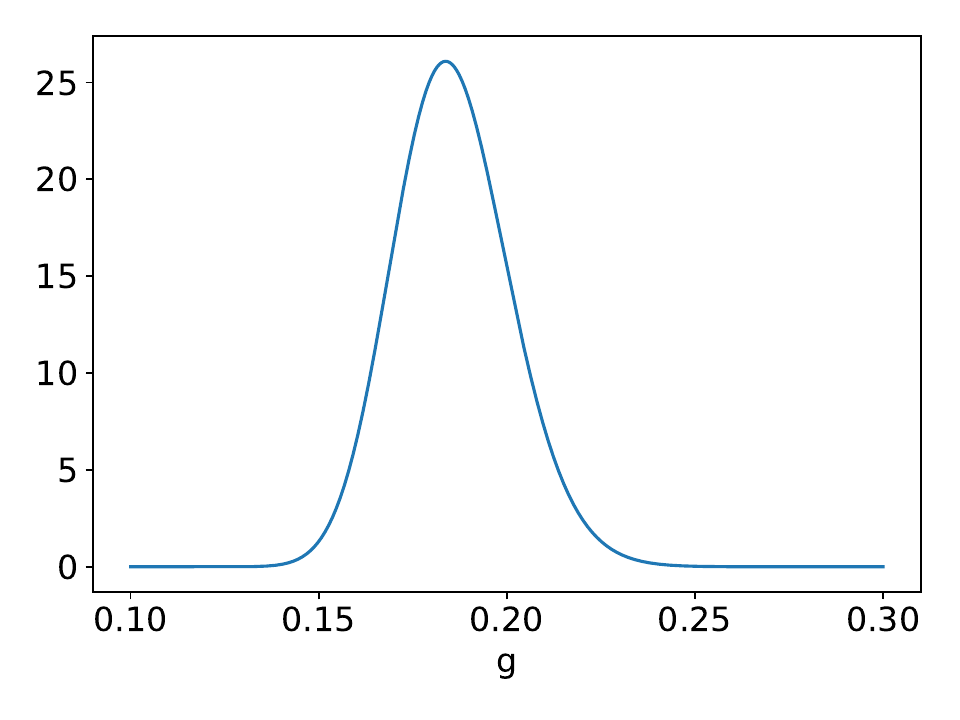}
    \caption{Approximate posterior density obtained by running a variational inference algorithm on the feather falling model with measured time $4.68\text{s}$, $6.97\text{s}$ and $9.69\text{s}$ for heights of $3\text{m}$, $5\text{m}$ and $7\text{m}$ respectively. The plotted variational distribution is a log normal distribution with parameters $\mu = -1.6873$ and $\sigma = 0.0829$.}
    \label{fig:vi_feather}
\end{figure}

Once the variational parameters have been optimized, the distribution $\pi_{\bphi^*}(\btheta)$ can be used as a drop-in replacement for $p(\btheta|\bx)$ in the downstream computations. By running variational inference on the feather falling simulator with the same observations as for point prediction and MCMC, we obtain the approximate posterior density shown in Figure \ref{fig:vi_feather}. We used a log normal distribution as variational family. It is parametrized by two parameters $\bphi = (\mu, \sigma)$ such that $\log g$ follows approximately a posteriori the distribution $\mathcal{N}(\mu, \sigma)$. By maximizing the ELBO, we obtain the parameters $\mu = -1.6873$ and $\sigma = 0.0829$. It should, however, be kept in mind that it only constitutes an approximation as, on the one hand, the posterior $p(\btheta|\bx)$ might not be contained in the variational family, and on the other hand, the optimization might not be perfect.

\section{Hypothesis testing}
Hypothesis testing is the task of assessing the compatibility between observed data and a hypothesis. The hypothesis to test is often called the null hypothesis and is denoted by $H_0$. The alternative hypothesis corresponds to cases where the null hypothesis does not hold and is denoted $H_1$. Of interest for this thesis is the fact that the hypothesis to test may be linked to a scientific theory. For example, one could want to test if the hypothesis $g=0.5$ could be rejected from data or not. In that case, the null hypothesis is $H_0: g = 0.5$, and the alternative hypothesis is $H_1: g \neq 0.5$. We may also want to test if the observations made are indeed coherent with our model or not.

In the frequentist setting, we compute the adequacy between the observations and the null hypothesis. There are two outcomes from this test. If the observations are likely under the null hypothesis, no conclusion can be drawn regarding this hypothesis. Otherwise, if the observations are unlikely under the null hypothesis, we can conclude that the null hypothesis is likely not to hold and hence reject it. To test if the observations are likely under the null hypothesis, we first compress the observations to a single real number through a summary statistic function that we will call $T(.)$ . That function should be designed to be, on average, higher if observations are drawn under the alternative hypothesis than under the null hypothesis. The null hypothesis is then rejected if that summary statistic is sufficiently high, i.e. above a given threshold $c$. The rejection region is then of the form
\begin{equation}
    R = \{\bx: T(\bx) > c\}.
\end{equation}
The threshold $c$ is chosen to match a given acceptable false rejection rate $\alpha$, often called the level of a test. A level $\alpha$ test is then a test such that the probability of erroneously rejecting the null hypothesis is lower than $\alpha$. The threshold $c$ must then satisfy
\begin{equation}
    P(\text{reject } H_0 | H_0) = P(T(\bX) > c | H_0) \leq \alpha.
\end{equation}
If the null hypothesis takes the form $H_0: \btheta \in \bTheta_0$, then this must hold for all possible values of $\btheta$ in $\bTheta_0$:
\begin{equation}
    P(T(\bX) > c | \btheta) \leq \alpha, \quad \forall \btheta \in \bTheta_0.
\end{equation}

For example, we could test if the observed measured times are indeed likely to be sampled from our model with $\btheta = \btheta_{\text{MLE}}$. If that is the case, then the residuals 
$$\text{measured time} -\sqrt{\frac{2h}{g_\text{MLE}}}$$
are sampled from the distribution $\mathcal{N}\left(0, \sigma_{\text{measurement}}\right)$. We can compute the following residuals for the observations made:
$$4.68 -\sqrt{\frac{6}{g_\text{MLE}}} = -1.016,\quad 6.97 -\sqrt{\frac{10}{g_\text{MLE}}} = -0.384,\quad 9.69 -\sqrt{\frac{14}{g_\text{MLE}}} = 0.988.$$
To test the hypothesis, we perform a Kolmogorov-Smirnov test that uses as test statistic, $T$, the maximum absolute difference between the empirical cumulative distribution function (c.d.f.) defined by the residuals and the c.d.f. of the distribution $\mathcal{N}\left(0, \sigma_{\text{measurement}}\right)$. We obtain a p-value of $0.47$, meaning that we could only reject the hypothesis that the samples come from that distribution at a level $\alpha$ equal or higher to $0.47$. Therefore, we cannot confidently reject that hypothesis from the observations at hand. Let us note that it does not mean that this hypothesis is correct. It could potentially be rejected by collecting more observations.

There is a direct link between hypothesis tests and confidence regions. Let us call by $R_\alpha(\btheta_0)$ the rejection region of a test $H_0: \btheta = \btheta_0$ and $H_1: \btheta \neq \btheta_0$ at level $\alpha$. Then the region
\begin{equation}
    \label{eq:neyman}
    C_{1-\alpha}(\bx) = \{\btheta: \bx \notin R_\alpha(\btheta) \}
\end{equation}
is a valid level $(1-\alpha)$ confidence region. Similarly, the rejection region
\begin{equation}
    R_\alpha(\btheta_0) = \{ \bx : \btheta_0 \notin C_{1-\alpha}(\bx) \}
\end{equation}
is a valid rejection region for the test. To prove it, let us first observe that
\begin{equation}
    P(\btheta_0 \in C_{1-\alpha}(\bX)|\btheta=\btheta_0) = P(\bX \notin R_\alpha(\btheta_0)|\btheta=\btheta_0) = 1 - P(\bX \in R_\alpha(\btheta_0)|\btheta=\btheta_0).
\end{equation}
As $P(\bX \in R_\alpha(\btheta_0)|\btheta=\btheta_0) \leq \alpha$ by definition of an hypothesis test, we have
\begin{equation}
    P(\btheta_0 \in C_{1-\alpha}(\bX)|\btheta=\btheta_0) \geq 1 - \alpha.
\end{equation}
Similarly, if $P(\btheta_0 \in C_{1-\alpha}(\bX)|\btheta=\btheta_0) \geq 1 - \alpha.$, then
\begin{equation}
    P(\bX \in R_\alpha(\btheta_0)|\btheta=\btheta_0) \leq \alpha.
\end{equation}

These properties then link the problem of finding a confidence region to the problem of testing a hypothesis. In addition to linking the two tasks, it also provides a way to approximate confidence regions by constructing a series of hypothesis test rejection regions. This method is called the Neyman construction \citep{Neyman1937Outline}. To determine if some parameter values $\btheta_0$ should be included in the confidence region at level $\alpha$, a critical value $c$ is computed such that $P(T(\bX) > c|\btheta = \btheta_0) \leq \alpha$. If the observation $\bx$ is such that $T(\bx) \leq c$, then $\btheta_0$ is included in the confidence region. Otherwise, it is not. By repeating this procedure for a fine grid of parameter values spanning the support of possible parameter values, an approximate confidence region can be constructed.

In the Bayesian setting, hypothesis testing takes a different form. As for the parameters of a statistical model, we can assign a probability to the different hypotheses corresponding to our belief that these hypotheses are true. The belief we have about a hypothesis is summarized by its posterior probability 
\begin{equation}
    p(H_0|\bx) = \frac{p(\bx|H_0)p(H_0)}{p(\bx)}.
\end{equation}
Hypothesis testing then boils down to choosing a hypothesis according to our belief. A common decision rule is then to favor $H_0$ over $H_1$ if
\begin{equation}
    P(H_0|\bx) \geq P(H_1|\bx).
\end{equation}
Depending on the application, rejecting $H_0$ while it is valid or accepting it while invalid might have different consequences, and we may want to avoid a given type of error as much as possible. An example that will drive this thesis is that refuting a scientific theory that is valid is much more harmful than failing to reject an invalid theory. In those cases, we can assign a cost to each error type and find the decision rule that minimizes the expected cost. We denote by $C_{10}$ the cost of rejecting $H_0$ (choosing $H_1$) while it is valid and by $C_{01}$ the cost of selecting $H_0$ while it is invalid. The expected cost is then expressed
\begin{equation}
    C = C_{10}P(\text{reject}(H_0 | H_0)) + C_{01}P(\text{not reject} H_0 | H_1).
\end{equation}
It is minimized when favoring $H_0$ over $H_1$ if
\begin{equation}
    P(H_0|\bx)C_{10} \geq P(H_1|\bx)C_{01}.
\end{equation}

  \chapter{Simulation-based inference}\label{c:sbi}
  Simulators are being used more and more in science to describe complex systems that evolve in an iterative manner. For example, simulators are used in astrophysics to describe the evolution of stellar streams with various dark matter masses \citep{Charles2017}, to model gravitational waves \citep{hannam2014simple}, or to model gravitational lensing \citep{wagner2024strong}. Simulators are also used in biology to model brain activity \citep{destexhe2000nonlinear}, in epidemiology \citep{su151411120}, and for particle physics \citep{atlas2012observation}. It allows for the simulation of phenomena under varying conditions and the comparison of those simulations to what is observed in reality. If we take back the example of a feather falling from the previous chapter, we can quickly identify that the proposed model is oversimplistic as it only takes gravity into account and not other forces that may be applied to the feather. We will improve that model by taking into account the drag due to the friction with air. Drag acts in the opposite direction to the fall, and its magnitude is given by the following equation
\begin{equation}
    F_D = \frac{1}{2}\ \rho \ v^2 \ C_D \ A,
\end{equation}
where $\rho$ is the air density, $v$ is the speed of the feather, $C_D$ is a drag coefficient, and $A$ is the cross-sectional area, i.e., the area of the projection of the object on the plane orthogonal to the fall direction. There are three quantities in this equation that are not constant over time. The speed evolves over time due to gravity, and the cross-sectional area evolves as the object's orientation changes during the fall. If we consider that there are air movements, air density under the feather is not constant either. It is then hard to find an analytical formula for the fall time. However, we can simulate the fall by discretizing time and by modeling the fall as a succession of time steps for which different forces apply. This is illustrated in Algorithm \ref{algo:feather_falling}. At every time step, forces are calculated based on current conditions. Speed and position are then updated based on the current forces applied, and a new object orientation and air density are computed. This is repeated until the feather has fallen from a given height, and the measurement of falling time is simulated.

\begin{algorithm}
   \caption{Feather falling simulator.}
   \label{algo:feather_falling}
   \begin{tabular}{ l l }
        {\itshape Inputs:} & gravity constant $g$, drag coefficient $C_D$, height $h$, mass $m$ and \\ & measurement error standard deviation $\sigma_{\text{measurement}}$. \\
        {\itshape Outputs:} & The measured fall time. \\
        {\itshape hyper-parameters:} & Duration of a step $\Delta t$ \\
  \end{tabular}
  \begin{algorithmic}
    \STATE $\text{distance} = 0$
    \STATE $\text{speed} = 0$
    \STATE $\text{elapsed\_time} = 0$
    \STATE $\rho = 1.225$
    \STATE $\text{orientation} = \text{initial\_orientation}$
    \REPEAT
        \STATE $A = \text{cross\_sectional\_area}(\text{orientation})$
        \STATE $\text{drag\_acceleration} = \frac{\text{speed}^2\ \rho\ C_D\ A}{2m}$
        \STATE $\text{next\_speed} = \text{speed} + \Delta t(g -\text{drag\_acceleration})$
        \STATE $\text{distance} = \text{distance} + \Delta t \frac{\text{speed} + \text{next\_speed}}{2}$
        \STATE $\text{orientation} = \text{rotation}(\text{orientation}, \Delta t)$
        \STATE $\rho = \text{air\_movement}(\rho, \Delta_t)$
        \STATE $\text{speed} = \text{next\_speed}$
        \STATE $\text{elapsed\_time} = \text{elapsed\_time} + \Delta t$
    \UNTIL{$\text{distance} \geq h$}
    \STATE $\text{measured\_time} \sim \mathcal{N}(\text{elapsed\_time}, \sigma_{\text{measurement}})$
    \STATE \textbf{return} measured\_time
   \end{algorithmic}
\end{algorithm} 

\section{Simulators as statistical models}
\begin{figure}[h!]
    \centering
    
    \begin{tikzpicture}[scale=0.9]
        \node [anchor=center, align=center] at (1.25,2.5) {$t=0\text{s}$};
        \node [anchor=center, align=center] at (3.25,2.5) {$t=1\text{s}$};
        \node [anchor=center, align=center] at (5.25,2.5) {$t=3\text{s}$};
        \node [anchor=center, align=center] at (7.25,2.5) {$t=5\text{s}$};
        \node [anchor=center, align=center] at (9.25,2.5) {$t=10\text{s}$};
        \node [anchor=center, align=center] at (11.25,2.5) {$t=20\text{s}$};
        \node [anchor=center, align=center] at (13.25,2.5) {$t=50\text{s}$};
        
        \node [anchor=east, align=center] at (0.25,0) {$g=1$ \\ $C_D = 1$};
        \node [anchor=west] at (0.5, 0) {\includegraphics[width=1.5cm]{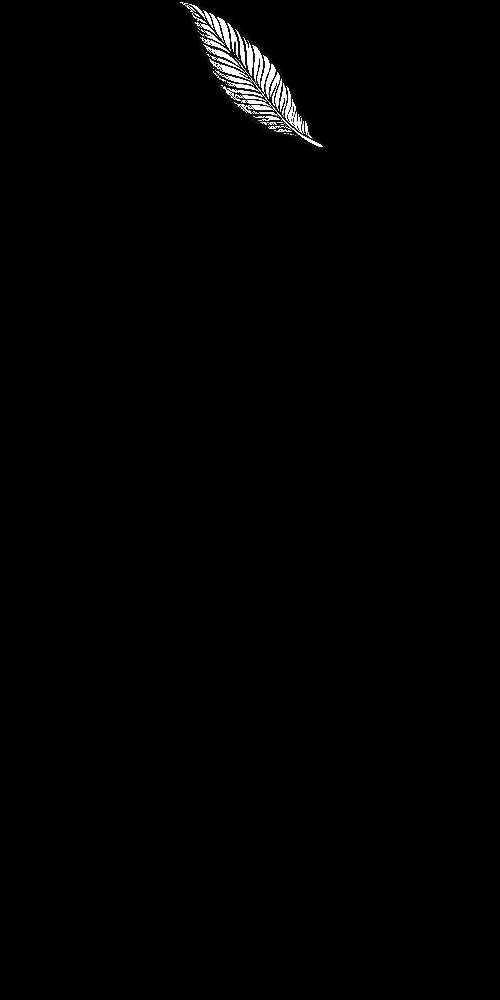}};
        \node [anchor=west] at (2.5, 0) {\includegraphics[width=1.5cm]{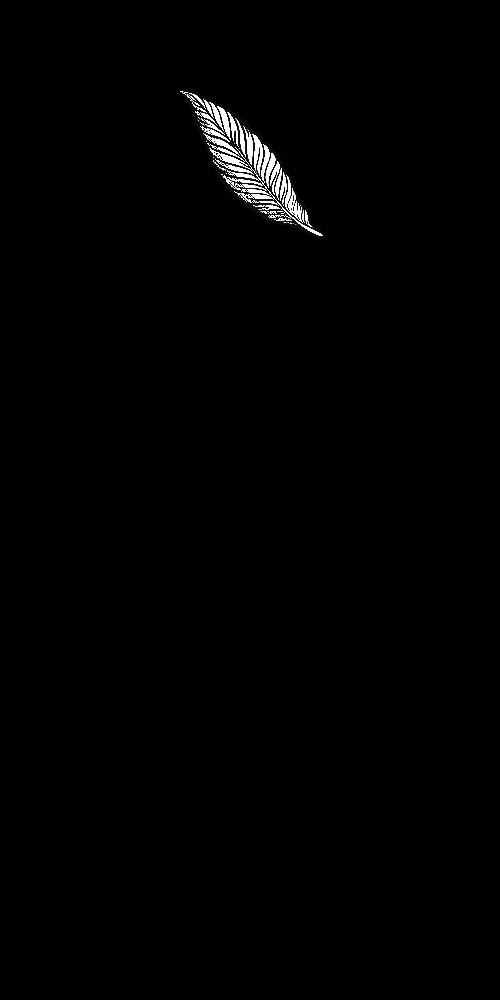}};
        \node [anchor=west] at (4.5, 0) {\includegraphics[width=1.5cm]{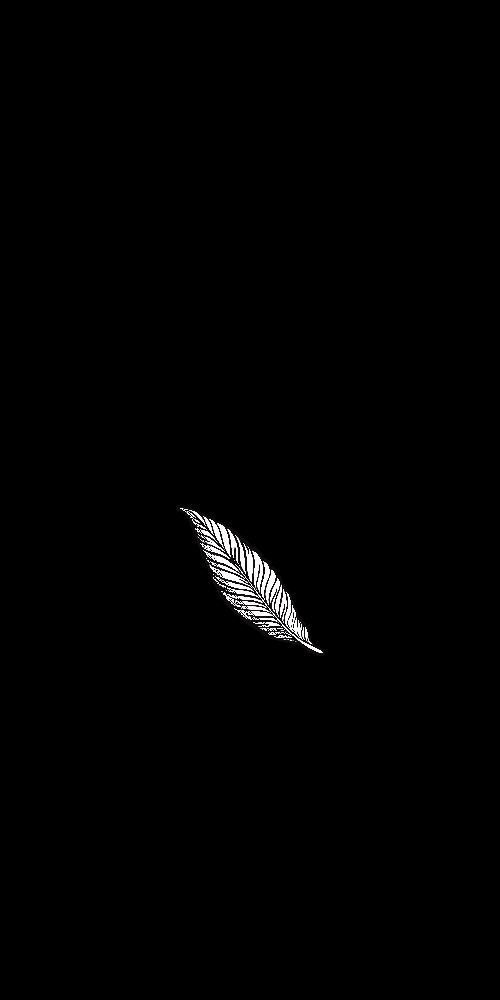}};
        \node [anchor=west] at (6.5, 0) {\includegraphics[width=1.5cm]{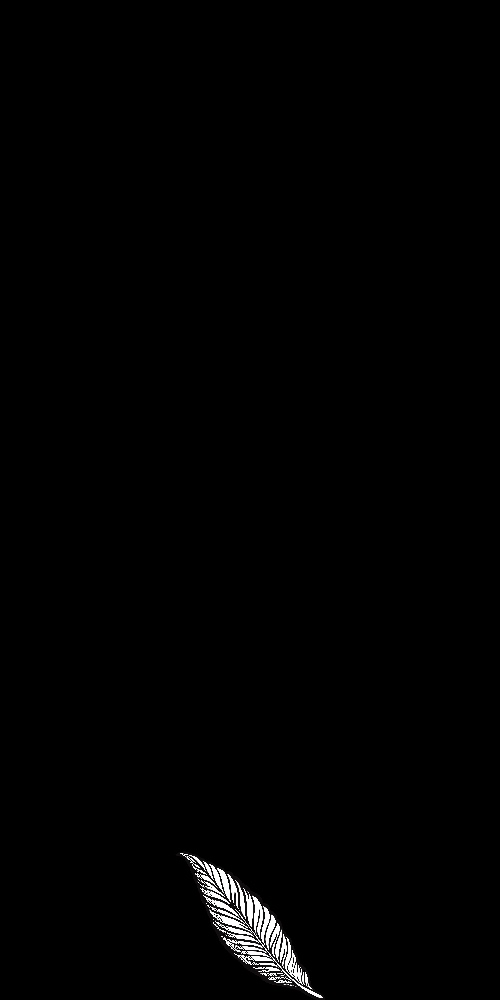}};
        \node [anchor=west] at (8.5, 0) {\includegraphics[width=1.5cm]{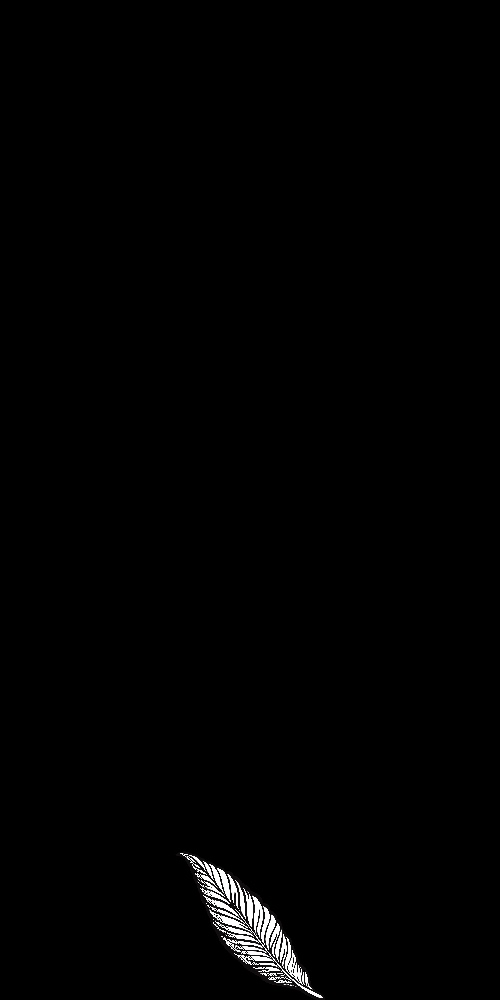}};
        \node [anchor=west] at (10.5, 0) {\includegraphics[width=1.5cm]{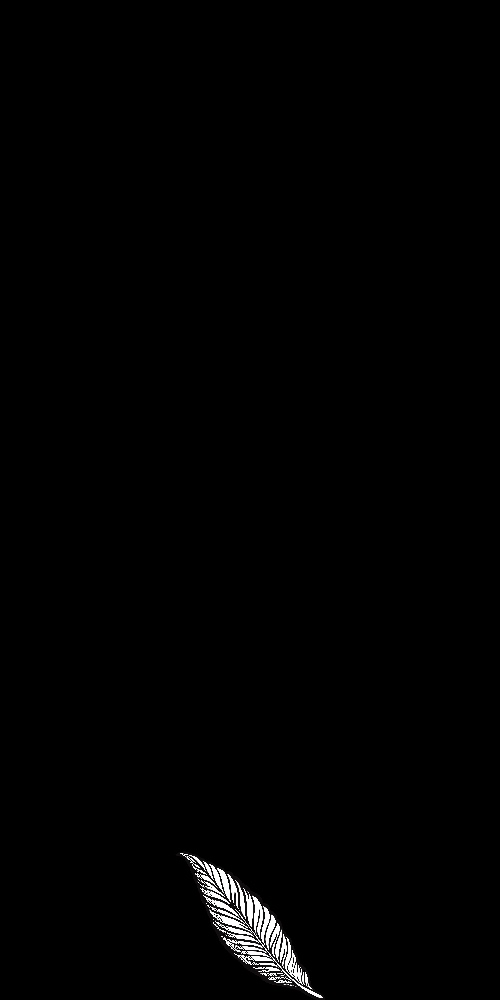}};
        \node [anchor=west] at (12.5, 0) {\includegraphics[width=1.5cm]{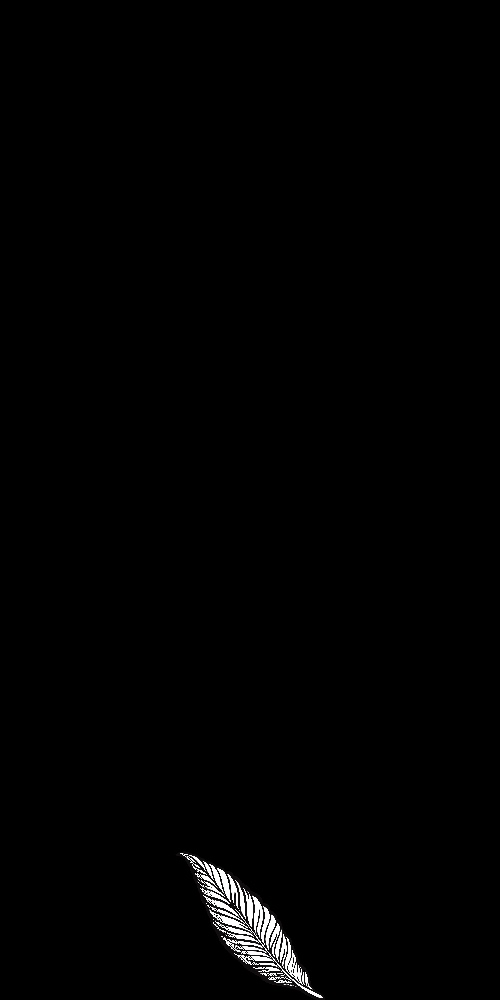}};

        \node [anchor=east, align=center] at (0.25,-4) {$g=9.81$ \\ $C_D = 1$};
        \node [anchor=west] at (0.5, -4) {\includegraphics[width=1.5cm]{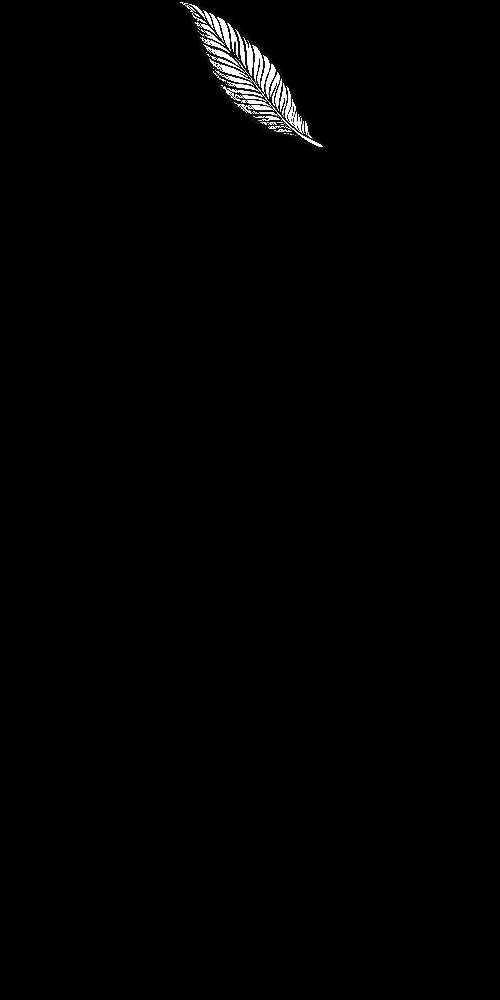}};
        \node [anchor=west] at (2.5, -4) {\includegraphics[width=1.5cm]{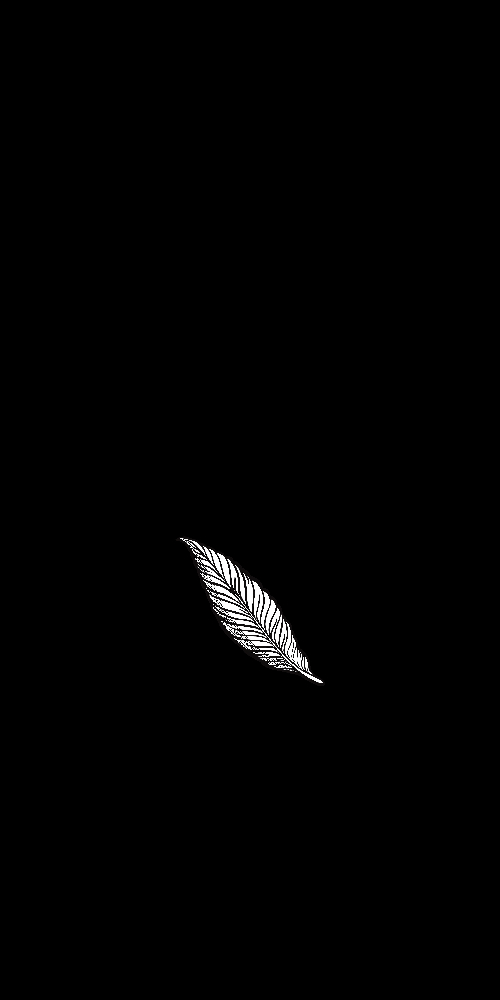}};
        \node [anchor=west] at (4.5, -4) {\includegraphics[width=1.5cm]{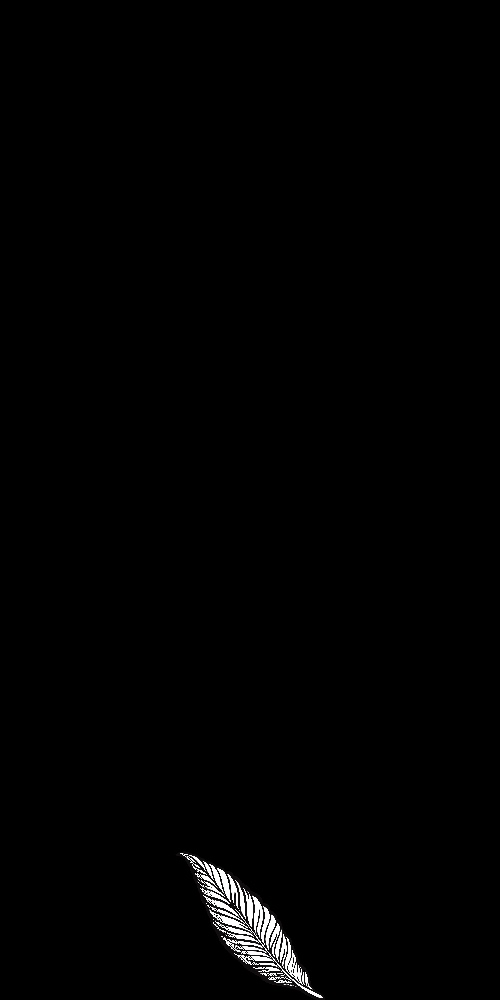}};
        \node [anchor=west] at (6.5, -4) {\includegraphics[width=1.5cm]{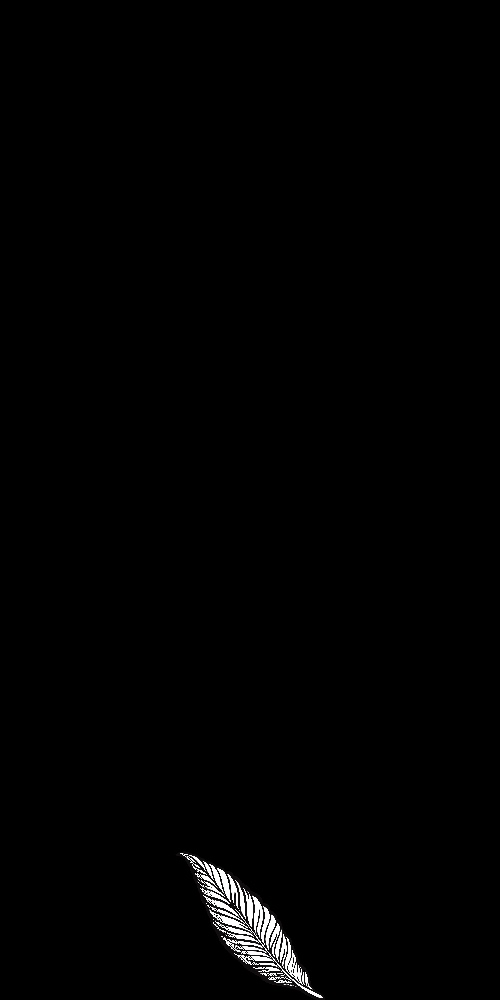}};
        \node [anchor=west] at (8.5, -4) {\includegraphics[width=1.5cm]{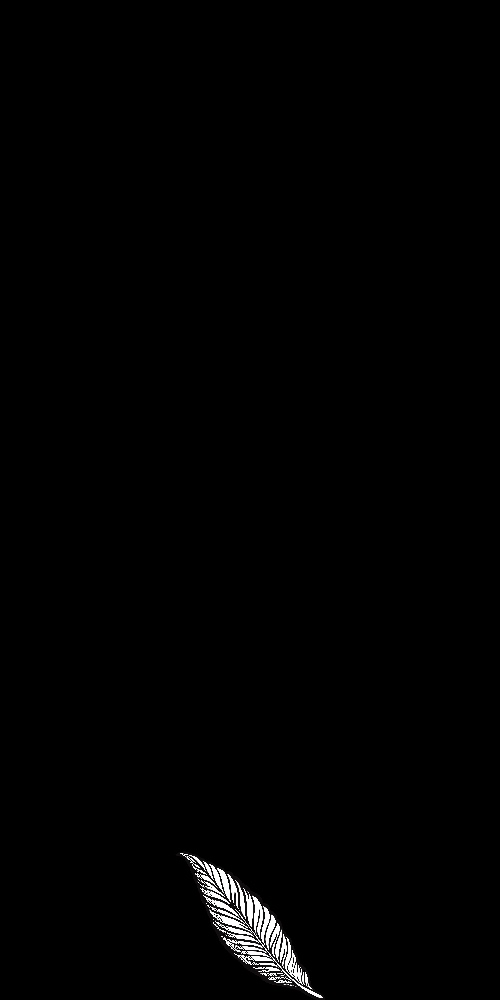}};
        \node [anchor=west] at (10.5, -4) {\includegraphics[width=1.5cm]{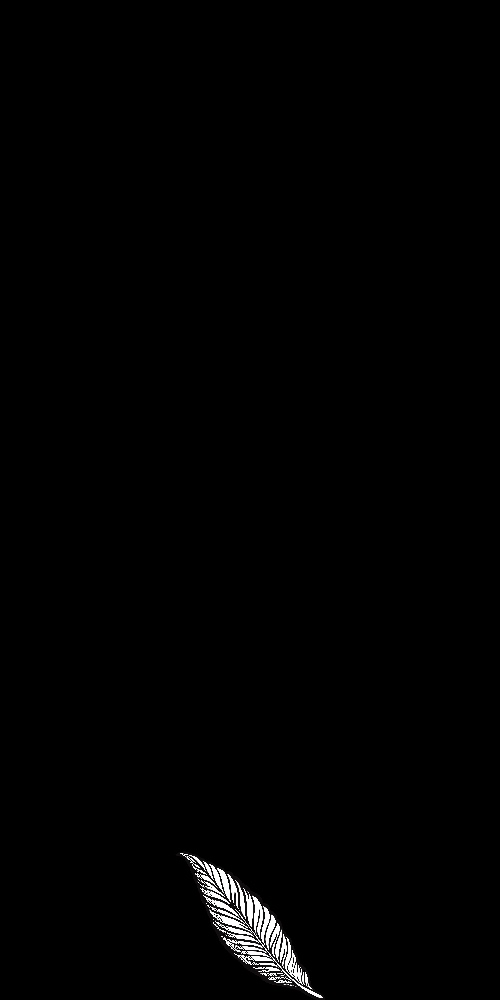}};
        \node [anchor=west] at (12.5, -4) {\includegraphics[width=1.5cm]{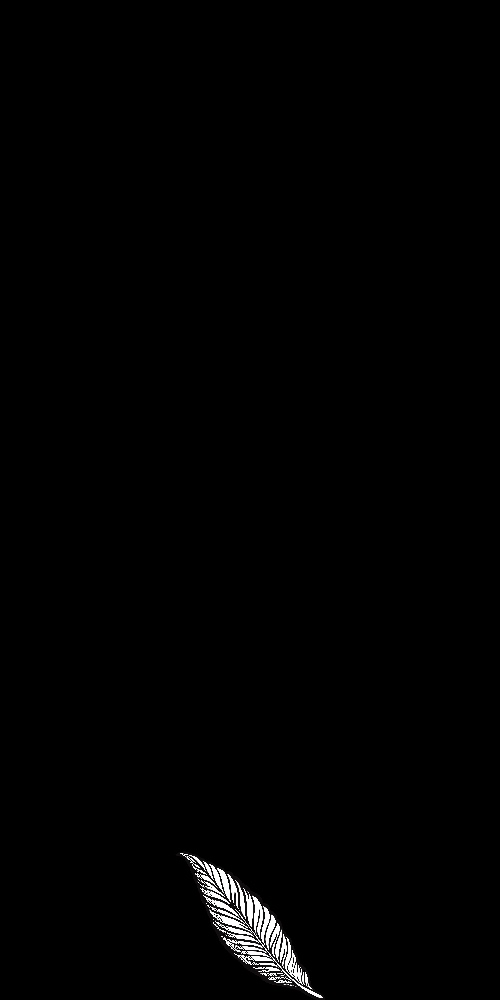}};

        \node [anchor=east, align=center] at (0.25,-8) {$g=9.81$ \\ $C_D = 10$};
        \node [anchor=west] at (0.5, -8) {\includegraphics[width=1.5cm]{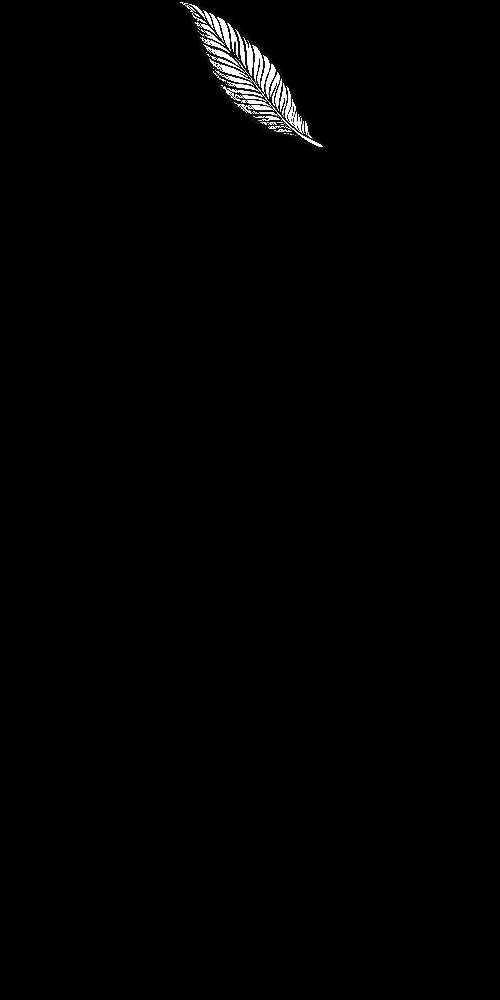}};
        \node [anchor=west] at (2.5, -8) {\includegraphics[width=1.5cm]{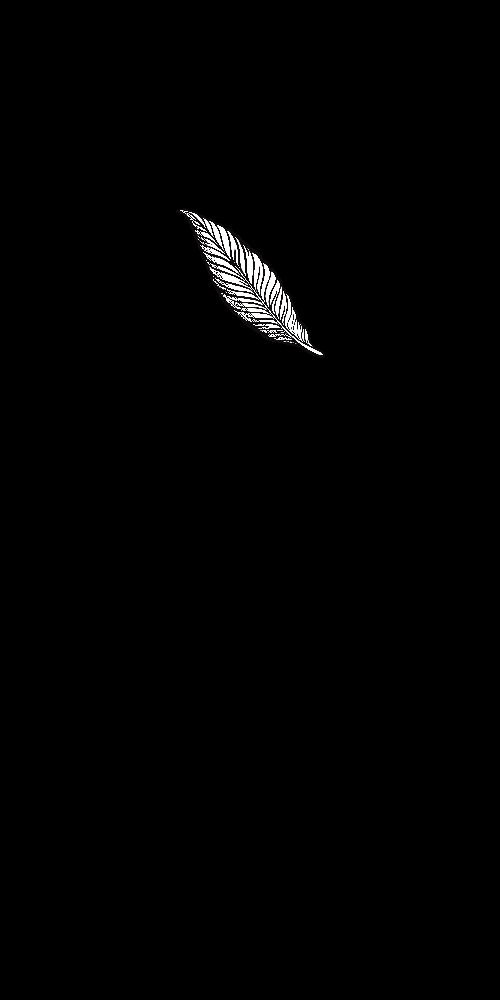}};
        \node [anchor=west] at (4.5, -8) {\includegraphics[width=1.5cm]{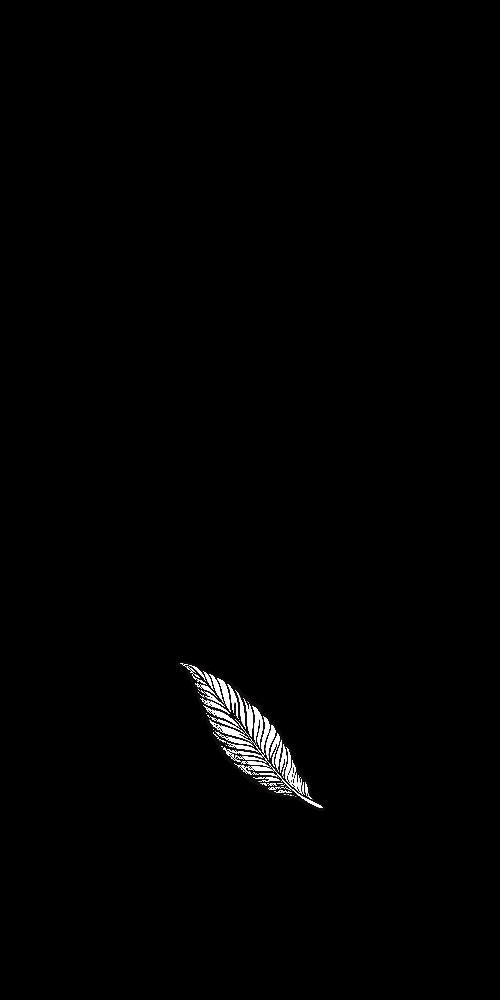}};
        \node [anchor=west] at (6.5, -8) {\includegraphics[width=1.5cm]{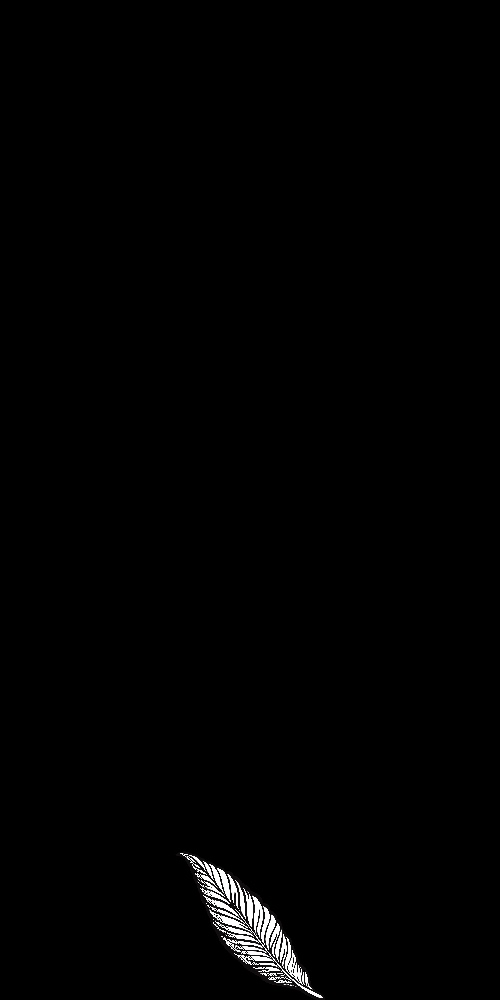}};
        \node [anchor=west] at (8.5, -8) {\includegraphics[width=1.5cm]{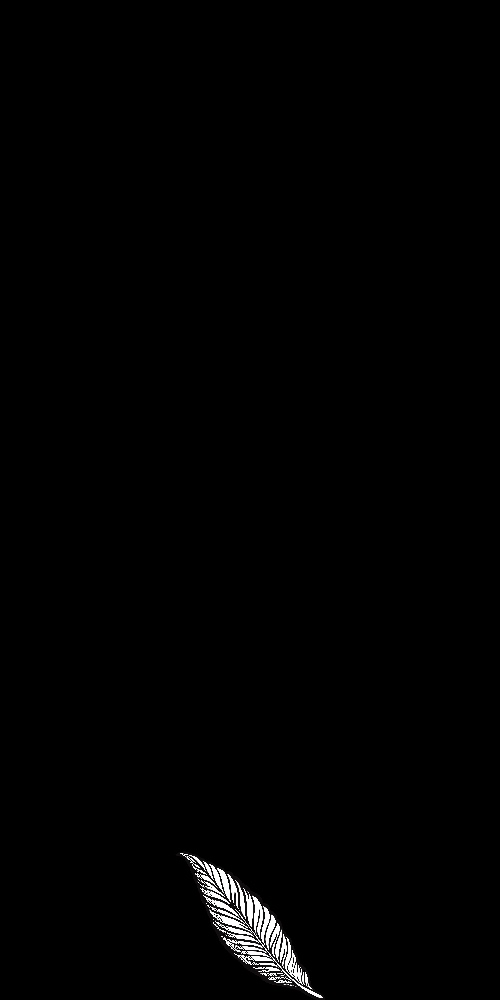}};
        \node [anchor=west] at (10.5, -8) {\includegraphics[width=1.5cm]{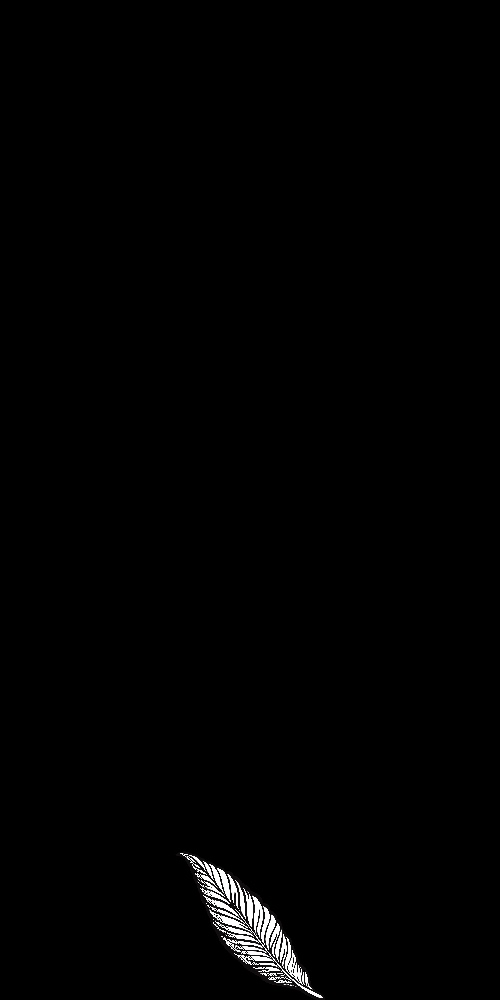}};
        \node [anchor=west] at (12.5, -8) {\includegraphics[width=1.5cm]{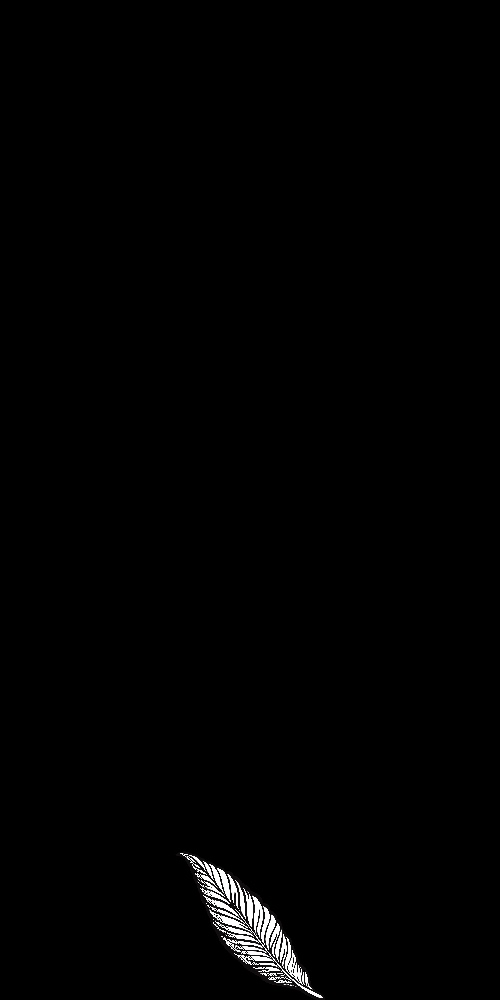}};
        
        \node [anchor=east, align=center] at (0.25,-12) {$g=9.81$ \\ $C_D = 1000$};
        \node [anchor=west] at (0.5, -12) {\includegraphics[width=1.5cm]{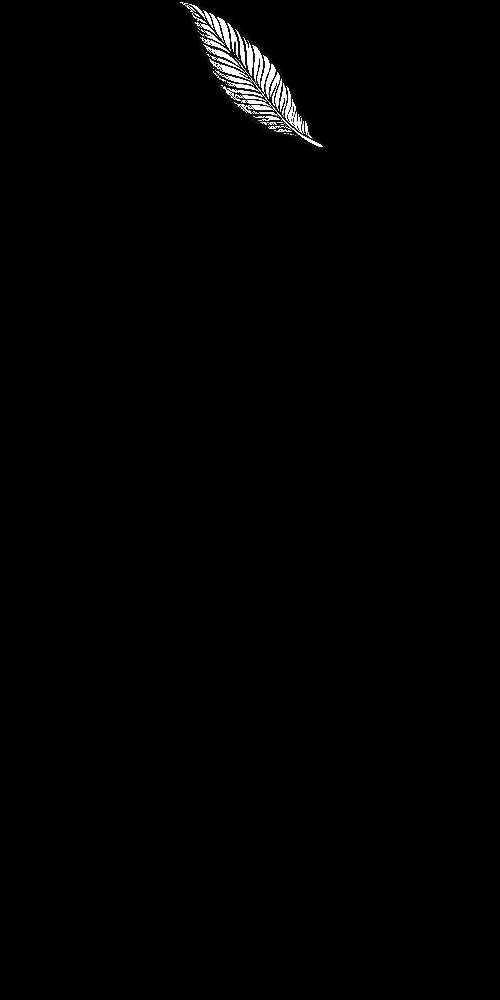}};
        \node [anchor=west] at (2.5, -12) {\includegraphics[width=1.5cm]{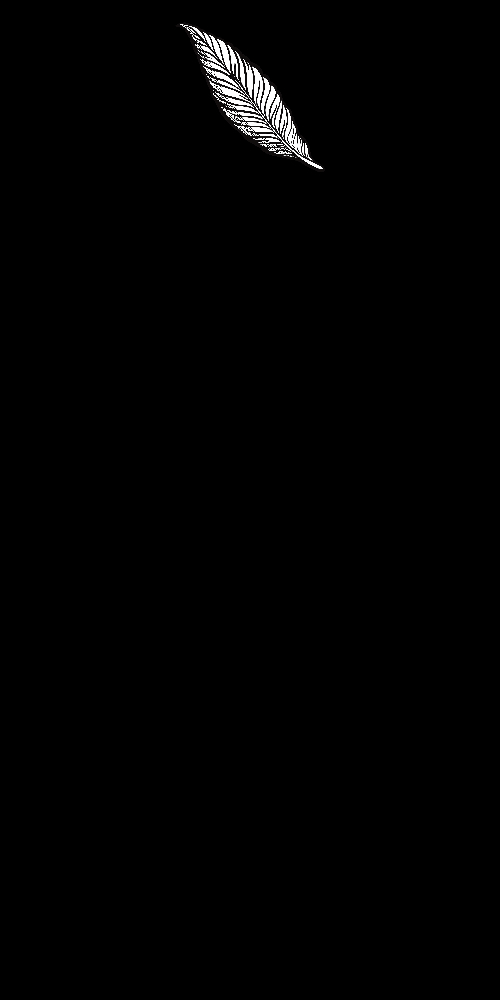}};
        \node [anchor=west] at (4.5, -12) {\includegraphics[width=1.5cm]{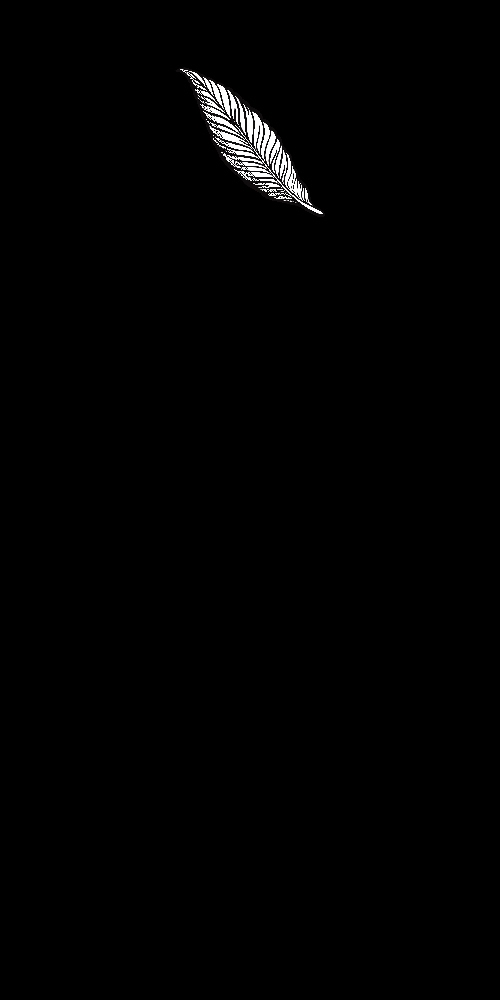}};
        \node [anchor=west] at (6.5, -12) {\includegraphics[width=1.5cm]{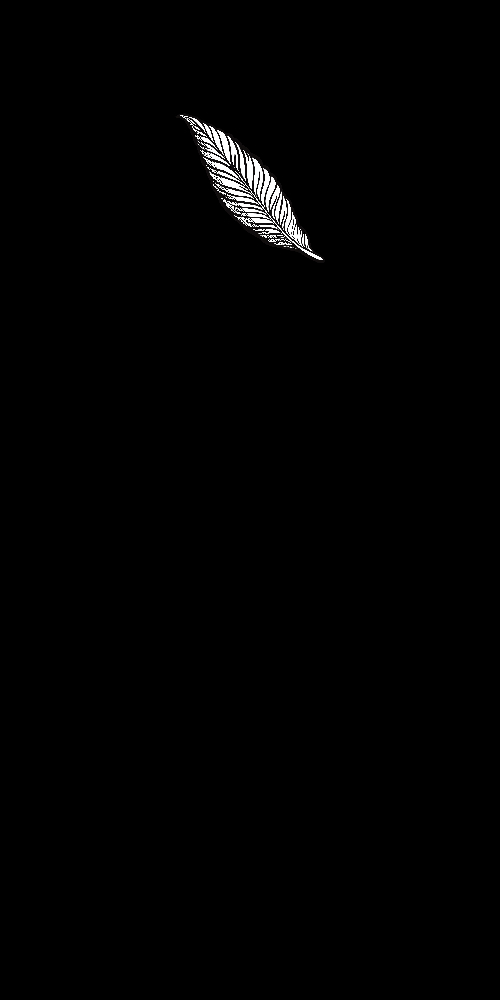}};
        \node [anchor=west] at (8.5, -12) {\includegraphics[width=1.5cm]{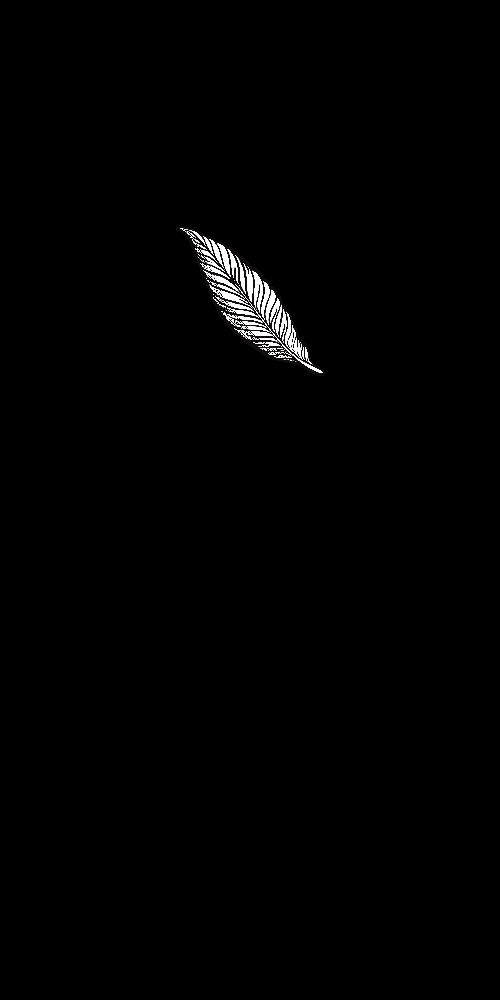}};
        \node [anchor=west] at (10.5, -12) {\includegraphics[width=1.5cm]{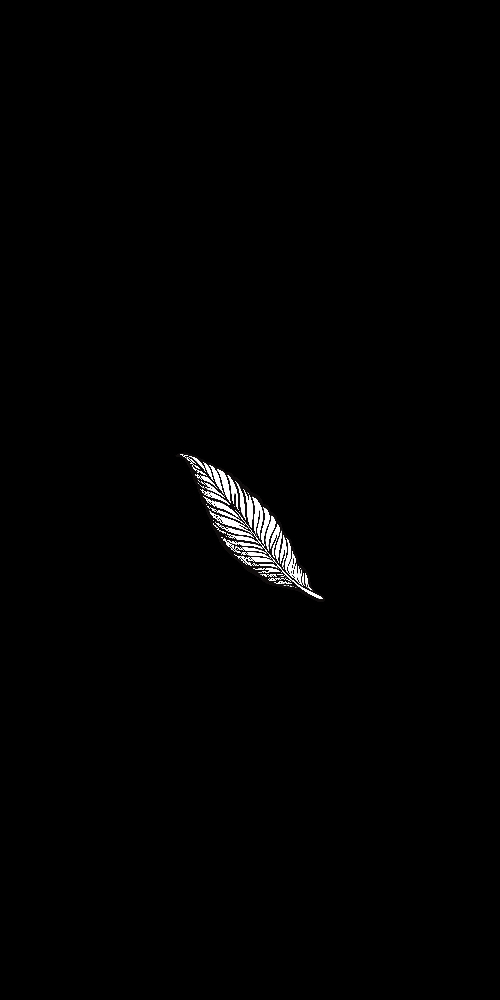}};
        \node [anchor=west] at (12.5, -12) {\includegraphics[width=1.5cm]{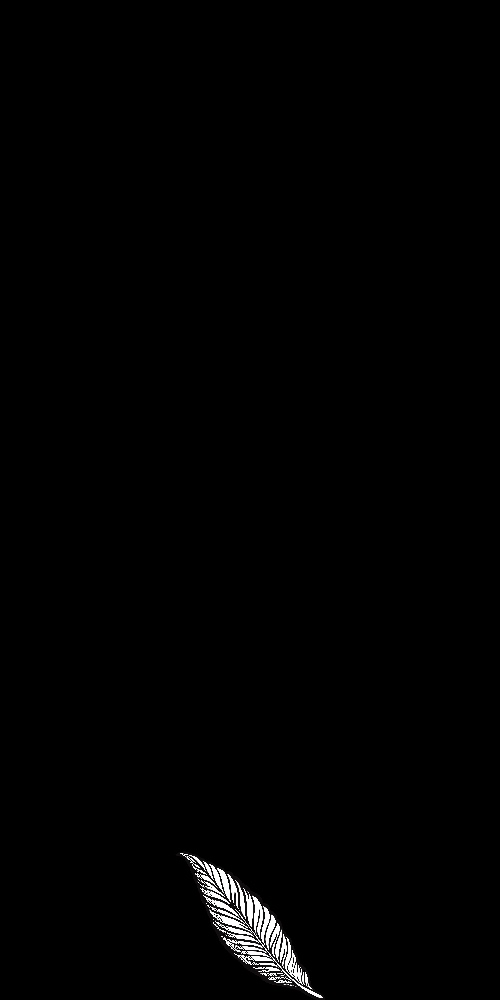}};

        \node [anchor=east, align=center] at (0.25,-16) {$g=100$ \\ $C_D = 1000$};
        \node [anchor=west] at (0.5, -16) {\includegraphics[width=1.5cm]{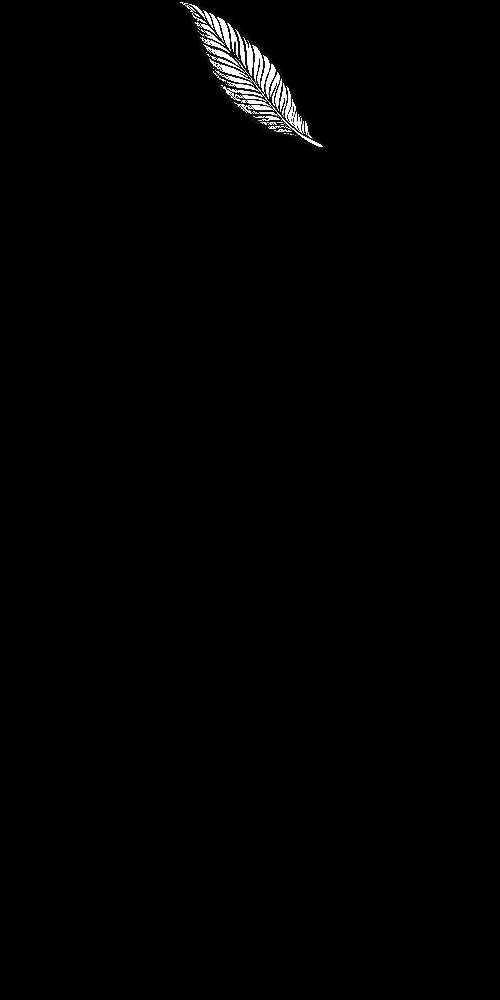}};
        \node [anchor=west] at (2.5, -16) {\includegraphics[width=1.5cm]{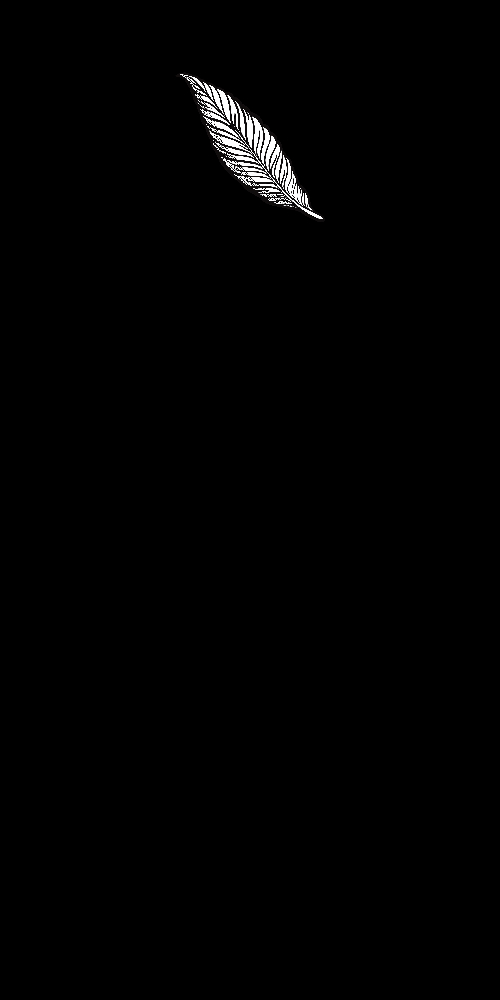}};
        \node [anchor=west] at (4.5, -16) {\includegraphics[width=1.5cm]{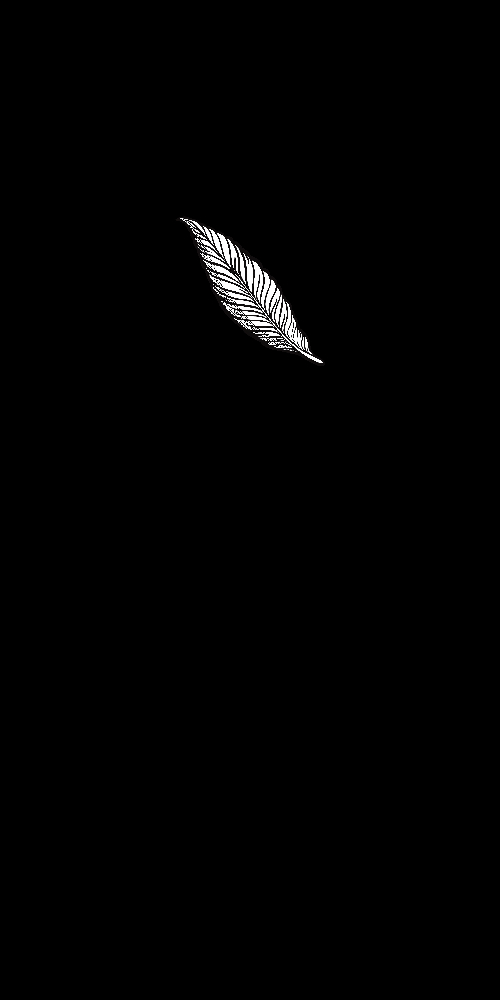}};
        \node [anchor=west] at (6.5, -16) {\includegraphics[width=1.5cm]{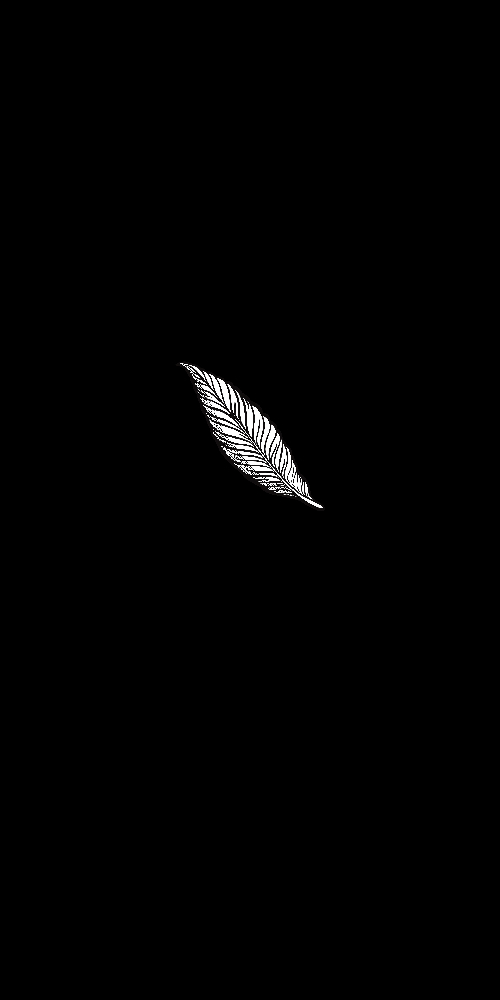}};
        \node [anchor=west] at (8.5, -16) {\includegraphics[width=1.5cm]{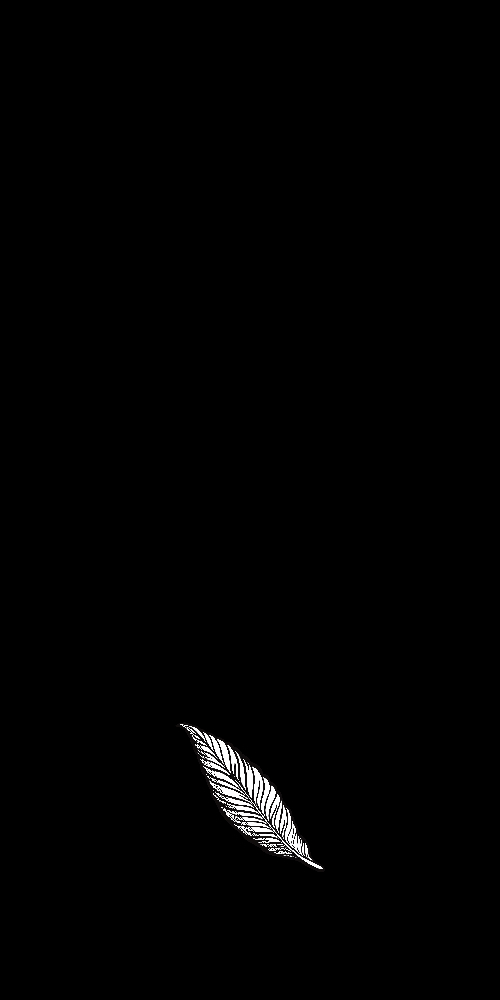}};
        \node [anchor=west] at (10.5, -16) {\includegraphics[width=1.5cm]{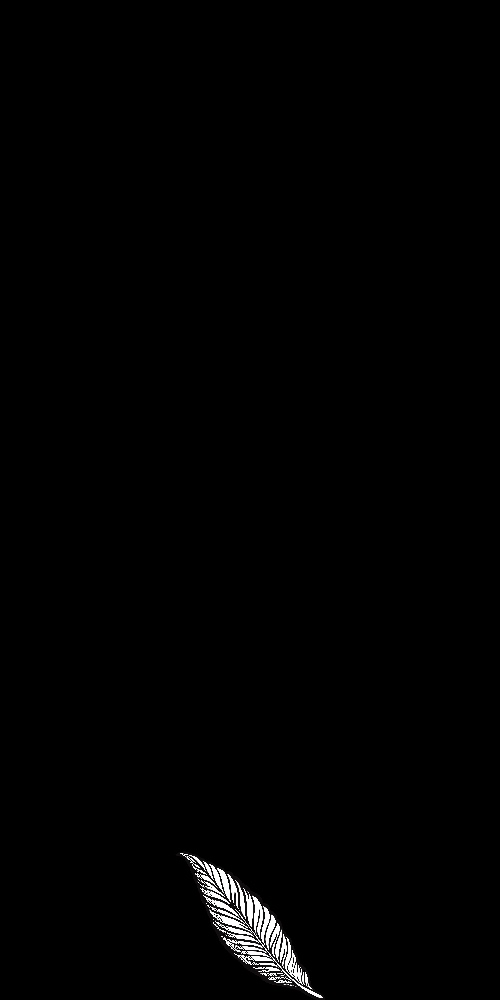}};
        \node [anchor=west] at (12.5, -16) {\includegraphics[width=1.5cm]{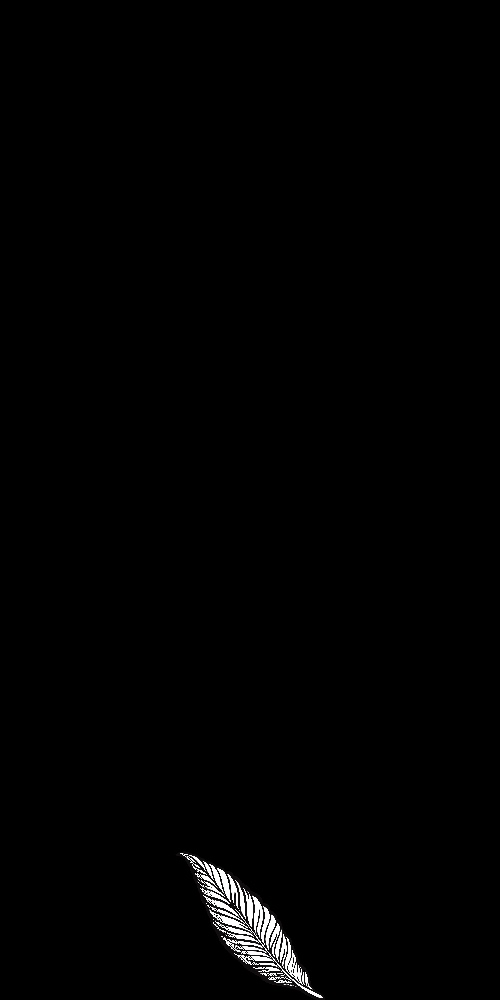}};
    \end{tikzpicture}

    \caption{Example of feather falling simulations using various gravity and friction parameters. The feather is dropped from the same height in all those configurations, and the feather position is reported at various time steps.}
    \label{fig:feather_simulations}
    
\end{figure}
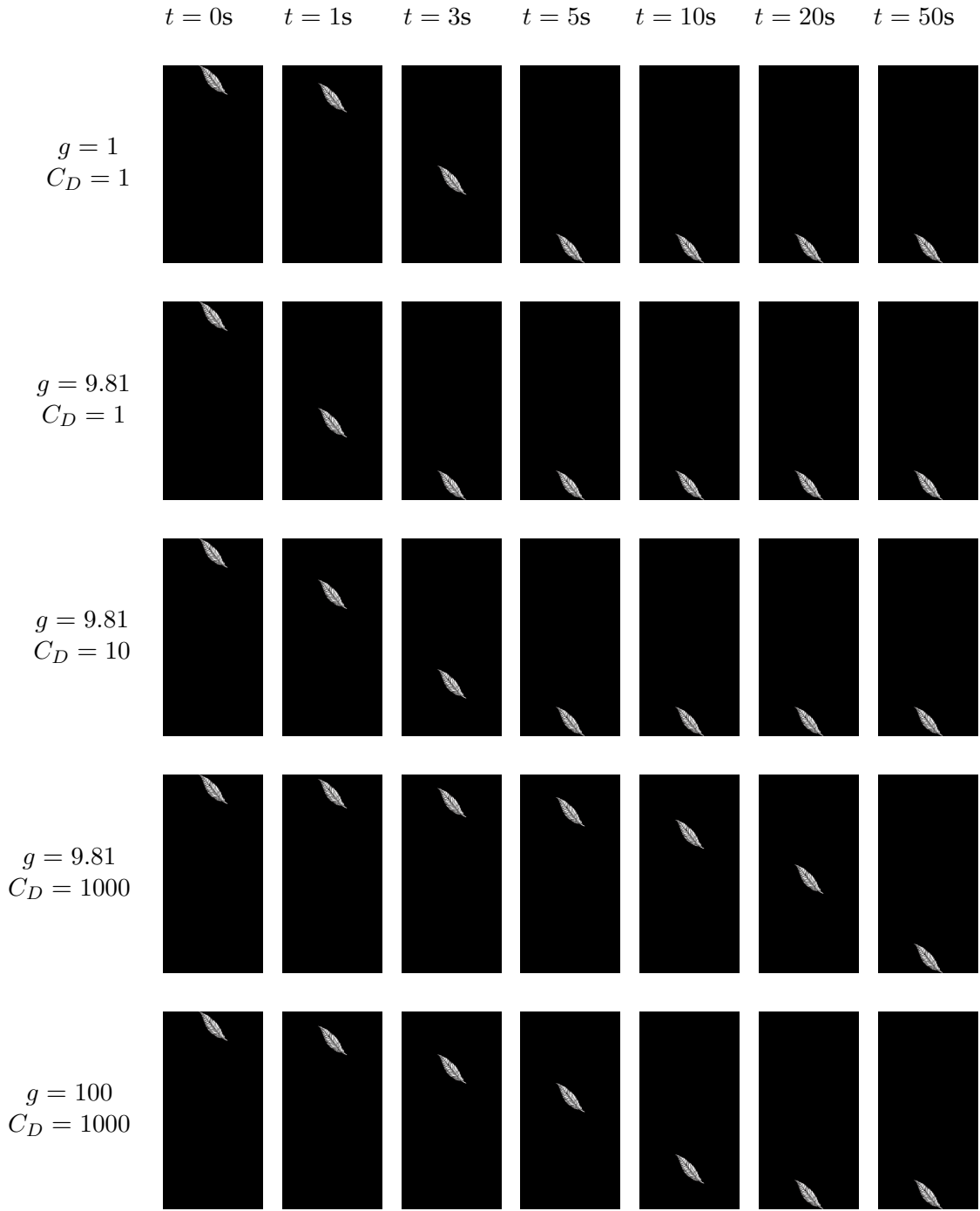

In the example of a feather falling, stochasticity arises in the measurement as a random measurement error is made. It may also be hard to model the change of orientation of the feather and air movement. We might want to model those as random processes where random rotations and air movements are performed at each step. Both these elements lead to a stochastic simulator that would generate different measured fall times if run several times with the same parameters. Examples of simulator runs are provided in Figure \ref{fig:feather_simulations} for various values of gravity constant $g$ and drag coefficient $C_D$.

A stochastic simulator is a form of statistical model that includes latent random variables corresponding to the stochastic steps during observation generation. We will denote those latent variables by $\bz$. In the example of the feather falling simulator, those latent variables correspond to the random rotations and air movements performed at each time step. The simulator is also conditioned on some parameters $\btheta$ that are $g$ and $C_D$ in the example. Formally, the simulator defines the distributions $p(\bx|\bz, \btheta)$ and $p(\bz|\btheta)$. Consequently, it also defines the distribution $p(\bx, \bz|\btheta) = p(\bx|\bz, \btheta)p(\bz|\btheta)$. The distribution $p(\bx | \btheta)$ cannot be computed analytically because it would require marginalizing and, hence, integrating over all possible latent variables, which is, in most cases, impossible. In the Bayesian setting, if we were to perform inference with variational inference or Markov chain Monte Carlo, we could use the known distributions $p(\bx | \bz, \btheta)$ and $p(\bz, \btheta) = p(\bz| \btheta)p(\btheta)$ to produce either a variational approximation or sample from $p(\bz, \btheta | \bx)$. If the variational distribution is such that analytical marginalization is possible, the distribution $p(\btheta|\bx)$ can then be obtained through marginalization 
\begin{equation}
    p(\btheta|\bx) = \int p(\bz, \btheta | \bx)\ d\bz.
\end{equation}
To obtain samples from the marginal, samples from $p(\bz, \btheta | \bx)$ can be used where the $\bz$ component is removed. However, to have a good enough simulator, it is usually necessary to have many internal latent variables. In the example of the feather falling, if we want an accurate simulator, we need to choose $\Delta t$ sufficiently small and hence perform many iterations and random rotations. This means that $\bz$ is typically high-dimensional, and this is an issue to ensure good optimization with VI and good posterior landscape coverage with MCMC.

\section{Simulation-based inference}

\begin{figure}
    \centering
    
    \begin{tikzpicture}[scale=0.8]
        \tikzstyle{box} = [draw=black, fill=white, thick, rectangle, anchor=west, rounded corners, minimum height=0.7cm]
        \tikzstyle{greyarrow} = [thick, ->, gray!70]
        \tikzstyle{blackarrow} = [thick, ->, black]
        \tikzstyle{bigarrow} = [thick, ->, black, double]
        \tikzstyle{line} = [thick, -, black]
        
        \node [anchor=west, align=center] at (-5, 0) (label_1) {Parameters \\ $\btheta \sim p(\btheta)$};
        \node [anchor=west, align=center] at (4.2, 0) (label_2) {Observations \\ $\bx$};
        \node [box, anchor=west, align=center] at (-0.5, 0) (label_3) {Simulator \\ $\bx \sim p(\bx|\btheta)$};
        \node [anchor=west, align=center] at (-5, -3) (label_4) {Estimation \\ $\hat{p}(\btheta\vert\bx_o)$};
        \node [anchor=west, align=center] at (4.0, -3) (label_5) {Real observation \\ $\bx_o$};
        \node [box, anchor=west] at (-0.3, -3) (label_6) {Inference};
        
        \draw [greyarrow] (label_1.east) -- (label_3.west);
        \draw [greyarrow] (label_3.east) -- (label_2.west);
        \draw [greyarrow] (label_5.west) -- (label_6.east);
        \draw [greyarrow] (label_6.west) -- (label_4.east);
        
        \draw [bigarrow] (0.95, -1.2) -- (0.95, -1.8);
        
        \draw [line] (-5, 1) -- (8, 1);
        \draw [line] (-5, -1) -- (8, -1);
        \draw [line] (8, 1) -- (8, -1);
        \draw [line] (-5, 1) -- (-5, -1);
        
        \draw [line] (-5, -2) -- (8, -2);
        \draw [line] (-5, -4) -- (8, -4);
        \draw [line] (8, -2) -- (8, -4);
        \draw [line] (-5, -2) -- (-5, -4);
    \end{tikzpicture}

    \caption{Bayesian simulation-based inference overview}
    \label{fig:sbi_overview}
    
\end{figure}
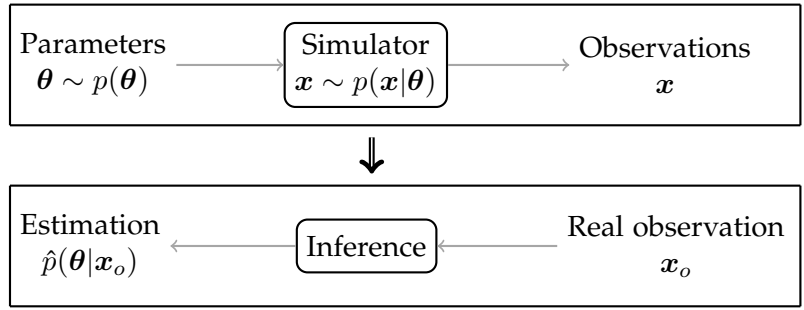

To obtain more accurate inferences, we can use the fact that sampling from the likelihood $p(\bx | \btheta)$ is possible by sampling from $p(\bz | \btheta)$ and then from $p(\bx|\btheta, \bz)$. In most settings, we are only interested in $p(\btheta|\bx)$, not in $p(\btheta, \bz |\bx)$. Then, if we have inference methods that only rely on samples and not on likelihood evaluation, we can completely ignore the latent variables $\bz$. This leads to the development of simulation-based inference (SBI) methods, also called likelihood-free inference (LFI), that only rely on simulations from the likelihood $p(\bx|\btheta)$ and not on likelihood evaluations. An overview of simulation-based inference in the Bayesian setting is provided in Figure \ref{fig:sbi_overview}.

\begin{figure}
    \centering
    
    \begin{tikzpicture}[scale=1.]
        \tikzstyle{box} = [draw=black, fill=white, thick, rectangle, anchor=west, rounded corners, minimum height=0.7cm]
        \tikzstyle{greyarrow} = [thick, ->, gray!70]
        \tikzstyle{blackarrow} = [thick, ->, black]
        \tikzstyle{bigarrow} = [thick, ->, black, double]
        \tikzstyle{line} = [thick, -, black]
        \tikzstyle{dashedline} = [dash pattern=on 3pt off 3pt, very thick, -, black]

        \node[] at (0,5) (label_1) {$\btheta = \btheta_1$};
        \node[] at (0,0) (label_2) {$\btheta = \btheta_2$};
        \node[] at (5.1, 5) (label_3) {\includegraphics[width=6cm]{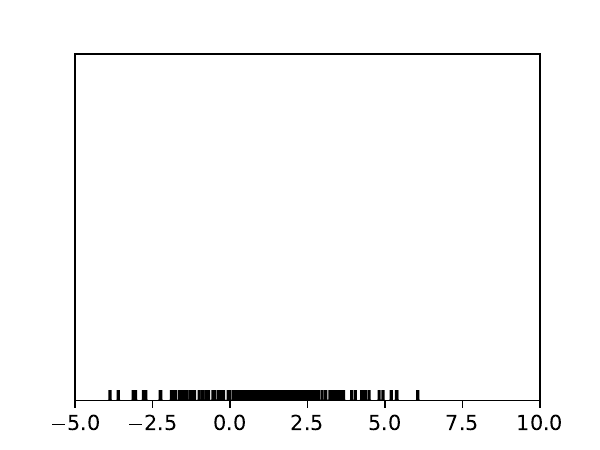}};
        \node[] at (12, 5) (label_4) {\includegraphics[width=6cm]{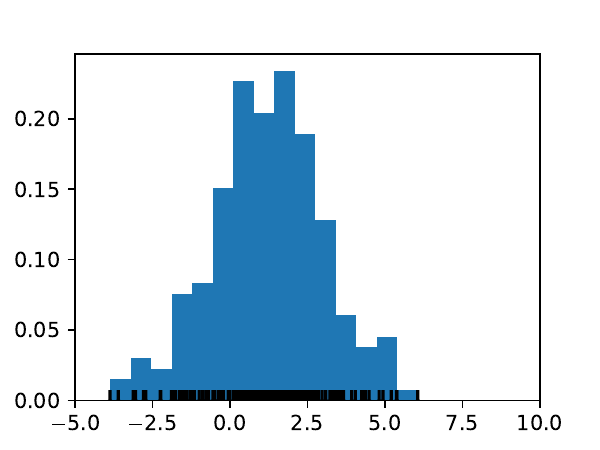}};
        \node[] at (5.1, 0) (label_5) {\includegraphics[width=6cm]{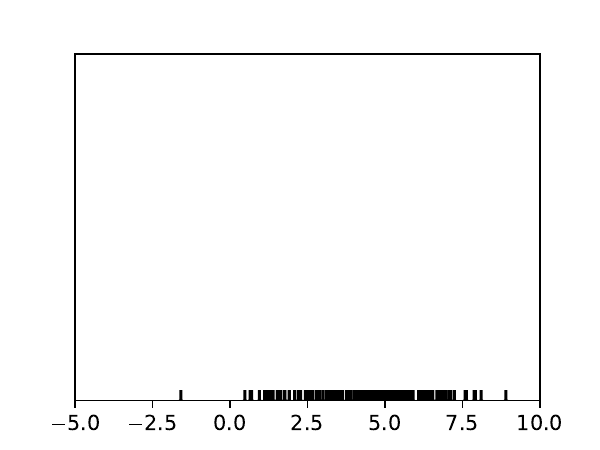}};
        \node[] at (12, 0) (label_6) {\includegraphics[width=6cm]{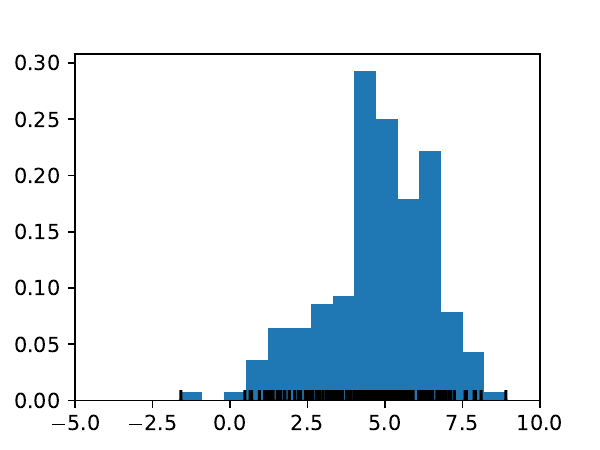}};

        \draw [blackarrow] (0.8,0) -- (2.7,0);
        \draw [blackarrow] (0.8,5) -- (2.7,5);
        \draw [blackarrow] (7.5, 7) to[out=45,in=135, distance=1cm] (9.5, 7);
        \draw [blackarrow] (0.5, 7) to[out=45,in=135, distance=1cm] (2.5, 7);

        \node[] at (1.7,5.5) {$\bx \sim p(\bx|\btheta_1)$};
        \node[] at (1.7,0.5) {$\bx \sim p(\bx|\btheta_2)$};

        \node[] at (1.5,8.2) {Run simulator};
        \node[] at (8.5,8.2) {Build histogram};

        \draw [dashedline] (12.7, 3.28) -- (12.7, 3.78);
        \draw [dashedline] (9.7, 3.78) -- (12.7, 3.78);
        \node[] at (8.4,3.78) {$\hat{p}(\bx_o|\btheta_1)$};
        \node[] at (12.7, 2.7) {$\bx_o$};

        \draw [dashedline] (12.7, -1.72) -- (12.7, 1.56);
        \draw [dashedline] (9.7, 1.56) -- (12.7, 1.56);
        \node[] at (8.4,1.56) {$\hat{p}(\bx_o|\btheta_2)$};
        \node[] at (12.7, -2.3) {$\bx_o$};
    \end{tikzpicture}

    \caption{Simulation-based inference with histograms. For both configurations $\btheta = \btheta_1$ and $\btheta = \btheta_2$, samples are drawn from the simulator. Those samples are then used to construct histograms approximating the densities $p(\bx_0 | \btheta_1)$ and $p(\bx_o | \btheta_2)$.}
    \label{fig:sbi_histograms}
    
\end{figure}
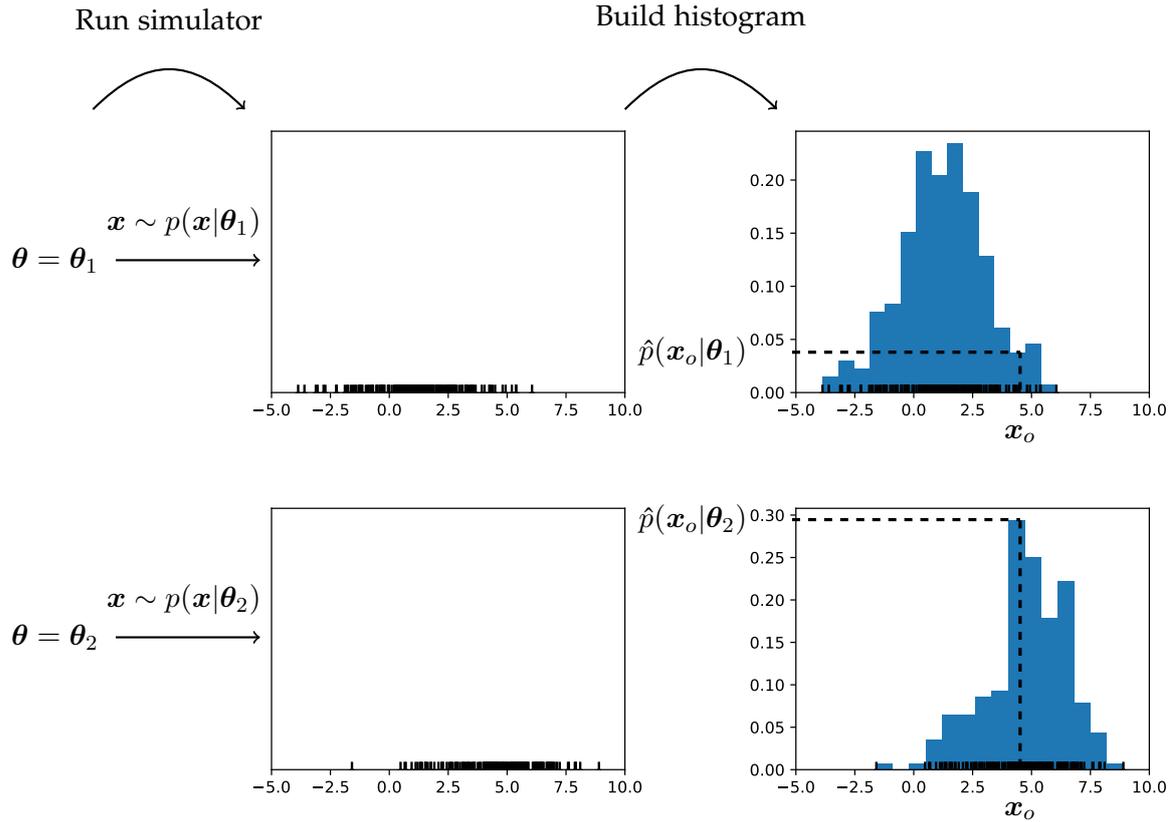

A common simulation-based inference method uses histograms. This method is illustrated in Figure \ref{fig:sbi_histograms}. The idea is to approximate a likelihood evaluation by building a histogram from samples from the likelihood. For any given configuration of parameters $\btheta_i$, a histogram can be constructed from samples $\bx \sim p(\bx | \btheta_i)$ to produce an approximation $\hat{p}(\bx | \btheta_i)$. For a given observation $\bx_o$. An approximate of the likelihood evaluated in $\btheta_i$ is then given by $\hat{p}(\bx_o | \btheta_i)$. Histograms provide approximate likelihood evaluations and can hence be used in any likelihood-based inference procedure. They are usually used for the Neyman construction of confidence regions. It uses the link between hypothesis tests and confidence intervals. Parameters $\btheta_i$ are included in the confidence region if $\bx_o$ is not in the rejection region of the test $H_0: \btheta = \btheta_i, H_1: \btheta \neq \btheta_i$. This rejection region can be approximated using the histograms $p(\bx |\btheta_i)$. By repeating this procedure for many $\btheta_i$, an approximated confidence region can be computed. They are also often used to compute approximate likelihood ratios $\frac{\hat{p}(\bx | \btheta_1)}{\hat{p}(\bx | \btheta_2)}$ that is a key ingredient for hypothesis tests. In some cases, the observable $\bx$ is high dimensional. Consequently, high-dimensional histograms must be constructed, and a too-large amount of samples might be needed to obtain good enough density approximations. To circumvent this issue, the observables can be replaced by low-dimensional summary statistics that can either be handcrafted or learned. Histograms have been used, for example, for the discovery of the Higgs boson \citep{atlas2012observation}.

In the Bayesian setting, a popular method for simulation-based inference is Approximate Bayesian Computation (ABC) \citep{rubin1984, pritchard1999population}. The main idea is to perform simulations with many parameter values $\btheta$ sampled from the prior and to retain parameter values that lead to synthetic observations close to the real observation. In the limit where synthetic observations precisely equal to the real observation are retained, then the retained parameters are samples from the posterior. The algorithm in its most simple form is presented in Algorithm \ref{algo:abc}.

\begin{algorithm}
   \caption{Approximate Bayesian Computation (ABC)}
   \label{algo:abc}
   \begin{tabular}{ l l }
        {\itshape Inputs:} & distance function $d$, distance threshold $\epsilon$, number of simulations $\text{n\_sim}$ \\ & and real observation $\bx_o$.\\
        {\itshape Outputs:} & The retained parameters, approximately sampled from the posterior. \\
  \end{tabular}
  \begin{algorithmic}
    \STATE $\text{retained\_parameters} = \text{List()}$
    \STATE $\text{n\_run\_sim} = 0$
    \WHILE {$\text{n\_run\_sim} < \text{n\_sim}$}
        \STATE $\btheta \sim p(\btheta)$
        \STATE $\bx \sim p(\bx | \btheta)$
        \IF {$d(\bx, \bx_o) < \epsilon$}
            \STATE retained\_parameters.append($\btheta$)
        \ENDIF
        \STATE $\text{n\_run\_sim} = \text{n\_run\_sim} + 1$
    \ENDWHILE
    \STATE \textbf{return} retained\_parameters
   \end{algorithmic}
\end{algorithm} 

Many improvements can be made to the ABC algorithm. We provide a non-exhaustive list of those. See \citet{beaumont2019approximate} for a more detailed review. If the observable $\bx$ is high-dimensional, it can be compressed into automatically constructed summary statistics before being passed to the distance function \citep{fearnhead2012constructing}. \citet{beaumont2002approximate} give more weight to parameters that lead to synthetic observation closer to the real observation. \citet{marjoram2003markov} extend MCMC algorithms to the simulation-based setting by replacing the likelihood-based acceptance rule with an acceptance rule computed from samples like in ABC. \citet{10.1093/bioinformatics/btp619, sisson2007sequential, beaumont2009adaptive} propose a sequential version of the ABC algorithm. An initial population of parameter values is generated from the prior $p(\btheta)$. The algorithm then iteratively refines that population to match the posterior distribution. At each step, synthetic observations are simulated using each of those parameter values. Parameter values that lead to the synthetic observations closest to the real observation are kept, and a new population is built by applying random perturbations to the kept parameter values. This procedure allows us to avoid doing simulations with parameters that are known to produce synthetic observations far from the real observation and to focus on more promising parts of the parameters space iteratively. 

\section{Machine learning for simulation-based inference}
ABC and histogram-based techniques suffer from several limitations. The main limitation of histograms is the inability to scale. If the observation $\bx$ is high dimensional, good summary statistics need to be constructed in order to use histograms. If the parameters $\btheta$ are high dimensional, then an unmanageable number of histograms are required to cover the parameter space. In its most simple form, ABC suffers from the same limitations as histograms. Combining sequential versions of ABC with the use of summary statistics alleviates those issues but makes the algorithm non-amortized, as opposed to amortized algorithms.
\begin{definition}
    An amortized inference algorithm is an algorithm that, once run, allows to perform inference for any observation $\bx$ without requiring heavy computations.
\end{definition}
When using sequential versions of ABC, a population is iteratively built to match the posterior for a given observation. If we were to then perform inference on another observation, we would need to start over again from a fresh population sampled from the prior and reiterate with that new population. It would then require heavy computations as the previous computations cannot be reused. Sequential versions of ABC are hence non-amortized. In addition, handcrafted summary statistics might not contain all the information of the observation necessary to constrain the parameters. In such cases, both ABC and histograms do not yield the correct posterior asymptotically. While there exist methods for learning summary statistics, machine learning-based methods do not require the explicit construction of summary statistics to deal with high-dimensional observations.

Amortized algorithms are desired for several reasons. The first obvious scenario is when inference for observations generated in different settings is needed. If amortized, the algorithm can be run only once and not for each observation. A second scenario where amortized inference is desired is when real-time inference is required. For example, we took advantage of amortization to study gravitational waves emitted by the merge of two black holes, where real-time inference would allow to set up measurement systems to capture other types of signals from this event \citep{delaunoy2020lightning}. Finally, the most important advantage of amortized algorithms is that their quality can be evaluated by performing inferences on many test synthetic observations. This is of high importance to increase the trust that can be given to those approximation techniques. 

Machine learning-based techniques can be designed to not suffer from the same limitations as ABC and histogram-based techniques. It allows the training of a model on several observations to then generalize to any observation and perform amortized inference. The development of neural network architectures that work with complex data such as images, time series, text, and graphs by providing appropriate inductive biases also allows the development of methods that can deal with high-dimensional observations and, to some extent, high-dimensional parameters. For those reasons, machine learning-based techniques are interesting for simulation-based inference but could also be of broad interest for statistical inference. In the remainder of this chapter, we will focus on machine learning-based methods.

\paragraph{Neural posterior and likelihood estimation} Neural Posterior Estimation (NPE) refers to all the methods that aim to build a model approximating the posterior distribution $p(\btheta | \bx)$. To that end, a parametric conditional density estimator $\hat{p}_{\bphi}(\btheta | \bx)$ is trained, where $\bphi$ are the parameters of the density estimator, which is usually a neural network. This density estimator can be trained through variational inference to minimize expected KL divergence between the approximate and true posteriors.
\begin{align}
    \bphi^* 
    & = \argmin_{\bphi} \mathbb{E}_{p(\bx)}\left[ \text{KL} \left[p(\btheta|\bx) || \hat{p}_{\bphi}(\btheta | \bx) \right]\right] \\ 
    &= \argmin_{\bphi} \mathbb{E}_{p(\bx)}\left[ \mathbb{E}_{p(\btheta|\bx)}\left[ \log \frac{p(\btheta|\bx)}{\hat{p}_{\bphi}(\btheta | \bx)}\right]\right] \\
    &= \argmin_{\bphi} \mathbb{E}_{p(\btheta, \bx)}\left[\log \frac{p(\btheta|\bx)}{\hat{p}_{\bphi}(\btheta | \bx)}\right] \\
    &= \argmax_{\bphi} \mathbb{E}_{p(\btheta, \bx)}\left[\log \hat{p}_{\bphi}(\btheta | \bx)\right]
\end{align}
This objective can be approximated using Monte-Carlo
\begin{equation}
    \mathbb{E}_{p(\btheta, \bx)}\left[ \log \hat{p}_{\bphi}(\btheta | \bx)\right] \simeq \frac{1}{N} \sum_{i=1}^N \log \hat{p}_{\bphi}(\btheta_i | \bx_i), \quad (\btheta_i, \bx_i) \sim p(\btheta, \bx).
\end{equation}
Samples from the joint distribution $p(\bx | \btheta)$ can be obtained by first sampling parameters from the prior $p(\btheta)$ and then observations using the simulator $p(\bx | \btheta)$. The density estimator can be of various nature. It can simply be a neural network that takes the observation $\bx$ as input and outputs the parameters of a normal distribution conditioned on that observation $\bx$. However, to have the flexibility to model complex distributions, normalizing flows \citep{dinh2015nice, rezende2015variational} are often used in practice. Normalizing flows are composed of a base probability distribution $q(\bz)$ and a neural network $f_{\bphi}$ defining a transformation from that base distribution to the target distribution. This distribution can be conditioned on $\bx$ by giving it as input to the neural network performing the transformation. By the change of variable formula, the approximate posterior distribution is computed as
\begin{equation}
    \hat{p}_{\bphi}(\btheta | \bx) = q(f_{\bphi}^{-1}(\btheta; \bx))\left| \text{det} \frac{\partial f_{\bphi}^{-1}(\btheta; \bx)}{\partial \btheta}  \right|
\end{equation}
Note that the neural network-based transformation must be chosen such that the inverse $f_{\bphi}^{-1}(.;\bx)$ exists and the Jacobian $\left| \text{det} \frac{\partial f_{\bphi}^{-1}(\btheta; \bx)}{\partial \btheta}  \right|$ can easily be computed. Normalizing flows then allow approximate posterior density evaluation, but it is also possible to sample parameters from that approximate posterior distribution. To do so, samples are first sampled from the base distribution $\bz \sim q(\bz)$ and then transformed into samples from the target distribution $f_{\bphi}(\bz; \bx) \sim \hat{p}_{\bphi}(\btheta | \bx)$.

Neural Likelihood Estimation (NLE) refers to all the methods that aim to build a model approximating the likelihood distribution $p(\bx | \btheta)$. Similarly to NPE, such models can be trained by minimizing the expected KL divergence
\begin{align}
    \bphi^* 
    & = \argmin_{\bphi} \mathbb{E}_{p(\btheta)}\left[ \text{KL} \left[p(\bx|\btheta) || \hat{p}_{\bphi}(\bx | \btheta) \right]\right] \\ 
    &= \argmin_{\bphi} \mathbb{E}_{p(\btheta)}\left[ \mathbb{E}_{p(\bx|\btheta)}\left[ \log \frac{p(\bx|\btheta)}{\hat{p}_{\bphi}(\bx | \btheta)}\right]\right] \\
    &= \argmin_{\bphi} \mathbb{E}_{p(\btheta, \bx)}\left[ \log \frac{p(\bx|\btheta)}{\hat{p}_{\bphi}(\bx | \btheta)}\right] \\
    &= \argmax_{\bphi} \mathbb{E}_{p(\btheta, \bx)}\left[ \log \hat{p}_{\bphi}(\bx | \btheta)\right].
\end{align}
The model $\hat{p}_{\bphi}(\bx | \btheta)$ can again be a normalizing flow or another density estimator. As NLE only provides the likelihood and not the posterior, likelihood-based inference techniques such as VI or MCMC must be applied using the approximate likelihood.

\paragraph{Neural ratio estimation} In the Bayesian setting, Neural Ratio Estimation (NRE) refers to the methods that aim to build a model for the ratio between two quantities that can then be used to approximate the posterior. The first version, often referred to as NRE-A or NRE \citep{2019arXiv190304057H}, builds a model for the likelihood to evidence ratio $r(\btheta, \bx) = \frac{p(\bx | \btheta)}{p(\bx)}$. If we have access to an approximation $\hat{r}(\btheta, \bx)$, we can construct an approximate posterior density by multiplying it by the prior 
\begin{equation}
    \hat{p}(\btheta | \bx) = \hat{r}(\btheta, \bx) p(\btheta),
\end{equation}
because
\begin{equation}
    p(\btheta | \bx) = \frac{p(\bx | \btheta) p(\btheta)}{p(\bx)} = r(\btheta, \bx) p(\btheta).
\end{equation}
To obtain an approximation of the likelihood-to-evidence ratio, a classifier $\hat{d}(\btheta, \bx)$ is trained to distinguish pairs $(\btheta, \bx)$ sampled from the joint distribution $(\btheta, \bx) \sim p(\btheta, \bx)$, with class label $y=1$, from pairs sampled from the marginal distributions $(\btheta, \bx) \sim p(\btheta) p(\bx)$, with class label $y=0$. It can be shown that the output of the optimal classifier trained with the binary cross-entropy loss is
\begin{equation}
    d(\btheta, \bx) = \frac{p(\btheta, \bx)}{p(\btheta, \bx) + p(\btheta) p(\bx)}.
\end{equation}
Consequently,
\begin{equation}
    \frac{d(\btheta, \bx)}{1-d(\btheta, \bx)} = \frac{p(\btheta, \bx)}{p(\btheta) p(\bx)} = \frac{p(\bx|\btheta)}{p(\bx)} = r(\btheta, \bx).
\end{equation}
An approximation of the likelihood-to-evidence ratio is then given by
\begin{equation}
    \hat{r}(\btheta, \bx) = \frac{\hat{d}(\btheta, \bx)}{1-\hat{d}(\btheta, \bx)} = \exp(\sigma^{-1}(\hat{d}(\btheta, \bx))),
\end{equation}
where $\sigma$ is the sigmoid function. To avoid numerically unstable computations, the approximate log posterior can be computed as
\begin{equation}
    \log(\hat{p}(\btheta | \bx)) = \log(\hat{r}(\btheta, \bx)) + \log(p(\btheta)) = \sigma^{-1}(\hat{d}(\btheta, \bx)) + \log(p(\btheta)),
\end{equation}
where $\sigma^{-1}(\hat{d}(\btheta, \bx))$ can be obtained by evaluating the neural network without the final sigmoid activation layer. This formulation allows us to evaluate the approximate log posterior density. To obtain samples from this distribution, running an additional MCMC algorithm is required.

NRE-B \citep{durkan2020contrastive} employs a contrastive learning scheme by considering several samples from the product of marginal distributions. They build a model $\hat{f}(\btheta, \bx)$ trained on the following contrastive loss
\begin{equation}
    L(\hat{f}) = - \frac{1}{B} \sum_{b=1}^B \log \frac{\exp(\hat{f}(\btheta^{(b)}, \bx^{(b)}))}{\sum_{k=1}^K \exp(\hat{f}(\btheta^{(k)}, \bx^{(b)}))},
\end{equation}
where $(\btheta^{(b)}, \bx^{(b)})$ are sampled from the joint distribution $p(\btheta, \bx)$ and the $K$ contrastive parameters $\btheta^{(k)}$ are each time composed of $\btheta^{(b)}$ and $K-1$ parameters set sampled independently from the prior $p(\btheta)$. When trained at the optimum, the output of the model is
\begin{equation}
    f(\btheta, \bx) = \log(p(\btheta | \bx)) - \log(p(\btheta)) + C(\bx),
\end{equation}
where $C(\bx)$ is an unknown constant. We can, therefore, recover an approximation for the unnormalized posterior density
\begin{equation}
    \log \hat{p}(\btheta | \bx)C(\bx) = \hat{f}(\btheta, \bx) + \log(p(\btheta)).
\end{equation}

NRE-C \citep{miller2022contrastive} aim to improve over NRE-B by removing the unknown constant. They do that by sampling all the pairs from the contrastive set independently from $\bx$ with probability $p_0$ instead of always having one of the parameter sets sampled jointly with $\bx$. They train their model on the following loss function
\begin{equation}
\begin{split}
    L(\hat{f}) = 
    & - \frac{1}{B} (1-p_0) \sum_{b=1}^{B} \log \frac{\frac{1-p_0}{p_0}\exp(\hat{f}(\btheta^{(b)}, \bx^{(b)}))}{K + \frac{1-p_0}{p_0}\sum_{k=1}^K \exp(\hat{f}(\btheta^{(k)} \bx^{(b)}))}\\ 
    & - \frac{1}{B} p_0 \sum_{b=B+1}^{2B} \log \frac{K}{K + \frac{1-p_0}{p_0}\sum_{k=1}^K \exp(\hat{f}(\btheta^{(k)}, \bx^{(b)}))},
\end{split}
\end{equation},
where $(\btheta^{(b)}, \bx^{(b)}) \sim p(\btheta, \bx)$ for $b = (1, ..., B)$ and $(\btheta^{(b)}, \bx^{(b)}) \sim p(\btheta)p(\bx)$ for $b = (B+1, ..., 2B)$. Again, the $K$ contrastive parameters $\btheta^{(k)}$ are each time composed of $\btheta^{(b)}$ and $K-1$ parameters set sampled independently from the prior $p(\btheta)$. In this case, when trained at the optimum, the output of the model is
\begin{equation}
    f(\btheta, \bx) = \log(p(\btheta | \bx)) - \log(p(\btheta)).
\end{equation}
We can recover an approximation for the normalized posterior density
\begin{equation}
    \log \hat{p}(\btheta | \bx) = \hat{f}(\btheta, \bx) + \log(p(\btheta)).
\end{equation}
Let us note that NRE-A is a special case of NRE-C when $p_0 = 1/2$ and $K=1$. In the limit of $p_0 \rightarrow 0$, NRE-C approaches NRE-B.

In the frequentist setting, \citet{cranmer2015approximating} trains a classifier to approximate likelihood ratios. Those can then be used for likelihood ratio hypothesis tests or to construct confidence intervals using Wilk's theorem or through a Neyman construction \citep{atlas2024implementation}. \citet{brehmer2019mining} show how to use the joint score $\nabla_{\btheta} \log p(\bx, \bz | \btheta)$, where $\bz$ are the latent variables to improve the inference quality. \citet{stoye2018likelihood} present an improved cross-entropy estimator to train the classifier. \citet{dalmasso2020confidence} show how other statistics could be used and present a framework for constructing confidence sets with bounded error and performing diagnostics.

\begin{figure}[h!]
    \centering
    \includegraphics[width=1.1\textwidth]{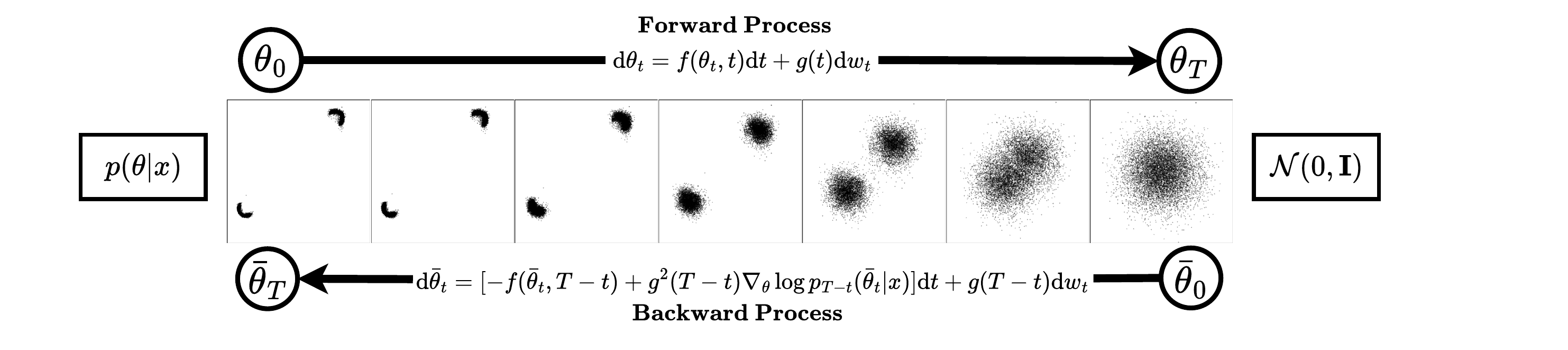}
    \caption{Visualization of a diffusion model applied on the two moons simulation-based inference benchmark. A backward process is learned to inverse a forward diffusion process. Applying the backward process to noise can then recover samples from the posterior distribution. Figure reproduced from \citet{sharrock2022sequential}.}
    \label{fig:nse_visualization}
\end{figure}

\paragraph{Neural posterior score estimation}
Neural Posterior Score Estimation (NPSE) \citep{geffner2022score, sharrock2022sequential, linhart2024diffusion} aim to compute a diffusion model score such that produced samples are approximately sampled from the posterior. As illustrated in Figure \ref{fig:nse_visualization}, diffusion models aim to inverse a diffusion process, also called a forward process. The diffusion process takes a sample from the target distribution as input and iteratively adds noise to it until there is only noise left. The inverse process, also called the backward process, takes a sample from the noise distribution as input and iteratively denoises it to obtain a sample from the target distribution. In our case, the target distribution is the posterior $p(\btheta | \bx)$. The denoising process uses a score function $\nabla_{\btheta} \log p_{T-t}(\bar{\btheta}|\bx)$ to control how the samples are denoised. To obtain samples from the approximate posterior $\hat{p}(\btheta | \bx)$, an approximation of the score is learned $s(\btheta_t, \bx, t)$ and this score is used to denoise samples drawn from the noise distribution.

Diffusion techniques have some advantages over NPE techniques. First, diffusion models work well at modeling high-dimensional distributions. Those models are then useful to obtain high-quality approximations when $\btheta$ is high-dimensional. Second, the score of multiple observations can be expressed as a composition of scores for a single observation \citep{geffner2022score, linhart2024diffusion}. The key observation is that
\begin{equation}
    p(\btheta | \bx_1, ..., \bx_n) \propto p(\btheta)^{1-n} \prod_{i=1}^n p(\btheta | \bx_i).
\end{equation}
\citet{geffner2022score} define the forward diffusion process such that the intermediate densities are
\begin{equation}
    p^f_t(\btheta | \bx_1, ..., \bx_n) \propto (p(\btheta)^{1-n})^{\frac{T-t}{T}} \prod_{i=1}^n p_t(\btheta | \bx_i),
\end{equation}
where $p_t(\btheta | \bx_i)$ is the distribution obtained after applying progressively some Gaussian noise for a time $t$. The score is then expressed
\begin{equation}
    \nabla_{\btheta} \log p^f_{t}(\btheta|\bx_1, ..., \bx_n) = \frac{(1-n)(T-t)}{T} \nabla_{\btheta} \log p(\btheta) + \sum_{i=1}^n \nabla_{\btheta} \log p_t(\btheta | \bx_i).
\end{equation}
As a consequence, score-based models trained on single observations could be used to approximate the posterior for any number of conditioning observations without having to retrain a model for each case. This method has the drawback of requiring Langevin dynamics to run the denoising process. This is due to the way intermediate densities are defined. To improve on that aspect, \citet{linhart2024diffusion} use the following diffusion process
\begin{equation}
    \btheta_t = \sqrt{\alpha_t} \btheta_0 + \sqrt{1-\alpha_t} \epsilon_t,
\end{equation}
where $\alpha_t$ is a parameter controlling noise intensity at different stages of the diffusion process, and $\epsilon_t$ are sampled from a Normal distribution. In this, case the score can be written as
\begin{equation}
    \nabla_{\btheta} \log p_{t}(\btheta|\bx_1, ..., \bx_n) = (1-n) \nabla_{\btheta} \log p_t^{\lambda}(\btheta) + \sum_{i=1}^n \nabla_{\btheta} \log p_t(\btheta | \bx_i) + \nabla_{\btheta} \log L_{\lambda}(\btheta, \bx1, ..., \bx_n),
\end{equation}
where, $\lambda$ is the prior, $p_t^{\lambda}$ is the diffused prior and $\nabla_{\btheta} \log L_{\lambda}(\btheta, \bx1, ..., \bx_n)$ is a term that needs to be approximated.

\citet{gloeckler2024compositional} showed that this compositional property of scores can be used when dealing with arbitrary length time series that satisfies the Markov property. Under that hypothesis, an arbitrary length time series can be viewed as a collection of one-step transitions of arbitrary size. The score for multiple transitions can then be recovered from the score for a single transition.

The main drawback of score-based methods compared to NPE with normalizing flows is that the denoising process only provides a way for drawing samples from an approximate posterior distribution but does not allow for approximate posterior density evaluation. This density can be computed by running the corresponding probability flow ordinary differential equation, but this has to be approximated using many neural network calls \citep{song2020score}.

\begin{figure}[h!]
    \centering
    \includegraphics[width=\textwidth]{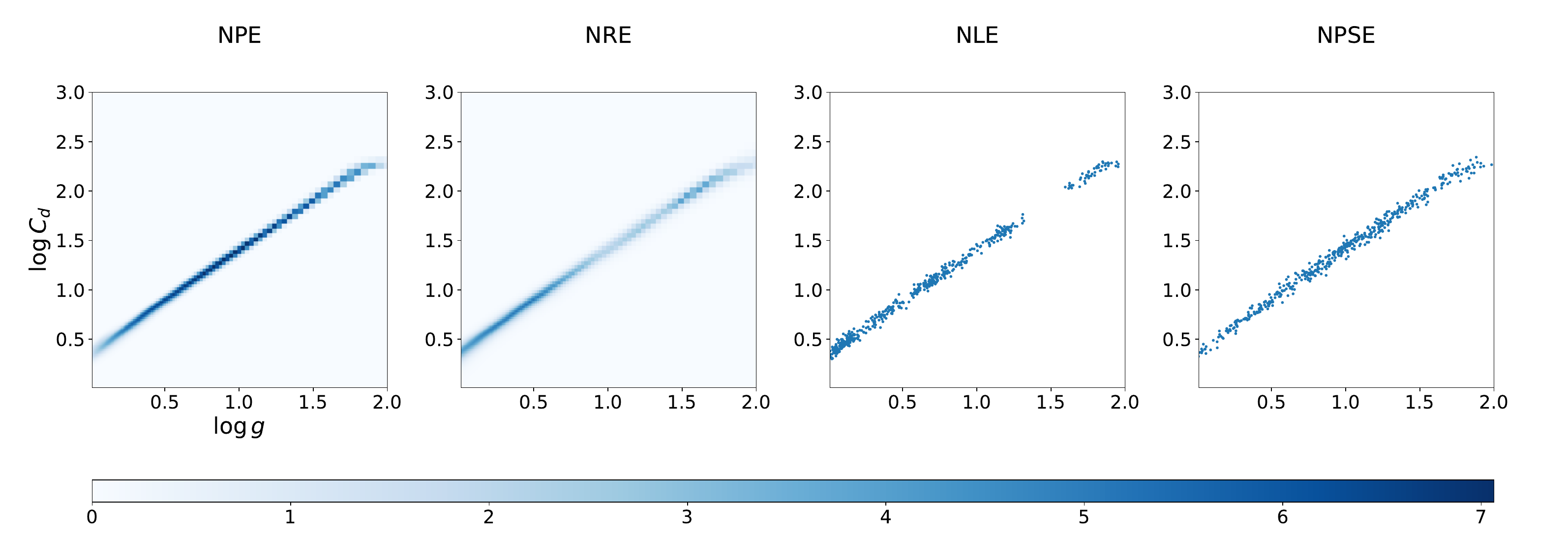}
    \caption{Approximate posteriors for the feather falling simulator with measured falling time of $4.68\text{s}$, $6.97\text{s}$ and $9.69\text{s}$ for heights $3\text{m}$, $5\text{m}$ and $7\text{m}$ respectively.}
    \label{fig:sbi:sbi_posteriors}
\end{figure}

\paragraph{Illustrative example}
We apply the NPE, NRE, NLE, and NPSE algorithms on the feather falling simulator to infer both the gravitational constant $g$ and the drag coefficient $C_d$. We consider the same observations as in Chapter \ref{c:Statistical_modeling}, being measured falling time of $4.68\text{s}$, $6.97\text{s}$ and $9.69\text{s}$ for heights $3\text{m}$, $5\text{m}$ and $7\text{m}$ respectively. Results are shown in Figure \ref{fig:sbi:sbi_posteriors}. Density plots are provided for NPE and NRE as those two methods allow for posterior density evaluation. For NLE and NPSE, $500$ samples are drawn from the posterior by running MCMC for NLE and diffusion for NPSE. From those plots, we can see that inferring the product of $g$ and $C_d$ seems to be easy, but it is hard to distinguish between both $g$ and $C_d$ being low or both being high. All methods provide similar results, but some visible differences remain. All methods are trained on the same $10,000$ samples from the simulator $(\btheta, \bx) \sim p(\bx | \btheta) p(\btheta)$.

\paragraph{Sequential machine learning techniques}
Sequential methods aim to reduce the amount of computations required when we are interested in the posterior for a single given observation $\bx_o$. In such cases, we aim to construct an approximation of the posterior $\hat{p}(\btheta | \bx)$ that is close to the true posterior when $\bx = \bx_o$ but can be a poor approximation when $\bx \neq \bx_0$. To efficiently construct this approximation, the main idea is to favor sampling in regions likely to provide observations close to $\bx_0$ in order for the model to be trained on many samples in the region of interest. To that end, the model training procedure operates in rounds where, at each round, the previously obtained approximate posterior is used to construct a proposal distribution $\tilde{p}(\btheta)$ that should lead to simulations close to $\bx_0$. In each round, a new training set is then sampled by drawing parameters $\btheta$ from the proposal $\tilde{p}(\btheta)$ and observations using the simulator $p(\bx|\btheta)$. An overview of sequential algorithms is shown in Algorithm \ref{algo:sequential}.

\begin{algorithm}[h]
   \caption{Sequential simulation-based inference}
   \label{algo:sequential}
   \begin{tabular}{ l l }
        {\itshape Inputs:} & A prior $p(\btheta)$, a simulator $p(\bx | \btheta)$ and a target $\bx_o$\\
        {\itshape Outputs:} & An approximate posterior $\hat{p}(\btheta | \bx)$ with focused training on $\bx_o$\\
        {\itshape Hyperparameter:} & Number rounds $\text{n\_rounds}$ and number of samples per round $\text{n\_samples}$\\
  \end{tabular}
  \begin{algorithmic}
    \STATE $\tilde{p}(\btheta) = p(\btheta)$
    \STATE $\text{current\_round} = 0$
    \WHILE {$\text{current\_round} < \text{n\_rounds}$}
        \STATE $\text{dataset} = \text{List}()$
        \STATE $\text{current\_sample} = 0$
        \WHILE {$\text{current\_sample} < \text{n\_samples}$}
            \STATE $\btheta \sim \tilde{p}(\btheta)$
            \STATE $\bx \sim p(\bx | \btheta)$
            \STATE $\text{dataset.append}(\bx, \btheta)$
            \STATE $\text{current\_sample} = \text{current\_sample} + 1$
        \ENDWHILE
        \STATE $\hat{p}(\btheta| \bx) = \text{train}(\text{dataset}, \tilde{p}(\btheta))$
        \STATE $\tilde{p}(\btheta) = \hat{p}(\btheta | \bx = \bx_o)$
        \STATE $\text{current\_round} = \text{current\_round} + 1$
    \ENDWHILE
    \STATE \textbf{return} $\hat{p}(\btheta| \bx)$
   \end{algorithmic}
\end{algorithm} 

Sequential training can be performed in combination with neural likelihood estimation \citep{papamakarios2019sequential}, neural ratio estimation \citep{2019arXiv190304057H, durkan2020contrastive, miller2022contrastive}, neural posterior estimation \citep{papamakarios2016fast, lueckmann2017flexible, greenberg2019automatic} and neural posterior score estimation \citep{sharrock2022sequential}. When directly approximating posteriors or posterior scores, the training procedure should take into account the fact that the parameters $\btheta$ are not sampled from the prior $p(\btheta)$ but from the proposal $\tilde{p}(\btheta)$. For neural posterior estimation, this can be taken into account by reweighting the approximate posterior after the training procedure \citep{papamakarios2016fast}, reweighting the importance of each sample in the loss \citep{lueckmann2017flexible} or training the model with a contrastive learning scheme \citep{greenberg2019automatic}. \citet{sharrock2022sequential} adapts those methods to the case of posterior scores.

The major drawback of sequential machine learning methods is that, as the sequential versions of the ABC algorithm, they are not amortized. This means that the trained model cannot be used to perform inference on observations $\bx \neq \bx_o$ as the model has not been trained on samples from that part of the observation space. Consequently, the quality of the model cannot be evaluated by performing inference on several synthetic test observations. In an attempt to get the best of both worlds, \citet{miller2021truncated} introduced truncated marginal neural ratio estimation. The key idea is to train the model to be valid for a small subspace around $\bx_o$ generated by a truncated version of the prior. The training is then still faster than training on the full observation space, and the resulting model is what they call locally amortized. The model can then be tested locally on synthetic observations generated using parameters drawn from the truncated prior. This idea has then been adapted to neural posterior estimation \citep{deistler2022truncated} and neural posterior score estimation \citep{sharrock2022sequential}.

\paragraph{Additional contributions to SBI}
In this paragraph, we non-exhaustively provide short descriptions of some other pieces of work from the SBI literature. We reviewed above the use of normalizing flows for likelihood and posterior estimation, classifiers for likelihood-to-evidence ratio estimation, and the use of posterior score estimation models. Other generative models can be used. \citet{ramesh2022gatsbi} and \citet{pacchiardi2022likelihood} explore the use of generative adversarial networks. \citet{glaser2022maximum} propose to use energy-based models, \citet{schmitt2023consistency} explore the use of consistency models, and \citet{wildberger2024flow} use flow matching. \citet{haggstrom2024fast} use an inverse regression approach via Gaussian locally linear mappings that has the benefit of being lightweight compared to neural network approaches.

\citet{glockler2021variational}, \citet{radev2023jana}, \citet{schmitt2023leveraging} and \citet{haggstrom2024fast} consider jointly approximating the posterior and likelihood. \citet{glockler2021variational} leverage both approximations to avoid the need for proposal corrections in sequential methods and still enable direct posterior estimation. \citet{radev2023jana} show that having both approximations allows one to run new types of tests to verify the quality of the approximations. \citet{schmitt2023leveraging} show that jointly training likelihood and posterior approximation by enforcing consistency between each other improves approximation quality in low-data regimes. \citet{haggstrom2024fast} show that an inverse regression approach via Gaussian local linear mapping provides both the posterior and likelihood.

\citet{rodrigues2021leveraging} and \citet{heinrich2023hierarchical} apply simulation-based inference to hierarchical Bayesian models. In those settings, both local and global parameters need to be inferred, where local parameters affect individual events and global parameters are shared by a set of events. \citet{rodrigues2021leveraging} consider the Bayesian case and do this by learning an approximation for the posterior over global parameters given a set of observations and an approximation for the posterior over local parameters given the local observation and global parameters. \citet{heinrich2023hierarchical} adapts this methodology to frequentist inference.

\citet{rozet2021arbitrary}, \citet{miller2021truncated}, \citet{rozet2023score}, and \citet{gloeckler2024all} aim to improve the flexibility of simulation-based inference methods by training models that can yield multiple distribution approximations. \citet{rozet2021arbitrary} train a neural ratio estimation model that can yield any parameter marginals on the fly. \citet{miller2021truncated} shows that using truncated sequential algorithms allows the training of an additional model for any parameter marginals using the truncated prior from the last round without restarting the sequential procedure from scratch. \citet{gloeckler2024all} use diffusion model with the transformer architecture to construct a model that can be queried to sample from all conditionals of the joint distribution $p(\btheta, \bx)$. \citet{rozet2023score} consider the problem of data assimilation where state trajectories are inferred from arbitrary observations.

\section{Diagnosing simulation-based inference}

We reviewed how different quantities, such as the posterior, likelihood, likelihood-to-evidence ratio, and posterior scores, could be approximated using samples from a simulator. Those approximations can then be used to make hypothesis tests or construct credible or confidence regions. However, those approximations can be of poor quality for many reasons. If the simulator is complex, many samples might be required to capture its behavior fully. If too few samples are used due to computational constraints, then the trained model might be of poor quality. Poor quality approximations could also happen if the machine learning model is not flexible enough or poorly optimized. For all those reasons, the quality of the obtained approximations should always be checked before relying on it in downstream tasks.

\begin{figure}[h]
    \centering
    \includegraphics[width=\textwidth]{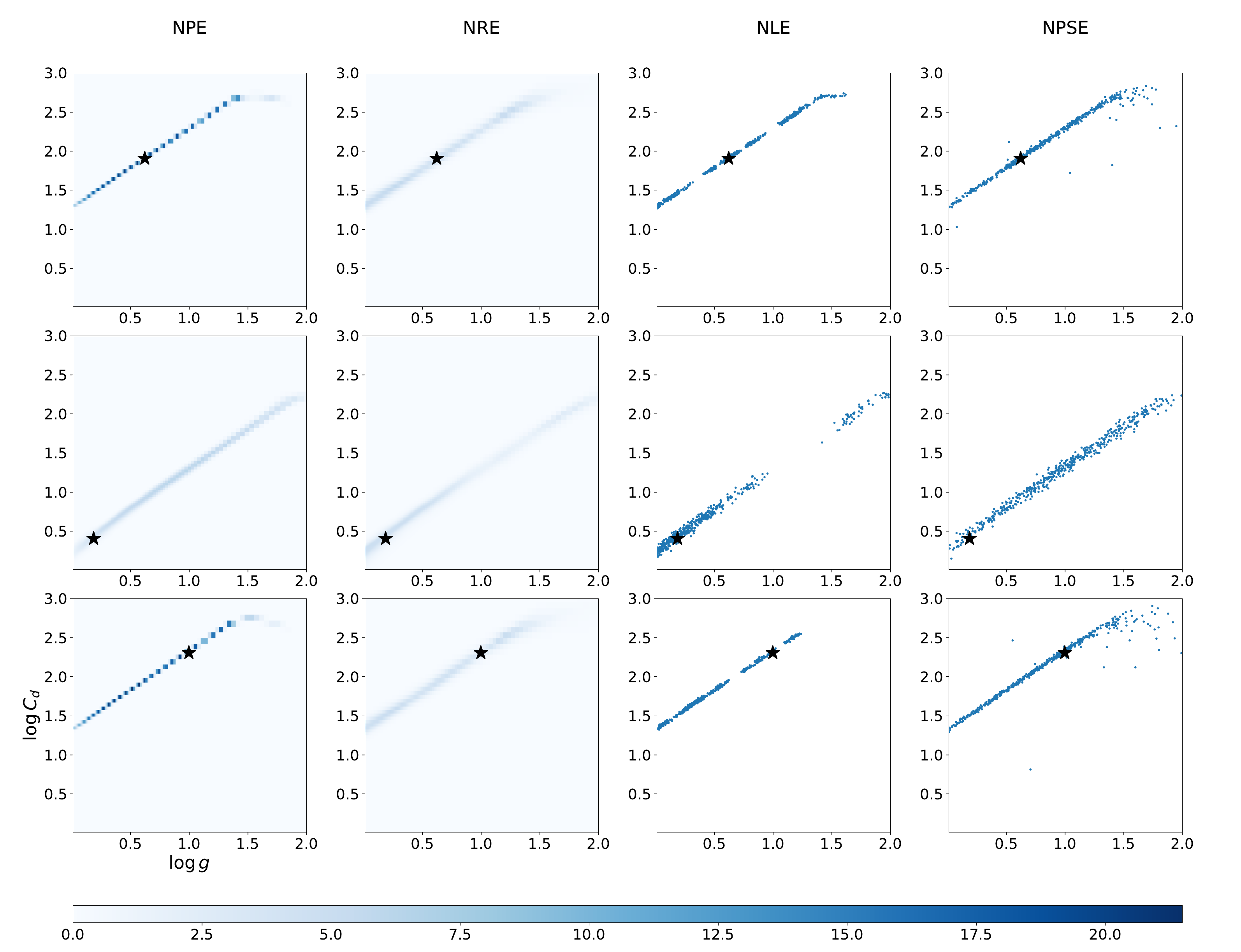}
    \caption{Visual inspection of the posterior quality. Either the approximate posterior density $\hat{p}(\btheta | \bx)$ or samples from it are shown. The black star represents the parameters $\btheta^*$ that were used to generate the synthetic test observation.}
    \label{fig:sbi:visual_inspection}
\end{figure}

The first thing to do is visually inspect the obtained posteriors on some test simulations for which the simulator's parameters are known. We sample parameters from the prior $\btheta^* \sim p(\btheta)$ and then synthetic observations from the simulator using those parameters $\bx^* \sim p(\bx | \btheta^*)$. We plot in Figure \ref{fig:sbi:visual_inspection} the approximates posteriors $\hat{p}(\btheta | \bx^*)$ and the parameters $\btheta^*$ used to simulate the observations $\bx^*$ for the feather falling problem. We observe that there is usually a high approximate posterior mass $\hat{p}(\btheta | \bx^*)$ around the parameters $\btheta^*$, indicating that the approximations do not look too bad.

The validity of an approximation can also be tested by performing a hypothesis test of the form 
\begin{itemize}

    \item $H_0: \hat{p}(\btheta | \bx) = p(\btheta | \bx),\ \forall \btheta, \bx$
    \item $H_1: \exists\ \btheta, \bx: \hat{p}(\btheta | \bx) \neq p(\btheta | \bx)$.
\end{itemize}

The Simulation-Based Calibration (SBC) algorithm \citep{sbc} constructs a histogram of ranks of samples from the true posterior compared to $L$ samples from the approximate posterior. Ranks are constructed as follow:

\begin{align*}
    \btheta^* &\sim p(\btheta) \\
    \bx^* &\sim p(\bx | \btheta^*) \\
    \{\btheta_1, ..., \btheta_L\} &\sim \hat{p}(\btheta | \bx^*) \\
    \text{rank} &= \sum_{i=1}^L \mathds{1}[f(\btheta_i) < f(\btheta^*)],
\end{align*}

where $f$ is a function compressing parameters $\btheta$ into a single real value. Ranks are then integer in $[0, L]$. It should be noted that sampling $\btheta^* \sim p(\btheta), \bx^* \sim p(\bx | \btheta^*)$ is equivalent to $\bx^* \sim p(\bx), \btheta^* \sim p(\btheta | \bx^*)$. The algorithm then compares a sample from $p(\btheta | \bx^*)$ to $L$ samples from $ \hat{p}(\btheta | \bx^*)$. If $\hat{p}(\btheta | \bx) = p(\btheta | \bx),\ \forall \btheta, \bx$, then the ranks should be distributed uniformly in $[0, L]$ as the two distributions are the same. If the histogram of ranks obtained empirically is too far from a uniform distribution, we can then conclude that $H_0$ can be rejected and that the approximation is not perfect. 

By running the SBC diagnostic for the different tested methods on the feather falling simulator, we obtain p-values of $0.000033$ for NPE, $0.22$ for NRE, $0.0086$ for NLE and $0.0087$ for NPSE. This means that we could reject the null hypothesis at a level $\alpha = 0.01$ for all the approximations but the one obtained with NRE. Note that it does not mean that NRE's approximation is trustworthy but only that the others are not. Being able to reject $3$ out of the $4$ approximations highlights the fact that approximations should not always be trusted and that careful analysis of their quality is required. For this analysis, $1000$ samples $(\btheta^*, \bx^*) \sim p(\btheta, \bx)$ were used, and for each observation $\bx^*$, $100$ samples were drawn from the approximate posterior. Parameters are compressed into a single real value through the function $f: g, C_d \rightarrow g \times C_d$.

The "Tests of Accuracy with Random Points" (TARP) algorithm \citep{lemos2023sampling} considers the special case of $f$ being the distance to a random reference point. They show that, in this case, asymptotic uniformity of the histogram implies that the posterior approximation is perfect. This is then not just a necessary condition but also a sufficient condition.

\citet{linhart2022validation} consider the special case of $\hat{p}(\btheta | \bx)$ being modeled by a normalizing flow with a normal distribution as base distribution. Instead of comparing the approximate and true posteriors in the space of parameters $\btheta$. They project those in the space of the base distribution of the normalizing flow by using the transformation defined by this flow. In this space, the approximate posterior is a normal distribution by definition. We can obtain samples from the true posterior in this space by applying the normalizing flow transformation on those, and they can then be compared to a normal distribution. They use the fact that the c.d.f. of a normal variable can be computed analytically to compute the c.d.f. for each dimension of the projected samples of the true posterior. If the approximation of the posterior is perfect, then those computed c.d.f. should be uniformly distributed, and all the dimensions should be independent of each other. Those two elements are then tested.

When the true posterior distribution is known, e.g., on toy problems for benchmarking purposes, the classifier two samples test (C2ST) can be used \citep{lopez2017revisiting}. It consists in training a classifier to discriminate between samples $\btheta \sim \hat{p}(\btheta | \bx)$ and $\btheta \sim p(\btheta | \bx)$. If the approximate posterior is equal to the true posterior, then even a perfectly trained classifier should have an accuracy of $50\%$. If we assume that a sufficient amount of samples are used to assess the accuracy and that the accuracy is not $50\%$, then we can conclude that the approximation is not perfect. \citet{linhart2024c2st} extends this algorithm to only require samples from the joint distribution $p(\btheta, \bx)$ and hence be applicable in a general simulation-based inference setting. In this setting, the classifier takes both observations $\bx$ and parameters $\btheta$ as input and is trained to discriminate between $\btheta, \bx \sim p(\btheta, \bx)$ and $\btheta, \bx \sim \hat{p}(\btheta | \bx) p(\bx)$. The deviation of the output of a classifier trained in such a way, from uniformity between both classes can then be used as a test statistic. Critical values for this test can be computed empirically by training classifiers on identical distributions, obtained by merging the samples from each distribution and randomly assigning a class to each sample.

Hypothesis tests provide a way to potentially reject the fact that the approximation is perfect. However, it should be kept in mind that failing to reject the null hypothesis does not necessarily mean that it is likely valid. Consequently, those tests do not guarantee the validity of the approximation. Moreover, the p-values associated with such tests should not be interpreted as distances between the true posterior and the approximation. To obtain a notion of distance, one could consider approximating the expected KL divergence between the true and approximate posteriors. 
\begin{align}
    \mathbb{E}_{p(\bx)}\left[ \text{KL} \left[p(\btheta|\bx) || \hat{p}(\btheta | \bx) \right]\right] 
    &=  \mathbb{E}_{p(\bx)}\left[ \mathbb{E}_{p(\btheta|\bx)}\left[\log \frac{p(\btheta|\bx)}{\hat{p}(\btheta | \bx)}\right]\right] \\
    &= \mathbb{E}_{p(\bx, \btheta)}\left[\log \frac{p(\btheta|\bx)}{\hat{p}(\btheta | \bx)}\right] \\
    &\propto - \mathbb{E}_{p(\bx, \btheta)}\left[\log \hat{p}(\btheta | \bx)\right] \\
    &\simeq - \frac{1}{N} \sum_{i=1}^N \log \hat{p}(\btheta_i | \bx_i), \quad (\btheta_i, \bx_i) \sim p(\btheta, \bx).
\end{align}
While $\mathbb{E}_{p(\bx, \btheta)}\left[ \log \hat{p}(\btheta | \bx)\right]$ values are hard to interpret, we can conclude that an approximation $\hat{p}_1(\btheta | \bx)$ is better on average than an approximation $\hat{p}_2(\btheta | \bx)$, according to the KL divergence, if $\mathbb{E}_{p(\bx, \btheta)}\left[ \log \hat{p}_1(\btheta | \bx)\right] > \mathbb{E}_{p(\bx, \btheta)}\left[ \log \hat{p}_2(\btheta | \bx)\right]$ and hence $\mathbb{E}_{p(\bx)}\left[ \text{KL} \left[p(\btheta|\bx) || \hat{p}_1(\btheta | \bx) \right]\right] < \mathbb{E}_{p(\bx)}\left[ \text{KL} \left[p(\btheta|\bx) || \hat{p}_2(\btheta | \bx) \right]\right] $. On the feather falling simulator, we have that $\mathbb{E}_{p(\bx, \btheta)}\left[ \log \hat{p}(\btheta | \bx)\right]$ is equal to $1.55$ for the NPE approximation and to $0.897$ for the NRE approximation. We can conclude that, according to the KL divergence, the NPE approximation is closer to the posterior than the one obtained with NRE, although we were able to reject NPE's approximation and not NRE's. It cannot be evaluated for NLE and NPSE as those only provide sampling from the approximate posterior and not density evaluation.

Alternatively, if the true posterior is known for benchmarking purposes, the Maximum Mean discrepancy (MMD) \citep{gretton2012kernel} can be used. The squared MMD can be expressed
\begin{equation}
\begin{aligned}
    \text{MMD}^2(p(\btheta | \bx), \hat{p}(\btheta | \bx)) &= \mathbb{E}_{\btheta, \btheta' \sim (p(\btheta | \bx)}\left[(k(\btheta, \btheta')\right] \\ &- 2 \mathbb{E}_{\btheta \sim (p(\btheta | \bx), \btheta' \sim \hat{p}(\btheta | \bx)}\left[(k(\btheta, \btheta')\right] \\ &+ \mathbb{E}_{\btheta, \btheta' \sim (\hat{p}(\btheta | \bx)}\left[(k(\btheta, \btheta')\right],
    \end{aligned}
\end{equation}
where $k$ is a kernel. The expectations can be approximated through Monte Carlo.
  
  \chapter{A crisis in simulation-based inference?}\label{c:crisis}
  \begin{prologuebox}
This chapter is based on the following publication: \emph{Hermans, J.$^*$, Delaunoy, A.$^*$, Rozet, F., Wehenkel, A., \& Louppe, G. (2022). A crisis in simulation-based inference? beware, your posterior approximations can be unfaithful. Transactions on Machine Learning Research.}

This paper is both a position and an empirical paper. We acknowledge the fact that posterior approximations in simulation-based inference will never be perfect and discuss the impact of using those for scientific reasoning. Scientific reasoning in the tradition of Popperian falsification does not aim to verify hypotheses but to disprove false ones. Our position is that, in this setting, failing to disprove a false theory is much less detrimental than erroneously disproving a theory that might be true. By translating this to posterior approximations, an underconfident posterior approximation is much less detrimental than an overconfident approximation. While perfect approximations of posteriors are an unfeasible goal, we argue that trying to avoid overconfidence as much as possible is more realistic and is a goal that should be pursued to allow the reliable use of such approximations for scientific reasoning.

We propose to quantify the overconfidence of an approximation through a quantity that we call expected coverage. To understand how existing methods perform in that regard, we perform a large-scale empirical evaluation of the expected coverage associated with approximations from state-of-the-art methods on a wide variety of benchmarks. The first version of this paper has been published in 2021. Consequently, only methods from the ABC, NLE, NRE, and NPE families are considered. Score-based methods and other types of generative models were introduced later in the context of simulation-based inference. Similarly, only classical sequential methods are considered, as truncated methods were introduced later on.
\end{prologuebox}

\section{Introduction}

Many scientific disciplines rely on computer simulations to study complex phenomena under various conditions. 
Although modern simulators can generate realistic synthetic observables through detailed descriptions of their data generating processes, they make statistical inference more challenging.
The computer code describing the data generating processes defines the likelihood function $p(\bx|\btheta)$ only implicitly, and its direct evaluation requires the often intractable integration of all stochastic execution paths.
In this problem setting, statistical inference based on the likelihood becomes impractical. 
However, approximate inference remains possible by relying on likelihood-free approximations thanks to the increasingly accessible and effective suite of methods and software from the field of simulation-based inference \citep{cranmer2020frontier}.

While simulation-based inference targets domain sciences, advances in the field are mainly driven from a machine learning perspective.
The field, therefore, inherits the quality assessments \citep{lueckmann2021benchmarking} customary to the machine learning literature, primarily targeting the exactness of the approximation.
In fact, domain sciences, and more specifically the physical sciences, are not necessarily interested in the exactness of an approximation. 
Instead, in the tradition of Popperian falsification, they often seek to constrain parameters of interest as much as possible at a given confidence level. 
Scientific examples include frequentist confidence intervals on the mass of the Higgs boson \citep{aad2012observation}, Bayesian credible regions on cosmological parameters \citep{gilman2018probing,Planck:2018vyg}, constraints on the intrinsic parameters of binary black hole coalescences \citep{abbott2016gw151226}, or characterizing the space of circuit configurations giving rise to rhythmic activity in the crustacean stomatogastric ganglion \citep{gonccalves2020training}.
Wrongly excluding plausible values could drive the scientific inquiry in the wrong direction, whereas failing to exclude implausible values because of too conservative estimations is much less detrimental. This implies that statistical approximations in simulation-based inference should ideally come with conservative guarantees to not produce credible regions smaller than they should be, even when the approximations are not faithful.
Despite recent developments of post hoc diagnostics to inspect the quality of likelihood-free \citep{cranmer2015approximating, Brehmer:2018eca, brehmer2019mining, Hermans:2020skz, lueckmann2021benchmarking, sbc, dalmasso2020confidence} and likelihood-based \citep{geweke1991evaluating,gelman1992inference,raftery1991many,dixit2017mcmc,sbc} approximations, assessing whether approximate inference results are sufficiently reliable for scientific inquiry remains largely unanswered whenever fitting criteria are not globally optimized or whenever the data is limited.

In this work, we measure and discuss the quality of the credible regions produced by various algorithms for Bayesian simulation-based inference.
We frame our main contribution as the collection of extensive empirical evidence requiring months of computation, demonstrating that all benchmarked techniques may produce non-conservative credible regions. Our results emphasize the need for a new class of methods: conservative approximate inference algorithms. In addition, we provide empirical evidence that using an ensemble of models in place of a single model tends to produce more reliable approximations.
The structure of the paper is outlined as follows.
Section \ref{sec:crisis:background} describes the statistical formalism, necessary background and includes a thorough motivation for coverage.
Section \ref{sec:crisis:experiments} highlights our main results.
Section \ref{sec:crisis:discussion} presents several avenues of future research to enable drawing reliable scientific conclusions with simulation-based inference. All code related to this manuscript is available at
\href{https://github.com/montefiore-ai/trust-crisis-in-simulation-based-inference}{\texttt{https://github.com/montefiore-ai/trust-crisis-in-simulation-based-inference}}.

\section{Background}
\label{sec:crisis:background}

\subsection{Statistical formalism}
\label{sec:statistical_formalism}
We evaluate posterior estimators that produce approximations $\hat{p}(\btheta|\bx)$ with the following semantics.

{\textbf{Target parameters} $\btheta$ denote the parameters of interest of a simulation model, and are sometimes referred to as free or model parameters. We make the reasonable assumption that the prior $p(\btheta)$ is tractable.

An \textbf{observable} $\bx$ denotes a synthetic realization of the simulator, or the observed data $\bx_o$ we would like to do inference on. We assume that the simulation model is correctly specified and hence is an accurate representation of the real data generation process.

The \textbf{likelihood} model $p(\bx|\btheta)$ is implicitly defined by the simulator's computer code. While we cannot evaluate the density $p(\bx|\btheta)$, we can draw samples through simulation.

The \textbf{ground truth} $\btheta^*$ specified to the simulation model whose forward evaluation produced the observable $\bx_o$, i.e., $\bx_o\sim p(\bx|\btheta=\btheta^*)$.

A \textbf{credible region} is a space $\Theta$ within the target parameters domain that satisfies
\begin{equation}
    \int_\Theta p(\btheta|\bx=\bx_o) d{\btheta} = 1 - \alpha
\end{equation}
for some observable $\bx_o$ and confidence level $1 - \alpha$.
Because many such regions exist, we compute the credible region with the smallest volume. In the literature this credible region is known as 
the highest posterior density region \citep{box2011bayesian,hyndman1996computing}. In our evaluations,
we determine the credible regions by evaluating the approximated posterior density function in a discretized and empirically normalized grid of the parameter space. The credible region is the set of parameters whose density is greater than a given threshold fitted to achieve the desired credibility level $1 - \alpha$.

\subsection{Statistical quality assessment}

\begin{figure}
    \centering
    \includegraphics[width=0.4\textwidth]{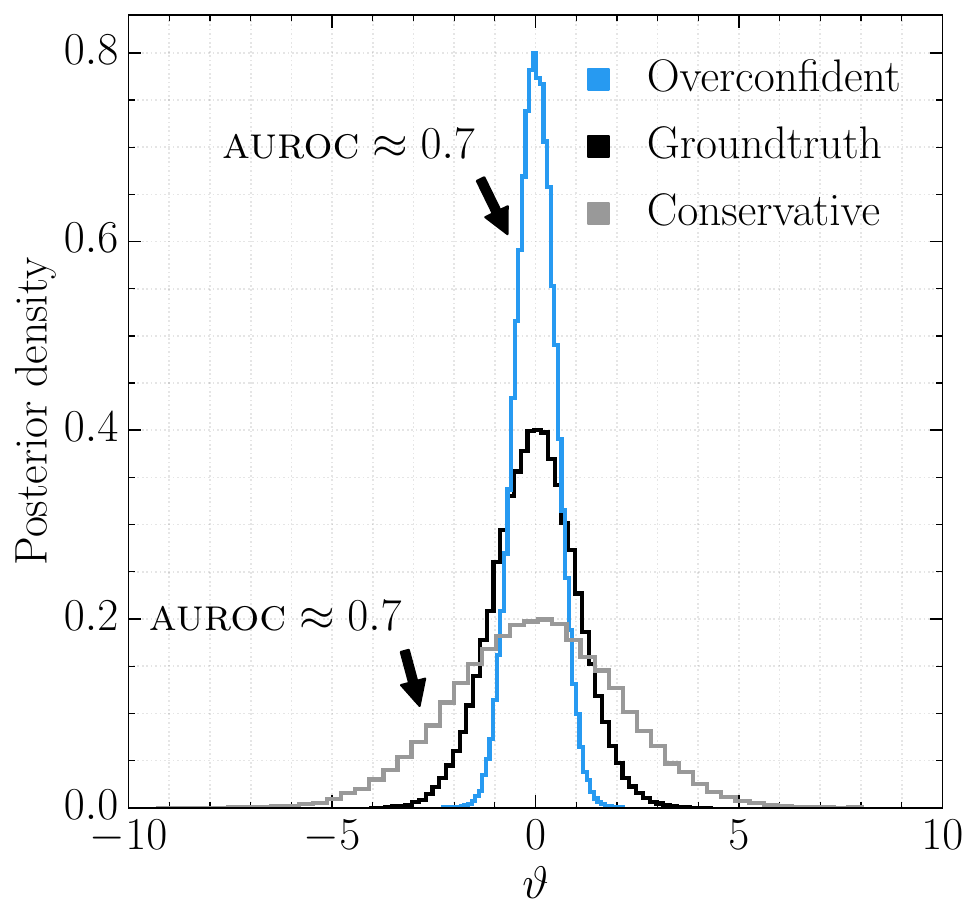}
    \caption{A classifier-based metric measures the divergence between posterior approximations and a ground truth by means of evaluating the classifier's discriminative performance through Area Under the Receiver Operating Characteristics curve (\textsc{auroc}). The metric indicates that both the overconfident and the conservative approximations are equally accurate as it yields \textsc{auroc} = 0.7 for both of them. From an inference perspective however, the conservative approximation is more suitable because it produces credible regions larger than they should be.}
    \label{fig:auc_deceitful}
\end{figure}

Common metrics for evaluating the quality of a posterior surrogate include the Classifier Two-sample Test \citep{lehmann2006testing,lopez2017revisiting} and Maximum Mean Discrepancy \citep{gretton2012kernel,bengio2014bounding,dziugaite2015training}. The main problem with these metrics is that they assess exactness of an approximation through a divergence with respect to the posterior. All approximations will diverge from the posterior and there are no criteria to what constitutes an acceptable estimator. 
For this reason, we argue that metrics evaluating the reliability for scientific inquiry should be used alongside the divergence evaluation when evaluating estimators.

To clarify this point, consider the demonstration in Figure \ref{fig:auc_deceitful}.
A binary classifier is trained to discriminate between samples from a posterior approximation and the true posterior, as in a classifier two-sample test.
The discriminative performance of the classifier is expressed through Area Under the Receiver Operating Characteristics curve (\textsc{auroc}) and serves as a measure for divergence between both densities. An \textsc{auroc} = 0.5 suggests an approximation that is indistinguishable from the true posterior, while \textsc{auroc} = 1.0 implies that both distributions do not overlap.
Although both the overconfident and the conservative approximations are of equal quality according to the \textsc{auroc} metric (\textsc{auroc} = 0.7), credible regions that are biased or smaller than they should be could result in the wrong exclusion of actually plausible parameter values for a given significance level, and hence to erroneous scientific conclusions. By contrast, conservative approximations, which leads to credible regions that are larger than they should be, are more scientifically reliable since they would not wrongly reject plausible parameters values but only fail to reject actually implausible parameter values. 
For this reason, we take the position that posterior approximations should produce overdispersed credible regions for any simulation budget. Posterior approximations do not have to closely match the true posterior to draw meaningful inferences, but they should however be conservative.

Instead of measuring the exactness of an approximation, this work directly assesses the quality of credible regions through the notion of expected coverage which probes the consistency of the posterior approximations and can be used to diagnose conservative and overconfident approximations. Similar usages of coverage diagnostics can be found in the context of standard statistical inference \citep{schall2012empirical} and approximate Bayesian computation \citep{prangle2014diagnostic}.

\begin{definition}
The {\bfseries expected coverage probability} of
the $1 - \alpha$ highest posterior density regions derived from the posterior estimator $\hat{p}(\btheta|\bx)$ is
\begin{equation}
    \label{eq:crisis:coverage}
    \E_{p(\btheta,\bx)}\left[\mathds{1}\left(\btheta \in \Theta_{\hat{p}(\btheta|\bx)}(1 - \alpha)\right)\right],
\end{equation}
where the function
$\Theta_{\hat{p}(\btheta|\bx)}(1 - \alpha)$ yields the $1 - \alpha$ highest posterior density region
of $\hat{p}(\btheta|\bx)$.
\end{definition}
Note that Equation \ref{eq:crisis:coverage} can be expressed either as
\begin{equation}
    \E_{p(\btheta)}\E_{p(\bx|\btheta)}\left[\mathds{1}\left(\btheta \in \Theta_{\hat{p}(\btheta|\bx)}(1 - \alpha)\right)\right],
\end{equation}
which is the expected frequentist coverage probability, or alternatively as the
expected Bayesian credibility
\begin{equation}
    \E_{p(\bx)}\E_{p(\btheta|\bx)}\left[\mathds{1}\left(\btheta \in \Theta_{\hat{p}(\btheta|\bx)}(1 - \alpha)\right)\right],
\end{equation}
whose inner expectation reduces to $1 - \alpha$ whenever the posterior estimator $\hat{p}(\btheta|\bx)$ is well-calibrated.

The expected coverage probability can be estimated empirically given a set of $n$ i.i.d. samples $(\btheta^*_i,\bx_i)\sim p(\btheta,\bx)$ as
\begin{equation}
\frac{1}{n} \sum_{i=1}^n \mathds{1}\left(\btheta^*_i \in \Theta_{\hat{p}(\btheta|\bx_i)}(1 - \alpha)\right).
\end{equation}

\begin{definition}
\label{def:threshold}
A {\bfseries conservative posterior estimator} is an estimator that has coverage at the credibility level of interest, i.e., whenever the expected coverage probability is larger or equal to the credibility level.
\end{definition}
While coverage is necessary to assess conservativeness, it is limited in its ability to determine the information gain a posterior (approximation) has over its prior. Consider an estimator whose posteriors are identical to the prior. In this case, there is no gain in information and the expected coverage probability is equal to the credibility level. For this reason, a complete analysis should be complemented with measures such as the expected information gain $\E_{p(\btheta,\bx)}\left[\log p(\btheta|\bx) - \log p(\btheta)\right]$ or classifier two-sample tests when the ground-truth posterior is available. 
This work is concerned with conservative inference and will, therefore, mainly focus on the evaluation of expected coverage. Finally, it should be noted that expected coverage is a statement about the credible regions in expectation and, therefore, does not provide any guarantee on the quality of a single posterior approximation. However, the quality of a single posterior approximation can itself be approximated with a local coverage test \citep{zhao2021diagnostics}.

\section{Empirical observations}
\label{sec:crisis:experiments}
This section covers our main contribution: the collection of empirical evidence to determine whether some simulation-based inference algorithms are conservative by nature. 
We are particularly interested in determining whether certain approaches should be
favored over others. We do so by estimating the expected coverage of posterior estimators produced by these approaches across a broad range of hyperparameters and benchmarks of varying complexity, including two real problems.
As in real use cases, the true posteriors are effectively intractable and, therefore, unknown.

\subsection{Methods}
\label{sec:methods}
We make the distinction between two paradigms.
\textit{Non-amortized} approaches are designed to approximate a single posterior, while \textit{amortized} methods aim to learn a general-purpose estimator that attempts to approximate all posteriors supported by the prior.
\subsubsection{Amortized}
\textbf{Neural Ratio Estimation} (NRE) is an established approach in the simulation-based inference literature both from a frequentist \citep{cranmer2015approximating} and Bayesian \citep{thomas2016likelihood,2019arXiv190304057H,durkan2020contrastive} perspective. In a Bayesian analysis, 
an amortized estimator $\hat{r}(\bx|\btheta)$
of the intractable likelihood-to-evidence ratio $r(\bx|\btheta)$ can be learned by training a binary classifier $\hat{d}(\btheta,\bx)$ to distinguish between samples of the
joint $p(\btheta,\bx)$ with class label $1$ and samples of the product of marginals $p(\btheta)p(\bx)$ with class label $0$, with equal label marginal probability. Similar to the density-ratio trick \citep{sugiyama2012density,goodfellow2014generative,cranmer2015approximating,2019arXiv190304057H}, the Bayes optimal classifier $d(\btheta,\bx)$ for the cross-entropy loss is
\begin{align}
  d(\btheta,\bx) = \frac{p(\btheta, \bx)}{p(\btheta,\bx) + p(\btheta)p(\bx)} = \sigmoid\left(\log\frac{p(\btheta,\bx)}{p(\btheta)p(\bx)}\right),
\end{align}
where $\sigmoid(\cdot)$ is the sigmoid function.
Given a target parameter $\btheta$ and an observable $\bx$ supported by $p(\btheta)$ and $p(\bx)$ respectively,
the learned classifier $\hat{d}(\btheta,\bx)$ approximates the log likelihood-to-evidence ratio $\log r(\bx|\btheta)$ through the logit function because
$\sigmoid^{-1}(\hat{d}(\btheta,\bx))\approx\log r(\bx|\btheta)$.
The approximate log posterior density function is $\log p(\btheta) + \log \hat{r}(\bx|\btheta)$.

\textbf{Neural Posterior Estimation} (NPE) \citep{rezende2015variational} is concerned with directly learning an amortized posterior estimator $\hat{p}_\psi(\btheta|\bx)$ with normalizing flows. Normalizing flows define a class of probability distributions $p_\psi(\cdot)$ built from neural network-based bijective transformations \citep{rezende2015variational, dinh2015nice, dinh2017density} parameterized by $\psi$. They are usually optimized using variational inference, by solving
$\argmin_\psi \mathds{E}_{p(\bx)} \left[\textsc{kl}(p(\btheta|\bx)~||~\hat{p}_\psi(\btheta|\bx)\right]$,
which is equivalent to $\argmax_\psi \E_{p(\btheta,\bx)}\left[\log \hat{p}_\psi(\btheta|\bx)\right]$.
Once trained, the density of the modeled distribution can directly be evaluated and sampled from. 

\textbf{Ensembles} of models constitute a standard method to improve predictive performance. In this work, we consider an ensemble model that averages the approximated posteriors of $n$ posterior estimators that are either trained independently on the same dataset (deep ensemble) \citep{lakshminarayanan2017simple} or on bootstrap replicates of the learning set (bagging) \citep{breiman1996bagging}. While this formulation is natural for NPE, averaging likelihood-to-evidence ratios is equivalent since
$\frac{1}{n}\sum_{i=1}^n \hat{p}_i(\btheta | \bx) = p(\btheta)\frac{1}{n} \sum_{i=1}^n \hat{r}_i(\bx | \btheta)$. We show in Section \ref{sec:crisis:experiments} that ensembles lead in average to a higher expected coverage than individual models and hence constitute a straightforward mitigation strategy against overconfidence.

\subsubsection{Non-amortized}
\textbf{Rejection Approximate Bayesian Computation} Rej-ABC \citep{rubin1984,pritchard1999population}
numerically estimates a single posterior by collecting samples $\btheta\sim p(\btheta)$ whenever
$\bx\sim p(\bx|\btheta)$ is similar to $\bx_o$. Similarity is expressed by means of a distance function $\rho$. For high-dimensional observables, the probability density of simulating an observable $\bx$ such that $\bx = \bx_o$ is extremely small. For this reason, ABC uses a summary statistic $s$ and an acceptance threshold $\epsilon$. Using these components, ABC accepts samples into the approximate posterior whenever $\rho(s(\bx),s(\bx_o))\leq\epsilon$. Improvements include regression adjustment \citep{beaumont2002approximate} of the sampled parameters using local linear regression, combining ABC with MCMC \citep{marjoram2003markov} and the automatic construction of summary statistics \citep{fearnhead2012constructing}. In our experiments, we apply Rej-ABC in its simplest form and use the identity function as a sufficient summary statistic, use no regression adjustment and set $\epsilon$ such that $\max(100, \text{simulation budget} / 100)$ samples are accepted.

Sequential methods for simulation-based inference aim to approximate a single posterior by iteratively improving a posterior approximation. These methods alternate between a simulation and an exploitation phase. The latter is designed to take current knowledge into account such that subsequent simulations can be focused on parameters that are more likely to produce observables $\bx$ similar to $\bx_o$.

{\bfseries Sequential Monte-Carlo ABC} (SMC-ABC) \citep{10.1093/bioinformatics/btp619, sisson2007sequential, beaumont2009adaptive}
iteratively updates a set of proposal states to match the posterior distribution. At each iteration, accepted proposals are ranked by distance. The rankings determine whether a proposal is propagated to the next iteration. New candidates are generated by perturbing the selected ranked proposals.

{\bfseries Sequential Neural Posterior Estimation} (SNPE) \citep{papamakarios2016fast, lueckmann2017flexible, greenberg2019automatic} iteratively improves a normalizing flow that models the posterior.
Our evaluations will specifically use the SNPE-C \citep{greenberg2019automatic} variant.

{\bfseries Sequential Neural Likelihood} (SNL) \citep{papamakarios2019sequential} models the likelihood $p(\bx|\btheta)$. A numerical approximation of the posterior is obtained by plugging the learned likelihood estimator into a Markov Chain Monte Carlo (MCMC) sampler as a surrogate likelihood. Similarly, \cite{price2018bayesian, frazier2022bayesian} construct a synthetic normal approximation of the likelihood over summary statistics.

{\bfseries Sequential Neural Ratio Estimation} (SNRE) \citep{2019arXiv190304057H,durkan2020contrastive} iteratively improves the modelled likelihood-to-evidence ratio.

\subsection{Benchmarks}

We consider 7 benchmarks, ranging from a toy problem to real scientific use cases covering various disciplines. All benchmarks and priors are available in the codebase.

The \textbf{SLCP} simulator models a fictive problem with 5 parameters. The observable $\bx \in \mathbb{R}^8$ represents the 2D-coordinates of 4 points. 
The coordinate of each point is sampled from the same multivariate Gaussian whose mean and covariance matrix are parametrized by $\btheta$. We consider an alternative version of the original task \citep{papamakarios2019sequential} by inferring the marginal posterior density of 2 of those parameters. In contrast to its original formulation, the likelihood is not tractable due to the marginalization.

The \textbf{Weinberg} problem \citep{weinberg} concerns a simulation of high energy particle collisions $e^+e^- \to \mu^+ \mu^-$. The angular distribution of the particles can be used to measure the Weinberg angle $\bx$
in the standard model of particle physics. From the scattering angle, we are interested in inferring Fermi's constant $\btheta$.

The \textbf{Spatial SIR} model involves a grid-world of susceptible,
infected, and recovered individuals. Based on initial conditions and the infection and recovery rate $\btheta$,
the model describes the spatial evolution of an infection.
The observable $\bx$ is a snapshot of the grid-world after some fixed amount of time. 

\textbf{M/G/1} \citep{blum2010non} models a processing and arrival queue. The problem is
described by 3 parameters $\btheta$ that influence the time it takes to serve a customer, and the time between their arrivals. The observable $\bx$ is composed of 5 equally spaced quantiles of inter-departure times.

The \textbf{Lotka-Volterra} population model \citep{lotka,volterra1926fluctuations} describes a process of interactions between a predator and a prey species. The model is conditioned on 4 parameters $\btheta$ which influence the reproduction and mortality rate of the predators and preys. We infer the marginal posterior of the predator parameters from time series representing the evolution of both populations over time.

Stellar \textbf{Streams} form due to the disruption of spherically packed clusters of stars by the Milky Way. Because of their distance from the galactic center and other visible matter, distant stellar streams are considered to be ideal probes to detect gravitational interactions with dark matter. The model \citep{banik2018probing} evolves the stellar density $\bx$ of a stream over several billion years and perturbs the stream over its evolution through gravitational interactions with dark matter subhaloes parameterized by the dark matter mass $\btheta$. 

\textbf{Gravitational Waves (GW)} are ripples in space-time emitted during events such as the collision of two black-holes. They can be detected through interferometry measurements $\bx$ and convey information about celestial bodies, unlocking new ways to study the universe. We consider inferring the masses $\btheta$ of two black-holes colliding through the observation of the gravitational wave as measured by \textsc{ligo}'s dual detectors \citep{lalsuite,Biwer:2018osg}.

\subsection{Setup}
Our evaluations consider simulation budgets ranging from $2^{10}$ up to $2^{17}$ samples and confidence levels from 0.05 up to 0.95. Within the amortized setting
we train, for every simulation budget, 5 posterior estimators for 100 epochs. The expected coverage probability is estimated on at least 5,000 unseen samples from the joint $p(\btheta,\bx)$ and for all confidence levels under consideration.
In addition, we repeat the expected coverage evaluation for ensembles of 5 estimators as well.
Special care for non-amortized approaches is necessary because they approximate a single posterior and can therefore not reasonably evaluate expected coverage in the same way.
Our experiments for non-amortized approaches estimate the expected coverage by repeating the inference procedure on 300 distinct observables for every simulation budget.
The expected coverage probabilities are subsequently estimated based on the resulting posterior approximations.
Our experiments with NPE, SNPE, SNL, SNRE, Rej-ABC and SMC-ABC rely on the implementation in the SBI package \citep{sbi}, while a custom implementation for NRE is used.

We emphasize the computational requirements necessary to generate our main contribution: the experimental observations, whose generation took months of computation. The average CPU time for evaluating an amortized procedure on all common benchmark problems is in order of 200 CPU days, while for a non-amortized approach this increases to 2800 CPU days. The bulk of the cost was associated with the repeated optimization procedure and execution of the simulator. While we considered benchmarks with low dimensional parameter spaces (up to 3 parameters), we stress that the cost for computing credible regions would grow exponentially with the number of parameters, making coverage diagnostics expensive for high-dimensional problems. Note that when using NPE, the procedure for computing the coverages could be modified to scale to higher dimensions \citep{rozet2021arbitrary, rozet2021arbitrarythesis}. However, this procedure cannot be applied to all simulation-based inference algorithms. Nevertheless, we expect the observed behaviours to be similar or accentuated in a high-dimensional setting.

\subsection{Results}

\begin{figure}
    \includegraphics[width=\textwidth]{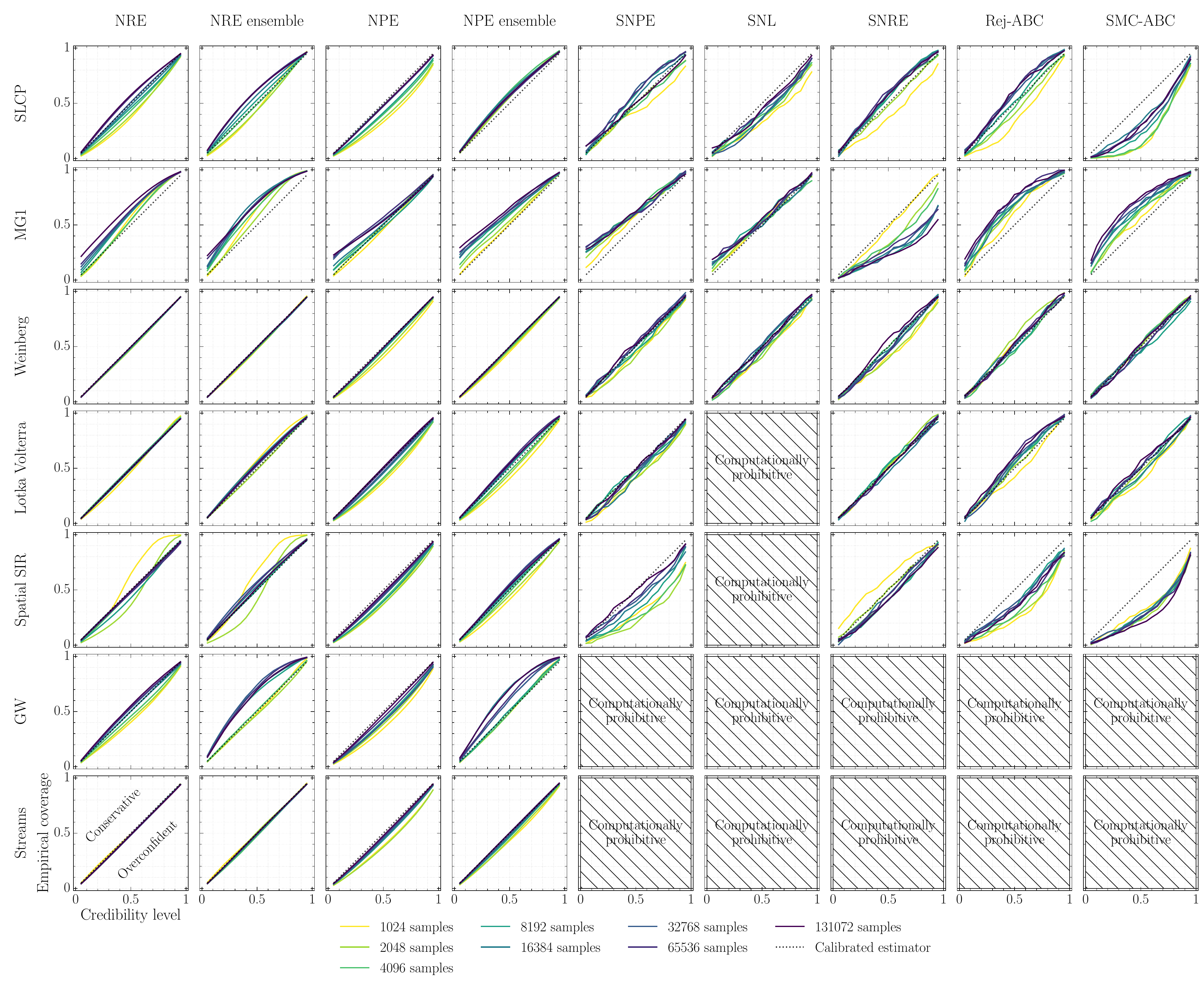}
    \caption{Evolution of the expected coverage w.r.t the simulation budget. A perfectly calibrated posterior has an expected coverage probability equal to the credibility level and produces a diagonal line. Conservative estimators on the other hand produce curves above the diagonal and overconfident models underneath. All algorithms can lead to non-conservative estimators. This pathology tends to be accentuated for small simulation budgets and non-amortized methods.
    Finally, the computationally prohibitive results indicate that the computational requirements did not allow for a coverage analysis. In the case of SNL, this was mostly due to high dimensional observables. 
    For the astronomy benchmarks, the simulation model was simply too expensive to reasonably evaluate coverage for non-amortized methods.
    }
    \label{fig:all_coverage_levels}
\end{figure}
\begin{figure}
    \centering
    \includegraphics[width=\linewidth]{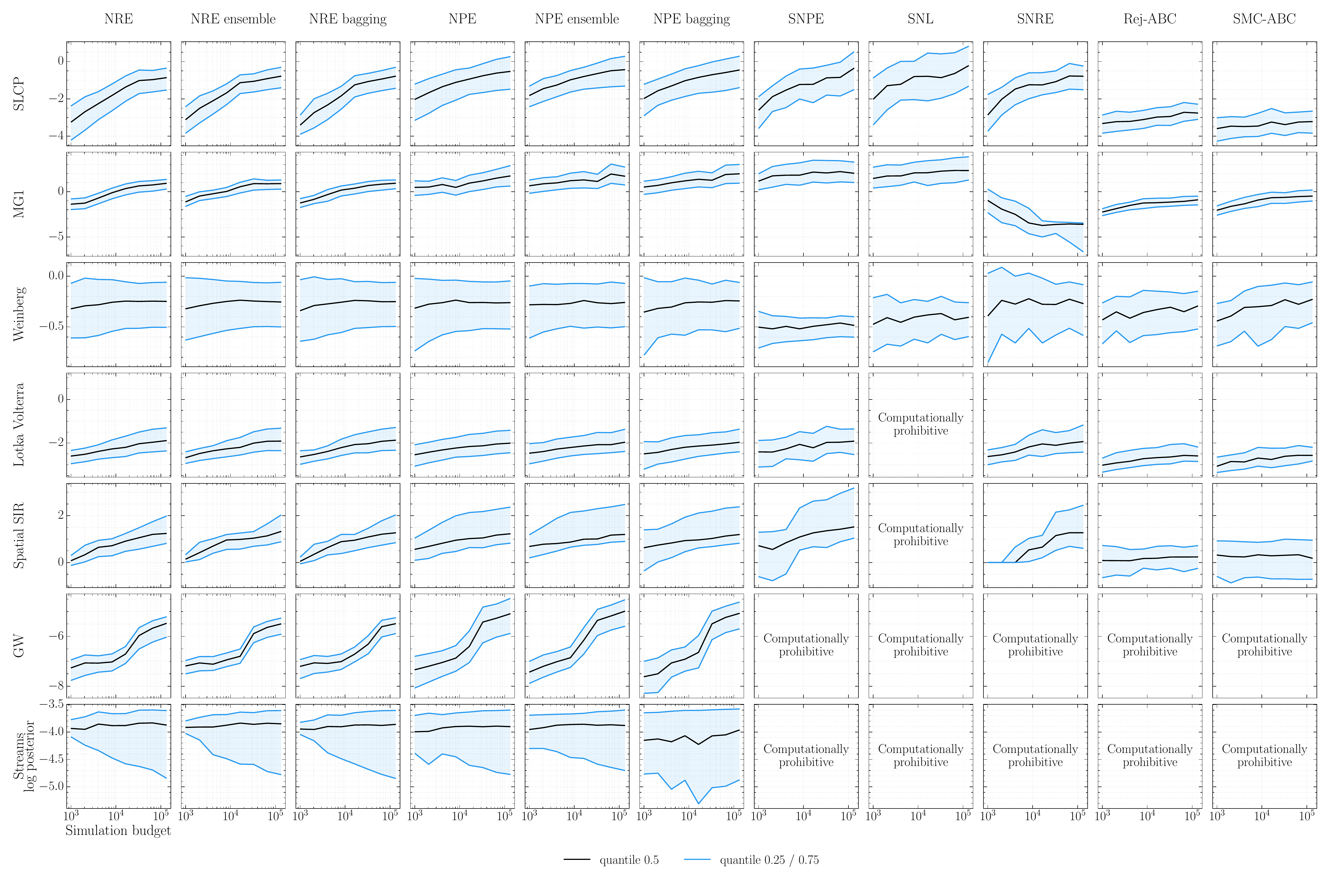}
    \caption{Evolution of the predicted log posterior probability of the nominal parameters with respect to the simulation budget. We evaluate the log posterior probability on $100,000$ test samples for amortized algorithms and $300$ for non-amortized algorithms. We report the $0.25; 0.5; 0.75$ quantiles of this quantity over the test set. We observe that the predictive performance increases in general with the simulation budget. Exceptions are the MG1 benchmark with the SNRE algorithm and the lowest quantile of the streams benchmark.}
    \label{fig:performance}
\end{figure}
\begin{figure}
    \includegraphics[width=\textwidth]{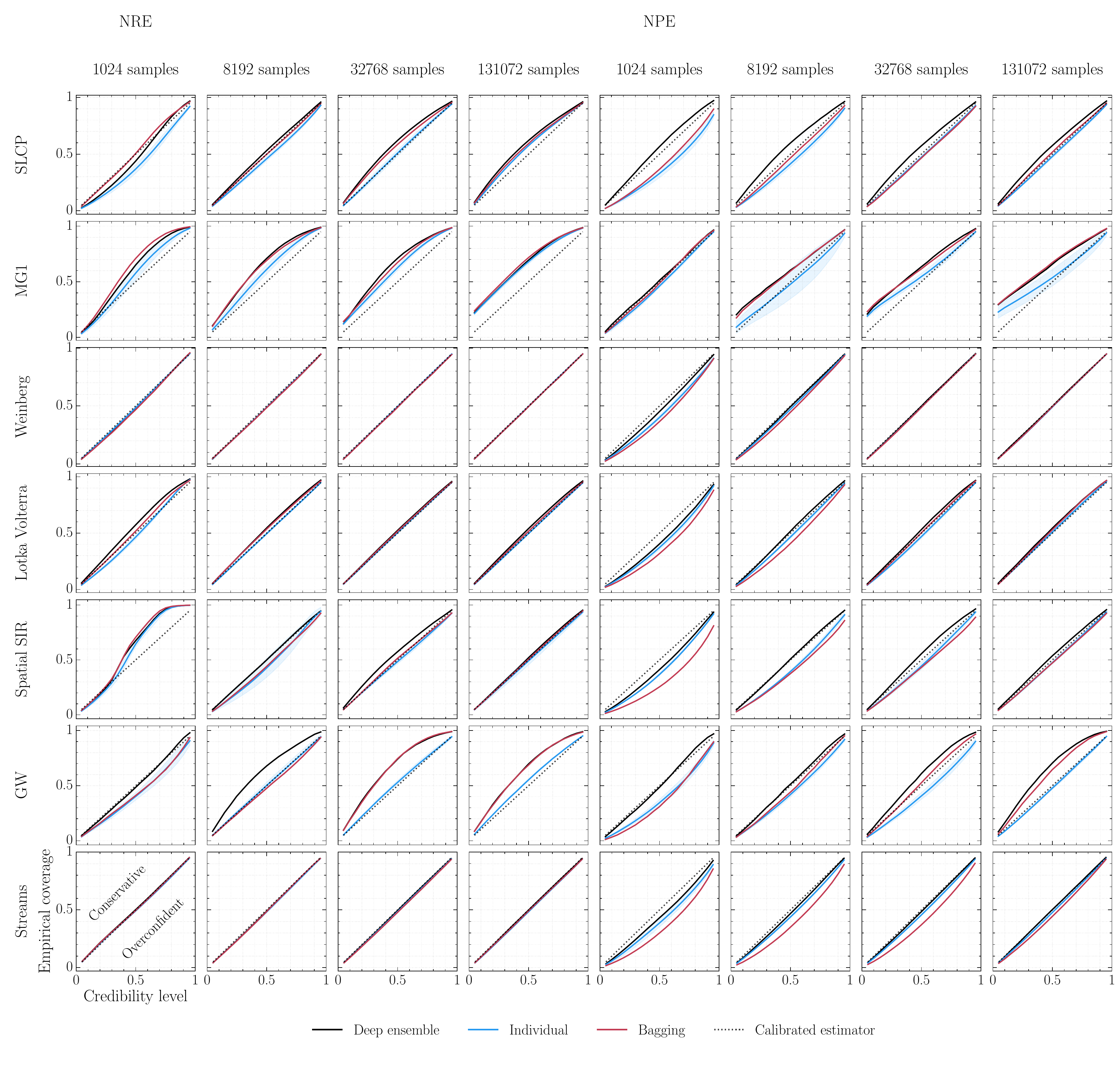}
    \caption{Analysis of coverage between ensemble and individual models w.r.t the various simulation budgets. The blue line represents the mean expected coverage of individual models over $5$ runs, the shaded area represents its standard deviation. The black line represents the expected coverage of a single ensemble composed of 5 models. We observe that deep ensembles consistently have a higher expected coverage probability compared to the average individual model. A similar effect is not always observed with bagging, indicated by the red line. Ensembles are only evaluated for amortized approaches such as NPE and NRE.}
    \label{fig:coverage_ensemble}
\end{figure}

Figures \ref{fig:all_coverage_levels} and \ref{fig:coverage_ensemble} highlight our main results. Through these plots, we can directly assess whether a posterior estimator is conservative at a given confidence level and simulation budget. The figures should be interpreted as follows: a perfectly calibrated posterior has an expected coverage probability equal to the credibility level. Plotting this relation produces a diagonal line. Conservative estimators on the other hand produce curves above the diagonal and overconfident models underneath.
The plots highlight an unsettling observation: all benchmarked approaches produce non-conservative posterior approximations on at least one problem setting. In general, this pathology is especially prominent in non-amortized approaches with a small simulation budget; a regime they have been specifically designed for. A large simulation budget does not guarantee that a posterior estimator is conservative either. More importantly, as the reliability of the approximations computed through sequential approaches cannot practically be determined, it leaves practitioners uncertain about the reliability of their approximations. To complement this analysis, an evaluation of the predictive performance of the approximate posteriors is provided in Figure \ref{fig:performance}. We observe that, on most benchmark/method pairs, the predictive performance increases with the simulation budget. However, this is not the case for SNRE applied to MG1 which explains the strange behaviour of the coverage curves.

In sequential approaches, overconfidence could be explained by the alternating simulation and exploitation phases. One potential failure mode is that a non-conservative posterior approximation at a previous iteration forces the subsequent simulation phases to not produce observables that should in fact be associated with a higher posterior density, causing the posterior estimator to increase its overconfidence at each iteration.

Despite the fact that all ABC approaches use a sufficient summary statistic (by definition, the identity function), our results demonstrate that this alone is no guarantee for conservative posterior approximations. In fact, using a sufficient summary statistic with $\epsilon > 0$ does not always correspond to conservative approximations either.
In such cases, ABC accepts
samples with larger distances, permitting the procedure to shift the mass of the approximated posterior elsewhere. In addition, a limited number of posterior samples can negatively affect the quality of the credible regions, e.g., when approximating the posterior density function with kernel density estimation.
Both cases can cause the observed behaviour.
ABC should therefore be applied with caution to scientific applications. Even though a handcrafted, albeit sufficient, summary statistic provides some insight into the approximated posterior, it does not imply that ABC approximations are conservative whenever the threshold $\epsilon > 0$.

In Figure \ref{fig:coverage_ensemble}, we observe that the expected coverage probability of deep ensemble models is consistently larger than the expected coverage probability of an individual posterior estimator. This highlights the fact that ensembling constitutes an immediately applicable and easy way to mitigate the overconfidence issue and build more reliable posterior estimators. However, the deep ensemble model can still be non-conservative. We hypothesize that the increase in coverage is linked to the added uncertainty captured by the ensemble model, leading to inflated credible regions. 
In fact, individual estimators only capture data uncertainty, while an ensemble is expected to partially capture the epistemic uncertainty as well.
Surprisingly, we find that ensembles built using bagging do not always produce higher coverage than individual models while they should also capture part of the epistemic uncertainty. We could potentially attribute this behaviour to the reduced effective dataset size used to train each member of the ensemble. Figures \ref{fig:coverage_ensemble_size} and \ref{fig:coverage_ensemble_size_bagging} show a positive effect with respect to ensemble size.

Not evident from the figures \ref{fig:all_coverage_levels} and \ref{fig:coverage_ensemble} are the computational consequences of a coverage analysis on non-amortized methods.
Although the figures mention a certain simulation budget, the total number of simulations for non-amortized methods should be multiplied by the number of approximated posteriors ($300$) to estimate the coverage. This highlights the simulation cost associated with diagnosing non-amortized approaches.
This issue is not limited to coverage. Simulation-Based Calibration (SBC) \citep{sbc} relies on samples of arbitrary posterior approximations as well.
Diagnosing non-amortized estimators with SBC therefore 
requires a similar approach as we have taken in our coverage analyses.
In fact, other benchmark papers such as \citet{lueckmann2021benchmarking} restrain from evaluating SBC
because the diagnostic is computationally prohibitive for non-amortized approaches.

Our results illustrate a clear distinction between the amortized and non-amortized paradigms. Amortized methods do not require retraining or new simulations to determine the empirical expected coverage probability of a posterior estimator, while non-amortized methods do.
For this reason, a global coverage analysis of non-amortized approaches is computationally prohibitive and mostly impractical. 
More importantly, the coverage analysis of a non-amortized approach only measures the quality of the training procedure, whereas a coverage analysis of an amortized approach diagnoses the posterior estimator itself. 
In addition, a global coverage analysis not only serves as diagnostic but also allows to partially alleviate the issue by performing post-training calibration. A simple way for calibrating level $\alpha$ credible regions is to replace those by credible regions at a level that has the desired expected coverage. 
Finally, non-amortized sequential algorithms have to repeat the entire simulation-training pipeline whenever architectural or hyperparameter changes are made, while amortized methods reuse previously simulated datasets. 
All of the above lead us to conclude that while sequential methods have the benefit of being faster to train, amortized methods should be considered for sensitive applications requiring detailed statistical validation.

\protect\begin{custombox}

{\bfseries Observation 1} All benchmarked algorithms may produce non-conservative posterior approximations. This pathology tends to be accentuated with small simulation budgets in both paradigms.

\medskip

{\bfseries Observation 2} Amortized approaches tend to be more conservative in contrast to non-amortized approaches. 

\smallskip

{\bfseries Observation 3} The expected coverage probability of an ensemble model is larger than the average individual model. The ensemble size positively affects the expected coverage probability as well.

\smallskip

{\bfseries Observation 4} Amortized methods are simulation-efficient, especially when taking hyper-parameter tuning and the evaluation of the expected coverage diagnostic into account.
\end{custombox}

\begin{figure}
    \centering
    \includegraphics[width=\linewidth]{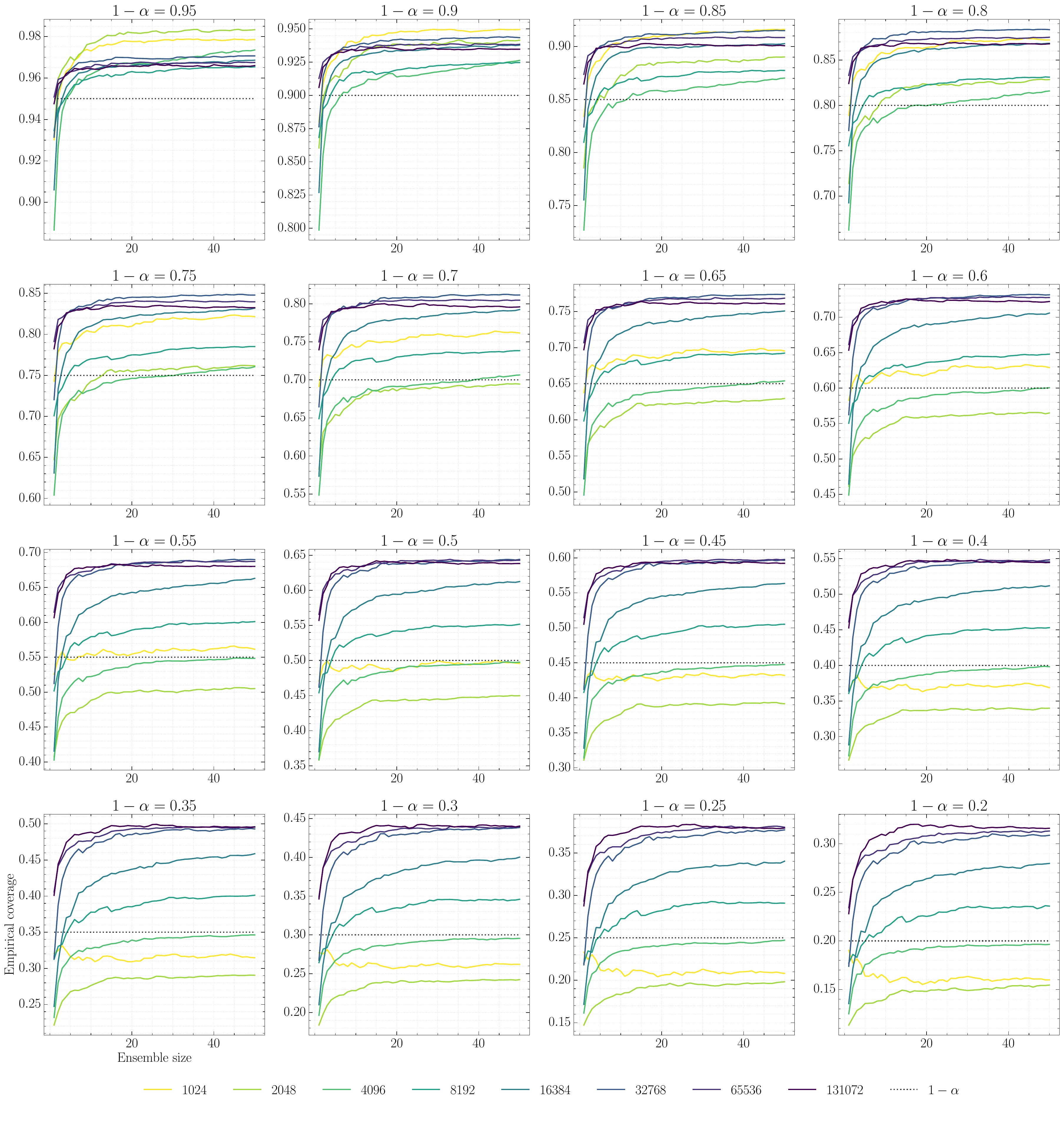}
    \caption{Evolution of the empirical expected coverage of deep ensembles with respect to ensemble size for various confidence levels. The results are obtained by training ratio estimators (\textsc{nre}) on the \textsc{slcp} benchmark. A positive effect is observed in terms of empirical expected coverage and ensemble size, i.e., a larger ensemble size correlates with a larger empirical expected coverage. This is unsurprising, because a larger ensemble is expected to capture more of the uncertainty that stems from the training procedure.}
    \label{fig:coverage_ensemble_size}
\end{figure}

\begin{figure}
    \centering
    \includegraphics[width=\linewidth]{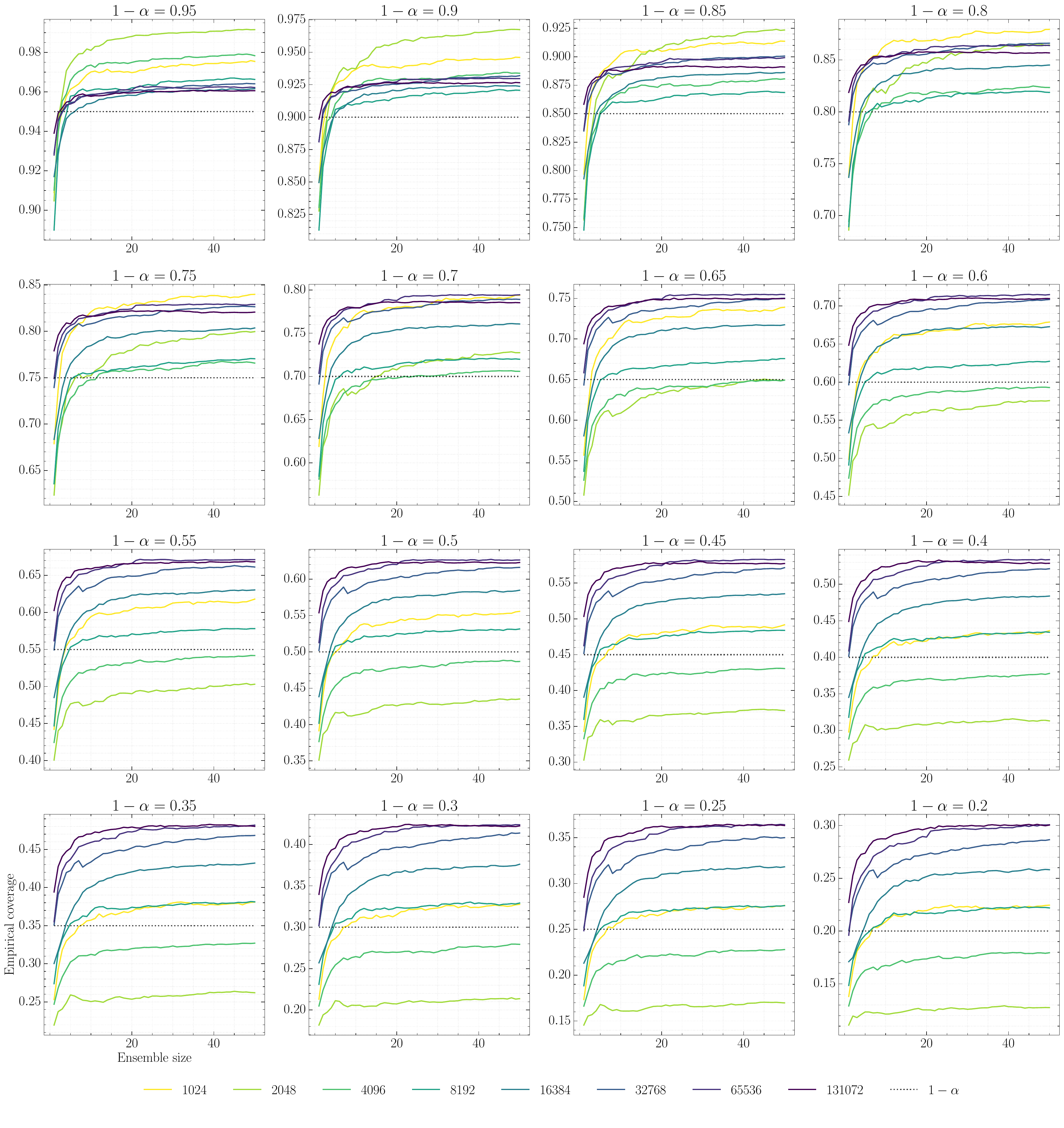}
    \caption{Evolution of the empirical expected coverage of bagging ensembles with respect to ensemble size for various confidence levels. The results are obtained by training ratio estimators (\textsc{nre}) on the \textsc{slcp} benchmark. A positive effect is observed in terms of empirical expected coverage and ensemble size, i.e., a larger ensemble size correlates with a larger empirical expected coverage. This is unsurprising, because a larger ensemble is expected to capture more of the uncertainty that stems from the training procedure.}
    \label{fig:coverage_ensemble_size_bagging}
\end{figure}

\section{Discussion}
\label{sec:crisis:discussion}

As demonstrated empirically, simulation-based inference can be unreliable, especially whenever its approximations cannot be diagnosed. 
The problem of determining whether a posterior approximation is correct, or rather, suitable to a use case, is in fact not restricted to simulation-based inference specifically; the concern occurs in all of approximate Bayesian inference. 
The MCMC literature deals with this exact same problem in the form of determining whether a set of Markov chain samples have converged to the target distribution \citep{lin2014integrated,hogg2018data}. 
In this regard, empirical diagnostic tools have been proposed over the years \citep{geweke1991evaluating,gelman1992inference,raftery1991many,dixit2017mcmc,sbc} and have helped practitioners to apply MCMC reliably. While, we are not aware of any coverage studies similar to ours for MCMC, we would expect similar results to be found. Nonetheless, there is currently no clear solution to determine convergence with absolute certainty \citep{dixit2018developments,roy2020convergence}, even if the likelihood function is tractable. This issue has also been studied in the ABC literature. \citet{prangle2014diagnostic} present a way for approximating the coverage in ABC by reusing the same simulations in different runs. \citet{xing2020distortion} provides a diagnostic for the quality of the approximation for a given observation by the mean of distortion maps. Finally, \citet{frazier2018asymptotic} provide condition on the threshold $\epsilon$ such as to produce posterior estimates that asymptotically have proper frequentist coverage. However, the quality of the posterior estimates remains uncertain in a practical non-asymptotic regime.

We are of the opinion that theoretical and methodological advances within the field of simulation-based inference will strengthen its reliability and thereby promote its applicability in sciences. 
First, although all benchmarked algorithms recover the true posterior under specific optimal conditions, it is generally not possible to know whether those conditions are satisfied in practice.
Therefore, the study of new objective functions that would force posterior estimators to always be conservative, regardless of optimality conditions, constitutes a valuable research avenue.
From a Bayesian perspective, \citet{rozet2021arbitrarythesis} propose using the focal and the peripheral losses to weigh down strongly classified samples as a means to tune the conservativeness of a posterior estimator. However, the technique is empirical and requires tuning to attain the desired properties in practice.
\citet{dalmasso2020confidence} on the other hand consider the frequentist setting and introduce a theoretically-grounded algorithm for the construction of confidence intervals that are guaranteed to have calibrated coverage, regardless of the quality of the used statistic. \citet{dalmasso2021likelihood} extends this work with finite sample guarantees.

Second, in light of our results that ensembles produce more conservative posteriors, model averaging constitutes another promising direction of study as a simple and directly applicable method to produce reliable posterior estimators. 
However, a deeper understanding of the behaviour we observe is certainly first required to further develop these methods.

Third, post-training calibration can be used to improve the reliability of posterior estimators and should certainly be considered as a way toward more conservative inference. To some extent, this has already been considered for amortized methods \citep{cranmer2015approximating,Brehmer:2018eca,Hermans:2020skz} and would be worth exploring further, especially for non-amortized approaches. 

In this work, we assumed that the simulator perfectly models the real data generating process and focused on the computational faithfulness of the inference engine. In practice, there might be a gap between the simulation and reality. This issue, orthogonal to our work, should also be taken into account for making reliable simulation-based inference, as already initiated by several authors \citep{schmon2020generalized, matsubara2021robust, pacchiardi2022score, dellaporta2022robust}.

In summary, we show that current algorithms for simulation-based inference can produce overconfident posterior approximations, making them possibly unreliable for scientific use cases and falsificationist inquiry.
Nevertheless, we remain confident and optimistic and advocate that our results are only a stepping stone toward more reliable simulation-based inference and its wider adoption in the sciences.

\FloatBarrier

\begin{epiloguebox}
In this work, we have introduced the notion of expected coverage to quantify the conservativeness of posterior approximations. From an extensive empirical evaluation, we observed that all the methods yielded overconfident posterior approximations on at least one benchmark. While not surprising, this observation illustrates some concerns with using simulation-based inference algorithms for scientific reasoning. When applying simulation-based inference algorithms in this setting, we advocate that coverage diagnostics are necessary. We observed that, nowadays, a lot of applicative simulation-based inference papers include expected coverage to analyze the quality of the produced posterior approximations \citep{dimitriou2023fast, vasist2023neural, berteaud2024simulation, reza2024constraining, swierc2024domain, morandini2024reconstructing}.

We highlighted that such diagnostics can only be performed with amortized algorithms and raised caution regarding the use of sequential algorithms. Nowadays, this statement is no longer valid with the introduction of truncated sequential algorithms that allow local coverage testing. However, similar caution can be raised regarding newly developed models that only allow sampling from the approximate posterior, such as score-based models and generative adversarial networks. Indeed, computing the expected coverage requires posterior density estimation, which is not immediately available for such models.

In addition to diagnosing posterior approximations, it would be desired to develop algorithms that yield approximations passing the coverage diagnostic. As an attempt to easily improve conservativeness, we showed that ensembling several models empirically leads more often to conservative models. However, there are no guarantees, and ensembling leads to heavier computations. This paper also serves as a call to develop more conservative algorithms for simulation-based inference to increase its applicability to scientific domains.
\end{epiloguebox}
  
  \chapter{Balanced neural ratio estimation}\label{c:bnre}
  \begin{prologuebox}
This chapter is based on the following publication: \emph{Delaunoy, A.$^*$, Hermans, J.$^*$, Rozet, F., Wehenkel, A., \& Louppe, G. (2022). Towards reliable simulation-based inference with balanced neural ratio estimation. Advances in Neural Information Processing Systems, 35, 20025-20037.}

This piece of work is an attempt to increase the conservativeness of neural ratio estimation. We aim to derive conditions under which the trained classifier yields conservative posterior approximations and enforce those conditions during training. We introduce the notion of balanced classifier and show empirically that a balanced classifier yields more conservative approximations than a non-balanced classifier. Additionally, we derive properties of balanced classifier, providing intuition about why more conservative approximations are observed. We introduce the Balanced Neural Ratio Estimation (BNRE) algorithm, a variant of NRE that aims to produce balanced classifiers. BNRE is implemented in the LAMPE \citep{rozet2023lampe} and SBI \citep{sbi} libraries.
\end{prologuebox}

\section{Introduction}
\label{sec:introduction}

Many areas of science and engineering use parametric computer simulations to describe complex stochastic generative processes. In this setting, Bayesian inference provides a principled framework to identify parameters matching empirical observations. 
Computer simulations, however, define the necessary likelihood function only implicitly, which prevents its evaluation and the use of classical inference algorithms.
To overcome this obstacle, recent simulation-based inference (SBI) algorithms \citep{cranmer2020frontier} build upon deep learning surrogates to approximate parts of the Bayes rule and enable approximate inference.
For example, \citet{papamakarios2019sequential, glockler2021variational} build a surrogate of the likelihood function while \citet{cranmer2015approximating, thomas2016likelihood, 2019arXiv190304057H, durkan2020contrastive,miller2021truncated} approximate the likelihood-to-evidence ratio. The posterior can also be targeted directly with variational inference, as proposed by \citet{papamakarios2016fast, greenberg2019automatic, glockler2021variational}.
These algorithms are either amortized or run sequentially to drive the training towards a target observation and improve the simulation efficiency of the procedure \citep{papamakarios2016fast, lueckmann2017flexible, 2019arXiv190304057H, greenberg2019automatic, papamakarios2019sequential, durkan2020contrastive, glockler2021variational}.
However, sequential methods have the drawback of being computationally expensive to diagnose as the surrogates are only valid for the target observation \citep{crisissbi}. 
Truncated marginal neural ratio estimation \citep{miller2021truncated} alleviates this issue by introducing a sequential algorithm that builds a surrogate valid in a local region around the target.

Since modern simulation-based inference algorithms rely on deep learning surrogates, concerns naturally arise regarding their computational faithfulness and whether they are sufficiently adequate for the inference task of interest.
In Bayesian inference, these concerns can be at least partially addressed with diagnostics designed to probe the correct behaviour of the inference method, such as $\hat{R}$ diagnostics for MCMC \citep{gelman1992inference}, or to assess the quality of posterior approximations directly. 
The latter include diagnostics such as simulation-based calibration (SBC) \citep{sbc} or coverage-based diagnostics \citep{zhao2021diagnostics, crisissbi}.
As discussed by \citet{crisissbi}, posterior approximations must be conservative to guarantee reliable inferences, even when approximations are not faithful.
For example, in the physical sciences, where the goal is often to constrain parameters of interest, wrongly excluding plausible values could drive the scientific inquiry in the wrong direction, whereas failing to exclude implausible values because of (too) conservative estimations is much less detrimental.
Unfortunately, the same authors also demonstrate that current simulation-based inference algorithms can lead to overconfident surrogates and therefore false inferences. 

Scientific use cases requiring conservative inference include for example the study of dark matter models in particle physics and astrophysics \citep{Hermans:2020skz}, which could be cold, warm, or hot dark matter. 
In general, thermal dark matter models are described by a single parameter, the dark matter thermal relic mass, which can be intuitively thought of as the energy the dark matter particle had in the Early Universe. Small values correspond to warm or hot dark matter, while high values are descriptive of cold dark matter. Applying an inference algorithm without diagnosing the learned estimator could lead to constraints that are tighter than they should be. For example, whenever an overconfident estimator produces posterior estimates that favor cold dark matter models, it could simultaneously reject alternative models, such as the extensively studied Sterile Neutrino, a potential candidate for the Warm Dark Matter particle. 
Making a scientific statement in this direction therefore requires the uttermost care to not wrongly exclude values of the thermal relic mass that are actually plausible.

In this work, we develop a novel algorithm that not only converges to exact inference as the simulation budget increases, but which is also more likely to produce conservative surrogates in small simulation budget regimes.
Toward this objective, we propose a variant of the NRE algorithm called Balanced Neural Ratio Estimation (BNRE), which enforces a balancing condition on the binary neural classifier to increase the reliability of its posterior approximations. 

The structure of the manuscript is outlined as follows. Section \ref{sec:bnre:background} describes the formalism and the necessary background. Section \ref{sec:method} describes BNRE and provides theoretical arguments towards its conservativeness and reliability. Section \ref{sec:bnre:experiments} illustrates our main results and provides insights regarding the behavior of the method. Finally, Section \ref{sec:related} discusses related work while Section \ref{sec:conclusion} summarizes our contributions and hints at future work. 

\section{Background}
\label{sec:bnre:background}

\subsection{Statistical formalism}
\label{sec:formalism}

This work is concerned with simulation-based inference algorithms that produce posterior approximations $\hat{p}(\btheta|\bx)$ under the following semantics.
Target parameters $\btheta$ denote the parameters of the model and we make the reasonable assumption that the prior $p(\btheta)$ is tractable. 
The model is generically expressed as a computer program, a simulator, that describes the forward dynamics of interest based on the input parameters $\btheta$. 
The simulator implicitly defines the likelihood function $p(\bx|\btheta)$. While we cannot directly evaluate the likelihood $p(\bx|\btheta)$, we can execute the computer program to generate synthetic observables $\bx\sim p(\bx|\btheta)$. Every observable $\bx_o$ is tied to ground truth parameters $\btheta^*$ whose forward evaluation within the simulator produced $\bx^*$.

Of special importance to Bayesians is the notion of a credible region, which is a domain $\bTheta$ within the target parameter space that satisfies
$\int_{\bTheta} p(\btheta|\bx=\bx^*) d{\btheta} = 1 - \alpha$
for some observable $\bx^*$ and confidence level $1 - \alpha$.
Because many such regions exist, we target the credible region with the smallest volume,
also known as the highest posterior density region \citep{hyndman1996computing,box2011bayesian}.

\subsection{Neural ratio estimation}
Neural Ratio Estimation (NRE) is an established approach in the simulation-based inference literature both from frequentist \citep{cranmer2015approximating} and Bayesian \citep{thomas2016likelihood,2019arXiv190304057H,durkan2020contrastive,miller2021truncated} perspectives. 
In essence, all protocols rely on the density-ratio trick \citep{sugiyama2012density,goodfellow2014generative,cranmer2015approximating} to construct a surrogate of the likelihood ratio.
In this work, we consider
an amortized estimator $\hat{r}(\bx|\btheta)$
of the intractable likelihood-to-evidence ratio $r(\bx|\btheta) = p(\btheta,\bx) / p(\btheta)p(\bx) = p(\bx|\btheta) / p(\bx)$ that can be learned by training a binary classifier $\hat{d}: \bX \times \bTheta \mapsto [0,1]$ to distinguish between samples of the
joint $p(\btheta,\bx)$ with class label $1$ and samples of the product of marginals $p(\btheta)p(\bx)$ with class label $0$, with equal label marginal probability. For the binary cross-entropy loss, the Bayes optimal classifier is
\begin{equation}\label{eqn:nre}
  d(\btheta,\bx) = \frac{p(\btheta, \bx)}{p(\btheta,\bx) + p(\btheta)p(\bx)} = \sigma\left(\log\frac{p(\btheta,\bx)}{p(\btheta)p(\bx)}\right),
\end{equation}
where $\sigma(\cdot)$ is the sigmoid function.
Given target parameters $\btheta$ and an observable $\bx$ supported by $p(\btheta)$ and $p(\bx)$ respectively,
the learned classifier $\hat{d}$ provides an approximation for the log likelihood-to-evidence ratio $\log r(\bx|\btheta)$ because
$\log r(\bx|\btheta) = \sigma^{-1}(d(\btheta,\bx)) \approx \sigma^{-1}(\hat{d}(\btheta,\bx)) = \log \hat{r}(\bx|\btheta)$.
The log posterior density function is approximated as $\log \hat{p}(\btheta|\bx)=\log p(\btheta) + \log \hat{r}(\bx|\btheta)$.

\section{Balanced binary classification for neural ratio estimation}
\label{sec:method}
Following \citet{crisissbi}, let us first define the expected coverage probability of the $1 - \alpha$ highest posterior density regions derived from the posterior estimator $\hat{p}(\btheta | \bx)$ as
\begin{equation}
    \label{eq:bnre:coverage}
    \mathbb{E}_{p(\btheta,\bx)}\left[\mathds{1}\left(\btheta \in \Theta_{\hat{p}(\btheta|\bx)}(1 - \alpha)\right)\right],
\end{equation}
where the function
$\Theta_{\hat{p}(\btheta|\bx)}(1 - \alpha)$ yields the $1 - \alpha$ highest posterior density region of $\hat{p}(\btheta|\bx)$. 
This diagnostic probes the conservativeness of the posterior estimator (or the lack thereof) and can be interpreted as the expected frequentist coverage  $\mathbb{E}_{p(\btheta)}\mathbb{E}_{p(\bx|\btheta)}\left[\mathds{1}\left(\btheta \in \Theta_{\hat{p}(\btheta|\bx)}(1 - \alpha)\right)\right]$.

In this work, a posterior estimator has coverage at the confidence level $1-\alpha$ whenever the expected coverage probability is larger or equal to the nominal coverage probability, $1-\alpha$. We say that a posterior estimator is conservative when it has coverage for all confidence levels. The expected coverage probability can be plotted for various levels $\alpha$, which allows to visually identify conservative posterior estimators. 

The expected coverage can also be shown to be a special case of the Simulation-based calibration (SBC) diagnostic~\citep{sbc}, further motivating the usage of expected coverage. SBC provides a way to diagnose the faithfulness of an approximate posterior distribution $\hat{p}(\btheta | \bx)$. Given an observation $\bx^* \sim p(\bx)$, \citet{sbc} prove that, for any one-dimensional statistic $f: \Theta \mapsto \mathbb{R}$, the rank statistic
\begin{equation}
    r(\btheta^*) = \mathbb{E}_{p(\btheta | \bx^*)} \big[ \mathds{1}[ f(\btheta) \leq f(\btheta^*) ] \big]
\end{equation}
of posterior samples $\btheta^* \sim p(\btheta | \bx^*)$ is uniformly distributed over the interval $[0, 1]$. Consequently, any deviation from the uniform distribution for the approximate rank statistic
\begin{equation}
    \hat{r}(\btheta^*) = \mathbb{E}_{\hat{p}(\btheta | \bx^*)} \big[ \mathds{1}[ f(\btheta) \leq f(\btheta^*) ] \big]
\end{equation}
indicates some error in the approximate posterior $\hat{p}(\btheta | \bx^*)$. As this holds for any statistic $f$, it also holds for $f(\btheta) = \hat{p}(\btheta | \bx^*)$. In this special case, if $\hat{r}(\btheta^*) = \alpha$, a proportion $1 - \alpha$ of samples $\btheta \sim \hat{p}(\btheta | \bx^*)$ have an approximate posterior density larger than $\btheta^*$. In other words, it means that $\btheta^*$ resides within the $1 - \alpha$ highest posterior density region $\Theta_{\hat{p}(\btheta | \bx^*)}(1 - \alpha)$ of $\hat{p}(\btheta | \bx^*)$. Therefore, we have
\begin{equation}
    P(\hat{r}(\btheta^*) \geq \alpha) = \mathbb{E}_{p(\btheta^* | \bx^*)} \left[\mathds{1}[\btheta^* \in \Theta_{\hat{p}(\btheta | \bx^*)}(1 - \alpha)] \right]
\end{equation}
and since $\hat{r}(\btheta^*)$ should be uniformly distributed, $P(\hat{r}(\btheta^*) \geq \alpha)$ should be equal to $1 - \alpha$. In practice, this test cannot be performed locally for a given $\bx^*$ as we cannot sample from the unknown posterior distribution $p(\btheta | \bx^*)$. Instead, SBC checks globally that $\hat{r}(\btheta^*)$ is uniformly distributed over pairs $(\btheta^*, \bx^*) \sim p(\btheta, \bx)$ sampled from the joint distribution, which, in the special case $f(\btheta) = \hat{p}(\btheta | \bx^*)$, comes down to check that
\begin{equation}
    \mathbb{E}_{p(\btheta^* \!,\, \bx^*)} \left[\mathds{1}[\btheta^* \in \Theta_{\hat{p}(\btheta | \bx^*)}(1 - \alpha)]\right] = 1 - \alpha
\end{equation}
is satisfied for all $\alpha \in [0, 1]$. We recognize here the expected coverage diagnostic used in \citet{crisissbi} and this work.

Our main objective is to restrict the hypothesis space of the approximate classifiers $\hat{d}$ to those leading to conservative posterior estimators, hence solving the reliability concerns of NRE. 
Towards this goal, we construct a hypothesis space of balanced classifiers and show both theoretically and empirically that they lead to posterior estimators that tend to be more conservative. 

\subsection{Balanced binary classification}

\begin{definition}
    A classifier $\hat{d}$ is balanced if $\mathbb{E}_{p(\btheta, \bx)} \left[\hat{d}(\btheta, \bx)\right] = \mathbb{E}_{p(\btheta)p(\bx)} \left[1 - \hat{d}(\btheta, \bx)\right]$,
    or 
    \begin{equation}
        \mathbb{E}_{p(\btheta, \bx)} \left[\hat{d}(\btheta, \bx)\right] + \mathbb{E}_{p(\btheta)p(\bx)} \left[\hat{d}(\btheta, \bx)\right] = 1.
    \end{equation}
\end{definition}

\begin{theorem}\label{thm:conservativeness_1}
Any balanced classifier $\hat{d}$ satisfies $\mathbb{E}_{p(\btheta, \bx)}\left[\displaystyle\frac{d(\btheta, \bx)}{\hat{d}(\btheta, \bx)} \right] \geq 1$.
\begin{proof}
    The integral form of the balancing condition
    \begin{equation}
        \iint \big( p(\btheta, \bx) + p(\btheta) p(\bx) \big)\hat{d}(\btheta, \bx) d\btheta d\bx = 1
    \end{equation}
    implies that $\big(p(\bx, \btheta) + p(\btheta) p(\bx)\big)\hat{d}(\btheta, \bx)$ is a valid density, both integrating to 1 and positive everywhere. Therefore, its Kullback-Leibler (KL) divergence with $p(\btheta, \bx)$ is positive. Through Jensen's inequality, we obtain
    \begin{align*}
        0 & \leq \text{KL}\left(p(\btheta, \bx) \, \big\vert\big\vert \big(p(\btheta, \bx) + p(\btheta) p(\bx)\big) \hat{d}(\btheta, \bx) \right) \\
        & \leq \mathbb{E}_{p(\btheta, \bx)}\left[\log\frac{p(\btheta, \bx)}{\big(p(\btheta, \bx) + p(\btheta) p(\bx)\big)\hat{d}(\btheta, \bx)} \right] \\
        & \leq \mathbb{E}_{p(\btheta, \bx)}\left[\log\frac{d(\btheta, \bx)}{\hat{d}(\btheta, \bx)} \right] \\
        \Rightarrow \quad 1 & \leq \mathbb{E}_{p(\btheta, \bx)}\left[\exp \left( \log \frac{d(\btheta, \bx)}{\hat{d}(\btheta, \bx) } \right) \right] = \mathbb{E}_{p(\btheta, \bx)}\left[\frac{d(\btheta, \bx)}{\hat{d}(\btheta, \bx)} \right]. \qedhere
    \end{align*}
\end{proof}
\end{theorem}

\begin{theorem}\label{thm:conservativeness_2}
Any balanced classifier $\hat{d}$ satisfies $\mathbb{E}_{p(\btheta) p(\bx)}\left[\displaystyle\frac{1-d(\btheta, \bx)}{1-\hat{d}(\btheta, \bx)} \right] \geq 1$.
\begin{proof}
    From the integral form of the balancing condition, we have
    \begin{align*}
        1 & = \iint \big( p(\btheta, \bx) + p(\btheta) p(\bx) \big)\hat{d}(\btheta, \bx) d\btheta d\bx \\
        & = 2 - \iint \big( p(\btheta, \bx) + p(\btheta) p(\bx) \big)\hat{d}(\btheta, \bx) d\btheta d\bx \\
        & = \iint p(\btheta, \bx) d\btheta d\bx + \iint p(\btheta) p(\bx) d\btheta d\bx - \iint \big( p(\btheta, \bx) + p(\btheta) p(\bx) \big)\hat{d}(\btheta, \bx) d\btheta d\bx \\
        & = \iint \big( p(\btheta, \bx) + p(\btheta) p(\bx) \big) \big(1 - \hat{d}(\btheta, \bx)\big) d\btheta d\bx,
    \end{align*}
    which implies that $\big(p(\bx, \btheta) + p(\btheta) p(\bx)\big) \big(1 - \hat{d}(\btheta, \bx)\big)$ is a valid density, integrating to 1 and positive everywhere. Therefore, its Kullback-Leibler divergence with $p(\btheta) p(\bx)$ is positive and, using Jensen's inequality, we have
    \begin{align*}
        0 & \leq \text{KL}\left(p(\btheta) p(\bx) \, \big\vert\big\vert \big(p(\btheta, \bx) + p(\btheta) p(\bx)\big) \big(1 - \hat{d}(\btheta, \bx)\big) \right) \\
        & \leq \mathbb{E}_{p(\btheta) p(\bx)}\left[\log\frac{p(\btheta) p(\bx)}{\big(p(\btheta, \bx) + p(\btheta) p(\bx)\big) \big(1 - \hat{d}(\btheta, \bx)\big)} \right] \\
        & \leq \mathbb{E}_{p(\btheta) p(\bx)}\left[\log\frac{1 - d(\btheta, \bx)}{1 - \hat{d}(\btheta, \bx)} \right] \\
        \Rightarrow \quad 1 & \leq \mathbb{E}_{p(\btheta) p(\bx)}\left[\exp \left( \log \frac{1 - d(\btheta, \bx)}{1 - \hat{d}(\btheta, \bx) } \right) \right] = \mathbb{E}_{p(\btheta) p(\bx)}\left[\frac{1 - d(\btheta, \bx)}{1 - \hat{d}(\btheta, \bx)} \right]. \qedhere
    \end{align*}
\end{proof}
\end{theorem}

Theorem \ref{thm:conservativeness_1} shows that, in expectation over the joint distribution $p(\btheta,\bx)$, a balanced classifier $\hat{d}$ tends to make predictions whose probability values $\hat{d}(\btheta, \bx)$ are smaller than the exact probability values $d(\btheta, \bx)$. 
In other words, a balanced classifier $\hat{d}$ tends to be less confident than the Bayes optimal classifier $d$.
Similarly, Theorem \ref{thm:conservativeness_2} shows that, in expectation over the product of the marginals $p(\btheta)p(\bx)$, a balanced classifier tends to make predictions whose probability values $1-\hat{d}(\btheta,\bx)$ are smaller than the exact probability values $1-d(\btheta, \bx)$, hence showing that a balanced classifier $\hat{d}$ tends to also be less confident than the Bayes optimal classifier $d$.
We note however that these two theorems hold only in expectation, which implies that neither $\hat{d}(\btheta,\bx) \leq d(\btheta, \bx)$ for all $\btheta,\bx$ nor $1-\hat{d}(\btheta,\bx) \leq 1-d(\btheta, \bx)$ for all $\btheta,\bx$ can generally be guaranteed.

\begin{theorem}\label{thm:optimal_balancing}
\vspace{0.25cm}
The Bayes optimal classifier $d(\btheta,\bx)$ is balanced.

\begin{proof}
    Replacing the Bayes optimal classifier
    \begin{equation}
        d(\btheta,\bx)\triangleq\frac{p(\btheta,\bx)}{p(\btheta,\bx) + p(\btheta)p(\bx)}
    \end{equation}
    in the integral form of the balancing condition, we have
    \begin{align*}
        & \iint (p(\btheta,\bx) + p(\btheta)p(\bx)) d(\btheta,\bx) d\btheta d\bx \\
        & = \iint \frac{\big(p(\btheta,\bx) + p(\btheta)p(\bx)\big) \, p(\btheta,\bx)}{p(\btheta,\bx) + p(\btheta)p(\bx)} d\btheta d\bx \\
        & = \iint p(\btheta,\bx) d\btheta d\bx = 1. \qedhere
    \end{align*}
\end{proof}
\end{theorem}

Theorem \ref{thm:optimal_balancing} states that the Bayes optimal classifier is balanced. Therefore, \textbf{minimizing the cross-entropy loss while restricting the model hypothesis space to balanced classifiers results in the same Bayes optimal classifier of Eqn. \ref{eqn:nre}.}

\subsection{Balanced neural ratio estimation}
\label{sec:bnre}
We now extend the NRE algorithm to enforce the balancing condition.
The previous results show that enforcing the condition should result in more conservative classifiers $\hat{d}$ and therefore to dispersed posterior approximations.
Let us first note that Theorem \ref{thm:conservativeness_1} can be expressed as $\mathbb{E}_{p(\bx)}[\mathbb{E}_{p(\btheta| \bx)}[d(\btheta, \bx) / \hat{d}(\btheta, \bx ]] \geq 1$, which can (ideally) be achieved when the inner expectation is larger than $1$ for all $\bx$.
In this case, the classifier $\hat{d}$ will be such that $\hat{d}(\btheta,\bx) \leq d(\btheta,\bx)$ in regions of high posterior density. Then, 
\begin{equation}
    \frac{\hat{d}(\btheta,\bx)}{1 - \hat{d}(\btheta,\bx)}\leq \frac{d(\btheta,\bx)}{1 - d(\btheta,\bx)}, ~\text{which is equivalent to}~ \hat{r}(\bx\vert\btheta) \leq r(\bx\vert\btheta),
\end{equation}
and $\hat{p}(\btheta|\bx) \leq p(\btheta|\bx)$ since $\hat{p}(\btheta\vert\bx) = p(\btheta)\hat{r}(\bx\vert\btheta)$.
Similarly, Theorem \ref{thm:conservativeness_2} implies $1-d(\btheta, \bx) \geq 1-\hat{d}(\btheta, \bx)$ in regions of high prior density, which results in $p(\btheta|\bx) \leq \hat{p}(\btheta|\bx)$. 
Between those two opposite effects, the constraint on $\hat{p}(\btheta|\bx)$ that will dominate depends on whether $p(\btheta|\bx) > p(\btheta)$ or $p(\btheta|\bx) < p(\btheta)$. If $p(\btheta|\bx) > p(\btheta)$, then $\hat{p}(\btheta|\bx) \leq p(\btheta|\bx)$, whereas if $p(\btheta|\bx) < p(\btheta)$ then $p(\btheta|\bx) \leq \hat{p}(\btheta|\bx)$.
Overall, imposing the balancing condition will therefore result in approximate posteriors that lie between the prior and the exact posterior, without being more confident than they should.

Practically, the balancing condition can be targeted through a regularization penalty. For the binary cross-entropy $\mathcal{L}[\hat{d}] \triangleq -\mathbb{E}_{p(\btheta, \bx)}[\log\hat{d}(\btheta, \bx)] - \mathbb{E}_{p(\btheta)p(\bx)}[\log(1 - \hat{d}(\btheta, \bx))]$
and given that the balancing condition only depends on samples from $p(\bx)p(\btheta)$ and $p(\bx, \btheta)$,
the full loss functional including the balancing condition can be expressed as
\begin{equation}
  \mathcal{L}_b\left[\hat{d}\right] \triangleq \mathcal{L}\left[\hat{d}\right] + \lambda \left( \mathbb{E}_{p(\btheta)p(\bx)}\left[\hat{d}(\btheta, \bx)\right] + \mathbb{E}_{p(\btheta, \bx)}\left[\hat{d}(\btheta, \bx)\right] - 1\right)^2,
\end{equation}
where $\lambda$ is a (scalar) hyper-parameter controlling the strength of the balancing condition's contribution. 
The training procedure is summarized in Algorithm \ref{algo:cnre}. 
Since a classifier is balanced if the balancing condition cancels out, $\lambda$ could, in principle, be set arbitrarily large. 
However, as the balancing condition is estimated via Monte Carlo sampling, setting $\lambda$ to a large value could impair the classifier's learning ability. We found that $\lambda = 100$ works well across many problem domains with varying simulation budgets.

\begin{algorithm}
   \caption{Training algorithm for Balanced Neural Ratio Estimation (BNRE).}
   \label{algo:cnre}
   \begin{tabular}{ l l }
        {\itshape Inputs:} & Implicit generative model $p(\bx\vert\btheta)$ (simulator) and prior $p(\btheta)$ \\
        {\itshape Outputs:} &  Approximate classifier $\hat{d}_\psi(\btheta,\bx)$ parameterized by $\psi$ \\
        {\itshape hyper-parameters:} & Balancing condition strength $\lambda$ (default = 100) and batch-size $n$
  \end{tabular}
  \begin{algorithmic}
    \REPEAT
        \STATE Sample data from the joint $\{\btheta_i,~\bx_i \sim p(\btheta,\bx), ~ y_i = 1\}^{n/2}_{i=1}$
        \STATE Sample data from the marginals $\{\btheta_i,~\bx_i \sim p(\btheta)p(\bx), ~ y_i = 0\}^{n}_{i= n/2 + 1}$
    
        \STATE $\mathcal{L}[\hat{d}_\psi] = -\frac{1}{n}\sum_{i=1}^n y_i\log\hat{d}_\psi(\btheta_i, \bx_i) + (1 - y_i)\log(1 - \hat{d}_\psi(\btheta_i, \bx_i))$ 
        
        \STATE $\mathcal{B}[\hat{d}_\psi] = \frac{2}{n}\sum^{n/2}_{i=1}\hat{d}_\psi(\btheta_i,\bx_i) + \frac{2}{n}\sum^{n}_{i= n/2 + 1}\hat{d}_\psi(\btheta_i,\bx_i)$
        
        \STATE $\psi = \texttt{minimizer\_step}(\texttt{params=}\psi,~\texttt{loss=}\mathcal{L}[\hat{d}_\psi] + \lambda(\mathcal{B}[\hat{d}_\psi] - 1)^2)$
    \UNTIL{{\bfseries convergence}}
    \STATE {\bfseries return} $\hat{d}_\psi(\btheta,\bx)$.
   \end{algorithmic}
\end{algorithm}

\section{Experiments}
\label{sec:bnre:experiments}
We start by providing an extensive validation of BNRE on a broad range of benchmarks demonstrating that the proposed method alleviates the problem. Section \ref{sec:illustrative_example} follows up  with  an illustrative demonstration on the behaviour of BNRE and its hyper-parameters.
Code is available at \href{https://github.com/montefiore-ai/balanced-nre}{\texttt{https://github.com/montefiore-ai/balanced-nre}}.

\subsection{Extensive validation}
\label{sec:extensive_validation}

\paragraph{Setup} 
We evaluate the expected coverage of posterior estimators produced by both NRE and BNRE on various problems. Those benchmarks cover a diverse set of problems from particle physics (Weinberg), epidemiology (Spatial SIR), queueing theory (M/G/1), population dynamics (Lotka Volterra, and astronomy (Gravitational Waves). They are representative of real scientific applications of simulation-based inference.

The \emph{SLCP} simulator models a fictive problem with 5 parameters. The observable $\bx$ is composed of 8 scalars which represent the 2D-coordinates of 4 points. 
The coordinate of each point is sampled from the same multivariate Gaussian whose mean and covariance matrix are parametrized by $\btheta$. We consider an alternative version of the original task \citep{papamakarios2019sequential} by inferring the marginal posterior density of 2 of those parameters. In contrast to its original formulation, the likelihood is not tractable due to the marginalization.

The \emph{Weinberg} problem \citep{weinberg} concerns a simulation of high energy particle collisions $e^+e^- \to \mu^+ \mu^-$. The angular distributions of the particles can be used to measure the Weinberg angle $\bx$
in the standard model of particle physics. From the scattering angle, we are interested in inferring Fermi's constant $\btheta$.

The \emph{Spatial SIR} model \citep{crisissbi} involves a grid-world of susceptible,
infected, and recovered individuals. Based on initial conditions and the infection and recovery rate $\btheta$,
the model describes the spatial evolution of an infection.
The observable $\bx$ is a snapshot of the grid-world after some fixed amount of time.

\emph{M/G/1} \citep{shestopaloff2014bayesian} models a processing and arrival queue. The problem is
described by 3 parameters $\btheta$ that influence the time it takes to serve a customer, and the time between their arrivals. The observable $\bx$ is composed of 5 equally spaced quantiles of inter-departure times.

The \emph{Lotka-Volterra} population model \citep{lotka,volterra1926fluctuations} describes a process of interactions between a predator and a prey species. The model is conditioned on 4 parameters $\btheta$ which influence the reproduction and mortality rate of the predator and prey species. We infer the marginal posterior of the predator parameters from time series representing the evolution of both populations over time. The specific implementation is based on a Markov Jump Process as in \citet{papamakarios2019sequential}.

\emph{Gravitational Waves (GW)} are ripples in space-time emitted during events such as the collision of two black-holes. They can be detected through interferometry measurements $\bx$ and convey information about celestial bodies, unlocking new ways to study the universe. We consider inferring the masses $\btheta$ of two black-holes colliding through the observation of the gravitational wave as measured by LIGO's dual detectors \citep{lalsuite,Biwer:2018osg}.

 Our evaluation considers simulation budgets of increasing size, ranging from $2^{10} = 1024$ to $2^{17} = 131,072$ samples, and credibility levels from $0.05$ to $0.95$. For every simulation budget, we train 5 posterior estimators for 500 epochs and determine the credible region by evaluating the approximated posterior density function in a discretized and empirically normalized grid of the parameter space with sufficient resolution. The subsequent credible region is the set of parameters whose estimated (and normalized) posterior density is higher or equal to an inclusion threshold fitted to obtain the desired credibility level $1 - \alpha$. The expected coverage probability is estimated on $10000$ unseen samples from the joint $p(\btheta,\bx)$, for each considered credibility level.

Table \ref{tab:bnre:architectures} summarizes the architectures and hyper-parameters used for each benchmark. The classifier architectures are separated into two parts: the embedding and the head networks. The embedding network $\phi$ compresses the observable into a set of features. The head network $f$ then uses those features $\phi(\bx)$ concatenated with the parameters $\btheta$ to predict the class,
\begin{equation*}
    \hat{d}(\btheta, \bx) = f(\btheta, \phi(\bx)).
\end{equation*}
The learning rate is scheduled during training. Table \ref{tab:bnre:architectures} provides the initial learning rates. Those are then divided by $10$ each time no improvement was observed on the validation loss for $10$ epochs. Further details can be found in the code repository attached to this manuscript.

\begin{table}[H]
    \caption{Architectures and training hyper-parameters}
    \centering
    \label{tab:bnre:architectures}
    \begin{tabular}{llllllp{2cm}}
        \toprule
        & SLCP & M/G/1 & Weinberg & Lotka-V. & Spatial SIR & GW \\
        \midrule
        \emph{Embedding network} & None & None & None & CNN & Resnet-18 & CNN \\
        \emph{Embedding layers} & / & / & / & $8$ & / & $13$ \\ 
        \emph{Embedding channels} & / & / & / & $8$ & /  & $16$ \\ 
        \emph{Convolution type} & / & / & / & Conv1D & Conv2D & Dilated \newline Conv1D \\ 
        \emph{Head network} & MLP & MLP & MLP & MLP & MLP & MLP \\
        \emph{Head layers} & $6$ & $6$ & $6$ & $3$ & $3$ & $3$  \\
        \emph{Head hidden neurons} & $256$ & $256$ & $256$ & $128$ & $256$ & $128$  \\
        \emph{Learning rate} & $0.001$ & $0.001$ & $0.001$ & $0.001$ & $0.001$ & $0.001$ \\ 
        \emph{Epochs} & $500$ & $500$ & $500$ & $500$ & $500$ & $500$ \\ 
        \emph{Batch size} & $256$ & $256$ & $256$ & $256$ & $256$ & $256$ \\ 
        \bottomrule
    \end{tabular}
\end{table}

\paragraph{Expected coverage}

The expected coverage curves and their interpretation are detailed in Figure~\ref{fig:bnre:coverage}. We observe that NRE often produces posterior estimators that are overconfident, especially for small simulation budgets. However, NRE's reliability increases with the availability of training data. By contrast, BNRE produces posterior estimators that are conservative on all benchmarks for all simulation budgets. Figure \ref{fig:auc} explores the same phenomena through a quantity which we call the coverage AUC, highlighting the effect of the simulation budget. Coverage AUC corresponds to the integrated signed area between the expected coverage curve and the diagonal of a particular simulation. From this quantity it is evident there is a clear distinction between NRE and BNRE with respect to the available simulation budget. Both methods have the tendency to converge towards 0, indicating both methods are moving closer to the Bayes optimal classifier. However, the difference between these methods lies with how this solution is approached. While NRE can approach this limit from both sides, BNRE consistently produces coverage AUC's above 0, corresponding to conservative posterior approximations, and therefore exhibits the desired behaviour (in expectation).

\begin{figure}[h]
    \centering
    \includegraphics[width=\textwidth]{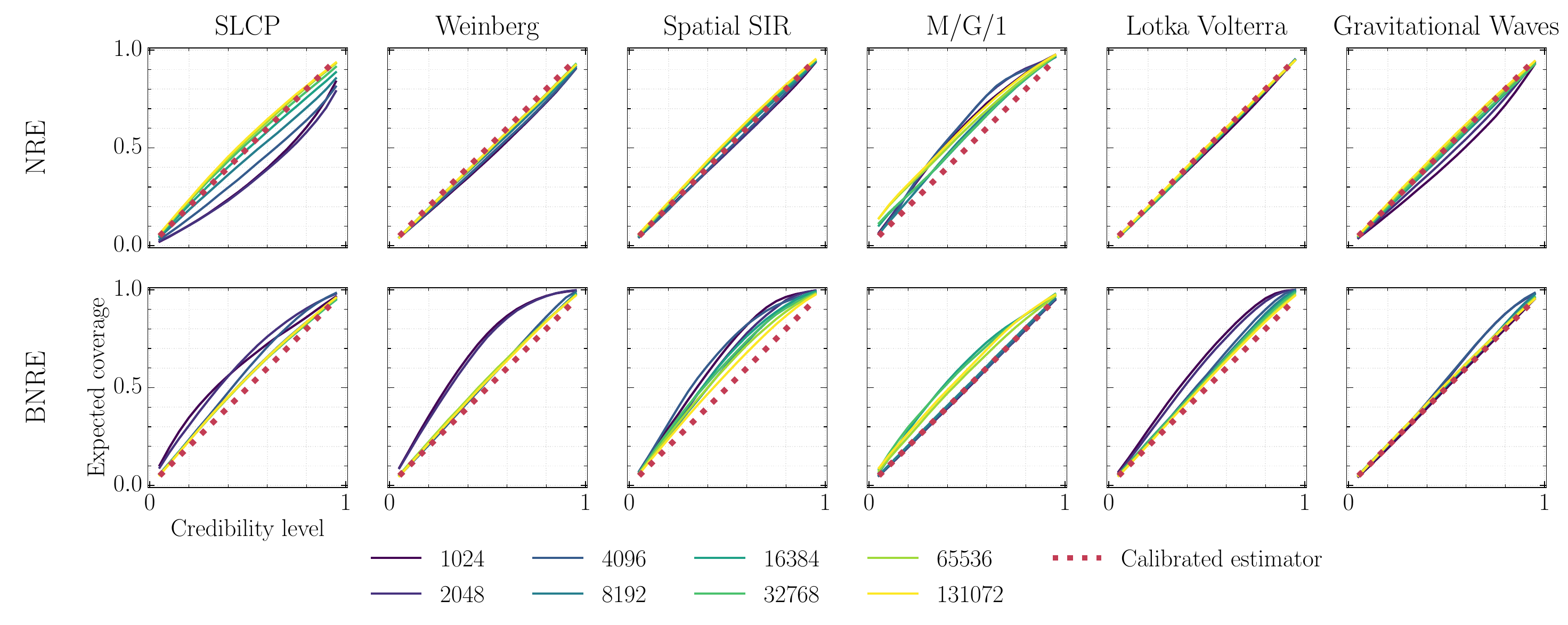}
    \caption{Expected coverage for increasing simulation budgets. A perfectly calibrated posterior has an expected coverage probability equal to the nominal coverage probability and hence produces a diagonal line. A conservative estimator has an expected coverage curve at or above the diagonal line, while an overconfident estimator produces curves below the diagonal line. The diagnostic therefore provides an immediate visual interpretation. We observe that NRE can produce overconfident estimators, while BNRE always produces coverage curves above the diagonal line and therefore the desired behaviour: conservative posterior approximations. The means over $5$ runs are reported.}
    \label{fig:bnre:coverage}
\end{figure}

\begin{figure}[h]
    \centering
    \includegraphics[width=.9\textwidth]{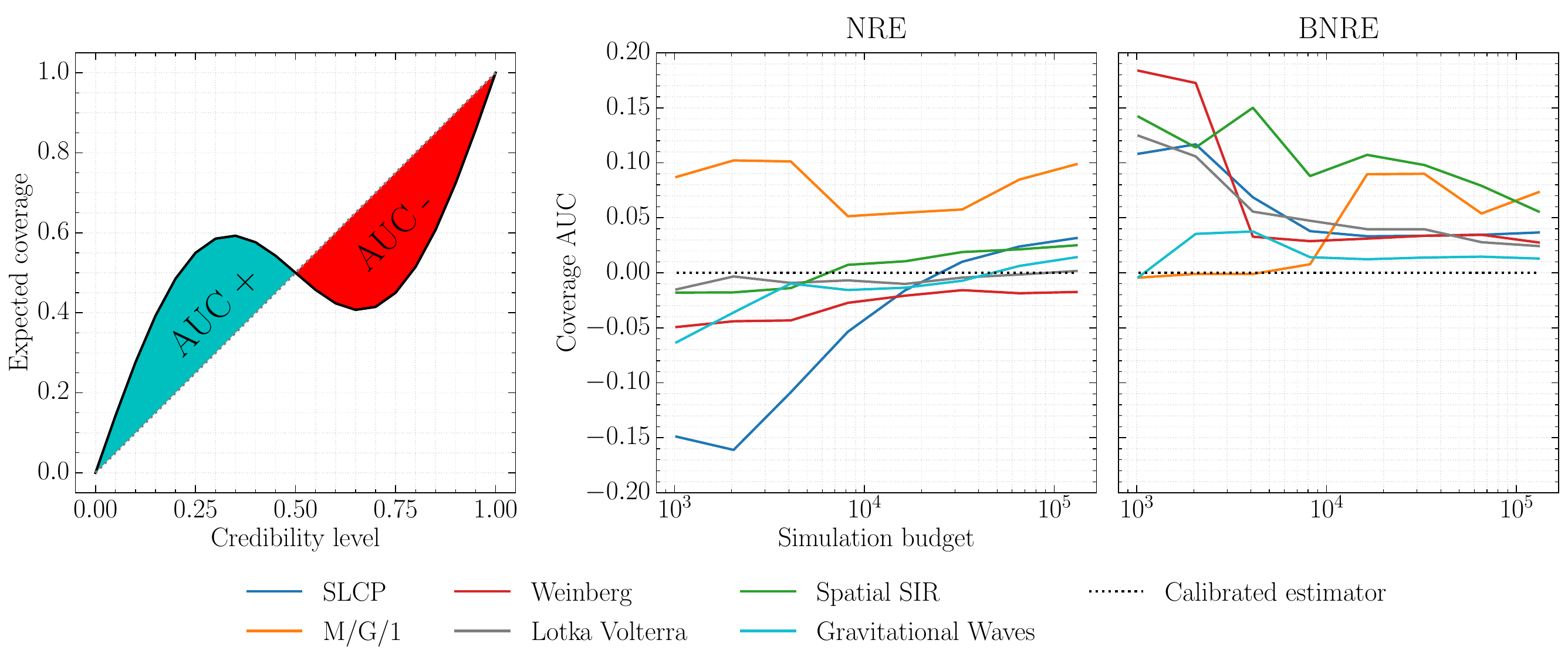}
    \caption{Coverage AUC measures the integrated signed area between the expected coverage curve and the diagonal. A perfectly calibrated posterior has an expected coverage probability equal to the nominal coverage probability, producing a diagonal line and has a coverage AUC of $0$, as shown on the left subplot. A conservative estimator on the other hand has a coverage AUC larger than $0$ and an overconfident estimator smaller than $0$. We observe that while NRE can produce coverage AUC both below or above $0$, BNRE always produces a coverage AUC larger than $0$, implying that its posterior approximations are conservative on average. The means over $5$ runs are reported. A complete overview, including standard deviations, are provided in Section \ref{sec:additional_figures}.}
    \label{fig:auc}
\end{figure}

\paragraph{Statistical performance}
In addition to the reliability of the posteriors, we evaluate and compare the statistical performance of the posterior approximations produced by NRE and BNRE.
We estimate the expected approximate log posterior density $\mathbb{E}_{p(\btheta,\bx)}\big[\log \hat{p}(\btheta|\bx)\big]$ over a large number of pairs $\btheta, \bx$. It captures how well the posterior surrogates $\hat{p}(\btheta|\bx)$ approximate the true posteriors $p(\btheta|\bx)$ since $\mathbb{E}_{p(\btheta,\bx)}\left[\log\hat{p}(\btheta\vert\bx)\right] = -\mathbb{E}_{p(\bx)}\text{KL}\left[p(\btheta\vert\bx)~\vert\vert~\hat{p}(\btheta\vert\bx)\right] + \mathbb{E}_{p(x)}\mathbb{E}_{p(\btheta\vert\bx)}\left[\log p(\btheta\vert\bx)\right]$ \citep{lueckmann2021benchmarking}.

Figure \ref{fig:bnre:log_posterior} shows our results. We observe that enforcing the balancing condition for $\lambda = 100$ is associated with a loss in statistical performance. However, the loss in statistical performance is eventually recovered by increasing the simulation budget. In fact, practitioners might be inclined to favor reliability over statistical performance \citep{crisissbi}, although it is always a trade-off that depends on the use case. Nevertheless, it is possible to improve the statistical performance by tuning the surrogate, or by increasing the available simulation budget as we have demonstrated.

\begin{figure}[h]
    \centering
    \includegraphics[width=\textwidth]{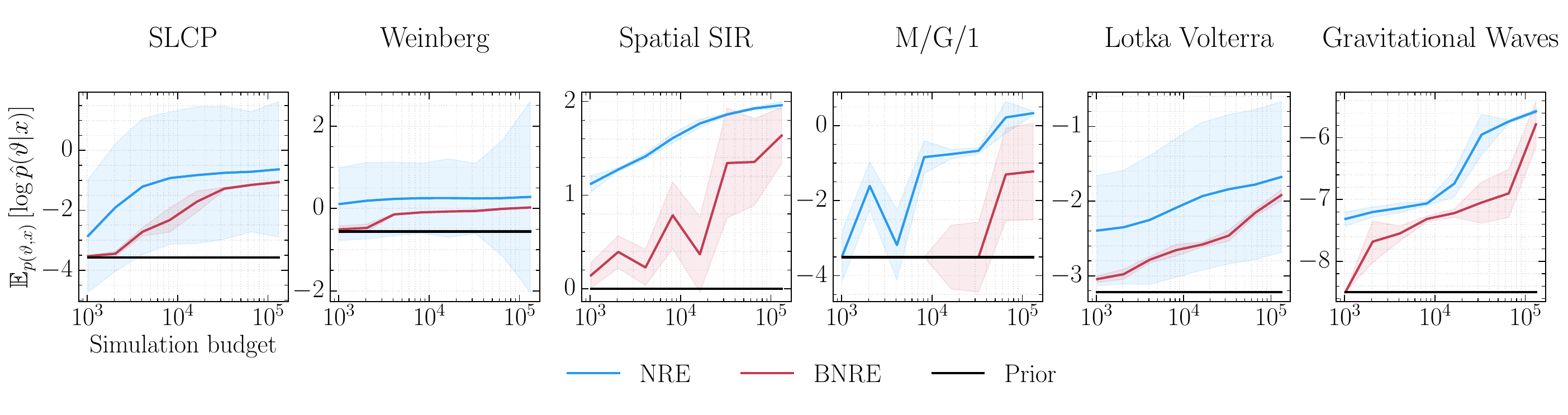}
    \caption{Expected value $\mathbb{E}_{p(\btheta,\bx)}\big[\log \hat{p}(\btheta|\bx)\big]$ of the approximate log posterior density of the nominal parameters with respect to the simulation budget. We observe that BNRE produces log posterior densities lower than NRE. This shows that enforcing the balancing condition to have more reliable posterior approximates comes at the price of a small loss in information gain. However, BNRE improves over the prior and eventually converges towards NRE as the simulation budget increases. Solid lines represent the mean over $5$ runs and shaded areas represent the standard deviation.}
    \label{fig:bnre:log_posterior} 
\end{figure}

\subsection{In-depth analysis}
\label{sec:illustrative_example}
In this section, we consider the Weinberg benchmark. The quality of the posterior approximations produced by BNRE is initially discussed with respect to the simulation budget. Afterwards, the effects
of the hyper-parameter $\lambda$ are studied.

\begin{figure}[t]
    \centering
    \includegraphics[width=\textwidth]{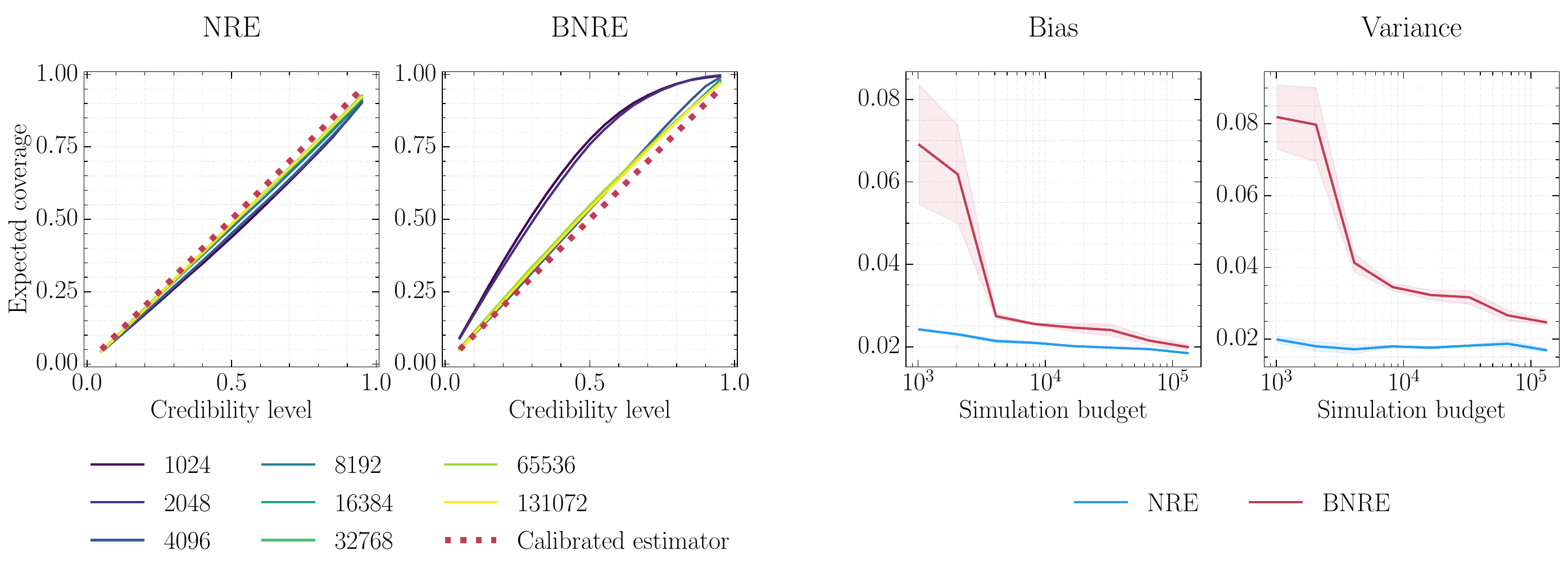}
    \caption{Comparison between NRE and BNRE in terms of expected coverage, bias and variance on the Weinberg benchmark. On the left side, the coverage is shown with respect to the simulation budget represented by the colormap. The bias and variance are represented on the right side of the plot. BNRE is run with $\lambda=100$. Consistent with our previous observations in Figure \ref{fig:bnre:log_posterior}, we observe that the gap in both bias and variance reduces as the simulation budget increases. Furthermore, in contrast with NRE, the posterior approximations of BNRE are tending towards being increasingly calibrated while at the same time being conservative. Solid lines represent the mean over $5$ runs and shaded areas represent the standard deviation.}
    \label{fig:illustrative_summary}
\end{figure}
\paragraph{Quality assessment}
Because the expected coverage does not capture the quality of an approximation in terms of information gain, we complement our assessment with a bias and variance analysis of the posterior approximations.
Let us consider the expected squared error over the approximate posterior $\mathbb{E}_{\hat{p}(\btheta | \bx)} \left[\left(\btheta - \btheta^* \right)^2 \right]$, where $\btheta^*$ is the ground truth parameter value. With $\bar{\btheta}(\bx) = \mathbb{E}_{\hat{p}(\btheta | \bx)} \left[\btheta\right]$, we decompose $\mathbb{E}_{\hat{p}(\btheta | \bx)} \left[\left(\btheta - \btheta^* \right)^2 \right]$ as
\begin{align*}
     & \mathbb{E}_{\hat{p}(\btheta | \bx)} \left[\left(\btheta -\bar{\btheta}(\bx) \right)^2\right] +2 \left(\bar{\btheta}(\bx) - \btheta^* \right) \underbrace{\mathbb{E}_{\hat{p}(\btheta | \bx)} \left[\left(\btheta -\bar{\btheta}(\bx) \right)\right]}_{=0} + \mathbb{E}_{\hat{p}(\btheta | \bx)} \left[\left(\bar{\btheta}(\bx) - \btheta^* \right)^2 \right] \\
     & = \mathbb{E}_{\hat{p}(\btheta | \bx)} \left[\left(\btheta -\bar{\btheta}(\bx) \right)^2\right] + \left(\bar{\btheta}(\bx) - \btheta^* \right)^2.
\end{align*}
The expectation over the joint distribution $p(\btheta^*\!, \bx)$ of the expected squared error can hence be decomposed in a bias term defined as
\begin{equation}
    \text{bias}(\hat{p}(\btheta \vert \bx)) \triangleq  \mathbb{E}_{p(\btheta^* \!,\,\bx)}\left[\left(\bar{\btheta}(\bx) - \btheta^*\right)^2\right],
\end{equation}
which can be interpreted as the expected discrepancy between the nominal value $\btheta^*$ and the expected posterior value $\bar{\btheta}$. 
The variance term is
\begin{equation}
    \text{variance}(\hat{p}(\btheta \vert \bx)) \triangleq \mathbb{E}_{p(\btheta^* \!,\, \bx)}\left[ \mathbb{E}_{\hat{p}(\btheta | \bx)} \left[\left(\btheta - \bar{\btheta}(\bx) \right)^2\right] \right]
\end{equation}
and measures the dispersion of the posterior approximations. Note that these terms differ from the typical statistical bias and variance of point estimators since we are considering full posterior estimators. In particular, the bias of the Bayes optimal model does not necessarily reduce to $0$.

Figure \ref{fig:illustrative_summary} shows the evolution of expected coverage, bias and variance with respect to the available simulation budget.
By taking all plots into consideration with respect to the simulation budget, we can validate that -- as suggested by theorems \ref{thm:conservativeness_1} and \ref{thm:conservativeness_2} -- the increase in
expected coverage is tied to an increase in variance.
However, this increase comes at the price of a slight increase in bias. Consistent with our previous observations in Figure \ref{fig:bnre:log_posterior}, we observe that the gap in both bias and variance reduces as the simulation budget increases. The bias gets close to $0$ for high simulation budgets, showing that the bias induced by BNRE vanishes as the simulation-budget increases.
A bias and variance analysis for all remaining benchmarks is discussed in Section \ref{sec:additional_figures}. 

\begin{figure}[t]
    \centering
    \includegraphics[width=\linewidth]{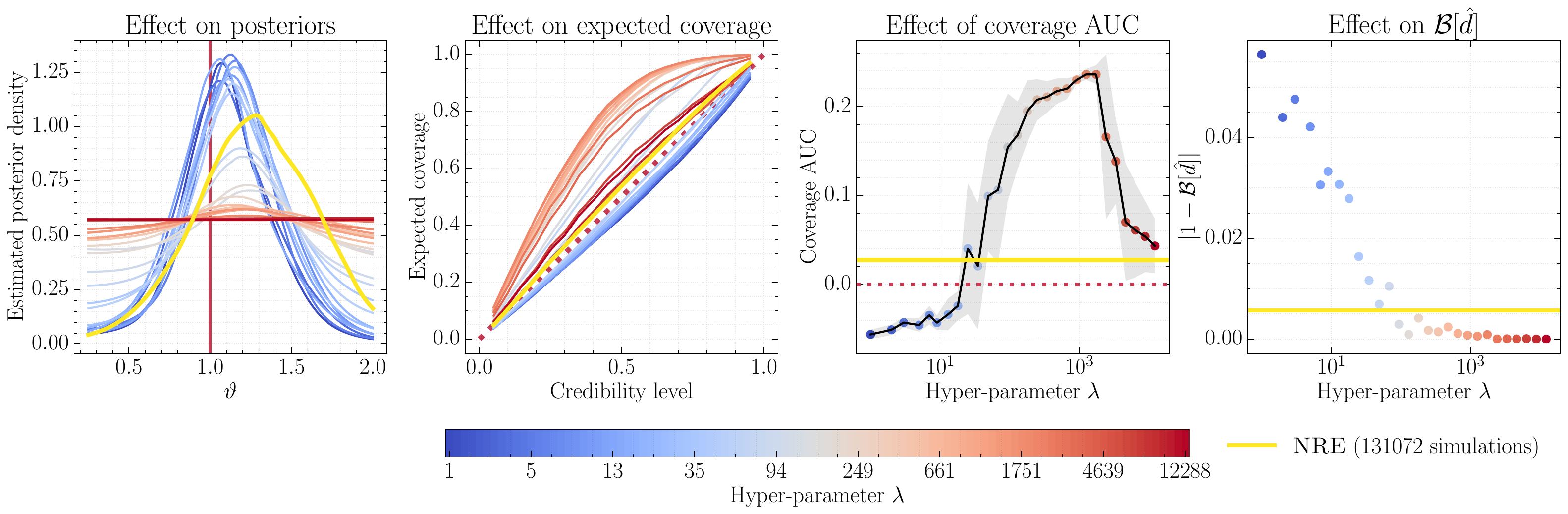}
    \caption{Effect of the hyper-parameter $\lambda$ for a fixed simulation budget of $1024$. The first plot from left to right shows the evolution of the approximate posterior for a given observation at a fixed $\btheta^*$, indicated by the red vertical line. This approximate posterior is compared to NRE trained on a large simulation budget, shown in yellow and serving as a proxy for the true posterior. The second plot illustrates the empirical expected coverage. The third plot provides a summarized view of the second plot using the coverage AUC as summary statistic. The fourth plot shows that classifiers are becoming increasingly more balanced as $\lambda$ increases. In addition, the plots show that $\lambda$ is directly tied to the statistical performance and reliability of the posterior approximations. Classifiers trained with small $\lambda$'s are associated with (relatively) tight posteriors and overconfident approximations, while classifiers trained with larger values of $\lambda$ are increasingly more dispersed and conservative until the posterior approximations reduce to the prior due to inflated statistical noise of the Monte Carlo estimation of the balancing condition. Furthermore, the expected coverage plot shows the estimator is almost perfectly calibrated and implicitly balanced. Immediately visible from the various posterior approximations in the leftmost subplot, is the fact that BNRE produces overconfident and biased approximations in the presence of a small simulation budget and a small $\lambda$, indicated by their dark blue color. However, the balancing condition can be applied to the underlying estimator to improve its reliability by increasing $\lambda$. 
    Ideally, $\lambda$ should be as small as possible to maximize predictive performance, while at the same time remain sufficiently large to guarantee coverage. From the 3\textsuperscript{th} subplot from the left, in this particular problem setting, that happens at the point where the coverage AUC transitions from being negative to positive ($\lambda \approx 25.0$).}
    \label{fig:hyper_parameter}
\end{figure}

\paragraph{Effects of $\lambda$}

Finally, Figure \ref{fig:hyper_parameter} shows the effect the hyper-parameter $\lambda$ on the posterior approximations, their expected coverage and the balancing condition. BNRE is run $5$ times for $\lambda$ ranging from $1$ to $2^{15}$ and for a fixed simulation budget of $1024$. 
Initially, the effect on the posterior approximations is limited for small values of $\lambda$. However, once $\lambda$ increases, the balancing condition forces the posterior approximations to become increasingly dispersed and conservative. Eventually, at least for this specific simulation budget, the posterior approximation reduces to the prior as
the balancing condition becomes dominant over the cross-entropy term. Although the global optimum remains unchanged as stated by Theorem \ref{thm:optimal_balancing}, large $\lambda$ values are likely to impair the training procedure. In particular, a large $\lambda$ can inflate the statistical noise of the Monte Carlo estimation of the balancing condition and make the classifier $\hat{d}$ degenerate to a classifier that is trivially balanced such as the random classifier $\hat{d}(\btheta, \bx)=0.5$ for all $\btheta, \bx$. In this case, $\hat{r}(\bx|\btheta)=1$ for all $\btheta, \bx$ and the approximate posterior degenerates to the prior. This effect is directly evident from Figure \ref{fig:hyper_parameter}, starting from $\lambda \simeq 1000$.
In practice, $\lambda$ should be sufficiently large such that the approximate classifier is balanced, while maximizing the statistical performance of the posterior estimator.
Therefore, we recommend to start with a small value for $\lambda$ and to gradually increase $\lambda$ until the posterior estimator becomes conservative. We empirically found $\lambda=100$ to be a reasonably good default value leading to good performance across all considered benchmarks with various model architectures.

\section{Related work}\label{sec:related}
In the Bayesian setting, BNRE improves the reliability of NRE by constraining the classifier hypothesis space to balanced classifiers, which results in more conservative posteriors.
Towards the same objective of conservative and reliable approximate posteriors, \citet{crisissbi} have shown empirically that ensembling posterior estimators increases their expected coverage. Since the two solutions are complementary, we suggest that ensembling BNRE is a safe practice to follow. To the best of our knowledge, no other related work exists to make Bayesian simulation-based inference algorithms more conservative and reliable.

In the frequentist setting, \citet{cranmer2015approximating} make use of neural ratio estimation to learn likelihood ratio test statistics. They show that the classifier $\hat{d}$ does not need to be exact for the statistic to remain the most powerful, provided that the approximate likelihood ratio is monotonic with exact likelihood ratio. When this is not the case, robust inference remains possible by calibrating the classifier, at the price of a loss in statistical power. Similarly, for frequentist likelihood-free inference, \citet{dalmasso2020confidence} use classifiers to estimate likelihood ratio statistics and propose a procedure for guaranteeing valid hypothesis tests and confidence sets. Finally, \citet{dalmasso2021likelihood} propose a practical procedure for the Neyman construction of confidence sets with finite-sample guarantees of nominal coverage as well as diagnostics that estimate conditional coverage over the entire parameter space.

In this work, we make the assumption that the simulator is well-specified, in the sense that it accurately models the real data generation process. However, this assumption is often violated. To overcome this issue, Generalized Bayesian inference (GBI) extends Bayesian inference by replacing the likelihood term by with arbitrary loss function \citep{bissiri2016general}. Those loss functions can be designed to mitigate specific types of misspecifications and enable robust inference, even with intractable likelihoods \citep{schmon2020generalized, matsubara2021robust, pacchiardi2021generalized}. Power likelihood losses have also been shown to increase robustness to model misspecification \citep{grunwald2017inconsistency}. It consists in raising the likelihood to a power to control the impact it has over the prior. The lower the power of likelihood, the lower the importance given to the data and the higher the uncertainty of the posterior. It can either be set based on practitioner knowledge or derived from observed data \citep{holmes2017assigning}. Following the same objective, \citet{miller2018robust} introduce coarsened posteriors that condition on a neighborhood of the empirical data distribution rather than on the data itself. This neighborhood is derived from a distance function that, when set to the relative entropy, allows the approximation of coarsened posteriors by a power posterior. Recently, \citet{dellaporta2022robust} applied  Bayesian non-parametric learning to SBI, making inference with misspecified simulator models both robust and computationally efficient. 

\section{Conclusions and future work}\label{sec:conclusion}

In this work, we introduced Balanced Neural Ratio Estimation (BNRE), a variation of neural ratio estimation designed to produce more conservative posterior estimators, even when the likelihood-to-evidence ratio estimator is not computationally faithful.
We provide theoretical arguments suggesting that enforcing the balancing condition should lead to more conservative posteriors without sacrificing exactness in the large simulation budget regime. Our theoretical results are experimentally validated on benchmarks of varying complexity.

Nevertheless, our inference algorithm comes with limitations that practitioners should keep in mind.
First, we emphasize that theorems \ref{thm:conservativeness_1} and \ref{thm:conservativeness_2} hold only in expectation, which means that we cannot provide any guarantee at the level of single inferences. Second, the balancing condition is enforced through a regularization penalty that is not estimated exactly.
This implies that the classifier $\hat{d}$ is rarely strictly balanced, although close to be, in which case theorems \ref{thm:conservativeness_1} and \ref{thm:conservativeness_2} do not hold. Third, the benefits of BNRE remain to be assessed in high-dimensional parameter spaces. In particular, the posterior density must be evaluated on a discretized grid over the parameter space to compute credibility regions, which currently prohibits the accurate computation of expected coverage in the high-dimensional setting.
In conclusion, BNRE should not be viewed as a way to obtain conservative posterior estimators with 100\% reliability, but rather as a way to increase the reliability of the posterior estimators with minimal effort and no computational overhead.

Looking forward, the balancing condition could potentially be applied to other simulation-based inference algorithms. Future works could include a generalization to neural posterior estimation (NPE). In fact, the likelihood-to-evidence ratio can be extracted from an approximate posterior by removing its dependence on the prior, $\log \hat{r}(\bx \vert \btheta) = \log \hat{p}(\btheta \vert \bx) - \log p(\btheta)$, which in turn can be expressed as a classifier $\hat{d}(\btheta, \bx) = \sigma(\log \hat{r}(\bx \vert \btheta))$ on which the balancing condition can be evaluated and enforced.
Although our work focuses on amortized approximate inference, the balancing condition could also be applied to sequential inference algorithms to increase their reliability.

Finally, although our initial motivation is framed within the field of simulation-based inference, our theoretical results are directly applicable to any binary classification task by replacing the joint and marginal distributions in the balancing condition with the distributions of the two considered classes. Therefore, it provides an easy-to-implement modification for high-risk classification problems.

\section{Additional results}\label{sec:additional_figures}
Figure \ref{fig:auc_grid} shows the coverage AUC for various simulation budgets. The mean and standard deviation over $5$ runs are reported.
\begin{figure}[H]
    \centering
    \includegraphics[width=\textwidth]{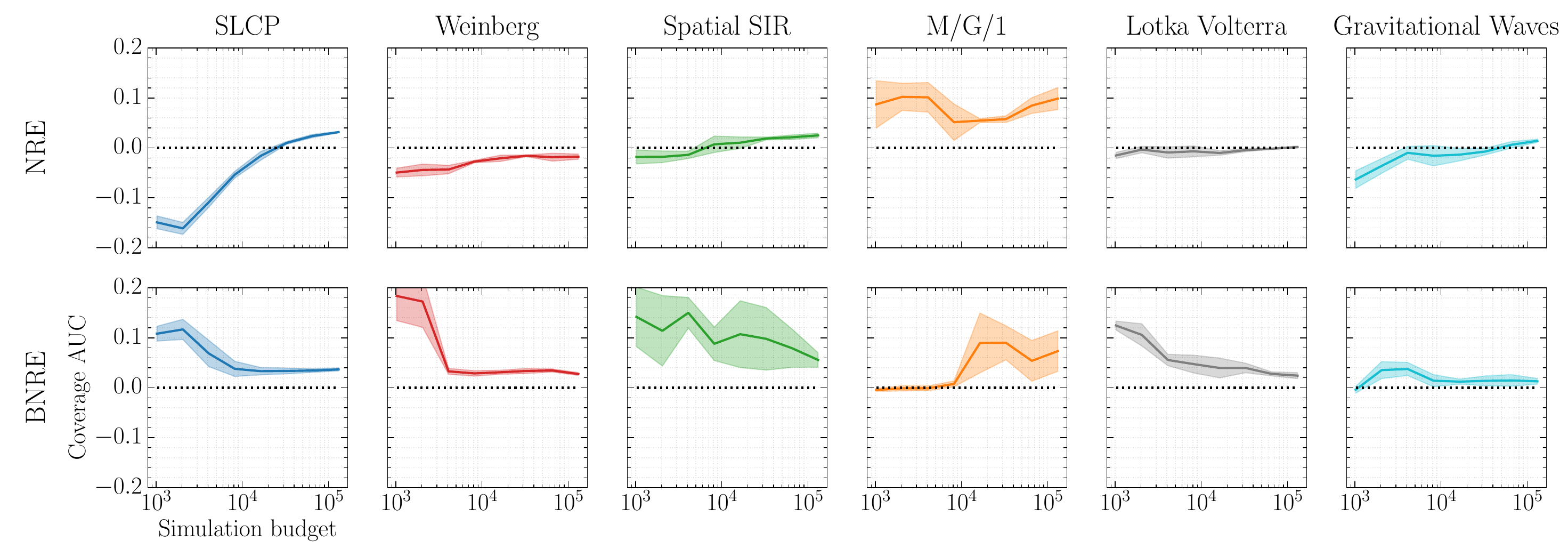}
    \caption{Coverage AUC measures the integrated signed area between the expected coverage curve and the diagonal. A perfectly calibrated posterior has an expected coverage probability equal to the nominal coverage probability, producing a diagonal line and has a coverage AUC of $0$, as shown on the left subplot. A conservative estimator on the other hand has a coverage AUC larger than $0$ and an overconfident estimator smaller than $0$. We observe that while NRE can produce coverage AUC both below or above $0$, BNRE always produces a coverage AUC larger than $0$, implying that its posterior approximations are conservative. Solid lines represent the mean over $5$ runs and shaded areas represent the standard deviation.} 
    \label{fig:auc_grid}
\end{figure}

Figure \ref{fig:bias_variance} shows the evolution of the bias and variance w.r.t. the simulation budget on a wide variety of benchmarks. We observe that observations made on Weinberg in Section \ref{sec:bnre:experiments} generalize to all benchmarks. The variance obtained with BNRE is always higher or equal than the one obtained with NRE as suggested by Theorems \ref{thm:conservativeness_1} and \ref{thm:conservativeness_2}. In addition, as suggested by Theorem \ref{thm:optimal_balancing}, the bias and variance obtained with BNRE converge, as NRE, to the Bayes optimal solution.
\begin{figure}[H]
    \centering
    \includegraphics[width=\textwidth]{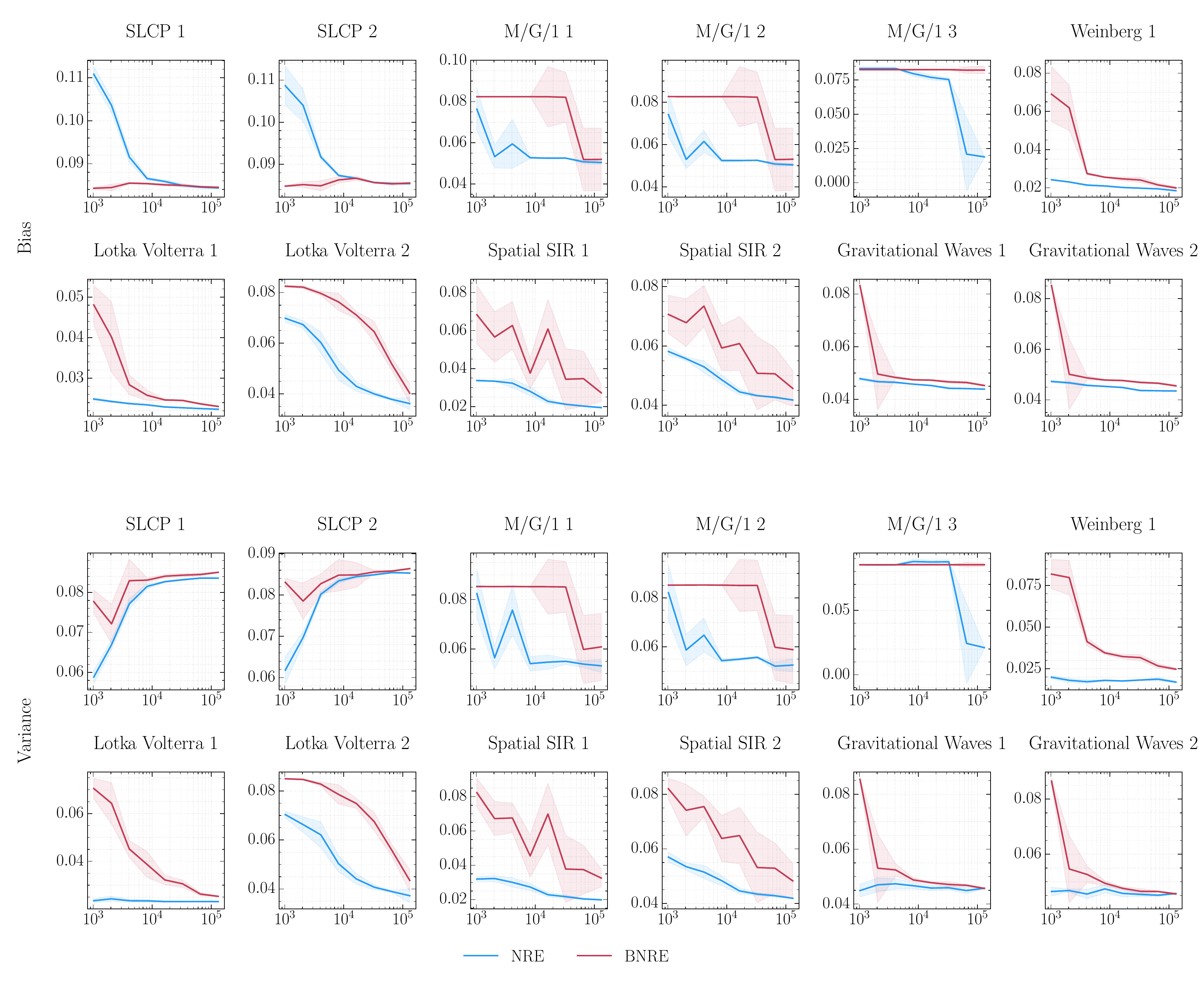}
    \caption{Evolution of the bias and variance w.r.t. the simulation budget. The bias and variance are estimated as described in Section \ref{sec:bnre:experiments} and are scaled to account for the prior's spread, permitting a direct comparison between the benchmarks. Marginals are considered when dealing with multidimensional parameter spaces. Those are denoted by an index following the benchmark name.} 
    \label{fig:bias_variance}
\end{figure}

\FloatBarrier

\begin{epiloguebox}
    We introduced the Balanced Neural Ratio Estimation (BNRE) algorithm that enforces the balancing condition through regularization. Empirical evaluation shows that enforcing the balancing condition leads to more conservative posterior approximations. This comes at the cost of slightly hurting the predictive performance, as computed by the expected nominal log posterior density. Nevertheless, we showed that enforcing the balancing condition does not prevent attaining the Bayes optimum as the Bayes optimal classifier is always balanced. This is also observed empirically as we see that BNRE achieves similar predictive performance as NRE as the simulation budget increases. In addition to empirical results, we present in Theorems \ref{thm:conservativeness_1} and \ref{thm:conservativeness_2} some properties of balanced classifiers that provide intuition about why enforcing balancing leads to more conservative approximations. However, those properties only give intuition and do not guarantee conservativeness as they only hold in expectation, are not statements on the coverage directly, and involve expectations of ratios that are not equal to ratios of expectations. While not guaranteeing conservativeness, BNRE is very easy to implement as it only involves adding a few lines of code in the loss function to switch from NRE to BNRE, and it does not introduce any computational overhead. BNRE is, hence, a simple fix to try to improve the expected coverage of posterior approximations. However, NRE is rarely used in practice as most applicative pieces of work approximate the posterior directly with normalizing flows.
\end{epiloguebox}

  \chapter{Balancing simulation-based inference}\label{c:balancing_sbi}
  \begin{prologuebox}
    This chapter is based on the following publication: \emph{Delaunoy, A.$^*$, Miller, B. K.$^*$, Forré, P., Weniger, C., \& Louppe, G. Balancing Simulation-based Inference for Conservative Posteriors. In Fifth Symposium on Advances in Approximate Bayesian Inference.}

    Previous work has shown that balancing classifiers in NRE algorithms leads to more conservative posterior approximations. However, the most widely used methods at the time of writing are NPE methods. This piece of work aims to extend the applicability of balancing to other types of algorithms, including NPE and NRE-C.
\end{prologuebox}

\section{Introduction}
\label{sec:intro}

Simulation-based inference (SBI) \citep{cranmer2020frontier} is a statistical inference framework that solves the inverse problem of identifying which parameter $\btheta$ generated observation $\bx$ by approximating the posterior $p(\btheta \mid \bx)$ with a surrogate model $\phat(\btheta \mid \bx)$. $\phat$ is constructed from simulated pairs $(\btheta, \bx)$ produced by a generative model where the likelihood $p(\bx \mid \btheta)$ is only implicitly defined. A classic method to produce samples from the surrogate is a rejection sampling technique called Approximate Bayesian Computation \citep{sisson2018overview}. Recently, there has been significant development of algorithms using machine learning for estimating the posterior \citep{papamakarios2016fast, lueckmann2017flexible, greenberg2019automatic, glockler2021variational, sharrock2022sequential, geffner2022score}, which we call Neural Posterior Estimation (NPE); the likelihood \citep{papamakarios2019sequential, gratton2017glass}; or the likelihood-to-evidence ratio \citep{thomas2016likelihood, tran2017hierarchical, 2019arXiv190304057H, durkan2020contrastive, miller2021truncated, miller2022contrastive}, which we call Neural Ratio Estimation (NRE).

It has been demonstrated that the estimated surrogate $\phat(\btheta \mid \bx)$ can be more confident than $p(\btheta \mid \bx)$ using common SBI algorithms \citep{crisissbi}. This poses a problem for the reliability of SBI in a scientific setting where surrogates must be conservative, i.e. avoid inaccurately excluding parameters at a given credibility level. There has been development in testing for overconfidence using empirical expected coverage and related methods \citep{cook2006validation, sbc, crisissbi}. \citet{lemos2023sampling} extend expected coverage testing to be a sufficient condition for posterior surrogate correctness, using only samples from the posterior surrogate. In an algorithmic approach to encourage conservativeness, \citet{delaunoytowards} found that the so-called balance condition regularizes overconfidence in expectation in surrogates trained with NRE \citep{2019arXiv190304057H}. Similarly, it has been shown that ensembling reduces overconfidence \citep{alsing2019fast, crisissbi}. \citet{zhao2021diagnostics} rather test for valid local coverage. \citet{linhart2022validation} focuses on normalizing flows and extends this method to the multivariate setting. In a similar fashion \citep{dalmasso2020confidence, dalmasso2021likelihood, masserano2022simulation} aim to produce valid frequentist coverage. \citet{cannon2022investigating} empirically show that model misspecification leads to overconfident posterior approximations and that this can be mitigated using ensembling or sharpness-aware minimization techniques.

\paragraph{Contribution}
After providing some background, we generalize the balancing condition to NPE methods and Contrastive Neural Ratio Estimation (NRE-C) \citep{miller2022contrastive}. We provide empirical evidence of the regularizing effect of the balance condition in all of these settings and the first expected coverage tests of NRE-C. Additionally, we relate the balance condition to the $\chi^2$-divergence. Code is available at \href{https://github.com/ADelau/balancing_sbi}{\texttt{https://github.com/ADelau/balancing\_sbi}}.

\section{Background}
\label{sec:bnpe:background}
\paragraph{Posterior estimation (NPE)}
A density estimator $q_{\bw}(\btheta \mid \bx)$ with weights $\bw$, such as a mixture density network \citep{bishop1994mixture} or normalizing flow \citep{papamakarios2019normalizing}, approximates $p(\btheta \mid \bx)$ when the expected Kullback-Leibler divergence
\begin{align}
    \label{eqn:npe-loss}
    \E_{p(\bx)}\left[ \DKL(p(\btheta \mid \bx) \mid q_{\bw}(\btheta \mid \bx) \right],
\end{align}
is minimized. In NPE, the surrogate model is directly $\phat(\btheta \mid \bx) \coloneqq q_{\bw}(\btheta \mid \bx)$.

\paragraph{Ratio estimation (NRE)}
The likelihood-to-evidence ratio $r(\btheta, \bx) \coloneqq \frac{p(\bx \mid \btheta)}{p(\bx)} = \frac{p(\btheta, \bx)}{p(\btheta) p(\bx)}$ is estimated through a supervised learning task using classifier $\varpi(y=1 \mid \btheta, \bx)$. The target conditional distribution $\pi(y=1 \mid \btheta, \bx)$ comes from  $\pi(\btheta, \bx, y) \coloneqq \pi(\btheta, \bx \mid y) \pi(y)$ where the marginals are set to ${\pi(y=0)} \coloneqq {\pi(y=1)} \coloneqq \frac{1}{2}$ and the remaining conditional is defined as
\begin{align}
    \pi(\btheta, \bx \mid y) \coloneqq
    \begin{cases}
        p(\btheta) p(\bx) & y=0 \\
        p(\btheta, \bx) & y=1
    \end{cases}.
\end{align}
The classifier $\varpi(y=1 \mid \btheta, \bx) \coloneqq \sigma \circ f_{\bw}(\btheta, \bx)$ is parameterized by a neural network $f_{\bw}(\btheta, \bx)$ and $\sigma$ is the sigmoid. $\varpi(y=1 \mid \btheta, \bx)$ approximates $\pi(y=1 \mid \btheta, \bx)$ when the NRE loss
\begin{align}
\label{eqn:nre-loss}
    \E_{\pi(\btheta, \bx)} \bigg[ \DKL(\pi(y \mid \btheta, \bx) \Mid \varpi (y \mid \btheta, \bx)) \bigg],
\end{align}
is minimized. Let $\phat(\btheta \mid \bx) \coloneqq \frac{\exp \circ f_{\bw}(\btheta, \bx)}{Z_{\bw}(\bx)} p(\btheta)$, with $Z_{\bw}(\bx) \coloneqq \int \exp \circ  f_{\bw}(\btheta, \bx) p(\btheta) d\btheta$, define the surrogate model. When the loss is zero $f_{\bw}(\btheta, \bx) = \log r(\btheta, \bx)$, recovering the posterior.

\paragraph{Contrastive Neural Ratio Estimation (NRE-C)}
\citep{miller2022contrastive} introduce NRE-C, an algorithm featuring a flexible, multiclass distribution $\pitilde(y)$ for $y = 0, 1, \ldots, K$.

 We let marginal distribution $\pitilde(y)$ with $y \in \{0, 1, \ldots, K\}$ have probabilities $\pitilde(y=k) \coloneqq \pitilde_{K}$ for all $k \geq 1$ and $\pitilde(y=0) \coloneqq \pitilde_{0}$ which yields the relationship $\pitilde_0 = 1 - K \pitilde_{K}$. The remaining conditional is
\begin{align}
    \pitilde(\bTheta, \bx \mid y = k) &\coloneqq
    \begin{cases}
        p(\btheta_1) \cdots p(\btheta_K) p(\bx) & k=0 \\
    	p(\btheta_1) \cdots p(\btheta_K) p(\bx \mid \btheta_k) & k = 1, \ldots, K
    \end{cases},
\end{align}
where $\bTheta \coloneqq (\btheta_1, ..., \btheta_K)$ are contrastive parameters, sampled from the prior $p(\btheta)$. We fit the variational, multiclass classifier 
\begin{align}
    \varpitilde(y = k \mid \bTheta, \bx) &\coloneqq
    \begin{cases}
    	\frac{K}{K + \sum_{i=1}^{K} \exp \circ h_{\bw}(\btheta_i, \bx)} & k = 0 \\
    	\frac{\exp \circ h_{\bw}(\btheta_k,\bx))}{K + \sum_{i=1}^{K} \exp \circ h_{\bw}(\btheta_i,\bx)} & k = 1, \ldots, K
    \end{cases}
\end{align}
by minimizing $\E_{\pitilde(\bTheta, \bx)} \left[ \DKL(\pitilde(y \mid \bTheta, \bx) \mid \varpitilde(y \mid \bTheta, \bx)) \right]$, where $h_{\bw}$ is a neural network parameterized by weights $\bw$. We write the loss function out explicitly:
\begin{align}
    \label{eqn:nre-c-loss}
    \begin{split}
    L[\varpitilde] \coloneqq &-\frac{1}{1 + \gamma} \mathbb{E}_{\pitilde(\bTheta, \bx \mid y=0)} \left[
    \log \varpitilde(y =0 \mid \bTheta, \bx) 
    \right] \\
    &- \frac{\gamma}{1 + \gamma} \mathbb{E}_{\pitilde(\bTheta, \bx \mid y=K)} \left[ 
    \log \varpitilde(y = K \mid \bTheta, \bx) 
    \right],
    \end{split}
\end{align}
where we introduced $\gamma \coloneqq \frac{\pitilde(y \geq 1)}{\pitilde(y=0)} = \frac{K \pitilde_{K}}{\pitilde_{0}}$. There are only two terms in this sum because we exploited symmetries in the conditionals $\pitilde(\bTheta, \bx \mid y = k)$ when $k \neq 0$. We define the surrogate model as $\phat(\btheta \mid \bx) \coloneqq \frac{\exp \circ h_{\bw}(\btheta,\bx)}{Z_{\bw}(\bx)} p(\btheta)$ with $Z_{\bw}(\bx) \coloneqq \int \exp \circ h_{\bw}(\btheta,\bx) p(\btheta) \, d\btheta$.

When the objective \eqref{eqn:nre-c-loss} becomes zero, the NRE-C surrogate model recovers the posterior. NRE-C represents a strict generalization of classifier-based, likelihood-to-evidence ratio estimation methods \citep{2019arXiv190304057H, durkan2020contrastive}. In the experiments we took $\gamma \coloneqq 1$ and $K \coloneqq 5$.

\paragraph{Conservative surrogates}
Undesirably, simulation-based inference algorithms can produce overconfident surrogate models \citep{crisissbi}. We define overconfidence in terms of the $(1-\alpha)$ expected coverage probability of the posterior surrogate $\phat(\btheta \mid \bx)$,
\begin{equation}
    \label{eqn:expected-coverage-probability}
    1 - \alphahat[\phat; \alpha] \coloneqq \E_{p(\btheta, \bx)} \left[ \mathds{1}(\btheta \in \Theta_{\phat(\btheta \mid \bx)}(1 - \alpha) ) \right],
\end{equation}
where $\indicator$ is an indicator function, and $\Theta_{\phat(\btheta \mid \bx)}(1 - \alpha)$ yields the $(1 - \alpha)$ highest posterior density region (HPDR) of $\phat(\btheta \mid \bx)$ with $\alpha \in [0, 1]$. The quantity $(1 - \alpha)$ is called the nominal coverage probability. When $\exists \alpha': 1 - \alphahat[\phat; \alpha'] < 1 - \alpha'$, we say that $\phat(\btheta \mid \bx)$ is overconfident. Overconfidence is problematic because the surrogate tends to exclude parameter values that are actually plausible at the considered credibility level. On the other hand, extremely underconfident surrogates are not informative.
Although there is a tradeoff, scientific applications take a cautious approach by favoring underconfidence. Therefore, we encourage conservative surrogates at credibility level $\alpha'$, which have $1 - \alphahat[\phat; \alpha'] \geq 1 - \alpha'$. We imprecisely define the relative conservativeness of one posterior to another: One surrogate is more conservative than another when there are ``more credibility levels'' at which it is conservative than the other.

\paragraph{Balance condition}
In an effort to produce conservative surrogates, \citet{delaunoytowards} introduced the balance condition. It holds for any classifier $\varpi(y=1 \mid \btheta, \bx)$ that satisfies
\begin{align}
\label{eqn:balance-condition}
\begin{aligned}
    1 = \E_{p(\btheta)p(\bx)}\left[ \varpi(y=1 \mid \btheta, \bx) \right] + \E_{p(\btheta, \bx)} \left[ \varpi(y=1 \mid \btheta, \bx) \right].
\end{aligned}
\end{align}
\citet{delaunoytowards} show that $\E_{p(\btheta, \bx)} \left[ \frac{\pi(y=1 \mid \btheta, \bx)}{\varpi(y=1 \mid \btheta, \bx)} \right] \geq 1$ and $\E_{p(\btheta)p(\bx)} \left[ \frac{\pi(y=0 \mid \btheta, \bx)}{\varpi(y=0 \mid \btheta, \bx)} \right] \geq 1$ for balanced classifiers. In expectation, it implies the balanced classifier's probabilities $\varpi(y \mid \btheta, \bx)$ are closer to uniform than the target, encouraging $\rhat(\btheta, \bx)$ to be closer to 1. This brings to surrogate closer to the prior, $p(\btheta)$, which is conservative.

\section{Extending the balance condition}
\label{sec:extending-balance-condition}

We clarify and generalize the balance criterion: Identifying it with the $\chi^2$-divergence, and applying it to models that can evaluate the approximate posterior density.

\paragraph{Balance as divergence}
We encourage balance during training by regularizing the loss with a Lagrange multiplier; penalizing solutions that do not satisfy the balance criterion
\begin{align}
    \label{eqn:balance-criterion}
    B[\varpi] \coloneqq B(\bw) \coloneqq
    \left( \E_{p(\btheta)p(\bx)}\left[ \varpi(y=1 \mid \btheta, \bx) \right] + \E_{p(\btheta, \bx)} \left[ \varpi(y=1 \mid \btheta, \bx) \right] - 1 \right)^2,
\end{align}
where $\bw$ are the classifier weights. The effects of balance are clearer when \eqref{eqn:balance-criterion} is rewritten in the form of a $\chi^2$-divergence. The $\chi^2$-divergence \citep{sason2016f} is defined as
\begin{align}
    \label{eqn:chi2-divergence}
    \Dchi(\pi(y) \Mid \varpi(y)) \coloneqq \int \left( \frac{\varpi(y)}{\pi(y)} - 1 \right)^2 \pi(y) \, dy,
\end{align}
where $\varpi(y) \coloneqq \int \varpi(y \mid \btheta, \bx) \pi(\btheta, \bx) d\btheta \, d\bx$ is the marginal classifier. With the following steps, we identify that $B[\varpi] = \Dchi(\pi(y) \Mid \varpi(y))$. 

\begin{align}
\label{eqn:balance-is-chi2}
\begin{aligned}
    \Dchi(\pi(y) \Mid \varpi(y)) &\coloneqq \int \left( \frac{\varpi(y)}{\pi(y)} - 1 \right)^2 \pi(y) \, dy \\
    &= \left( \frac{\varpi(y=0)}{\pi(y=0)} - 1 \right)^2 \pi(y=0) + \left(\frac{\varpi(y=1)}{\pi(y=1)} - 1 \right)^2 \pi(y=1) \\
    &= \frac{(\varpi(y=0)-\pi(y=0))^2}{\pi(y=0)} + \frac{(\varpi(y=1) - \pi(y=1))^2}{\pi(y=1)} \\
    &= \frac{(\varpi(y=1)-\pi(y=1))^2}{\pi(y=0)} + \frac{(\varpi(y=1) - \pi(y=1))^2}{\pi(y=1)} \\
    &= (\varpi(y=1)-\pi(y=1))^2 \left( \frac{1}{\pi(y=0)} + \frac{1}{\pi(y=1)} \right) \\
    &= 4 (\varpi(y=1)-\pi(y=1))^2 \\
    &= (2 \varpi(y=1) - 1)^2 \\
    &= \left( 2 \int \varpi(y=1 \mid \btheta, \bx) \pi(\btheta, \bx) d\btheta \, d\bx - 1 \right)^2 \\
    &= \left( 2 \int \varpi(y=1 \mid \btheta, \bx) \left( \int \pi(\btheta, \bx \mid y) \pi(y) dy \right) d\btheta \, d\bx - 1 \right)^2 \\
    &= \left( \int \varpi(y=1 \mid \btheta, \bx) (\pi(\btheta, \bx \mid y=0) + \pi(\btheta, \bx \mid y=1)) d\btheta \, d\bx - 1 \right)^2 \\
    &= \left( \E_{p(\btheta)p(\bx)}\left[ \varpi(y=1 \mid \btheta, \bx) \right] + \E_{p(\btheta, \bx)} \left[ \varpi(y=1 \mid \btheta, \bx) \right] - 1 \right)^2
\end{aligned}
\end{align}
where we used $\pi(y=0)=\pi(y=1)=\frac{1}{2}$ and $\varpi(\btheta, \bx) \coloneqq \pi(\btheta, \bx)$. Therefore, a balanced classifier satisfies $\varpi(y) = \pi(y)$, i.e. balancing regularizes the marginal classifier towards the target distribution for $y$.

\paragraph{Balance criterion for alternative models}
The balance criterion regularizes marginal distribution $\varpi(y)$; this makes sense for NRE which defines $\varpi(y)$ and target marginal $\pi(y)$.
However, we are interested in regularizing objectives $L(\bw)$ that either do not introduce a binary auxiliary variable $y$, or use an alternative. In order to apply the balance criterion, we propose to use the same target distribution $\pi(\btheta, \bx, y)$ as NRE does and define a classifier in terms of the variational (unnormalized) posterior approximant $\qhat_{\bw}(\btheta \mid \bx)$. We approximate $r(\btheta, \bx) \coloneqq \frac{p(\btheta, \bx)}{p(\btheta)p(\bx)} = \frac{p(\btheta \mid \bx)}{p(\btheta)}$ with $\frac{\qhat_{\bw}(\btheta \mid \bx)}{p(\btheta)}$ which yields the classifier $\varpi(y=1 \mid \btheta, \bx; \qhat_{\bw}) \coloneqq \frac{\qhat_{\bw}(\btheta \mid \bx) / p(\btheta)}{1+\qhat_{\bw}(\btheta \mid \bx) / p(\btheta)}$ and regularize $L(\bw)$ with $B(\bw)$ and Lagrange multiplier $\lambda$
\begin{equation}
    \label{eqn:arbitrary-balanced-loss}
    L(\bw) + \lambda B(\bw).
\end{equation}
The main contribution is reformulating $B(\bw)$ to be expressed in terms of $\qhat_{\bw}$, which generalizes the balance condition to models which allow for approximate density evaluation!
We consider losses $L(\bw)$ that go to zero if and only if $\qhat_{\bw}(\btheta \mid \bx) = p(\btheta \mid \bx)$. When this is true, the balance condition \eqref{eqn:balance-condition} also holds since $B(\bw)$ becomes zero:
\begin{align}
\begin{aligned}
    B(\bw) &\coloneqq \left( \int (\pi(\btheta, \bx \mid y=0) + \pi(\btheta, \bx \mid y=1)) \varpi(y=1 \mid \btheta, \bx; \qhat_{\bw}) d\btheta \, d\bx - 1 \right)^2 \\
    &= \left(\int (p(\btheta)p(\bx) + p(\btheta, \bx)) \frac{\qhat_{\bw}(\btheta \mid \bx)/p(\btheta)}{1 + \qhat_{\bw}(\btheta \mid \bx)/p(\btheta)} d\btheta \, d\bx - 1 \right)^2 \\
\end{aligned}
\end{align}
That means for loss $L(\bw) = 0$ the regularized version \eqref{eqn:arbitrary-balanced-loss} also goes to zero; just like BNRE. As advised in \citet{delaunoytowards}, we use $\lambda=100$. In practice, this value should be tuned for best performance.

\paragraph{Balanced Neural Posterior Estimation (BNPE)}
We propose BNPE which regularizes NPE's objective \eqref{eqn:npe-loss} with the balance criterion to train a normalized density estimator. We have $\qhat_{\bw}(\btheta \mid \bx) \coloneqq q_{\bw}(\btheta \mid \bx)$. The corresponding classifier becomes $\varpi(y=1 \mid \btheta, \bx; \qhat_{\bw}) \coloneqq \frac{q_{\bw}(\btheta \mid \bx) / p(\btheta)}{1+q_{\bw}(\btheta \mid \bx) / p(\btheta)}$. Training minimizes \eqref{eqn:arbitrary-balanced-loss}, substituting the appropriate loss and classifier.

\paragraph{Balanced Contrastive Neural Ratio Estimation (BNRE-C)}
We propose BNRE-C which regularizes NRE-C's objective \eqref{eqn:nre-c-loss} with the balance criterion to train a ratio estimator. The density estimator is $\qhat_{\bw}(\btheta \mid \bx) \coloneqq \exp \circ h_{\bw}(\btheta, \bx) p(\btheta)$. The corresponding classifier becomes $\varpi(y=1 \mid \btheta, \bx; \qhat) \coloneqq \frac{\exp \circ h_{\bw}(\btheta , \bx) p(\btheta) / p(\btheta)}{1+\exp \circ h_{\bw}(\btheta , \bx) p(\btheta) / p(\btheta)} = \sigma \circ h_{\bw}(\btheta, \bx)$. Training minimizes \eqref{eqn:arbitrary-balanced-loss}, substituting the appropriate loss and classifier.

\section{Refining Balanced Neural Posterior Estimation}
\label{sec:refining}

We observe that BNPE sometimes struggles to minimize the balance criterion for low simulation budgets. In this section, we provide a way to simplify learning.

\subsection{Improving Balanced Neural Posterior Estimation}

Let us first observe that the prior $p(\btheta)$ is balanced:
\begin{align}
    & \int (\pi(\btheta, \bx \mid y=0) + \pi(\btheta, \bx \mid y=1)) \varpi(y=1 \mid \btheta, \bx) d\btheta \, d\bx \\
    =& \int (p(\btheta)p(\bx) + p(\btheta, \bx)) \frac{\hat{p}(\btheta| \bx)/p(\btheta)}{1+\hat{p}(\btheta| \bx)/p(\btheta)} d\btheta \, d\bx \\
    =& \int (p(\btheta)p(\bx) + p(\btheta, \bx)) \frac{p(\btheta)/p(\btheta)}{1+p(\btheta)/p(\btheta)} d\btheta \, d\bx = 1
\end{align}

BNRE can easily model the prior as this is achieved for $\varpi(y=1 \mid \btheta, \bx) = 0.5 \quad \forall \btheta, \bx$. In opposition, BNPE has to explicitly learn a balanced distribution. In order to ease the task of learning a balanced distribution, we propose to initialize the posterior surrogate to the prior distribution. We will refer to this algorithm as Initialized Balanced Neural Posterior Estimation (BNPE Init). 

\subsection{Intializing neural spline flows to the prior distribution}
\label{sec:init}

In this work, we use Neural Spline Flows \citep{durkan2019neural}. We here discuss how to initialize this architecture to the prior. A normalizing flow models a complex distribution as a sequence of transformations of some base distribution. Consequently, the flow models the prior in the two following scenarios. The transformations are all identity transformations and the base distribution is the prior, or the transformations are all identity transformations and a fixed transformation that maps the base distribution to the prior is added at the end of the flow.

In the following, we describe how to obtain the two core components: identity transformations and a transformation that maps the base distribution to the prior. We then discuss the advantages and drawbacks of the two methods. In the experiments, we used a transformation from the base distribution to the prior.

\paragraph{Neural spline transformations} \citep{durkan2019neural} are transformations defined by $K$ rational-quadratic functions, with boundaries set by $K+1$ knots denoted by $(x^{(k)}, y^{k})_{k=0}^K$. The boundaries are defined by $(x^{(0)}, y^{0}) = (-B, -B)$ and $(x^{(K)}, y^{K}) = (B, B)$. The knots are parametrized by two vectors $\theta^w$ and $\theta^h$ of length $K$. Those vectors are passed through a softmax and multiplied by $2B$ to define the bins' width and height. 

The rational-quadratic in the $k^{\text{th}}$ bin is defined as 
\begin{equation}
    \frac{\alpha^{(k)}(\xi)}{\beta^{(k)}(\xi)} = y^{(k)} + \frac{(y^{(k+1)} - y^{(k)})[s^{(k)}\xi^2 + \delta^{(k)}\xi(1-\xi)]}{s^{(k)} + [\delta^{(k+1)} + \delta^{(k)} - 2s^{(k)}] \xi(1-\xi)}.
\end{equation}
The terms $\left\{\delta^{(k)}\right\}_{k=1}^{K-1}$ define the derivatives at the internal points and are parametrized by the vector $\theta^d$ of length $K-1$. The other terms are defined as $s_k = (y^{k+1} - y^k)/(x^{k+1} - x^k)$ and $\xi = (x - x^k)/(x^{k+1} - x^k)$.

\paragraph{Initializing transformations to identity}
To let the spline transformation be close to identity, we need to initialize the knots such that all the weights and widths are the same. This can be done by initializing the vectors $\theta^w$ and $\theta^h$ to zeros for all conditionings. In addition, all $s^{(k)}$ are then equal to $1$. We also need $\delta^{(k)} = 1, \ \forall k$. In the implementation used \citep{Rozet_Zuko_2022}, the vector $\theta^d$ models the log derivatives and hence must be initialized to a vector full of zeros.

To achieve the identity transform, the outputs of the neural network $\theta^h$, $\theta^w$ and $\theta^d$ must be zeros at initialization for all conditionings. This is achieved when biases and weights are all set $0$. However, this initialization is not optimal for learning via stochastic gradient descent, so we aim for a trade-off between ease of training and closeness to the prior. We initialize the neural network using some standard initialization, set the biases to $0$ and divide the weights by $5$ to obtain a function close to an identity transformation.

\paragraph{Mapping the base distribution to the prior}
In our case, the base distribution is a normal distribution $\mathcal{N}(0,1)$ and the priors are uniform in all the benchmarks considered. Therefore, we need to define a mapping from a Uniform distribution $\mathcal{U}(a, b)$, which p.d.f. is denoted by $p_u(u)$, to a normal distribution $\mathcal{N}(0, 1)$, which p.d.f. is denoted by $p_n(n)$ and c.d.f denoted $F_n(n)$. Let us first consider an intermediate mapping to a Uniform distribution $\mathcal{U}(0, 1)$ which p.d.f is denoted $p_{\tilde{u}}(\tilde{u})$. 
Going from $u$ to $\tilde{u}$ can be achieved with the following transformation
\begin{equation}
    \tilde{u} = \frac{u + a}{b - a}.
\end{equation}
The Jacobian linked to this transformation is then
\begin{equation}
    \frac{1}{b - a}
\end{equation}

Going from $\tilde{u}$ to $n$ can be achieved with the following transformation
\begin{equation}
    \tilde{u} = F_n(n) \Leftrightarrow n = F_n^{-1}(\tilde{u}).
\end{equation}
The Jacobian linked to this transformation is then
\begin{align}
    \left| (F_n^{-1})'(\tilde{u}) \right| &= \left| \frac{1}{F_n'(F_n^{-1}(\tilde{u}))} \right|\\
    &= \frac{1}{p_n(F_n^{-1}(\tilde{u}))}.
\end{align}

\paragraph{Comparison of the different initialization schemes} 
We have considered two initialization schemes: using the prior as base distribution or adding a transformation that maps the base distribution to the prior distribution. Using the prior distribution as base distribution has the advantage to be easy to implement as it does not require defining a mapping between both distributions. However, we have empirically observed that modifying the base distribution can lead to worse performance. We hypothesize that this is due to the fact that all the considered benchmarks have a uniform prior while flows work better with a Gaussian base. Let us note that using such a transformation could be beneficial in the more general setting as it solves leakage! Leakage is the fact that NPE algorithms can lead to a posterior surrogate that has density outside of the prior support. This is a problem in sequential settings where this surrogate is used as a proposal for simulating new data points. The transformation from a normal distribution to the prior is a transformation that maps any distribution with infinite support to a distribution that has the same support as the prior. Therefore, leakage cannot happen when such a transformation is applied.

\section{Experiments}
\label{sec:bnpe:experiments}
We investigate whether our proposed algorithms BNPE and BNRE-C lead to more conservative surrogate models compared to their non-balanced counterparts. Our diagnostics test the empirical expected coverage probability \eqref{eqn:expected-coverage-probability} of considered algorithms trained with ($\lambda > 0$) and without ($\lambda = 0$) the balance criterion. We also benchmark BNRE \citep{delaunoytowards}, an algorithm that trains a neural network to minimize \eqref{eqn:arbitrary-balanced-loss} using \eqref{eqn:nre-loss} as $L(\bw)$. The empirical expected coverage tests are performed on the following set of benchmarks of varying difficulty.

The \emph{SLCP} simulator models a fictive problem with 5 parameters. The observable $\bx$ is composed of 8 scalars which represent the 2D-coordinates of 4 points. 
The coordinate of each point is sampled from the same multivariate Gaussian whose mean and covariance matrix are parametrized by $\btheta$. We consider an alternative version of the original task \citep{papamakarios2019sequential} by inferring the marginal posterior density of 2 of those parameters. In contrast to its original formulation, the likelihood is not tractable due to the marginalization.

The \emph{Weinberg} problem \citep{weinberg} concerns a simulation of high energy particle collisions $e^+e^- \to \mu^+ \mu^-$. The angular distributions of the particles can be used to measure the Weinberg angle $\bx$
in the standard model of particle physics, leading to an observable composed of 20 scalars. From the scattering angle, we are interested in inferring Fermi's constant $\btheta$.

The \emph{Spatial SIR} model \citep{crisissbi} involves a grid-world of susceptible,
infected, and recovered individuals. Based on initial conditions and the infection and recovery rate $\btheta$,
the model describes the spatial evolution of an infection.
The observable $\bx$ is a snapshot of the grid-world after some fixed amount of time. The grid used is of size 50 by 50. 

The \emph{Lotka-Volterra} population model \citep{lotka,volterra1926fluctuations} describes a process of interactions between a predator and a prey species. The model is conditioned on 4 parameters $\btheta$ which influence the reproduction and mortality rate of the predator and prey species. We infer the marginal posterior of the predator parameters from time series of $2001$ steps representing the evolution of both populations over time. The specific implementation is based on a Markov Jump Process as in \citet{papamakarios2019sequential}.

The \emph{Two Moons} \citep{greenberg2019automatic} simulator models a fictive problem with 2 parameters. The observable $\bx$ is composed of 2 scalars which represent the 2D-coordinates of a random point sampled from a crescent-shaped distribution shifted and rotated around the origin depending on the parameters' values. Those transformations involve the absolute value of the sum of the parameters leading to a second crescent in the posterior and hence making it multi-modal.

Table \ref{tab:bnpe:architectures} summarizes the architectures and hyper-parameters used for each benchmark. The architectures are separated into two parts: the embedding and the head networks. The embedding network compresses the observable into a set of features. The head network then uses those features concatenated with the parameters to produce the target quantity. The head can either be a classifier or a normalizing flow depending on the algorithm considered. The learning rate is scheduled during training. Table \ref{tab:bnpe:architectures} provides the initial learning rates. Those are then divided by $10$ each time no improvement was observed on the validation loss for $10$ epochs.

\begin{table}[h]
    \centering
    \begin{tabular}{lllllll}
        \toprule
        & SLCP & Weinberg & Lotka-V. & Spatial SIR & Two Moons \\
        \midrule
        \emph{Embedding network} & None &  None & CNN & CNN & None\\
        \emph{Embedding layers} & / &  / & $8$ & $8$ & / \\ 
        \emph{Embedding channels} & / &  / & $8$ & $16$ & / \\ 
        \emph{Convolution type} & / & / & Conv1D & Conv2D & / \\ 
        \emph{Classifier Head} & MLP & MLP & MLP & MLP & MLP \\
        \emph{Classifier layers} & $6$ & $6$ & $6$ & $6$ & $6$ \\
        \emph{Classifier hidden neurons} & $256$ &  $256$ & $256$ & $256$  & $256$ \\
        \emph{Flow Head} & NSF & NSF & NSF & NSF & NSF \\
        \emph{Flow layers} & $3$ & $3$ & $3$ & $3$ & $3$ \\
        \emph{Flow hidden neurons} & $256$ &  $256$ & $256$ & $256$ & $256$ \\
        \emph{Learning rate} & $0.001$ &  $0.001$ & $0.001$ & $0.001$ & $0.001$ \\ 
        \emph{Epochs} & $500$ &  $500$ & $500$ & $500$ & $500$ \\ 
        \emph{Batch size} & $256$ & $256$ & $256$ & $256$ & $256$ \\ 
        \emph{$\lambda$ (balanced algorithms)} & $100$ & $100$ & $100$ & $100$ & $100$ \\
        \emph{$\gamma$ (NRE-C)} & $1$ & $1$ & $1$ & $1$ & $1$ \\
        \emph{$K$ (NRE-C)} & $5$ & $5$ & $5$ & $5$ & $5$ \\
        \bottomrule
    \end{tabular}
    \caption{Architectures and training hyper-parameters}
    \label{tab:bnpe:architectures}
\end{table}

\begin{figure}
    \vspace{-1.4em}
    \centering
    \includegraphics[width=\textwidth]{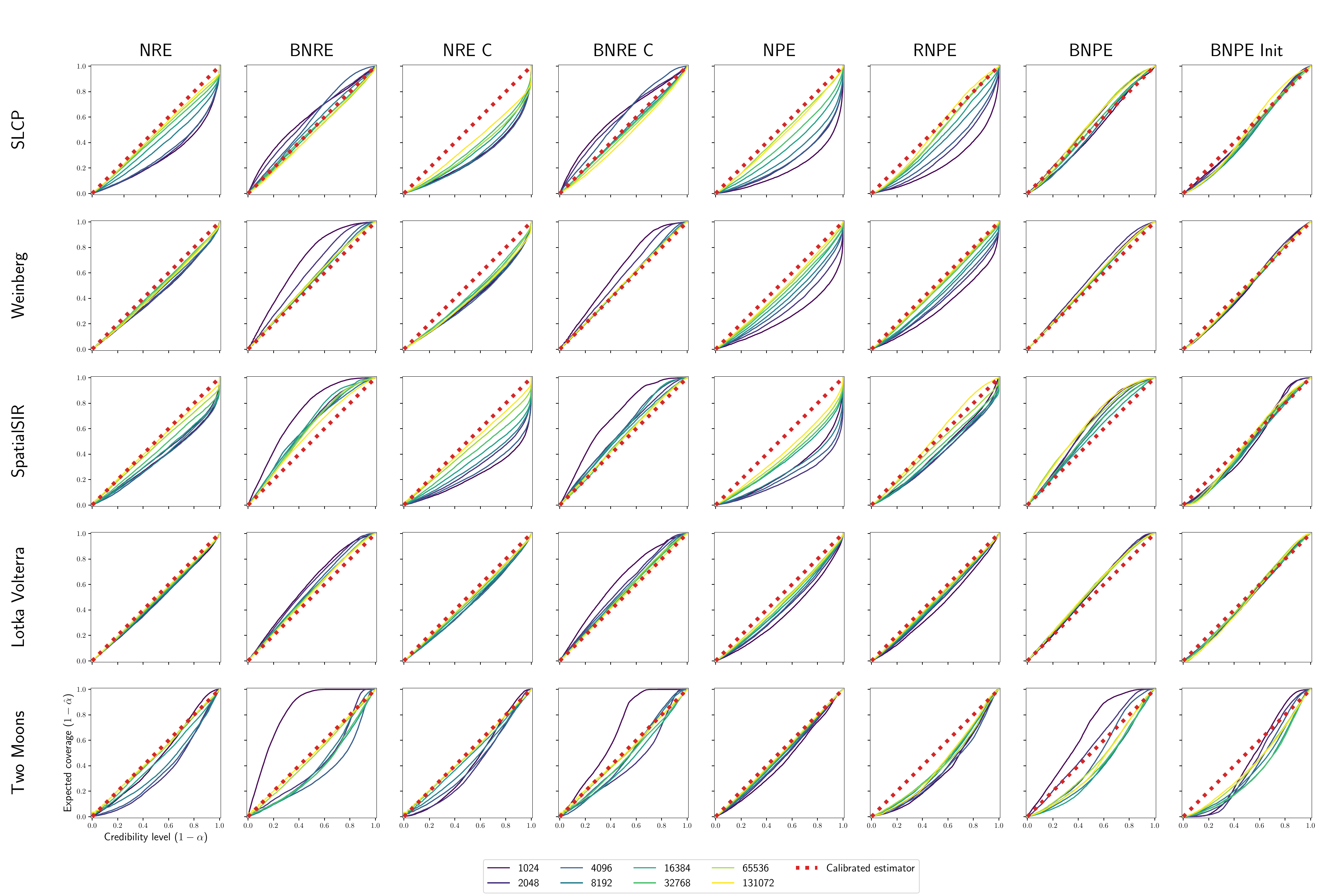}
    \caption{Empirical expected coverage for increasing simulation budgets. A perfectly calibrated surrogate has an expected coverage probability equal to the nominal coverage probability and 
    produces a diagonal line. A conservative surrogate has an expected coverage curve at or above the diagonal line.
    An overconfident surrogate produces curves below the diagonal line.
    Balanced algorithms tend to produce conservative surrogates. 
    5 runs are performed for each simulation budget with the median $\alphahat$ at each nominal credibility reported.
    }
    \label{fig:bnpe:coverage}
\end{figure}

\begin{figure}
    \vspace{-1.4em}
    \centering
    \includegraphics[width=\textwidth]{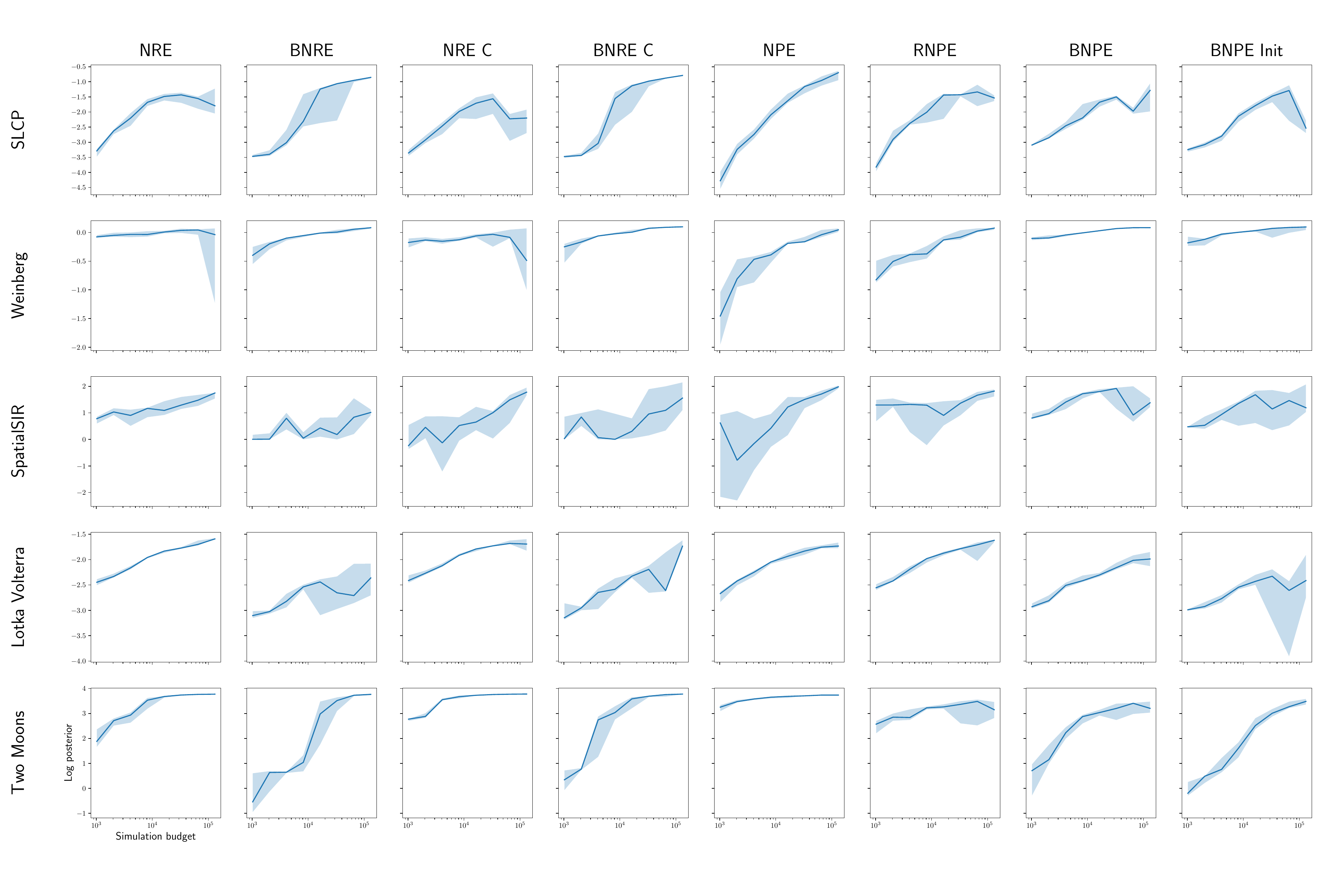}
    \caption{Nominal log posterior for increasing simulation budgets. 5 runs are performed for each simulation budget. Solid lines represent the median and the shaded areas represent the minimum and maximum. Larger values are desirable.}
    \label{fig:bnpe:log_posterior}
\end{figure}

\begin{figure}
    \vspace{-1.4em}
    \centering
    \includegraphics[width=\textwidth]{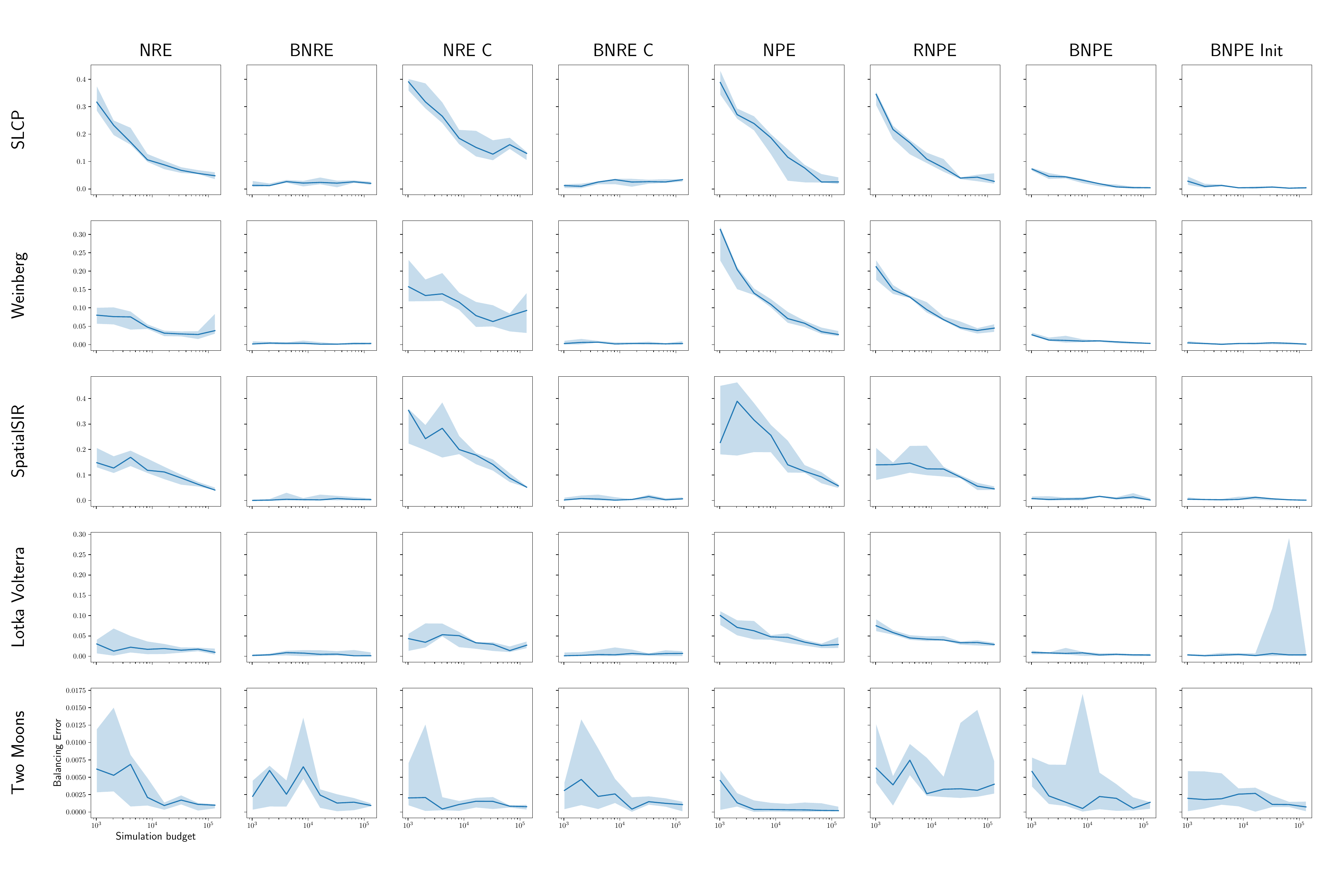}
    \caption{Balancing error for increasing simulation budgets. 5 runs are performed for each simulation budget. Solid lines represent the median and the shaded areas represent the minimum and maximum. Smaller values are desirable.}
    \label{fig:balancing_error_2}
\end{figure}

Expected coverage for these benchmarks are shown in \ref{fig:bnpe:coverage}. For both BNPE and BNRE-C, enforcing the balance criterion tends to produce conservative surrogates on most of the benchmarks, just as it did with BNRE. The only exception was Two Moons, which consistently produced overconfident surrogates for all algorithms. Taken as a whole, this provides some evidence that regularizing models allowing posterior density evaluation with the balance criterion makes them more conservative.
Our results suggest that enforcing the balance criterion did not negatively impact the informativeness of the surrogate for higher simulation budgets.
This is quantified in Figure \ref{fig:bnpe:log_posterior} where we estimate the information contained in the surrogates through the nominal log posterior, defined as
\begin{equation*}
    \E_{p(\btheta, \bx)} \left[ \log \phat(\btheta \mid \bx) \right].
\end{equation*}
We observe that balanced algorithms have a lower nominal log posterior than their corresponding non-balanced algorithms. However, they have similar nominal log posterior as the simulation budget increases. Figure \ref{fig:balancing_error_2} shows the balancing error, defined as
\begin{align*}
    & \left| \int (p(\btheta)p(\bx) + p(\btheta, \bx)) \varpi(y=1 \mid \btheta, \bx) \, d\btheta \, d\bx - 1 \right| \\
    =& \left| \int (p(\btheta)p(\bx) + p(\btheta, \bx)) \frac{\hat{p}(\btheta| \bx)/p(\btheta)}{1+\hat{p}(\btheta| \bx)/p(\btheta)} \, d\btheta \, d\bx - 1 \right|.
\end{align*}
Without surprise, we observe that enforcing the balancing condition leads to more balanced surrogates. Non-balanced algorithms show a high balancing error for low simulation budgets but this balancing error diminishes as the simulation budget gets higher. This is due to the fact that the posterior surrogate gets closer to the true posterior as the simulation increases and that the true posterior is always balanced. We also observe that normalizing flows struggle to minimize the balance criterion on the SLCP and Weinberg benchmarks for low simulation budgets. Normalizing flows must learn the prior as a multiplicative component in their density estimate, while likelihood-to-evidence ratio methods use the ground truth prior to construct the surrogate. It can be observed that initializing the posterior surrogate to the prior (BNPE init) indeed leads to a lower balancing error. Let us note that although the posterior surrogate is initialized to the prior, with a high enough simulation budget, it is able to learn a balanced surrogate that carries information about the parameter as seen from the log posterior quantity in Figure \ref{fig:bnpe:log_posterior}.

\section{Conclusions}

In this work, we have shown that balancing can be applied to any simulation-based inference algorithm that yields a posterior density estimator, hence extending the applicability of balancing. We have also shown that the balancing condition can be expressed as a $\chi^2$ divergence. We believe that this reformulation is a stepping stone towards a better understanding of the balancing condition and could inspire novel algorithms.

Let us note that there exist algorithms that do not directly provide the posterior density and hence do not fall into this framework. Some algorithms only aim to provide an unnormalized posterior approximation. This is the case of algorithms that model the likelihood \citep{papamakarios2019sequential} or an unnormalized version of the likelihood-to-evidence ratio \citep{durkan2020contrastive}. A direct consequence is that those algorithms are not necessarily balanced at optimum. Enforcing the balancing condition may hence prevent reaching the optimal solution. Some algorithms also only allow sampling from the posterior surrogate. This is the case of score-based methods \citep{song2020score, geffner2022score}. In such a setting, the balancing condition cannot be computed. Future work would then include a reformulation of the balancing condition that uses only samples from the posterior surrogate.

\FloatBarrier

\begin{epiloguebox}
    We showed that the balancing condition can be applied to any algorithm providing explicit posterior density estimation. This includes NPE methods and NRE variants that produce normalized posterior densities, such as NRE-A and NRE-C. However, balancing is more difficult to enforce for NPE methods. To circumvent this issue, we proposed a new initialization procedure for normalizing flows, leading to balanced approximations at initialization. We observed empirically that using this initialization led to more balanced approximations. With the rise of diffusion models allowing the modeling of high-dimensional distributions, adapting balancing to methods that only provide posterior sampling could be beneficial for reliable simulation-based inference with high-dimensional parameter spaces. 

    While balancing has gained attention from the community and is cited in subsequent methodological papers, most applicative papers that report overconfidence do not modify the used algorithms to address the issue \citep{reza2024constraining, swierc2024domain, morandini2024reconstructing}. This issue is not only about balancing, but all papers aiming at increasing conservativeness are at the time of writing, not used in practice. However, this line of work is still very recent. We hope that field scientists will become aware of it in the future and make use of those new tools to produce more reliable results.
\end{epiloguebox}

  \chapter{Simulation-based inference with Bayesian neural networks}\label{c:bnn}
  \begin{prologuebox}
    This chapter is based on the following publication: \emph{Delaunoy, A.$^*$, Bonardeaux, M. D. L. B.$^*$, Mishra-Sharma, S., \& Louppe, G. (2024). Low-Budget Simulation-Based Inference with Bayesian Neural Networks. arXiv preprint arXiv:2408.15136.}

    While balancing provides a simple way to improve the conservativeness of many simulation-based inference algorithms, it needs to be enforced through regularization. We show that this can be problematic in settings where very few simulations are available for training. In addition, the demonstrated properties of balanced classifiers only provide intuition about why increased conservativeness is observed, but no direct link with coverage or posterior uncertainty has been established.

    In this piece of work, we take a different approach and hypothesize that overconfidence is observed in classical methods because the uncertainty linked to approximation errors is not taken into account. We propose to use Bayesian neural networks to explicitly take this uncertainty into account. As opposed to the balancing line of work, this works with very low simulation budgets and, hence, has an increased range of applicability.
\end{prologuebox}

\section{Introduction}

Simulation-based inference aims at identifying the parameters of a stochastic simulator that best explain an observation.
In its Bayesian formulation, simulation-based inference approximates the posterior distribution of the model parameters given an observation. This approximation usually takes the form of a neural network trained on synthetic data generated from the simulator. 
In the context of scientific discovery, \citet{crisissbi} stressed the need for posterior approximations that are conservative -- not overconfident -- in order to make reliable downstream claims. They also showed that common simulation-based inference algorithms can produce overconfident approximations that may lead to erroneous conclusions. 

In the data-poor regime \citep{villaescusa2020quijote, zhang2023sensitivity, zeng2023probabilistic}, where the simulator is expensive to run and only a small number of simulations are available, training a neural network to approximate the posterior can easily lead to overfitting.
With small amounts of training data, the neural network weights are only loosely constrained, leading to high computational uncertainty~\citep{wenger2022posterior}.
That is, many neural networks can fit the training data equally well, yet they may have very different predictions on test data.
For this reason, the posterior approximation is uncertain and, in the absence of a proper quantification of this uncertainty, potentially overconfident.
Fortunately, computational uncertainty in a neural network can be quantified using Bayesian neural networks (BNNs) \citep{gal2016uncertainty}, which account for the uncertainty in the neural network weights. Therefore, in the context of simulation-based inference, BNNs can provide a principled way to quantify the computational uncertainty of the posterior approximation.

\citet{crisissbi} showed empirically that using ensembles of neural networks, a crude approximation of BNNs~\citep{lakshminarayanan2017simple}, does improve the calibration of the posterior approximation. A few studies have also used BNNs as density estimators in simulation-based inference \citep{cobb2019ensemble, walmsley2020galaxy, lemos2023robust}. However, these studies have remained empirical and limited in their evaluation.
This lack of theoretical grounding motivates the need for a more principled understanding of BNNs for simulation-based inference.
In particular, the choice of prior on the neural network weights happens to be crucial in this context, as it can strongly influence the resulting posterior approximation. Yet, arbitrary priors that convey little or undesired information about the posterior density have been used so far.

Our contributions are twofold. Firstly, we provide an improved understanding of BNNs in the context of simulation-based inference by empirically analyzing their effect on the resulting posterior approximations. Secondly, we introduce a principled way of using BNNs in simulation-based inference by designing meaningful priors. These priors are constructed to produce calibrated posteriors even in the absence of training data. We show that they are conservative in the small-data regime, for very low simulation budgets.  
The code is available at \url{https://github.com/ADelau/low_budget_sbi_with_bnn}.

\section{Background}\label{sec:bnn:background}
\paragraph{Simulation-based inference}\label{sec:sbi}
We consider a stochastic simulator that takes parameters $\btheta$ as input and produces synthetic observations $\bx$ as ouput. 
The simulator implicitly defines the likelihood $p(\bx | \btheta)$ in the form of a forward stochastic generative model but does not allow for direct evaluation of its density due to the intractability of the marginalization over its latent variables. In this setup, Bayesian simulation-based inference aims at approximating the posterior distribution $p(\btheta | \bx)$ using the simulator. Among possible approaches, \emph{neural} simulation-based inference methods train a neural network to approximate key quantities from simulated data, such as the posterior, the likelihood, the likelihood-to-evidence ratio, or a score function \citep{cranmer2020frontier}.

Recently, concerns have been raised regarding the calibration of the approximate posteriors obtained with neural simulation-based inference. \citet{crisissbi} showed that, unless special care is taken, common inference algorithms can produce overconfident posterior approximations. They quantify the calibration using the expected coverage
\begin{equation}
    \text{EC}(\hat{p}, \alpha) = \mathbb{E}_{p(\btheta, \bx)}[\mathds{1}(\btheta \in \bTheta_{\hat{p}}(\alpha))]
\end{equation}
where $\bTheta_{\hat{p}}(\alpha)$ denotes the highest posterior credible region at level $\alpha$ computed using the posterior approximate $\hat{p}(\btheta|\bx)$. The expected coverage is equal to $\alpha$ when the posterior approximate is calibrated, lower than $\alpha$ when it is overconfident and higher than $\alpha$ when it is underconfident or conservative.

The calibration of posterior approximations has been improved in recent years in various ways.
\citet{delaunoytowards, delaunoy2023balancing} regularize the posterior approximations to be balanced, which biases them towards conservative approximations. Similarly, \citet{falkiewicz2024calibrating} regularize directly the posterior approximation by penalizing miscalibration or overconfidence. \citet{masserano2022simulation} use Neyman constructions to produce confidence regions with approximate Frequentist coverage. \citet{patel2023variational} combine simulation-based inference and conformal predictions. \citet{schmitt2023leveraging} enforce the self-consistency of likelihood and posterior approximations to improve the quality of approximate inference in low-data regimes.

\paragraph{Bayesian deep learning} 
Bayesian deep learning aims to account for both the aleatoric and epistemic uncertainty in neural networks.
The aleatoric uncertainty refers to the intrinsic randomness of the variable being modeled, typically taken into account by switching from a point predictor to a density estimator.
The epistemic uncertainty, on the other hand, refers to the uncertainty associated with the neural network itself and is typically high in small-data regimes.
Failing to account for this uncertainty can lead to high miscalibration as many neural networks can fit the training data equally well, yet they may have very different predictions on test data.

Bayesian deep learning accounts for epistemic uncertainty by treating the neural network weights as random variables and considering the full posterior over possible neural networks instead of only the most probable neural network \citep{papamarkou2024position}. Formally, let us consider a supervised learning setting in all generality, where $\bx$ denotes inputs, $\by$ outputs, $\bD$ a dataset of $N$ pairs $(\bx, \by)$, and $\bw$ the weights of the neural network. The likelihood of a given set of weights is 
\begin{equation}
    p(\bD|\bw) \propto \prod_{i=1}^N p(\by_i | \bx_i, \bw),
\end{equation}
where $p(\by_i | \bx_i, \bw)$ is the output of the neural network with weights $\bw$ and inputs $\bx_i$.
The resulting posterior over the weights is
\begin{equation}
    p(\bw|\bD) = \frac{p(\bD|\bw)p(\bw)}{p(\bD)},
\end{equation}
where $p(\bw)$ is the prior. Once estimated, the posterior over the neural network's weights can be used for predictions through the Bayesian model average
\begin{equation}
    p(\by|\bx, \bD) = \int p(\by|\bx, \bw) p(\bw|\bD) d\bw \simeq \frac{1}{M} \sum_{i=1}^M p(\by|\bx, \bw_i), \bw_i \sim p(\bw|\bD).
    \label{eq:bayes_model_average}
\end{equation}
In practice, the Bayesian model average can be approximated by Monte Carlo sampling, with $M$ samples from the posterior over the weights. The quality of the approximation depends on the number of samples $M$, which should be chosen high enough to obtain a good enough approximation but small enough to keep reasonable the computational costs of predictions.

Estimating the posterior over the neural network weights is a challenging problem due to the high dimensionality of the weights.
Variational inference \citep{blundell2015weight} optimizes a variational family to match the true posterior, which is typically fast but requires specifying a variational family that may restrict the functions that can be modeled. Markov chain Monte Carlo methods \citep{welling2011bayesian, chen2014stochastic}, on the other hand, are less restrictive in the functions that can be modeled but require careful tuning of the hyper-parameters and are more computationally demanding. The Bayesian posterior can also be approximated by an ensemble of neural networks \citep{lakshminarayanan2017simple, pearce2020uncertainty, he2020bayesian}. Laplace methods leverage geometric information about the loss to construct an approximation of the posterior around the maximum a posteriori \citep{mackay1992bayesian}. Similarly, \citet{maddox2019simple} use the training trajectory of stochastic gradient descent to build an approximation of the posterior. 

\section{Bayesian neural networks for simulation-based inference}
In the context of simulation-based inference, treating the weights of the inference network as random variables enables the quantification of the computational uncertainty of the posterior approximation.
Considering neural networks taking observations $\bx$ as input and producing parameters $\theta$ as output, the posterior approximation $\hat{p}(\btheta|\bx)$ can be modeled as the Bayesian model average
\begin{equation}
    \hat{p}(\btheta|\bx) =  \int p(\btheta|\bx, \bw) p(\bw|\bD) d\bw,
\end{equation}
where $p(\btheta|\bx, \bw)$ is the posterior approximation parameterized by the weights $\bw$ and evaluated at $(\btheta, \bx)$, and $p(\bw|\bD)$ is the posterior over the weights given the training set $\bD$.

Remaining is the choice of prior $p(\bw)$. While progress has been made in designing better priors in Bayesian deep learning \citep{fortuin2022priors}, we argue that none of those are suitable in the context of simulation-based inference. To illustrate our point, let us consider the case of a normal prior $p(\bw) = \mathcal{N}(\boldsymbol{0}, \sigma^2 \boldsymbol{I})$ on the weights, in which case
\begin{equation}
    \hat{p}_{\text{normal prior}}(\btheta| \bx) = \int p(\btheta|\bx, \bw)\ \mathcal{N}(\bw|\boldsymbol{\mu} = \boldsymbol{0}, \boldsymbol{\Sigma} = \sigma^2 \boldsymbol{I}) d\bw.
\end{equation}
As mentioned in Section \ref{sec:bnn:background}, a desirable property for a posterior approximation is to be calibrated. Therefore we want $\text{EC}(\hat{p}_{\text{normal prior}}, \alpha) = \alpha, \forall \alpha$. Although it might be possible for this property to be satisfied in particular settings, this is obviously not the case for all values of $\sigma$ and all neural network architectures. Therefore, and as illustrated in Figure \ref{fig:prior_coverage}, the Bayesian model average is not even calibrated a priori when using a normal prior on the weights. As the Bayesian model average is not calibrated a priori, it cannot be expected that updating the posterior over weights $p(\bw|\bD)$ with a small amount of data would lead to a calibrated a posteriori Bayesian model average.

\subsection{Functional priors for simulation-based inference}\label{sec:func_prior}
We design a prior that induces an a priori-calibrated Bayesian model average. To achieve this, we work in the space of posterior functions instead of the space of weights. We consider the space of functions taking a pair ($\btheta$, $\bx$) as input and producing a posterior density value $f(\btheta, \bx)$ as output. Each function $f$ is defined by the joint outputs it associates with any arbitrary set of inputs, such that a posterior over functions can be viewed as a distribution over joint outputs for arbitrary inputs. Formally, let us consider $M$ arbitrary pairs $(\btheta, \bx)$ represented by the matrices $\bTheta = [\btheta_1, ..., \btheta_M]$ and $\bX = [\bx_1, ..., \bx_M]$ and let $\boldf = [f_1, ..., f_M]$ be the joint outputs associated with those inputs. The distribution $p(\boldf | \bTheta, \bX)$ then represents a distribution over posteriors $\boldf = [\tilde{p}(\btheta_1 | \bx_1), ..., \tilde{p}(\btheta_M | \bx_M)]$. The functional posterior distribution over posteriors for parameters $\bTheta$ and observations $\bX$ given a training dataset $\bD$ is then $p(\boldf | \bTheta, \bX, \bD)$ and the Bayesian model average is obtained through marginalization, that is
\begin{equation}
    p(\btheta_i | \bx_i, \bD) = \int f_i\ p(\boldf |\bTheta, \bX, \bD) d\boldf, \quad \forall i.
\end{equation}

Computing the posterior over functions requires the specification of a prior over functions. We first observe that the prior over the simulator's parameters is a calibrated approximation of the posterior. That is, for the prior function $p_{\text{prior}}: (\btheta, \bx) \to p(\btheta)$, we have that $\text{EC}(p_{\text{prior}}, \alpha) = \alpha, \forall \alpha$ \citep{delaunoy2023balancing}. It naturally follows that the a priori Bayesian model average with a Dirac delta prior around the prior on the simulator's parameters is calibrated
\begin{equation}
\begin{aligned}
    \hat{p}(\btheta_i | \bx_i) 
    &= \int f_i\ \delta([f_j = p_{\text{prior}}(\btheta_j, \bx_j)])\ d\boldf, \forall i \\
    &= \int f_i\ \delta(f_i = p(\btheta_i))\ df_i, \forall i \Rightarrow \text{EC}(\hat{p}, \alpha) = \alpha, \forall \alpha.
\end{aligned}
\end{equation}
However, this prior has limited support, and the Bayesian model average will not converge to the posterior $p(\btheta | \bx)$ as the dataset size increases. We extend this Dirac prior to include more functions in its support while retaining the calibration property, which we propose defining as a Gaussian process centered at $p_{\text{prior}}$.

A Gaussian process (GP) defines a joint multivariate normal distribution over all the outputs $\boldf$ given the inputs ($\bTheta$, $\bX$). It is parametrized by a mean function $\mu$ that defines the mean value for the outputs given the inputs and a kernel function $K$ that models the covariance between the outputs. If we have access to no data, the mean and the kernel jointly define a prior over functions as they define a joint prior over outputs for an arbitrary set of inputs. In order for this prior over functions to be centered around the prior $p_{\text{prior}}$, we define the mean function as $\mu(\btheta, \bx) = p(\btheta)$.
The kernel $K$, on the other hand, defines the spread around the mean function and the correlation between the outputs $\boldf$. Its specification is application-dependent and constitutes a hyper-parameter of our method that can be exploited to incorporate domain knowledge on the structure of the posterior. For example, periodic kernels could be used if periodicity is observed. Kernel's hyperparameters can also be chosen such as to incorporate what would be a reasonable deviation of the approximated posterior from the prior. We denote the Gaussian process prior over function outputs as $p_{\text{GP}}(\boldf|\mu(\bTheta, \bX), K(\bTheta, \bX))$. Proposition \ref{thm:gp_calibration} shows that a functional prior defined in this way leads to a calibrated Bayesian model average.

\begin{proposition}\label{thm:gp_calibration}
The Bayesian model average of a Gaussian process centered around the prior on the simulator's parameters is calibrated. Formally, let $p_\text{GP}$ be the density probability function defined by a Gaussian process, $\mu$ its mean function, and $K$ the kernel. Let us consider $M$ arbitrary pairs $(\btheta, \bx)$ represented by the matrices $\bTheta = [\btheta_1, ..., \btheta_M]$ and $\bX = [\bx_1, ..., \bx_M]$ and represent by the vector $\boldf = [f_1, ..., f_M]$ the joint outputs associated with those inputs. The Bayesian model average on the $i^{\text{th}}$ pair is expressed
\begin{equation*}
    \hat{p}(\btheta_i | \bx_i) = \int f_i\ p_{\text{GP}}(\boldf|\mu(\bTheta, \bX), K(\bTheta, \bX))\ d\boldf
\end{equation*}
If $\mu(\btheta, \bx) = p(\btheta), \forall \btheta, \bx$, then,
\begin{equation*}
    \mathrm{EC}(\hat{p}, \alpha) = \alpha, \forall \alpha,
\end{equation*}
for all kernel $K$.
\begin{proof}
As $p_\text{GP}$ is, by definition of a Gaussian process, a multivariate normal, the expectations of the marginals are equal to the mean parameters
\begin{equation*}
    \hat{p}(\btheta_i | \bx_i) = \mu(\btheta_i, \bx_i) = p(\btheta_i).
\end{equation*}
The joint evaluation of the Bayesian model average of the Gaussian process is hence equivalent to the joint evaluation of the prior for any matrices $\bTheta$ and $\bX$. We can therefore conclude that $\hat{p}$ is equivalent to $p_{\text{prior}}: (\btheta, \bx) \rightarrow p(\btheta)$. Since $\text{EC}(p_{\text{prior}}, \alpha) = \alpha, \forall \alpha$ \citep{delaunoy2023balancing}, then, $\text{EC}(\hat{p}, \alpha) = \alpha, \forall \alpha$.
\end{proof}
\end{proposition}

\subsection{From functional to parametric priors}
In this section, we now discuss how the functional GP prior over posterior density functions proposed in Section 3.1 can be used in the simulation-based inference setting. We follow \citet{flam2017mapping} and \citet{sun2018functional} for mapping the functional prior to a distribution over neural network weights. We note that other methods for functional Bayesian deep learning have been proposed in the literature. These methods could also be used in our setting and are discussed at the end of the section.

Let us first observe that a neural network architecture and a prior on weights jointly define a prior over functions. We parameterize the prior on weights by $\bphi$ and denote this probability density over function outputs by
\begin{equation}
\begin{aligned}
    p_{\text{BNN}}(\boldf \mid \bphi, \bTheta, \bX) 
    &= \int p(\boldf \mid \bw, \bTheta, \bX)\ p(\bw | \bphi)\ d\bw \\
    &= \int \delta([f_i = p(\btheta_i | \bx_i, \bw)])\ p(\bw | \bphi)\ d\bw.
\end{aligned}
\end{equation}
To obtain a prior on weights that matches the target GP prior, we optimize $\bphi$ such that $p_{\text{BNN}}(\boldf \mid \bphi, \bTheta, \bX)$ matches $p_{\text{GP}}(\boldf|\mu(\bTheta, \bX), K(\bTheta, \bX))$. Following \citet{flam2017mapping}, given a measurement set $\mathcal{M} = \{\btheta_i, \bx_i\}_{i=1}^M$ at which we want the distributions to match, the KL divergence between the two priors can be expressed as
\begin{equation}
\begin{aligned}
    & \text{KL} \left[p_\text{BNN}(\boldf \mid \bphi, \mathcal{M}) \dmid p_{\text{GP}}(\boldf \mid  \mu(\mathcal{M}), K(\mathcal{M})) \right] \\
    =\ &\int p_\text{BNN}(\boldf \mid \bphi, \mathcal{M}) \log\frac{p_\text{BNN}(\boldf \mid \bphi, \mathcal{M})}{p_{\text{GP}}(\boldf \mid  \mu(\mathcal{M}), K(\mathcal{M}))} d\by\\
    =\ &-\mathbb{H}\left[p_\text{BNN}(\boldf \mid \bphi, \mathcal{M})\right] - \mathbb{E}_{p_\text{BNN}(\boldf \mid \bphi, \mathcal{M})}\left[\log p_{\text{GP}}(\boldf \mid  \mu(\mathcal{M}), K(\mathcal{M}))\right],
\end{aligned}
\label{eq:kl_divergence}
\end{equation}
where the second term $\mathbb{E}_{p_\text{BNN}(\boldf \mid \bphi, \mathcal{M})}\left[\log p_{\text{GP}}(\boldf \mid  \mu(\mathcal{M}), K(\mathcal{M}))\right]$ can be estimated using Monte-Carlo. The first term $\mathbb{H}\left[p_\text{BNN}(\boldf \mid \bphi, \mathcal{M})\right]$, however, is harder to estimate as it requires computing $\log p_\text{BNN}(\boldf \mid \bphi, \mathcal{M})$, which involves the integration of the output over all possible weights combinations. To bypass this issue, \citet{sun2018functional} propose to use Spectral Stein Gradient Estimation (SSGE) \citep{shi2018spectral} to approximate the gradient of the entropy as
\begin{equation}
    \nabla \mathbb{H}\left[p_\text{BNN}(\boldf \mid \bphi, \mathcal{M})\right] \simeq \text{SSGE}\left(\boldf_1, ..., \boldf_N \sim p_\text{BNN}(\boldf \mid \bphi, \mathcal{M})\right).
\end{equation}

We note that the measurement set $\mathcal{M}$ can be chosen arbitrarily but should cover most of the support of the joint distribution $p(\btheta, \bx)$. If data from this joint distribution are available, those can be leveraged to build the measurement set. To showcase the ability to create a prior with limited data, in this work, we derive boundaries of the support of each marginal distribution and draw parameters and observations independently and uniformly over this support. If the support is known a priori, this procedure can be performed without (expensive) simulations. We draw a new measurement set at each iteration of the optimization procedure. If a fixed measurement set is available, a subsample of this measurement set should be drawn at each iteration. 

As an illustrative example, we chose independent normal distributions as a variational family $p(\bw | \bphi)$ over the weights and minimize (\ref{eq:kl_divergence}) w.r.t. $\bw$. In Figure \ref{fig:prior_coverage}, we show the coverage of the resulting a priori Bayesian model average using the tuned prior, $p(\bw \mid \bphi)$, and normal priors for increasing standard deviations $\sigma$, for the SLCP benchmark. We observe that while none of the normal priors are calibrated, the trained prior achieves near-perfect calibration. This prior hence guides the obtained posterior approximation towards more calibrated solutions, even in low simulation-budget settings.

\begin{figure}
    \centering
    \includegraphics[width=\textwidth]{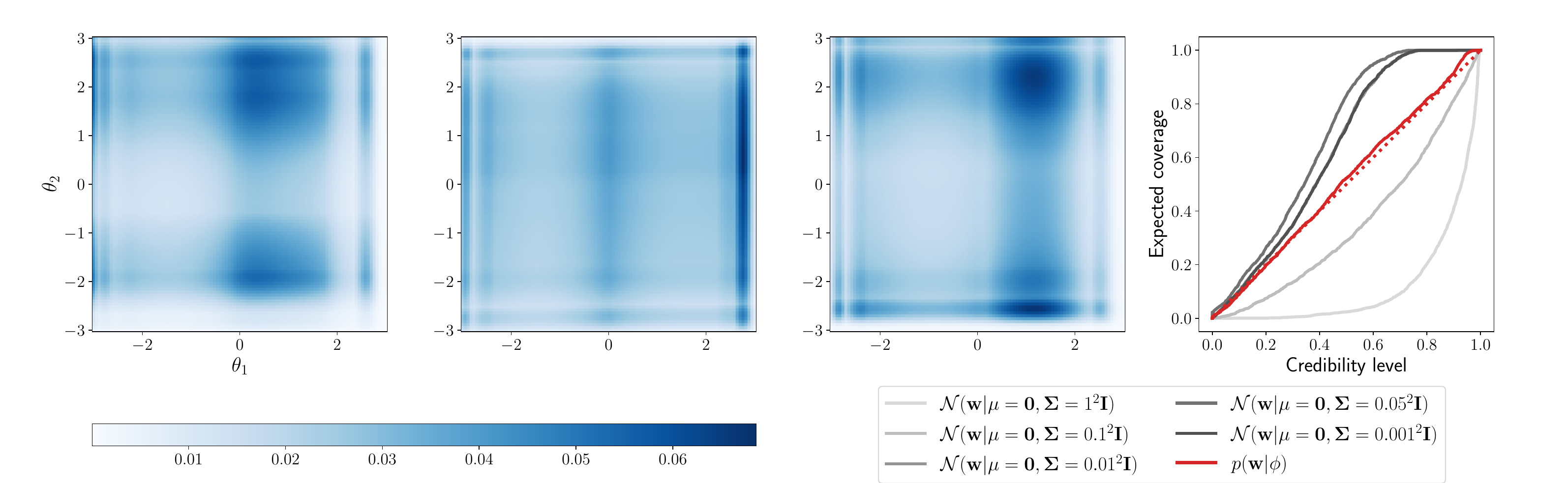}
    \vspace{-1.5em}
    \caption{Visualization of the prior tuned to match the GP prior on the SLCP benchmark. Left: examples of posterior functions sampled from the tuned prior over neural network's weights. Right: expected coverage of the prior Bayesian model average with the tuned prior and normal priors for varying standard deviations.}
    \label{fig:prior_coverage}
\end{figure}

The attentive reader might have noticed that $p_{\text{BNN}}(\boldf \mid \bphi, \bTheta, \bX)$ and $p_{\text{GP}}(\boldf|\mu(\bTheta, \bX), K(\bTheta, \bX))$ do not share the same support, as the former distribution is limited to functions that represent valid densities by construction, while the latter includes arbitrarily shaped functions. This is not an issue here as the support of the first distribution is included in the support of the second distribution, and functions outside the support of the first distribution are ignored in the computation of the divergence.

Note that there are various methods that can be used to perform inference on the neural network's weights with our GP prior. Instead of minimizing the KL-divergence, the parameters $\bphi$ can be optimized using an adversarial training procedure by treating both priors as function generators and training a discriminator between the two \citep{tran2022all}. Another approach to performing inference using a functional prior is to directly use it during inference by modifying the inference algorithm to work in function space. Variational inference can be performed in the space of function \citep{sun2018functional, rudner2022tractable}. The stochastic gradient Hamiltonian Monte Carlo algorithm \citep{chen2014stochastic} could also be modified to include a functional prior \cite{kozyrskiy2023imposing}. Alternatively, a variational implicit process can be learned to express the posterior in function space \citep{ma2021functional}.

\section{Experiments}
In this section, we empirically demonstrate the benefits of replacing a regular neural network with a BNN equipped with the proposed prior for simulation-based inference.
We consider both Neural Posterior Estimation (NPE) with neural spline flows \citep{durkan2019neural} and Neural Ratio Estimation (NRE) \citep{2019arXiv190304057H}, along with their balanced versions (BNRE and BNPE) \citep{delaunoytowards, delaunoy2023balancing} and ensembles \citep{lakshminarayanan2017simple, crisissbi}. BNNs-based methods are trained using mean-field variational inference \citep{blundell2015weight}. 
As advocated by \citet{wenzel2020good}, we also consider cold posteriors to achieve good predictive performance. More specifically, the variational objective function is modified to give less weight to the prior by introducing a temperature parameter $T$, 
\begin{equation}
    \mathbb{E}_{\mathbf{w} \sim p(\bw | \boldsymbol{\tau})}\left[\sum_{i} \log p(\btheta_i | \bx_i, \bw)\right] - T\ \text{KL}[p(\bw | \boldsymbol{\tau}) || p(\bw | \bphi)],
\end{equation}
where $\boldsymbol{\tau}$ are the parameters of the posterior variational family and $T$ is a parameter called the temperature that weights the prior term. In the following, we call BNN-NPE, a Bayesian Neural Network posterior estimator trained without temperature ($T=1$), and BNN-NPE ($T=0.01$), an estimator trained with a temperature of $0.01$, assigning a lower weight to the prior.

\paragraph{Setup}
All the NPE-based methods use a Neural Spline Flow (NSF) \citep{durkan2019neural} with 3 transforms of 6 layers, each containing 256 neurons. Meanwhile, all the NRE-based methods employ a classifier consisting of 6 layers of 256 neurons. For the spatialSIR and Lotka Volterra benchmarks, the observable is initially processed by an embedding network. Lotka Volterra's embedding network is a 10 layers 1D convolutional neural network that leads to an embedding of size $512$. On the other hand, SpatialSIR's embedding network is an 8 layers 2D convolutional neural network resulting in an embedding of size $256$. All the models are trained for $500$ epochs which we observed to be enough to reach convergence.

Bayesian neural network-based methods use independent normal distributions as a variational family. During inference, $100$ neural networks are sampled to approximate the Bayesian model average. Ensemble methods involve training $5$ neural networks independently.

We evaluate the different methods of the SLCP, Two Moons, Lotka-Volterra and Spatial SIR benchmarks.

The \emph{SLCP} (Simple Likelihood Complex Posterior) benchmark \citep{papamakarios2019sequential} is a fictive benchmark that takes $5$ parameters as input and produces an 8-dimensional synthetic observable. The observation corresponds to the 2D coordinates of 4 points that are sampled from the same multivariate normal distribution. We consider the task of inferring the marginal over $2$ of the $5$ parameters.

The \emph{Two Moons} simulator \citep{greenberg2019automatic} models a fictive problem with 2 parameters. The observable x is composed of 2 scalars, which represent the 2D coordinates of a random point sampled from a crescent-shaped distribution shifted and rotated around the origin depending on the parameters’ values. Those transformations involve the absolute value of the sum of the parameters leading to a second crescent in the posterior and, hence
making it multi-modal.

The \emph{Lotka-Volterra} population model \citep{lotka, volterra1926fluctuations} describes a process of interactions between a predator and a prey species. The model is conditioned on 4 parameters that influence the reproduction and mortality rate of the predator and prey species. We infer the marginal posterior of the predator parameters from a time series of 2001 steps representing the evolution of both populations over time. The specific implementation is based on a Markov Jump Process, as in \citet{papamakarios2019sequential}.

The \emph{Spatial SIR} model \citep{crisissbi} involves a grid world of susceptible, infected, and recovered individuals. Based on initial conditions and the infection and recovery rate, the model describes the spatial evolution of an infection. The observable is a snapshot of the grid world after some fixed amount of time. The grid used is of size 50 by 50.

In the experiment, The kernel $K$ used in the GP prior is a combination of two Radial Basis Function (RBF) kernels. If more information on the structure of the target posterior is available, more informed kernels may be used to leverage this prior knowledge. The kernel is defined by
\begin{equation}
    K(\btheta_1, \btheta_2, \bx_1, \bx_2) = \sqrt{\text{RBF}(\btheta_1, \btheta_2)} * \sqrt{\text{RBF}(\bx_1, \bx_2)}.
\end{equation}
such that the correlation between outputs is high only if $\btheta_1$ and $\btheta_2$ as well as $\bx_1$ and $\bx_2$ are close. The RBF kernel is defined as
\begin{equation}
    \text{RBF}(\bx_1, \bx_2) = \sigma^2 \exp\left( -\frac{1}{N}\sum_{i}^N\frac{ (x_{1, i} - x_{2, i})^2}{2l_i^2}\right),
\end{equation}
where $\sigma$ is the standard deviation and $l_i$ is the lengthscale associated to the $i^\text{th}$ feature. The lengthscale is derived from the measurement set. To determine $l_i$, we query observations $\bx$ from the measurement set and compute the $0.1$ quantile of the squared distance between different observations for each feature. We then set $l_i$ such that $2l_i^2$ equals this quantile. All the benchmarks have a uniform prior over the simulator's parameters. The mean function is then equal to a constant $C$ for all input values. The standard deviation is chosen to be $C/2$.

Following \citet{delaunoytowards}, we evaluate the quality of the posterior approximations based on the expected nominal log posterior density and the expected coverage area under the curve (coverage AUC). The expected nominal log posterior density $\mathbb{E}_{\btheta, \bx \sim p(\btheta, \bx)}\left[\log \hat{p}(\btheta| \bx) \right]$  quantifies the amount of density allocated to the nominal parameter that was used to generate the observation. The coverage AUC $\int_0^1 (\text{EC}(\hat{p}, \alpha) - \alpha)\ d\alpha$ quantifies the calibration of the expected posterior. A calibrated posterior approximation exhibits a coverage AUC of $0$. A positive coverage AUC indicates conservativeness, and a negative coverage AUC indicates overconfidence. 

\begin{figure}[h!]
    \centering
    \includegraphics[width=\textwidth]{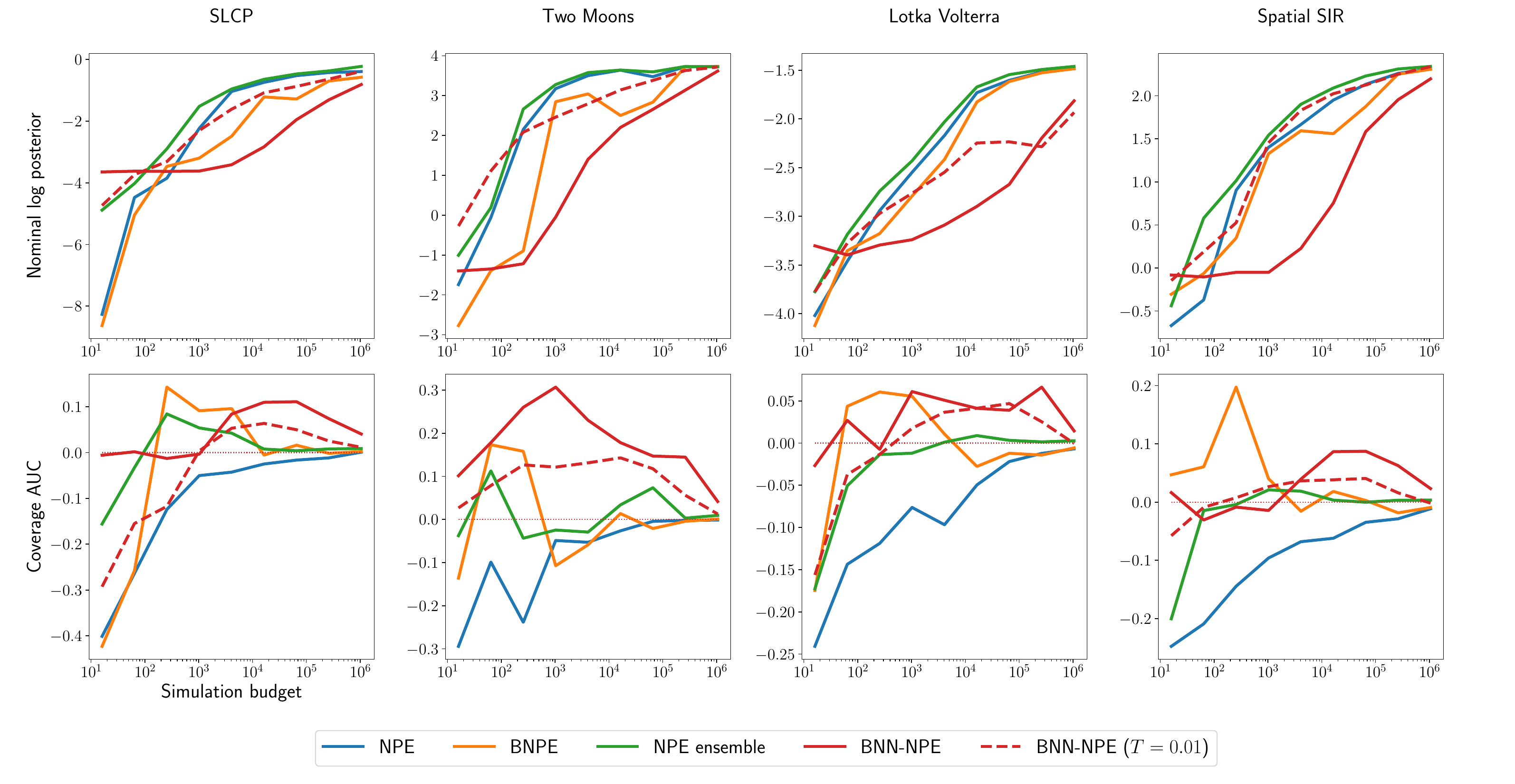}
    \includegraphics[width=\textwidth]{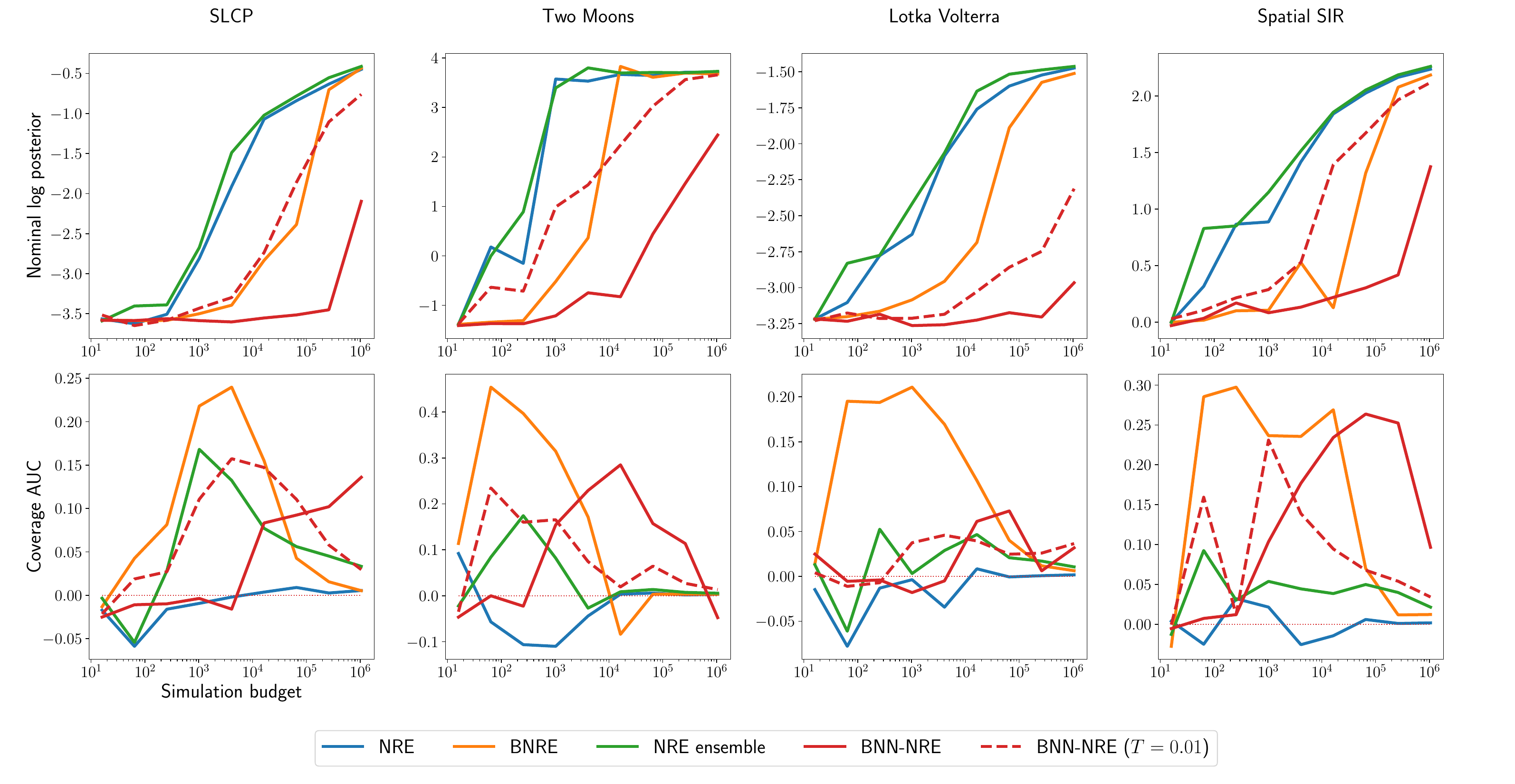}
    \vspace{-1.5em}
    \caption{Comparison of simulation-based inference methods through the nominal log probability and coverage area under the curve. The higher the nominal log probability, the more performant the method is. A calibrated posterior approximation exhibits a coverage AUC of $0$. A positive coverage AUC indicates conservativeness, and a negative coverage AUC indicates overconfidence. 3 runs are performed, and the median is reported. The plot at the top shows the results for NPE simulation-based inference methods, and the one at the bottom shows NRE methods.}
    \label{fig:performance_metrics}
\end{figure}

\paragraph{BNN-based simulation-based inference} Figure \ref{fig:performance_metrics} compares simulation-based inference methods with and without accounting for computational uncertainty. 
We observe that BNNs equipped with our prior and without temperature show positive, or only slightly negative, coverage AUC even for simulation budgets as low as $O(10)$. Negative coverage AUC is still observed, and hence conservativeness is not strictly guaranteed. However, this constitutes a significant improvement over the other methods in that regard.
The coverage curves are reported in Section \ref{sec:additional_experiments}.
We conclude that BNNs can hence be more reliably used than the other benchmarked methods when the simulator is expensive and few simulations are available. We observe that increasing the reliability comes with the drawback of requiring more simulations than the other methods to reach similar nominal log posterior density values. Without temperature, a few orders of magnitude more samples might be needed. However, in theory, as the amount of sample increases, the effect of the prior diminishes, and BNNs should reach the same nominal log posterior density as standard methods. By adding a temperature to the prior, its effect is diminished and better nominal log posterior density values are observed
From these observations, general guidelines to set the temperature include increasing $T$ if overconfidence is observed and decreasing it if low predictive performance is observed.

Examples of posterior approximations obtained with and without using a Bayesian neural network are shown in Figure \ref{fig:posterior_examples}. Wide posteriors are observed for low budgets for BNN-NPE, while NPE produces an overconfident approximation and excludes most of the relevant parts of the posterior. As the simulation budget increases, BNN-NPE converges slowly towards the same posterior as NPE. BNN-NPE ($T=0.01$) converges faster than BNN-NPE but, for low simulation budgets, excludes parts of the region that should be accepted according to high budget posteriors. Yet, the posterior approximate is still less overconfident than NPE's. Finally, Figure \ref{fig:performance_metrics} shows that BNN-NRE is more conservative than BNN-NPE. This comes at the cost of lower nominal log posterior density for a given simulation budget while still reaching comparable values as the simulation budgets grow.

\begin{figure}[h!]
    \centering
    \includegraphics[width=\textwidth]{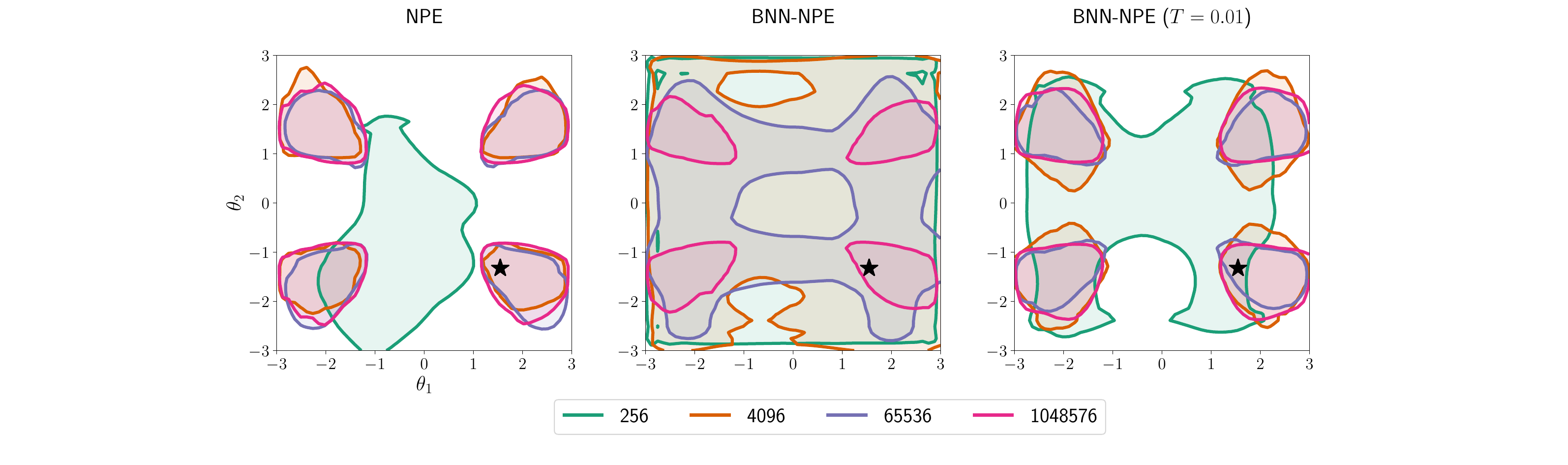}
    \vspace{-1.5em}
    \caption{Examples of $95 \%$ highest posterior density regions obtained with various algorithms and simulation budgets on the SLCP benchmark for a single observation. The black star represents the ground truth used to generate the observation.}
    \label{fig:posterior_examples}
\end{figure}

\paragraph{Comparison of different priors on weights} We analyze the effect of the prior on the neural network's weights on the resulting posterior approximation. The posterior approximations obtained using our GP prior are compared to the ones obtained using independent normal priors on weights with zero means and increasing standard deviations. In Figure \ref{fig:prior_comparison}, we observe that when using a normal prior, careful tuning of the standard deviation is needed to achieve results close to the prior designed for simulation-based inference. The usage of an inappropriate prior can lead to bad calibration for low simulation budgets or can prevent learning if it is too restrictive. 

\begin{figure}[h!]
    \centering
    \includegraphics[width=\textwidth]{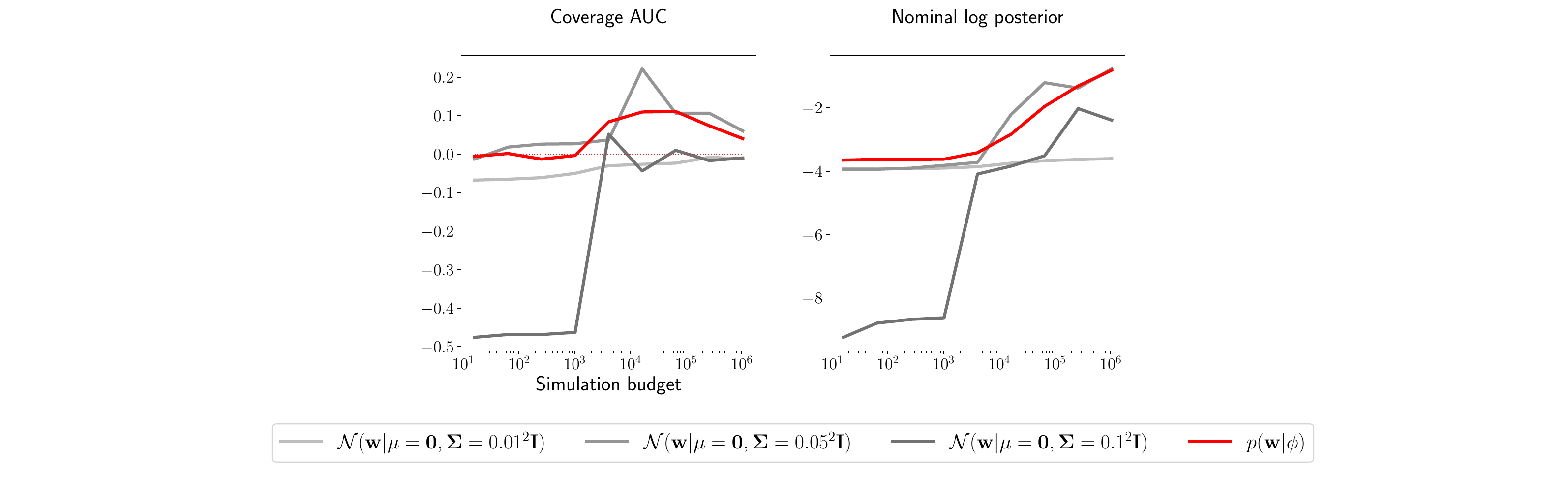}
    \vspace{-1.5em}
    \caption{Comparison of posterior approximations obtained using a prior tuned to match the Gaussian process-based prior and using independent normal priors on weights with zero means and various standard deviations on the SLCP benchmark. 3 runs are performed, and the median is reported.}
    \label{fig:prior_comparison}
\end{figure}

\paragraph{Uncertainty decomposition} We decompose the uncertainty quantified by the different methods. Following \citet{depeweg2018decomposition}, the uncertainty can be decomposed as 
\begin{equation}
    \mathbb{H}\left[\hat{p}(\btheta |\bx)\right] = \mathbb{E}_{q(\bw)}\left[ \mathbb{H}\left[\hat{p}(\btheta |\bx, \bw)\right]\right] + \mathbb{I}(\btheta, \bw),
\end{equation}
where $\mathbb{E}_{q(\bw)}\left[ \mathbb{H}\left[\hat{p}(\btheta |\bx, \bw)\right]\right]$ quantifies the aleatoric uncertainty, $\mathbb{I}(\btheta, \bw)$ quantifies the epistemic uncertainty, and the sum of those terms is the predictive uncertainty. 
Figure \ref{fig:uncertainty} shows the decomposition of the two sources of uncertainty, in expectation, on the SLCP benchmark. Other benchmarks can be found in Section \ref{sec:additional_experiments}. We observe that BNN-NPE and NPE ensemble methods account for the epistemic uncertainty while other methods do not. BNPE artificially increases the aleatoric uncertainty to be better calibrated. The epistemic uncertainty of BNN-NPE is initially low because most of the models are slight variations of $p_{\bTheta}$. The epistemic uncertainty then increases as it starts to deviate from the prior and decreases as the training set size increases. BNN-NPE ($T=0.01$) exhibits a higher epistemic uncertainty for low budgets as the effect of the prior is lowered.

\begin{figure}[t]
    \centering
    \includegraphics[width=\textwidth]{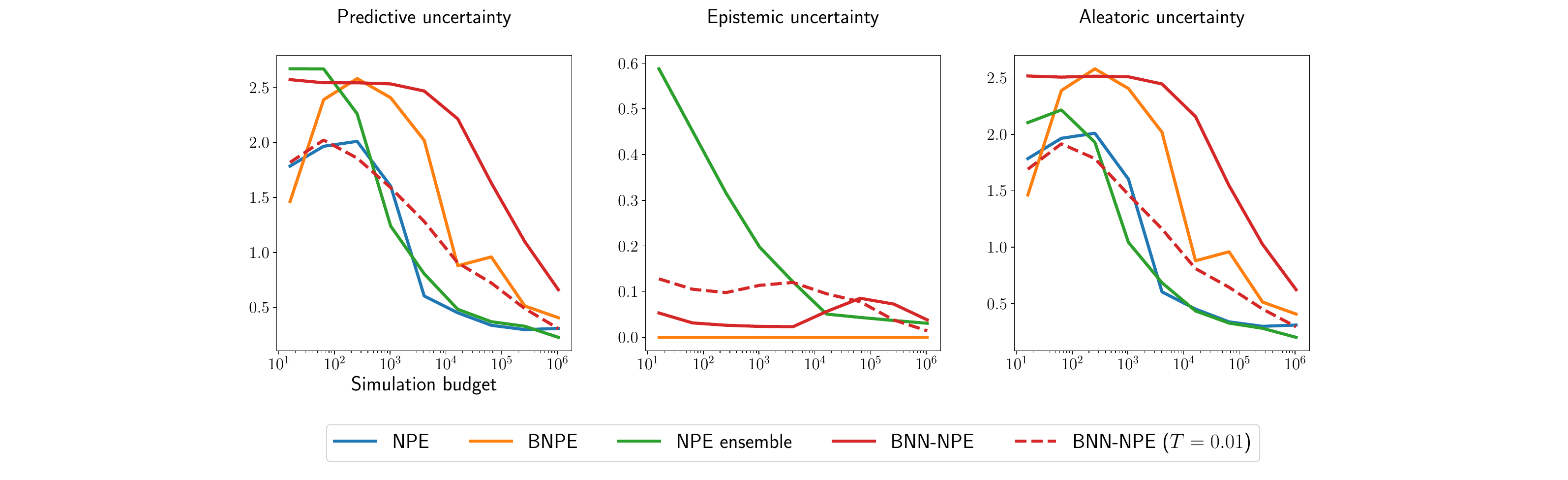}
    \vspace{-1.5em}
    \caption{Quantification of the different forms of uncertainties captured by the different NPE-based methods on the SLCP benchmark. 3 runs are performed, and the median is reported.}
    \label{fig:uncertainty}
\end{figure}
\begin{figure}[t]
    \centering
    \includegraphics[width=\textwidth]{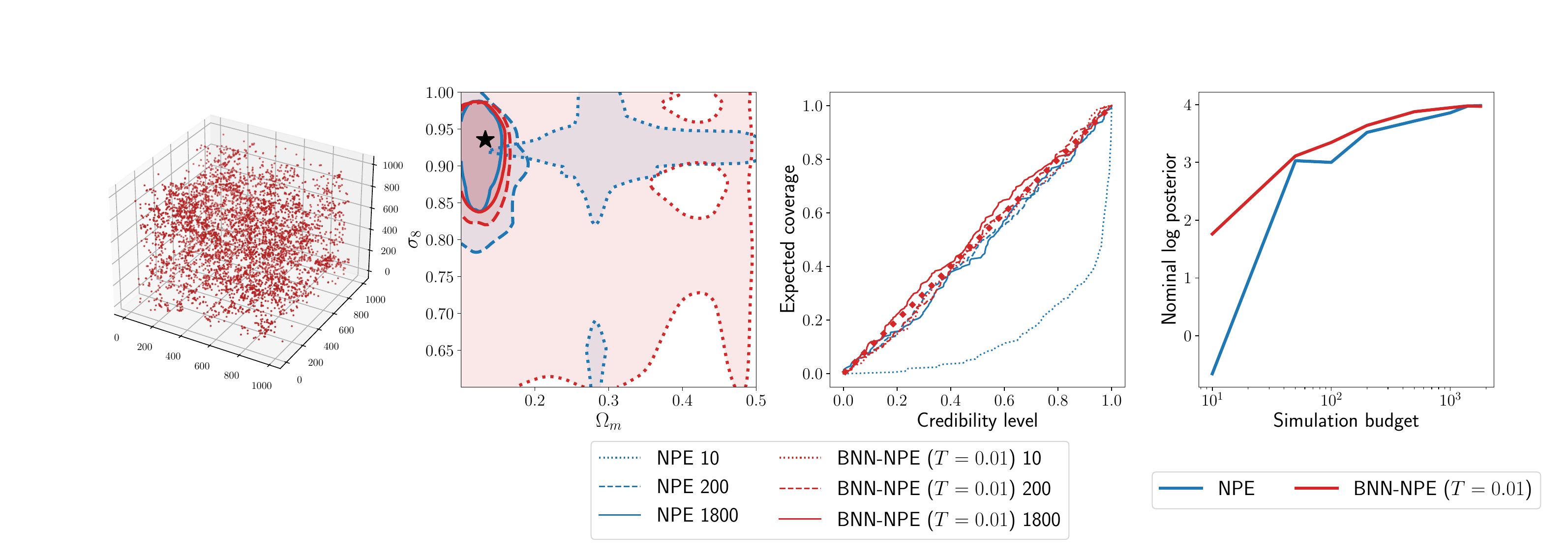}
    \vspace{-1.5em}
    \caption{Comparison of the posterior approximations obtained with and without a Bayesian neural network on the cosmological application. First plot: An example observation: particles representing galaxies in a synthetic universe. Second plot: example of $95 \%$ highest posterior density regions for increasing simulation budgets. The black star represents the ground truth used to generate the observation. Third plot: Expected coverage with and without using a Bayesian neural network for increasing simulation budgets. Fourth plot: The nominal log posterior.}
    \label{fig:galaxies}
\end{figure}

\paragraph{Infering cosmological parameters from $N$-body simulations}
To showcase the utility of Bayesian deep learning for simulation-based inference in a practical setting, we consider a challenging inference problem from the field of cosmology. We consider \emph{Quijote} $N$-body simulations \citep{villaescusa2020quijote} tracing the spatial distribution of matter in the Universe for different underlying cosmological models. The resulting observations are particles with different masses, corresponding to dark matter clumps, which host galaxies. 
We consider the canonical task of inferring the matter density (denoted $\Omega_m$) and the root-mean-square matter fluctuation averaged over a sphere of radius $8h^{-1}$ Mpc (denoted $\sigma_8$) from an observed galaxy field. Robustly inferring the values of these parameters is one of the scientific goals of flagship cosmological surveys. These simulations are very computationally expensive to run, with over $35$ million CPU hours required to generate $44100$ simulations at a relatively low resolution. Generating samples at higher resolutions, or a significantly larger number of samples, is challenging due to computational constraints. These constraints necessitate methods that can be used to produce reliable scientific conclusions from a limited set of simulations -- when few simulations are available, not only is the amount of training data low, but so is the amount of test data that is available to assess the calibration of the trained model. 

In this experiment, we use $2000$ simulations processed as described in \citet{cuesta2023point}. These simulations form a subset of the full simulation suite run with a uniform prior over the parameters of interest. $1800$ simulations are used for training and $200$ are kept for testing. We use the two-point correlation function evaluated at $24$ distance bins as a summary statistic. The observable is, hence, a vector of $24$ features. We observed that setting a temperature lower than $1$ was needed to reach reasonable predictive performance with Bayesian neural networks in this setting. Figure \ref{fig:galaxies} compares the posterior approximations obtained with a single neural network against those obtained with a BNN trained with a temperature of $0.01$. We observe from the coverage plots that while a single neural network can lead to overconfident approximations in the data-poor regime, the BNN leads to conservative approximations. BNN-NPE also exhibits higher nominal log posterior probability. Additionally, we observe that it provides posterior approximations that are calibrated and have a high nominal log probability with only a few hundred samples. 

\section{Conclusion}
In this work, we use Bayesian deep learning to account for the computational uncertainty associated with posterior approximations in simulation-based inference. We show that the prior on neural network's weights should be carefully chosen to obtain calibrated posterior approximations and develop a prior family with this objective in mind. The prior family is defined in function space as a Gaussian process and mapped to a prior on weights. Empirical results on benchmarks show that incorporating Bayesian neural networks in simulation-based inference methods consistently yields conservative posterior approximations, even with limited simulation budgets of $\mathcal O(10)$. As Bayesian deep learning continues to rapidly advance \citep{papamarkou2024position}, we anticipate that future developments will strengthen its applicability in simulation-based inference, ultimately enabling more efficient and reliable scientific applications in domains with computationally expensive simulators.

Using BNNs for simulation-based inference also comes with limitations. The first observed limitation is that the Bayesian neural network based methods might need orders of magnitude more simulated data in order to reach a predictive power similar to methods that do not use BNNs, such as NPE. While we showed that this limitation can be mitigated by tuning the temperature appropriately, this is something that might require trials and errors. A second limitation is the computational cost of predictions. When training a BNN using variational inference, the training cost remains on a similar scale as standard neural networks. At prediction time, however, the Bayesian model average described in Equation~\ref{eq:bayes_model_average} must be approximated, and this requires a neural network evaluation for each Monte Carlo sample in the approximation. The computational cost of predictions then scales linearly with the number $M$ of Monte Carlo samples. Finally, although our method significantly improves the reliability over standard methods for low simulation budgets, conservativeness is not strictly guaranteed. There are no theoretical guarantees and negative coverage AUC may still be observed.

\section{Additional experiments}\label{sec:additional_experiments}
In this section, we provide complementary results. Figures \ref{fig:coverage_grid_npe} and \ref{fig:coverage_grid_nre} display the coverage curves, demonstrating that a higher positive coverage AUC corresponds to coverage curves above the diagonal line. Figures \ref{fig:uncertainty_full_npe} and \ref{fig:uncertainty_full_nre} present the uncertainty decomposition of all methods on all the benchmarks. Figures \ref{fig:summary_npe_error_bars} and \ref{fig:summary_nre_error_bars} illustrate how the performance of the different algorithms varies across different runs.

\begin{figure}[h!]
    \centering
    \includegraphics[width=\textwidth,height=\textheight,keepaspectratio]{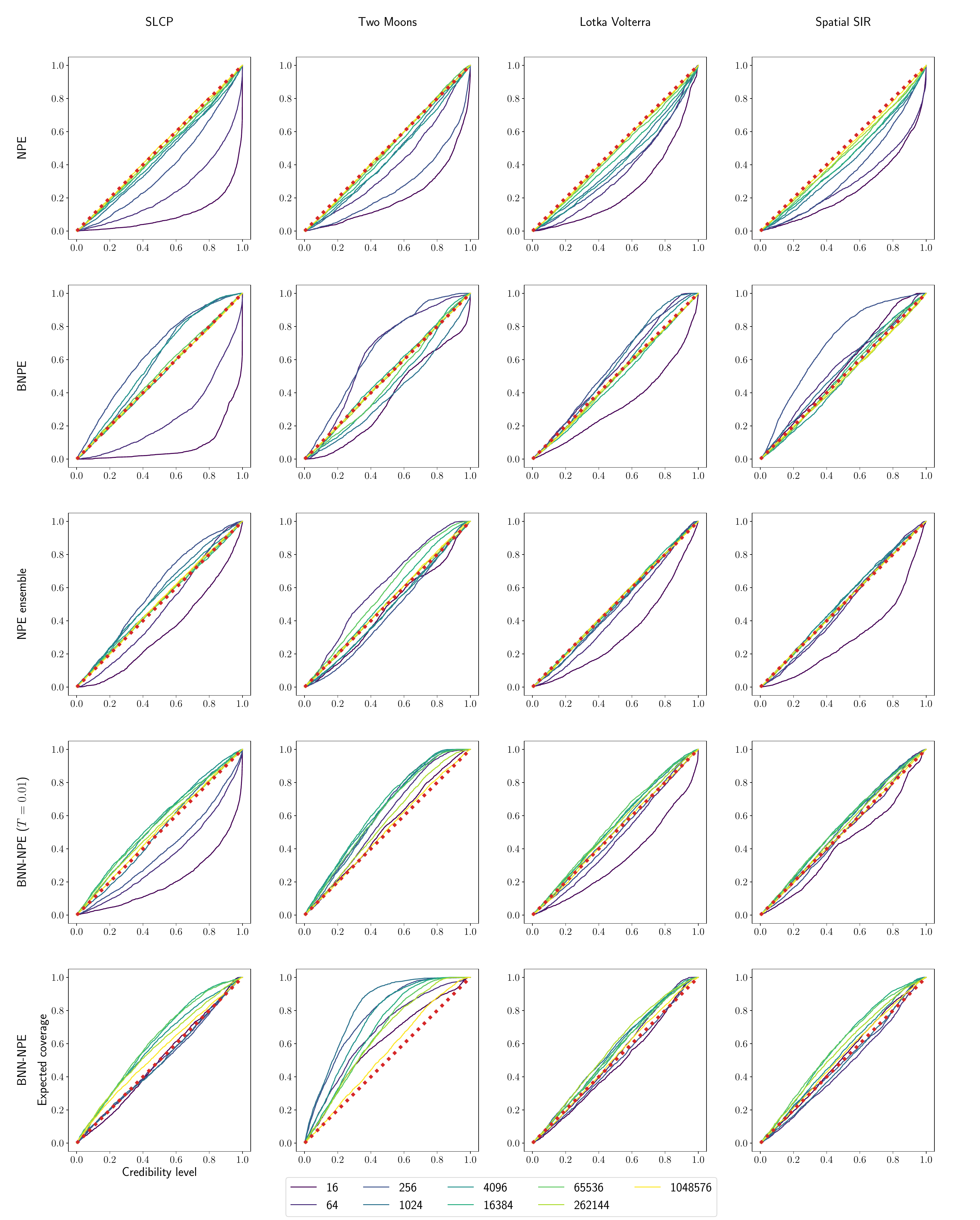}
    \vspace{-1.5em}
    \caption{Coverage of different NPE simulation-based inference methods. A calibrated posterior approximation exhibits a coverage AUC of $0$. A coverage curve above the diagonal indicates conservativeness and a curve below the diagonal indicates overconfidence. 3 runs are performed, and the median is reported.}
    \label{fig:coverage_grid_npe}
\end{figure}
\begin{figure}[h!]
    \centering
    \includegraphics[width=\textwidth,height=\textheight,keepaspectratio]{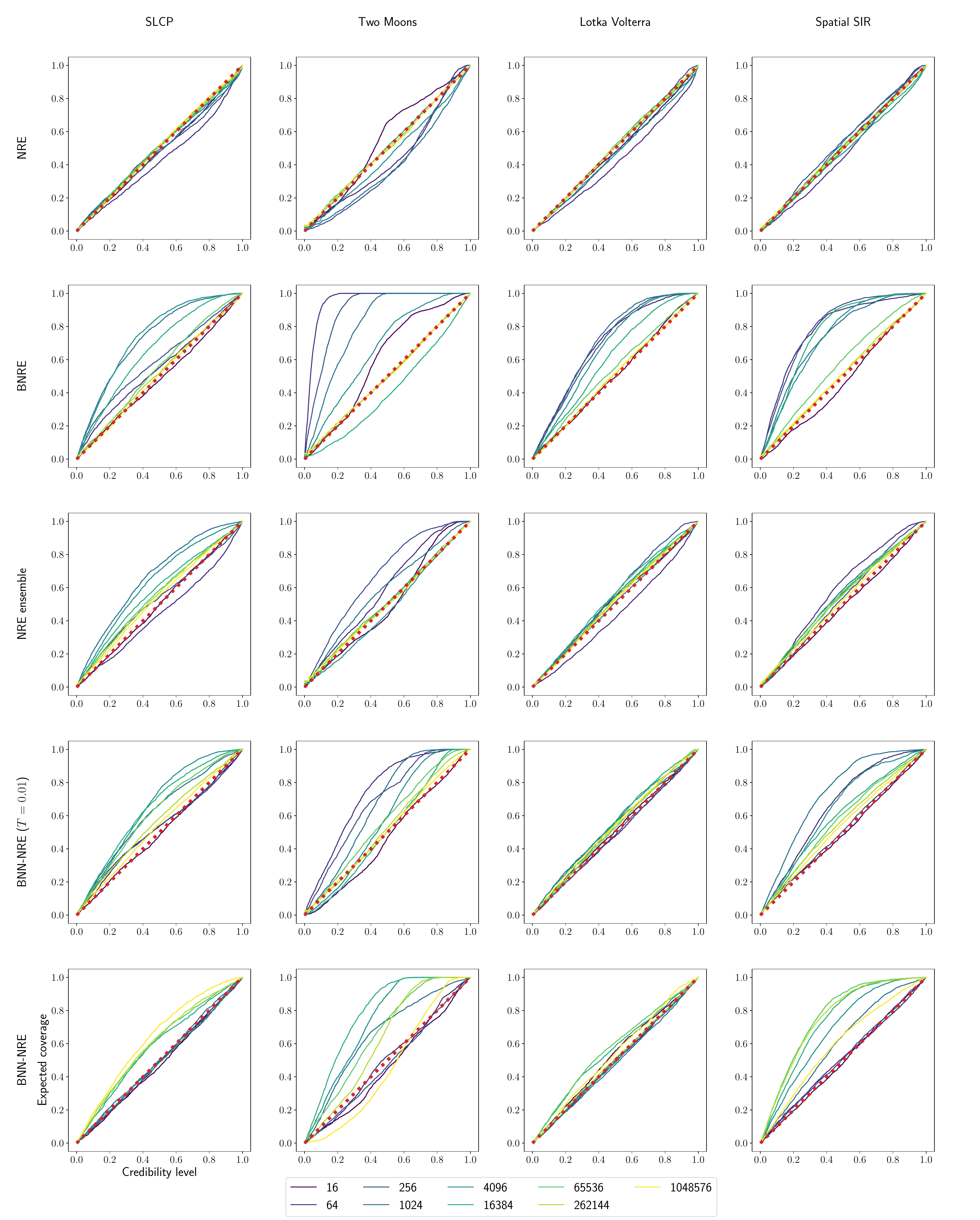}
    \vspace{-1.5em}
    \caption{Coverage of different NRE simulation-based inference methods. A calibrated posterior approximation exhibits a coverage AUC of $0$. A coverage curve above the diagonal indicates conservativeness and a curve below the diagonal indicates overconfidence. 3 runs are performed, and the median is reported.}
    \label{fig:coverage_grid_nre}
\end{figure}
\begin{figure}[h!]
    \centering
    \includegraphics[width=\textwidth]{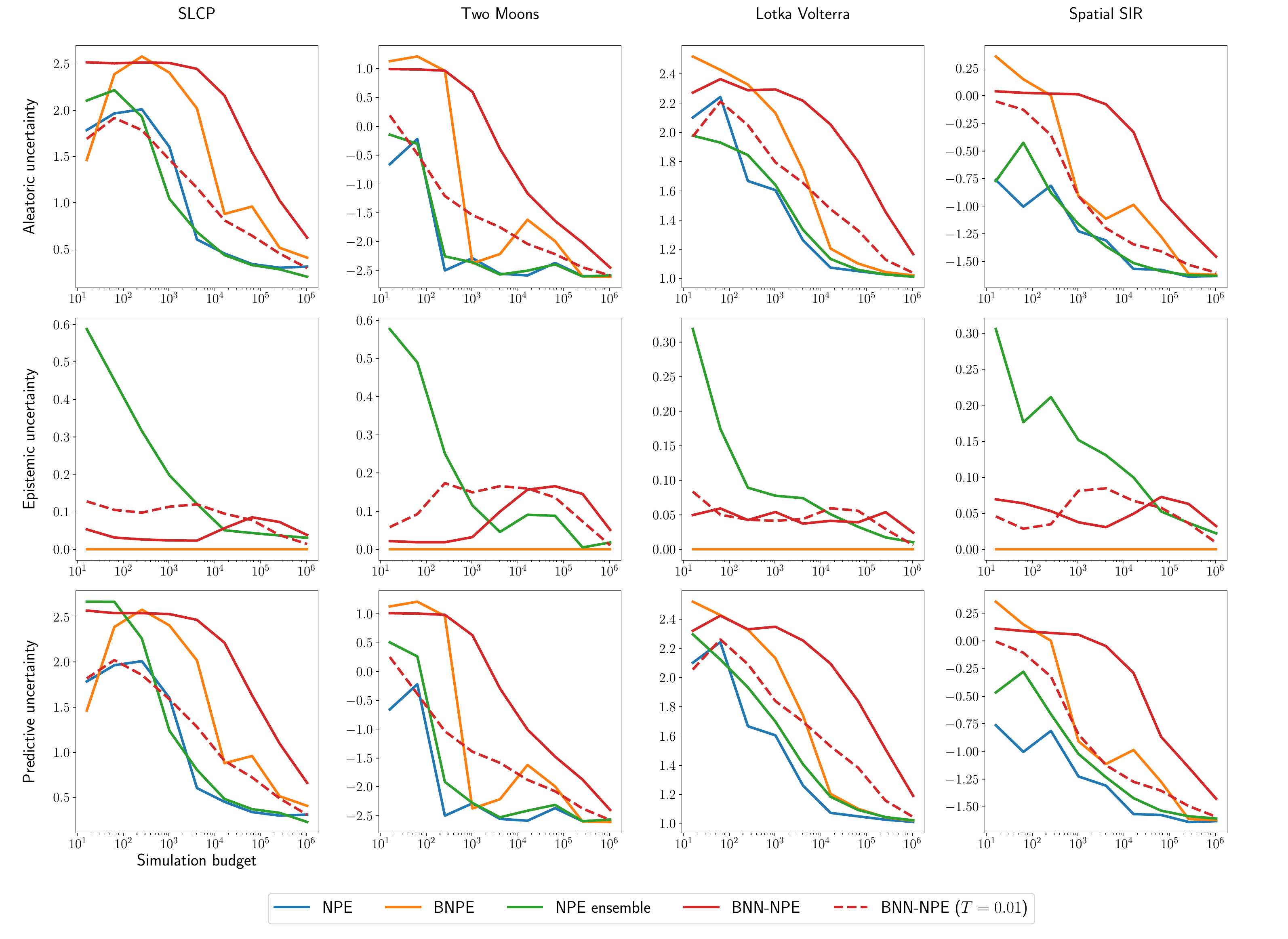}
    \vspace{-1.5em}
    \caption{Quantification of the different forms of uncertainties captured by the different NPE-based methods. 3 runs are performed, and the median is reported.}
    \label{fig:uncertainty_full_npe}
\end{figure}
\begin{figure}[h!]
    \centering
    \includegraphics[width=\textwidth]{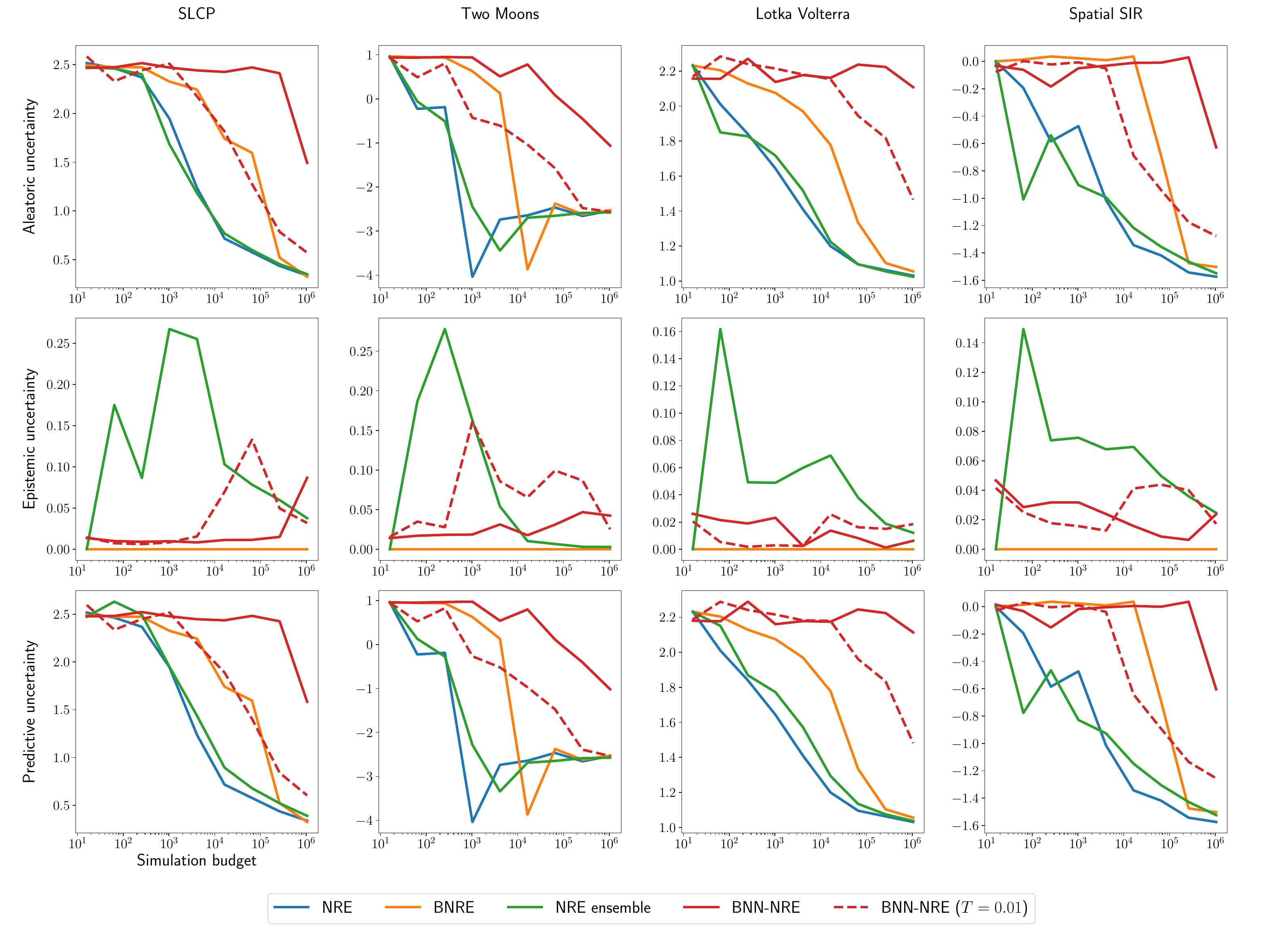}
    \vspace{-1.5em}
    \caption{Quantification of the different forms of uncertainties captured by the different NRE-based methods. 3 runs are performed, and the median is reported.}
    \label{fig:uncertainty_full_nre}
\end{figure}
\begin{figure}[h!]
    \centering
    \includegraphics[width=\textwidth]{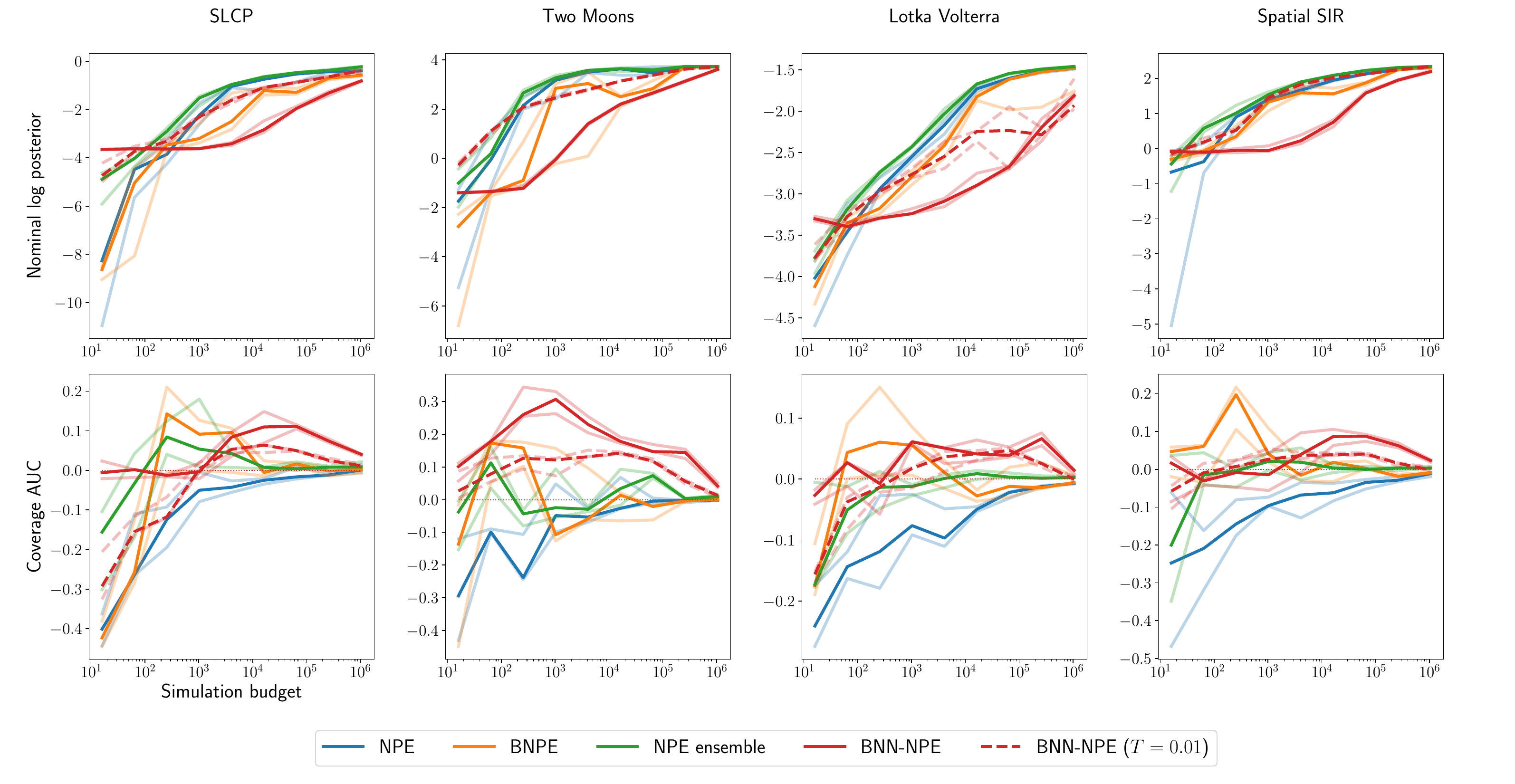}
    \vspace{-1.5em}
    \caption{Comparison of different NPE simulation-based inference methods through the nominal log probability and coverage area under the curve. The higher the nominal log probability, the more performant the method is. A calibrated posterior approximation exhibits a coverage AUC of $0$. A positive coverage AUC indicates conservativeness, and a negative coverage AUC indicates overconfidence. 3 runs are performed. The median run is reported in plain, and the shaded lines represent the other two runs.}
    \label{fig:summary_npe_error_bars}
\end{figure}
\begin{figure}[h!]
    \centering
    \includegraphics[width=\textwidth]{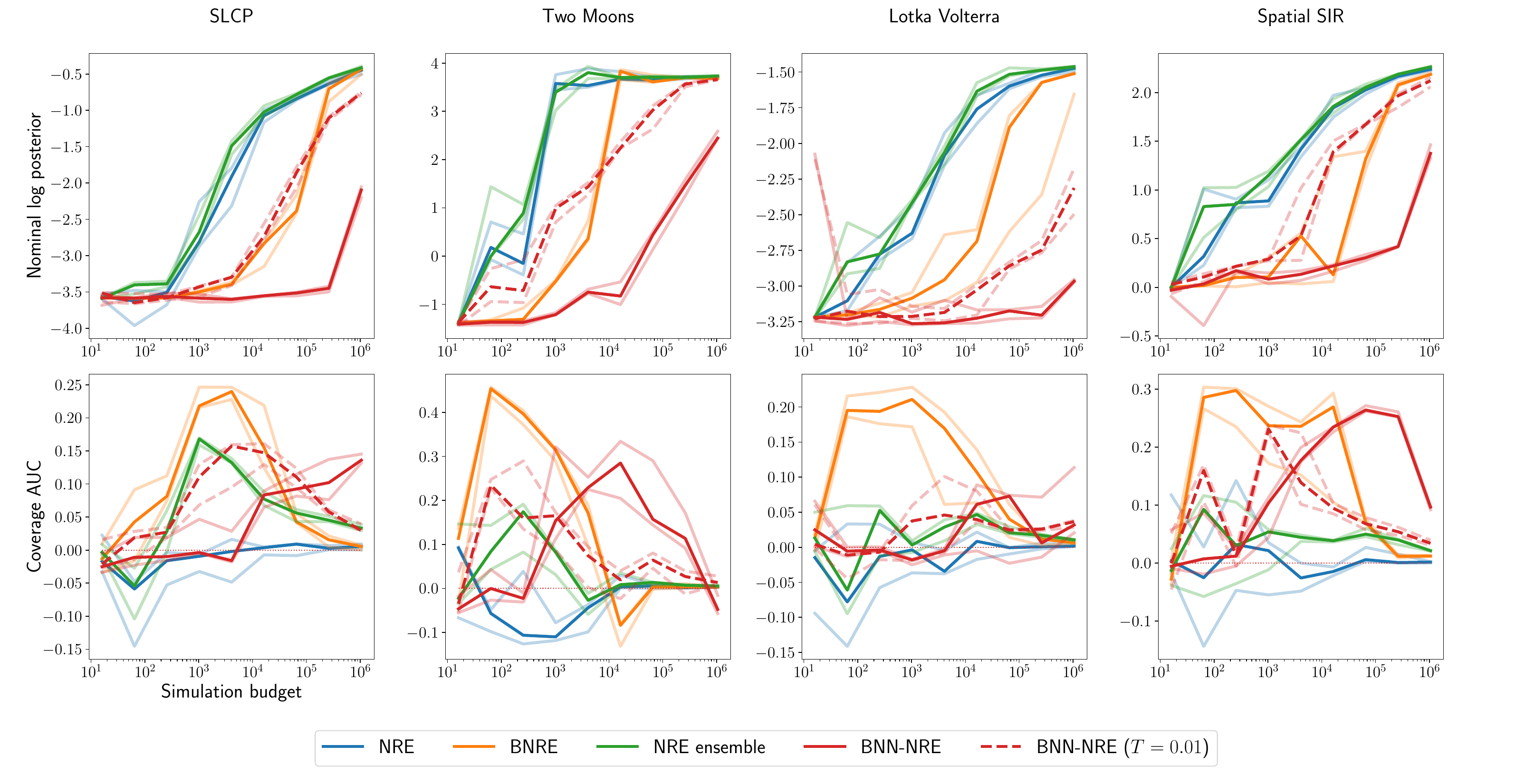}
    \vspace{-1.5em}
    \caption{Comparison of different NRE simulation-based inference methods through the nominal log probability and coverage area under the curve. The higher the nominal log probability, the more performant the method is. A calibrated posterior approximation exhibits a coverage AUC of $0$. A positive coverage AUC indicates conservativeness, and a negative coverage AUC indicates overconfidence. 3 runs are performed. The median run is reported in plain, and the shaded lines represent the other two runs.}
    \label{fig:summary_nre_error_bars}
\end{figure}

\FloatBarrier

\begin{epiloguebox}
    We showed that using Bayesian neural networks allows to take computational uncertainty into account and leads to more conservative posterior approximations. We illustrated the importance of the prior choice for Bayesian neural networks in the context of simulation-based inference and defined a family of priors tailored for this field. Limitations of this work include the increased computational cost as a prior must first be optimized, a posterior over neural networks must be approximated, and several neural network evaluations are required to make predictions. However, in settings where the simulator is very expensive to run, this cost will be negligible compared to the cost of making many more simulations. Another limitation is the loss of predictive power, requiring much more data to reach a predictive power similar to NPE. We showed that this can be improved by tuning the prior temperature. In addition, as the focus was on discussing the use of Bayesian deep learning for simulation-based inference and the design of an appropriate prior, we used mean field variational inference, a rather simple Bayesian deep learning algorithm that requires little tuning. Using more advanced inference algorithms and careful tuning could improve results in that regard. We also hope that BNNs' predictive power improves with new advances in the field of Bayesian deep learning.
\end{epiloguebox}

  \chapter{Discussion}\label{c:discussion}
  Throughout this thesis, we motivated the need for conservative posterior approximations, introduced the expected coverage to diagnose conservativeness, and introduced two ways to improve conservativeness: balancing and the use of Bayesian neural networks. In this chapter, we discuss other lines of work related to the conservativeness of posterior approximations and some limitations that remain with current simulation-based inference algorithms. We end the discussion with a general conclusion.

\section{Related work}

Another way of improving conservativeness in cases where the expected coverage can be approximated is to perform explicit calibration on the coverage. If we are solely interested in credible regions and not in the posterior itself, this can be done post-training. The expected coverage of a posterior approximation can be computed for a range of levels using a calibration set sampled from $p(\btheta, \bx)$, yielding a coverage plot. Given a posterior approximation $\hat{p}(\btheta | \bx)$, to obtain credible intervals with expected coverage approximately of $\alpha$, credible intervals at level $\tilde{\alpha}$ can be used, where $\tilde{\alpha}$ is such that
\begin{equation}
    \text{EC}(\hat{p}, \tilde{\alpha}) = \alpha,
\end{equation}
where $\text{EC}$ denotes the expected coverage. If an infinite amount of calibration samples are used, then the produced credible regions at level $\tilde{\alpha}$ have an expected coverage of exactly $\alpha$. For a finite amount of calibration samples, correct coverage is not guaranteed. 

\citet{patel2023variational} proposed to use conformal prediction as a way to produce regions with some guarantees about the expected coverage for a finite calibration set. Specifically, they use inductive conformal prediction that selects points to include in a region by comparing their non-conformity score to the ones of others in the calibration set. In this case, the authors propose to use
\begin{equation}
    s(\bx, \btheta) = \frac{1}{\hat{p}(\btheta | \bx)}
\end{equation}
as non-conformity score. This score is computed for all samples from the calibration set $(\btheta, \bx) \sim p(\btheta, \bx)$. If we aim for a region generator that leads to an expected coverage of $\alpha$, the threshold $s_C(\alpha)$ is the $\lceil(N_c + 1)/(1-\alpha)\rceil/N_c$ score quantile on the calibration set, where $N_c$ is the size of the calibration set. The prediction region for an observation $\bx$ is then constructed as 
\begin{equation}
    C(\bx) = \{\btheta: s(\btheta, \bx) \leq s_C(\alpha) \}.
\end{equation}
If the calibration samples and the observation used for prediction are all sampled independently from the same distribution (i.i.d.), the expected coverage of this prediction region generator is guaranteed to not be lower than $\epsilon$ with some probability above $\delta$, with $\epsilon$ and $\delta$ being varying with the size of the calibration set $N_c$ \citep{vovk2012conditional}. Inductive conformal prediction then bounds miscoverage with some probability. Note that this statement is about expected coverage and that those prediction regions might not be valid credible regions for all observations $\bx$ or valid confidence region generators for all parameters $\btheta$.

In order to obtain not only calibrated credibility regions in expectation but posterior approximations leading to such regions, \citet{falkiewicz2024calibrating} propose to add a regularization term based on the coverage that penalizes for either overconfidence or miscalibration. This is then similar to what is done for balanced algorithms, with the exception that regularization is directly performed on coverage instead of enforcing the balancing condition. While regularizing for coverage is more desirable than balancing, approximating the coverage at each batch introduces a computational overhead, while the balancing condition is cheap to approximate.

\citet{cranmer2015approximating} show that an approximate likelihood ratio can be calibrated to converge to the real likelihood ratio provided that the approximation is a monotonic function of the likelihood ratio. Under this condition, they show that
\begin{equation}
    r(\bx; \btheta_0, \btheta_1) = \frac{p(\bx | \btheta_0)}{p(\bx | \btheta_1)} = \frac{p(\hat{r}(\bx; \btheta_0, \btheta_1)| \btheta_0)}{p(\hat{r}(\bx; \btheta_0, \btheta_1) | \btheta_1)},
\end{equation}
where $p(\hat{r}(\bx; \btheta_0, \btheta_1)| \btheta_0)$ and $p(\hat{r}(\bx; \btheta_0, \btheta_1) | \btheta_1)$ can be approximated on a calibration set.

\citet{dalmasso2020confidence, masserano2022simulation} aim for tractable valid frequentists coverage for all parameter values $\btheta$. They base their method on the Neyman construction that uses the relationship between confidence regions and hypothesis tests presented in Equation \ref{eq:neyman}. A classical Neyman construction requires estimating critical values of a hypothesis test for each parameter value on a fine grid, which can be computationally intensive. Instead of estimating critical values for each parameter value in the grid independently, they propose to train a quantile regressor on test statistics computed on $(\btheta, \bx)$ pairs sampled from $p(\btheta, \bx)$. The regressor then takes parameters $\btheta$ as input and outputs approximate quantiles of the distribution of test statistics for various observations $\bx \sim p(\bx | \btheta)$. If the regressor is assumed to be perfectly trained, those quantiles are valid critical values, and the Neyman construction using such critical values yields valid Confidence regions. \citet{zhao2021diagnostics} introduce a local coverage test. Similarly to expected coverage computation, test pairs are sampled $\btheta^*, \bx^* \sim p(\btheta, \bx)$, and the rank, i.e., the minimal confidence level $\alpha$ required for the approximate credible region based on $\bx^*$ to include the nominal parameters $\btheta^*$, is computed. Instead of reporting the marginal rank distributions, they train regressors for various confidence levels $\alpha$, taking $\bx^*$ as input and outputting $1$ if the rank is lower than $\alpha$, $0$ otherwise. They use this regressor to construct a statistic for testing local coverage.

\section{Limitations}\label{sec:discussion:limitations}
In this thesis, we focused on the conservativeness aspect of approximate posteriors. While we provided ways to improve the conservativeness of current methods, limitations remain both on these aspects and on other aspects to obtain reliable simulation-based inference. One of those limitations is the lack of theoretical guarantees regarding conservativeness. Of all the discussed methods, conformal prediction \citep{patel2023variational} is the only one providing guarantees in the finite samples regime. These guarantees are, however, in expectation, and the produced regions are neither valid credible regions for all $\bx$ nor valid confidence regions for all $\btheta$. \citet{vovk2012conditional} showed that object conditional coverage cannot be guaranteed for rich observational spaces, such as $\mathbb{R}$, without any distributional assumptions. In the remainder of this section, we expand on two other limitations, namely the non-invariance to reparametrization of expected coverage based on high posterior density regions and possible misspecification of the simulator.

High posterior density regions are not invariant to reparametrizations. Let us assume a transformation $f$ applied to parameters $\btheta$, then $p(f(\btheta) | \bx)$ is not necessarily equivalent to $p(\btheta | \bx)$. For example, if we consider the transformation $f: \btheta \rightarrow 2\btheta$, then we have
\begin{equation}
    p(f(\btheta) | \bx) = \frac{p(\btheta | \bx)}{2}, \quad \forall \btheta.
\end{equation}
In this case, the density is divided by $2$ but the resulting high posterior density region would remain the same. This is not the case if the transformation $f$ is such that some parts of the space are more dilated or contracted than others. In those cases, we might have $p(f(\btheta) | \bx) > p(\btheta | \bx)$ for some parameters $\btheta$ and $p(f(\btheta) | \bx) < p(\btheta | \bx)$ for some other parameters $\btheta$. Consequently, high posterior density regions might be different after applying the reparametrization. Figure \ref{fig:reparametrization} shows an example of this for the transformation $f: \btheta \rightarrow \exp(\btheta)$. The $95\%$ high posterior density credible interval for $\btheta$ is $[-1.95,1.95]$ and the one for $\exp(\btheta)$ is $[0.03, 5.23]$. We have that $\exp(-1.95) = 0.14 \neq 0.03$ and $\exp(1.95) = 7.02 \neq 5.23$. Different parts of the parameter space are then included in the high posterior density credible interval depending on the used parametrization.

\begin{figure}
    \centering
    \begin{subfigure}[b]{0.49\textwidth}
    \includegraphics[width=\linewidth]{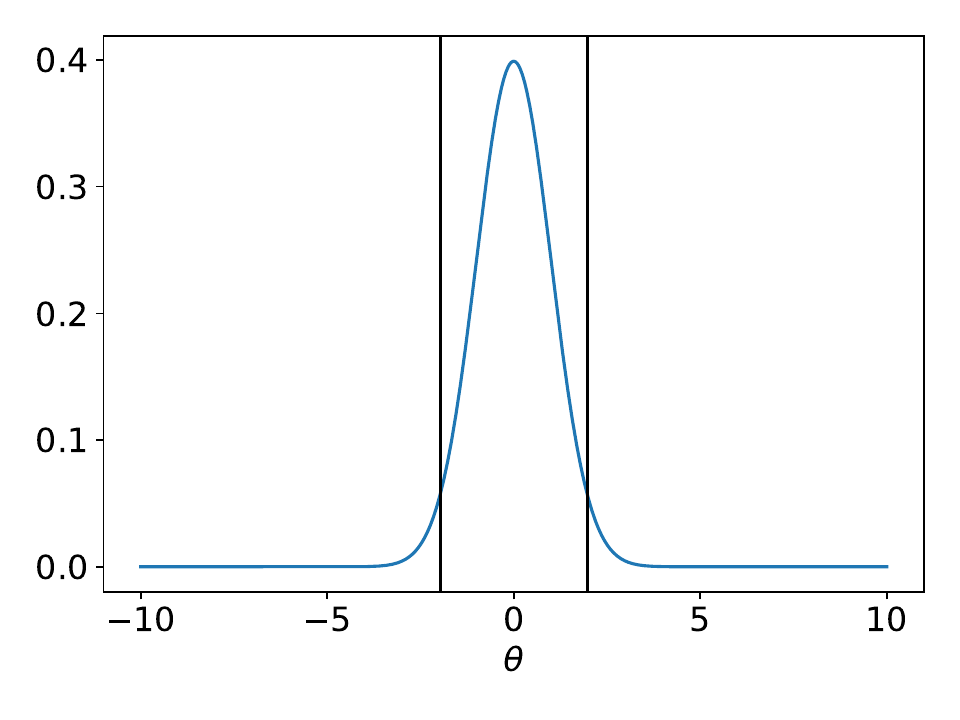}
    \end{subfigure}
    \begin{subfigure}[b]{0.49\textwidth}
    \includegraphics[width=\linewidth]{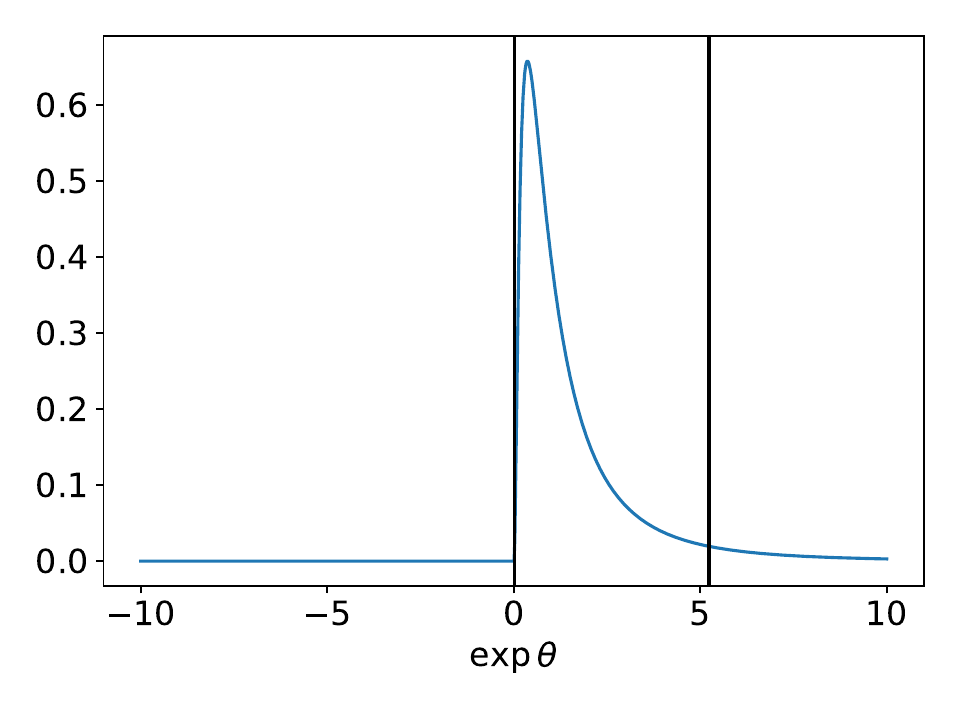}
    \end{subfigure}
    
    \caption{Comparison of high posterior density credible intervals obtained using various parameterizations. We consider that one parametrization is the exponential of the other. The blue lines represent the p.d.f. and the black lines represent the $95\%$ high posterior density credible intervals.}
    \label{fig:reparametrization}
\end{figure}

This raises some questions regarding the definition of a conservative posterior approximation. Indeed, any conservative posterior approximation can be turned into an overconfident approximation by reparametrizing $\btheta$, according to the definition based on high posterior density region expected coverage. To make sense of this definition of conservativeness, we argue that parametrization should be chosen by taking domain knowledge into account such that flat distributions correspond to distributions that domain experts would consider uncertain. To obtain quantities that are invariant to reparametrization, ideas can be taken from the objective Bayesian inference literature. In particular, high-posterior density regions can be replaced by low-loss credible regions that favor parts of the parameters space that have a low loss value according to some loss function \citep{bernardo2005intrinsic}. If that loss function is invariant to a reparametrization, then so are the resulting credible regions. 

In addition to issues linked to posterior approximations, scientific models can be misspecified, in the sense that it does not perfectly model reality. Let us consider that real observations are drawn from the data-generating process $p^*(\bx)$. Following \citet{cannon2022investigating}, a model is misspecified if there are no parameters $\btheta$ such that $p^*(\bx) = p(\bx | \btheta)$. A consequence of misspecification when using machine learning models for approximate inference is that the machine learning model is used to make predictions on observations from the distribution $p^*(\bx)$ while being trained on samples from the distribution $\int p(\bx|\btheta) p(\btheta) d\btheta$. The model might then perform poorly on observations from $p^*(\bx)$. This is illustrated by \citet{gloeckler2023adversarial}, showing that machine learning models are vulnerable to adversarial attacks in the context of simulation-based inference. Adversarial attacks are small perturbations to the observation $\bx$ such that the resulting posterior approximation changes drastically. Those can then be seen as a worst-case scenario of model misspecification.
Model misspecification may be detected by making posterior predictive checks, i.e., by comparing samples $\bx_o \sim p^*(\bx)$ to samples from $\int p(\bx | \btheta) p(\btheta | \bx_o)$

\citet{wehenkel2024addressing} consider an alternative definition. They assume that this data generation process can be expressed as $p^*(\bx) = \int p(\btheta) p^*(\bx | \btheta) d\btheta$, where $p(\btheta)$ is a known prior and $p^*(\bx | \btheta)$ is an unknown process. They then define a misspecified model as a model such that 
\begin{equation}
    p^*(\btheta | \bx) = \frac{p^*(\bx | \btheta) p(\btheta)}{p^*(\bx)} \neq \frac{p(\bx | \btheta) p(\btheta)}{p(\bx)} = p(\btheta | \bx).
\end{equation}
They assume having some samples from the process $p^*(\bx | \btheta)$ through, for example, more costly procedures and use those samples to correct the posterior approximation.

\section{Conclusion}
The use of approximate methods for statistical inference raises questions about the reliability of the obtained results for scientific reasoning. In particular, it might lead to the incorrect refutation of scientific models and lead scientists in the wrong direction. In chapter \ref{c:crisis}, we shed light on the impact of approximate methods for scientific reasoning and provided the expected coverage diagnostic as a way to identify issues with approximations. We defined approximations passing the expected coverage diagnostic, i.e., having an expected coverage higher than the desired credibility level as conservative, and argued that conservative approximations are desired for more reliable scientific reasoning. An empirical evaluation showed that state-of-the-art methods, at that time, did not always lead to conservative approximations. We introduced in Chapters \ref{c:bnre} and \ref{c:balancing_sbi} balancing as a way to increase the conservativeness of the resulting approximations. Balancing is a technique that is simple to implement and does not highly increase computational requirements. It is a simple first thing to try to improve conservativeness if overconfident approximations are observed. Balancing can fall short in cases where very few samples from a simulator model are available to approximate the desired quantities. In Chapter \ref{c:bnn}, we used Bayesian neural networks to improve the conservativeness of simulation-based inference algorithms in such regimes. 

While we proposed methods for improving conservativeness, the limitations shown in Section \ref{sec:discussion:limitations} highlight that the resulting methods still do not guarantee reliable Popperian falsification. The proposed methods aim for conservativeness as defined by the expected coverage but expected coverage only provides information about produced credible regions in expectation and is not invariant to reparametrization unless special care is taken. While we believe improving expected coverage conservativeness increases the reliability, those considerations suggest taking a step back and defining stronger objectives to aim for, with the objective of achieving reliable scientific reasoning with numerical approximations. For example, conservativeness should ideally not only be achieved in expectation. It should be ensured for the particular observation at hand.

It might, however, be unrealistic to aim for conservativeness guarantees for a given observation and, in general, to create algorithms having sufficiently strong guarantees to be used for Popperian falsification while dealing with numerical approximations. Most current methods aiming for conditional coverage guarantees involve another machine learning model for calibration that itself leads to another layer of approximations. A more realistic aim could be to make the scientific method evolve to deal with the presence of approximations while at the same time improving those approximations. A way would be to no longer consider that models can be definitively refuted, as the procedure leading to refutation involves approximations. Such models that would be considered as refuted under the hypothesis that approximations are perfect could, however, be considered as improbable. The quality of the approximations could then themselves be criticized, leading to considering those models again.

This thesis focused on the implications of numerical approximations on the process of scientific discovery and aimed to improve the reliability of approximate statistical inference methods in that regard. It should be noted that what constitutes an appropriate scientific methodology is a complex open philosophical debate even when no approximations are made in the statistical pipeline. Our work is only a stepping stone for improving an aspect of the scientific methodology. Drawing reliable scientific conclusions goes beyond having high-quality approximations, and the definition of reliable is up for debate.
  
  \cleardoublepage
  \appendix
  
  \singlespacing
\label{app:Bibliography} 

\manualmark 
\markboth{\spacedlowsmallcaps{\bibname}}{\spacedlowsmallcaps{\bibname}} 
\refstepcounter{dummy}

\addtocontents{toc}{\protect\vspace{\beforebibskip}} 
\addcontentsline{toc}{chapter}{\tocEntry{\bibname}}
\printbibliography

@article{rubin1984,
    author = {Donald B. Rubin},
    title = {{Bayesianly Justifiable and Relevant Frequency Calculations for the Applied Statistician}},
    volume = {12},
    journal = {The Annals of Statistics},
    number = {4},
    publisher = {Institute of Mathematical Statistics},
    pages = {1151 -- 1172},
    year = {1984},
}

@article{beaumont2002approximate,
  title={Approximate Bayesian computation in population genetics},
  author={Beaumont, Mark A and Zhang, Wenyang and Balding, David J},
  journal={Genetics},
  volume={162},
  number={4},
  pages={2025--2035},
  year={2002},
  publisher={Oxford University Press}
}

@article{marjoram2003markov,
  title={Markov chain Monte Carlo without likelihoods},
  author={Marjoram, Paul and Molitor, John and Plagnol, Vincent and Tavar{\'e}, Simon},
  journal={Proceedings of the National Academy of Sciences},
  volume={100},
  number={26},
  pages={15324--15328},
  year={2003},
  publisher={National Acad Sciences}
}

@article{sisson2007sequential,
  title={Sequential monte carlo without likelihoods},
  author={Sisson, Scott A and Fan, Yanan and Tanaka, Mark M},
  journal={Proceedings of the National Academy of Sciences},
  volume={104},
  number={6},
  pages={1760--1765},
  year={2007},
  publisher={National Acad Sciences}
}

@article{beaumont2009adaptive,
  title={Adaptive approximate Bayesian computation},
  author={Beaumont, Mark A and Cornuet, Jean-Marie and Marin, Jean-Michel and Robert, Christian P},
  journal={Biometrika},
  volume={96},
  number={4},
  pages={983--990},
  year={2009},
  publisher={Oxford University Press}
}

@article{10.1093/bioinformatics/btp619,
    author = {Toni, Tina and Stumpf, Michael P. H.},
    title = "{Simulation-based model selection for dynamical systems in systems and population biology}",
    journal = {Bioinformatics},
    volume = {26},
    number = {1},
    pages = {104-110},
    year = {2009},
    month = {10},
    issn = {1367-4803},
    doi = {10.1093/bioinformatics/btp619},
}

@article{blum2010non,
  title={Non-linear regression models for Approximate Bayesian Computation},
  author={Blum, Michael GB and Fran{\c{c}}ois, Olivier},
  journal={Statistics and computing},
  volume={20},
  number={1},
  pages={63--73},
  year={2010},
  publisher={Springer}
}

@article{fearnhead2012constructing,
  title={Constructing summary statistics for approximate Bayesian computation: semi-automatic approximate Bayesian computation},
  author={Fearnhead, Paul and Prangle, Dennis},
  journal={Journal of the Royal Statistical Society: Series B (Statistical Methodology)},
  volume={74},
  number={3},
  pages={419--474},
  year={2012},
  publisher={Wiley Online Library}
}

@article{cranmer2015approximating,
  title={Approximating likelihood ratios with calibrated discriminative classifiers},
  author={Cranmer, Kyle and Pavez, Juan and Louppe, Gilles},
  journal={arXiv preprint arXiv:1506.02169},
  year={2015}
}

@article{thomas2016likelihood,
  title={Likelihood-free inference by ratio estimation},
  author={Thomas, Owen and Dutta, Ritabrata and Corander, Jukka and Kaski, Samuel and Gutmann, Michael U and others},
  journal={Bayesian Analysis},
  year={2016},
  publisher={International Society for Bayesian Analysis}
}

@inproceedings{papamakarios2016fast,
  title={Fast $\varepsilon$-free inference of simulation models with bayesian conditional density estimation},
  author={Papamakarios, George and Murray, Iain},
  booktitle={Advances in neural information processing systems},
  pages={1028--1036},
  year={2016}
}

@article{gratton2017glass,
  title={GLASS: A General Likelihood Approximate Solution Scheme},
  author={Gratton, Steven},
  journal={arXiv preprint arXiv:1708.08479},
  year={2017}
}

@article{lueckmann2017flexible,
  title={Flexible statistical inference for mechanistic models of neural dynamics},
  author={Lueckmann, Jan-Matthis and Goncalves, Pedro J and Bassetto, Giacomo and {\"O}cal, Kaan and Nonnenmacher, Marcel and Macke, Jakob H},
  journal={Advances in Neural Information Processing Systems},
  volume={30},
  year={2017}
}

@article{tran2017hierarchical,
  title={Hierarchical implicit models and likelihood-free variational inference},
  author={Tran, Dustin and Ranganath, Rajesh and Blei, David M},
  journal={arXiv preprint arXiv:1702.08896},
  year={2017}
}

@incollection{sisson2018overview,
  title={Overview of ABC},
  author={Sisson, Scott A and Fan, Yanan and Beaumont, Mark A},
  booktitle={Handbook of approximate Bayesian computation},
  pages={3--54},
  year={2018},
  publisher={Chapman and Hall/CRC}
}

@article{stoye2018likelihood,
  title={Likelihood-free inference with an improved cross-entropy estimator},
  author={Stoye, Markus and Brehmer, Johann and Louppe, Gilles and Pavez, Juan and Cranmer, Kyle},
  journal={arXiv preprint arXiv:1808.00973},
  year={2018}
}

@article{price2018bayesian,
  title={Bayesian synthetic likelihood},
  author={Price, Leah F and Drovandi, Christopher C and Lee, Anthony and Nott, David J},
  journal={Journal of Computational and Graphical Statistics},
  volume={27},
  number={1},
  pages={1--11},
  year={2018},
  publisher={Taylor \& Francis}
}

@article{frazier2018asymptotic,
  title={Asymptotic properties of approximate Bayesian computation},
  author={Frazier, David T and Martin, Gael M and Robert, Christian P and Rousseau, Judith},
  journal={Biometrika},
  volume={105},
  number={3},
  pages={593--607},
  year={2018},
  publisher={Oxford University Press}
}

@inproceedings{greenberg2019automatic,
  title={Automatic posterior transformation for likelihood-free inference},
  author={Greenberg, David and Nonnenmacher, Marcel and Macke, Jakob},
  booktitle={International Conference on Machine Learning},
  pages={2404--2414},
  year={2019},
  organization={PMLR}
}

@inproceedings{papamakarios2019sequential,
  title={Sequential neural likelihood: Fast likelihood-free inference with autoregressive flows},
  author={Papamakarios, George and Sterratt, David and Murray, Iain},
  booktitle={The 22nd International Conference on Artificial Intelligence and Statistics},
  pages={837--848},
  year={2019},
  organization={PMLR}
}

@article{brehmer2019mining,
  title={Mining for Dark Matter Substructure: Inferring subhalo population properties from strong lenses with machine learning},
  author={Brehmer, Johann and Mishra-Sharma, Siddharth and Hermans, Joeri and Louppe, Gilles and Cranmer, Kyle},
  journal={The Astrophysical Journal},
  volume={886},
  number={1},
  pages={49},
  year={2019},
  publisher={IOP Publishing}
}

@InProceedings{2019arXiv190304057H,
  title = 	 {Likelihood-free {MCMC} with Amortized Approximate Ratio Estimators},
  author =       {Hermans, Joeri and Begy, Volodimir and Louppe, Gilles},
  booktitle = 	 {Proceedings of the 37th International Conference on Machine Learning},
  pages = 	 {4239--4248},
  year = 	 {2020},
  editor = 	 {III, Hal Daum{\'e} and Singh, Aarti},
  volume = 	 {119},
  series = 	 {Proceedings of Machine Learning Research},
  publisher =    {PMLR},
}

@article{beaumont2019approximate,
  title={Approximate bayesian computation},
  author={Beaumont, Mark A},
  journal={Annual review of statistics and its application},
  volume={6},
  number={1},
  pages={379--403},
  year={2019},
  publisher={Annual Reviews}
}

@inproceedings{durkan2020contrastive,
  title={On contrastive learning for likelihood-free inference},
  author={Durkan, Conor and Murray, Iain and Papamakarios, George},
  booktitle={International Conference on Machine Learning},
  pages={2771--2781},
  year={2020},
  organization={PMLR}
}

@inproceedings{dalmasso2020confidence,
  title={Confidence Sets and Hypothesis Testing in a Likelihood-Free Inference Setting},
  author={Dalmasso, Niccol{\`o} and Izbicki, Rafael and Lee, Ann},
  booktitle={International Conference on Machine Learning},
  pages={2323--2334},
  year={2020},
  organization={PMLR}
}

@article{cranmer2020frontier,
  title={The frontier of simulation-based inference},
  author={Cranmer, Kyle and Brehmer, Johann and Louppe, Gilles},
  journal={Proceedings of the National Academy of Sciences},
  year={2020},
  publisher={National Acad Sciences}
}

@article{schmon2020generalized,
  title={Generalized posteriors in approximate Bayesian computation},
  author={Schmon, Sebastian M and Cannon, Patrick W and Knoblauch, Jeremias},
  journal={arXiv preprint arXiv:2011.08644},
  year={2020}
}

@inproceedings{xing2020distortion,
  title={Distortion estimates for approximate Bayesian inference},
  author={Xing, Hanwen and Nicholls, Geoff and Lee, Jeong Kate},
  booktitle={Conference on Uncertainty in Artificial Intelligence},
  pages={1208--1217},
  year={2020},
  organization={PMLR}
}

@article{dalmasso2021likelihood,
  title={Likelihood-Free Frequentist Inference: Bridging Classical Statistics and Machine Learning in Simulation and Uncertainty Quantification},
  author={Dalmasso, Niccol{\`o} and Zhao, David and Izbicki, Rafael and Lee, Ann B},
  journal={arXiv preprint arXiv:2107.03920},
  year={2021}
}

@article{miller2021truncated,
  title={Truncated marginal neural ratio estimation},
  author={Miller, Benjamin K and Cole, Alex and Forr{\'e}, Patrick and Louppe, Gilles and Weniger, Christoph},
  journal={Advances in Neural Information Processing Systems},
  volume={34},
  pages={129--143},
  year={2021}
}

@inproceedings{glockler2021variational,
  title={Variational methods for simulation-based inference},
  author={Gl{\"o}ckler, Manuel and Deistler, Michael and Macke, Jakob H},
  booktitle={International Conference on Learning Representations},
  year={2021}
}

@mastersthesis{rozet2021arbitrarythesis,
  title={Arbitrary Marginal Neural Ratio Estimation for Likelihood-free Inference},
  author={Rozet, Fran{\c{c}}ois and Louppe, Gilles},
  year={2021},
  school={University of Li{\`e}ge, Belgium}
}

@article{rozet2021arbitrary,
  title={Arbitrary marginal neural ratio estimation for simulation-based inference},
  author={Rozet, Fran{\c{c}}ois and Louppe, Gilles},
  journal={arXiv preprint arXiv:2110.00449},
  year={2021}
}

@article{matsubara2021robust,
  title={Robust generalised Bayesian inference for intractable likelihoods},
  author={Matsubara, Takuo and Knoblauch, Jeremias and Briol, Fran{\c{c}}ois-Xavier and Oates, Chris J},
  journal={Journal of the Royal Statistical Society: Series B},
  year={2022},
  publisher={Wiley}
}

@article{pacchiardi2021generalized,
  title={Generalized Bayesian likelihood-free inference using scoring rules estimators},
  author={Pacchiardi, Lorenzo and Dutta, Ritabrata},
  journal={arXiv preprint arXiv:2104.03889},
  year={2021}
}

@article{crisissbi,
  title={A crisis in simulation-based inference? beware, your posterior approximations can be unfaithful},
  author={Hermans, Joeri and Delaunoy, Arnaud and Rozet, Fran{\c{c}}ois and Wehenkel, Antoine and Louppe, Gilles},
  journal={Transactions on Machine Learning Research},
  year={2022},
  publisher={OpenReview, Amherst, United States-Massachusetts}
}

@inproceedings{dellaporta2022robust,
  title={Robust Bayesian Inference for Simulator-based Models via the MMD Posterior Bootstrap},
  author={Dellaporta, Charita and Knoblauch, Jeremias and Damoulas, Theodoros and Briol, Fran{\c{c}}ois-Xavier},
  booktitle={International Conference on Artificial Intelligence and Statistics},
  pages={943--970},
  year={2022},
  organization={PMLR}
}

@article{pacchiardi2022score,
  title={Score matched neural exponential families for likelihood-free inference},
  author={Pacchiardi, Lorenzo and Dutta, Ritabrata},
  journal={Journal of Machine Learning Research},
  volume={23},
  number={38},
  pages={1--71},
  year={2022}
}

@article{frazier2022bayesian,
  title={Bayesian inference using synthetic likelihood: asymptotics and adjustments},
  author={Frazier, David T and Nott, David J and Drovandi, Christopher and Kohn, Robert},
  journal={Journal of the American Statistical Association},
  number={just-accepted},
  pages={1--28},
  year={2022},
  publisher={Taylor \& Francis}
}

@article{sharrock2022sequential,
  title={Sequential Neural Score Estimation: Likelihood-Free Inference with Conditional Score Based Diffusion Models},
  author={Sharrock, Louis and Simons, Jack and Liu, Song and Beaumont, Mark},
  journal={arXiv preprint arXiv:2210.04872},
  year={2022}
}

@article{ramesh2022gatsbi,
  title={GATSBI: Generative adversarial training for simulation-based inference},
  author={Ramesh, Poornima and Lueckmann, Jan-Matthis and Boelts, Jan and Tejero-Cantero, {\'A}lvaro and Greenberg, David S and Gon{\c{c}}alves, Pedro J and Macke, Jakob H},
  journal={arXiv preprint arXiv:2203.06481},
  year={2022}
}

@article{glaser2022maximum,
  title={Maximum Likelihood Learning of Unnormalized Models for Simulation-Based Inference},
  author={Glaser, Pierre and Arbel, Michael and Hromadka, Samo and Doucet, Arnaud and Gretton, Arthur},
  journal={arXiv preprint arXiv:2210.14756},
  year={2022}
}

@article{deistler2022truncated,
  title={Truncated proposals for scalable and hassle-free simulation-based inference},
  author={Deistler, Michael and Goncalves, Pedro J and Macke, Jakob H},
  journal={Advances in Neural Information Processing Systems},
  volume={35},
  pages={23135--23149},
  year={2022}
}

@article{geffner2022score,
  title={Score Modeling for Simulation-based Inference},
  author={Geffner, Tomas and Papamakarios, George and Mnih, Andriy},
  journal={arXiv preprint arXiv:2209.14249},
  year={2022}
}

@article{miller2022contrastive,
  title={Contrastive Neural Ratio Estimation},
  author={Miller, Benjamin K and Weniger, Christoph and Forr{\'e}, Patrick},
  journal={Advances in Neural Information Processing Systems},
  volume={35},
  pages={3262--3278},
  year={2022}
}

@article{masserano2022simulation,
  title={Simulator-Based Inference with WALDO: Confidence Regions by Leveraging Prediction Algorithms and Posterior Estimators for Inverse Problems},
  author={Masserano, Luca and Dorigo, Tommaso and Izbicki, Rafael and Kuusela, Mikael and Lee, Ann B},
  journal={Proceedings of Machine Learning Research},
  volume={206},
  year={2023}
}

@article{cannon2022investigating,
  title={Investigating the impact of model misspecification in neural simulation-based inference},
  author={Cannon, Patrick and Ward, Daniel and Schmon, Sebastian M},
  journal={arXiv preprint arXiv:2209.01845},
  year={2022}
}

@article{pacchiardi2022likelihood,
  title={Likelihood-free inference with generative neural networks via scoring rule minimization},
  author={Pacchiardi, Lorenzo and Dutta, Ritabrata},
  journal={arXiv preprint arXiv:2205.15784},
  year={2022}
}

@article{rodrigues2021leveraging,
  title={Leveraging Global Parameters for Flow-based Neural Posterior Estimation},
  author={Rodrigues, Pedro LC and Moreau, Thomas and Louppe, Gilles and Gramfort, Alexandre},
  journal={arXiv preprint arXiv:2102.06477},
  year={2021}
}

@inproceedings{delaunoytowards,
  title={Towards Reliable Simulation-Based Inference with Balanced Neural Ratio Estimation},
  author={Delaunoy, Arnaud and Hermans, Joeri and Rozet, Fran{\c{c}}ois and Wehenkel, Antoine and Louppe, Gilles},
  year={2022},
  booktitle={Advances in Neural Information Processing Systems}
}

@inproceedings{delaunoy2023balancing,
  title={Balancing Simulation-based Inference for Conservative Posteriors},
  author={Delaunoy, Arnaud and Miller, Benjamin Kurt and Forr{\'e}, Patrick and Weniger, Christoph and Louppe, Gilles},
  booktitle={Fifth Symposium on Advances in Approximate Bayesian Inference},
  year={2023}
}

@article{patel2023variational,
  title={Variational Inference with Coverage Guarantees},
  author={Patel, Yash and McNamara, Declan and Loper, Jackson and Regier, Jeffrey and Tewari, Ambuj},
  journal={arXiv preprint arXiv:2305.14275},
  year={2023}
}

@inproceedings{gloeckler2023adversarial,
  title={Adversarial robustness of amortized Bayesian inference},
  author={Gloeckler, Manuel and Deistler, Michael and Macke, Jakob H},
  booktitle={Proceedings of the 40th International Conference on Machine Learning},
  pages={11493--11524},
  year={2023}
}

@inproceedings{schmitt2023leveraging,
  title={Leveraging Self-Consistency for Data-Efficient Amortized Bayesian Inference},
  author={Schmitt, Marvin and Ivanova, Desi R and Habermann, Daniel and Koethe, Ullrich and B{\"u}rkner, Paul-Christian and Radev, Stefan T},
  booktitle={Forty-first International Conference on Machine Learning},
  year={2024}
}

@article{schmitt2023consistency,
  title={Consistency Models for Scalable and Fast Simulation-Based Inference},
  author={Schmitt, Marvin and Pratz, Valentin and K{\"o}the, Ullrich and B{\"u}rkner, Paul-Christian and Radev, Stefan T},
  journal={CoRR},
  year={2023}
}

@article{rozet2023score,
  title={Score-based data assimilation},
  author={Rozet, Fran{\c{c}}ois and Louppe, Gilles},
  journal={Advances in Neural Information Processing Systems},
  volume={36},
  pages={40521--40541},
  year={2023}
}

@inproceedings{radev2023jana,
  title={JANA: Jointly amortized neural approximation of complex Bayesian models},
  author={Radev, Stefan T and Schmitt, Marvin and Pratz, Valentin and Picchini, Umberto and K{\"o}the, Ullrich and B{\"u}rkner, Paul-Christian},
  booktitle={Uncertainty in Artificial Intelligence},
  pages={1695--1706},
  year={2023},
  organization={PMLR}
}

@article{heinrich2023hierarchical,
  title={Hierarchical Neural Simulation-Based Inference Over Event Ensembles},
  author={Heinrich, Lukas and Mishra-Sharma, Siddharth and Pollard, Chris and Windischhofer, Philipp},
  journal={arXiv preprint arXiv:2306.12584},
  year={2023}
}

@article{falkiewicz2024calibrating,
  title={Calibrating Neural Simulation-Based Inference with Differentiable Coverage Probability},
  author={Falkiewicz, Maciej and Takeishi, Naoya and Shekhzadeh, Imahn and Wehenkel, Antoine and Delaunoy, Arnaud and Louppe, Gilles and Kalousis, Alexandros},
  journal={Advances in Neural Information Processing Systems},
  volume={36},
  year={2024}
}

@article{linhart2024diffusion,
  title={Diffusion posterior sampling for simulation-based inference in tall data settings},
  author={Linhart, Julia and Cardoso, Gabriel Victorino and Gramfort, Alexandre and Corff, Sylvain Le and Rodrigues, Pedro LC},
  journal={arXiv preprint arXiv:2404.07593},
  year={2024}
}

@article{gloeckler2024compositional,
  title={Compositional simulation-based inference for time series},
  author={Gloeckler, Manuel and Toyota, Shoji and Fukumizu, Kenji and Macke, Jakob H},
  journal={arXiv preprint arXiv:2411.02728},
  year={2024}
}

@article{haggstrom2024fast,
  title={Fast, accurate and lightweight sequential simulation-based inference using Gaussian locally linear mappings},
  author={H{\"a}ggstr{\"o}m, Henrik and Rodrigues, Pedro LC and Oudoumanessah, Geoffroy and Forbes, Florence and Picchini, Umberto},
  journal={arXiv preprint arXiv:2403.07454},
  year={2024}
}

@article{wildberger2024flow,
  title={Flow matching for scalable simulation-based inference},
  author={Wildberger, Jonas and Dax, Maximilian and Buchholz, Simon and Green, Stephen and Macke, Jakob H and Sch{\"o}lkopf, Bernhard},
  journal={Advances in Neural Information Processing Systems},
  volume={36},
  year={2024}
}

@article{linhart2024c2st,
  title={L-c2st: Local diagnostics for posterior approximations in simulation-based inference},
  author={Linhart, Julia and Gramfort, Alexandre and Rodrigues, Pedro},
  journal={Advances in Neural Information Processing Systems},
  volume={36},
  year={2024}
}

@article{wehenkel2024addressing,
  title={Addressing Misspecification in Simulation-based Inference through Data-driven Calibration},
  author={Wehenkel, Antoine and Gamella, Juan L and Sener, Ozan and Behrmann, Jens and Sapiro, Guillermo and Cuturi, Marco and Jacobsen, J{\"o}rn-Henrik},
  journal={arXiv preprint arXiv:2405.08719},
  year={2024}
}

@article{gloeckler2024all,
  title={All-in-one simulation-based inference},
  author={Gloeckler, Manuel and Deistler, Michael and Weilbach, Christian and Wood, Frank and Macke, Jakob H},
  journal={arXiv preprint arXiv:2404.09636},
  year={2024}
}

@article{atlas2024implementation,
  title={An implementation of neural simulation-based inference for parameter estimation in ATLAS},
  author={ATLAS, Collaboration, and others},
  journal={arXiv preprint arXiv:2412.01600},
  year={2024}
}

@book{lehmann2006testing,
  title={Testing statistical hypotheses},
  author={Lehmann, Erich L and Romano, Joseph P},
  year={2006},
  publisher={Springer Science \& Business Media}
}

@article{cook2006validation,
  title={Validation of software for Bayesian models using posterior quantiles},
  author={Cook, Samantha R and Gelman, Andrew and Rubin, Donald B},
  journal={Journal of Computational and Graphical Statistics},
  volume={15},
  number={3},
  pages={675--692},
  year={2006},
  publisher={Taylor \& Francis}
}

@article{gretton2012kernel,
  title={A kernel two-sample test},
  author={Gretton, Arthur and Borgwardt, Karsten M and Rasch, Malte J and Sch{\"o}lkopf, Bernhard and Smola, Alexander},
  journal={The Journal of Machine Learning Research},
  volume={13},
  number={1},
  pages={723--773},
  year={2012},
  publisher={JMLR. org}
}

@inproceedings{bengio2014bounding,
  title={Bounding the test log-likelihood of generative models},
  author={Bengio, Yoshua and Yao, Li and Cho, Kyunghyun},
  booktitle={International Conference on Learning Representations},
  year={2014}
}

@article{prangle2014diagnostic,
  title={Diagnostic tools for approximate Bayesian computation using the coverage property},
  author={Prangle, Dennis and Blum, Michael GB and Popovic, G and Sisson, SA},
  journal={Australian \& New Zealand Journal of Statistics},
  volume={56},
  number={4},
  pages={309--329},
  year={2014},
  publisher={Wiley Online Library}
}

@inproceedings{dziugaite2015training,
  title={Training generative neural networks via maximum mean discrepancy optimization},
  author={Dziugaite, Gintare Karolina and Roy, Daniel M and Ghahramani, Zoubin},
  booktitle={Proceedings of the Thirty-First Conference on Uncertainty in Artificial Intelligence},
  pages={258--267},
  year={2015}
}

@inproceedings{lopez2017revisiting,
  title={Revisiting classifier two-sample tests},
  author={Lopez-Paz, David and Oquab, Maxime},
  booktitle={International Conference on Learning Representations},
  year={2017}
}

@article{Brehmer:2018eca,
    author = "Brehmer, Johann and Cranmer, Kyle and Louppe, Gilles and Pavez, Juan",
    title = "{A Guide to Constraining Effective Field Theories with Machine Learning}",
    eprint = "1805.00020",
    archivePrefix = "arXiv",
    primaryClass = "hep-ph",
    doi = "10.1103/PhysRevD.98.052004",
    journal = "Phys. Rev. D",
    volume = "98",
    number = "5",
    pages = "052004",
    year = "2018"
}

@article{sbc,
  title={Validating Bayesian inference algorithms with simulation-based calibration},
  author={Talts, Sean and Betancourt, Michael and Simpson, Daniel and Vehtari, Aki and Gelman, Andrew},
  journal={arXiv preprint arXiv:1804.06788},
  year={2018}
}

@inproceedings{zhao2021diagnostics,
  title={Diagnostics for conditional density models and Bayesian inference algorithms},
  author={Zhao, David and Dalmasso, Niccol{\`o} and Izbicki, Rafael and Lee, Ann B},
  booktitle={Uncertainty in Artificial Intelligence},
  pages={1830--1840},
  year={2021},
  organization={PMLR}
 }

@article{linhart2022validation,
  title={Validation Diagnostics for SBI algorithms based on Normalizing Flows},
  author={Linhart, Julia and Gramfort, Alexandre and Rodrigues, Pedro LC},
  journal={arXiv preprint arXiv:2211.09602},
  year={2022}
}

@article{lemos2023sampling,
  title={Sampling-Based Accuracy Testing of Posterior Estimators for General Inference},
  author={Lemos, Pablo and Coogan, Adam and Hezaveh, Yashar and Perreault-Levasseur, Laurence},
  journal={arXiv preprint arXiv:2302.03026},
  year={2023}
}

@article{destexhe2000nonlinear,
  title={Nonlinear thermodynamic models of voltage-dependent currents},
  author={Destexhe, Alain and Huguenard, John R},
  journal={Journal of Computational Neuroscience},
  volume={9},
  pages={259--270},
  year={2000},
  publisher={Springer}
}

@article{atlas2012observation,
  title={Observation of a new particle in the search for the Standard Model Higgs boson with the ATLAS detector at the LHC},
  author={ATLAS, Collaboration and others},
  year={2012},
  publisher={Elsevier 1}
}

@article{hannam2014simple,
  title={Simple model of complete precessing black-hole-binary gravitational waveforms},
  author={Hannam, Mark and Schmidt, Patricia and Boh{\'e}, Alejandro and Haegel, Le{\"\i}la and Husa, Sascha and Ohme, Frank and Pratten, Geraint and P{\"u}rrer, Michael},
  journal={Physical review letters},
  volume={113},
  number={15},
  pages={151101},
  year={2014},
  publisher={APS}
}

@misc{Charles2017,
  author = {Jo, Bovy and Jason, Sanders},
  title = {streamgap-pepper},
  year = {2017},
  publisher = {GitHub},
  journal = {GitHub repository},
  howpublished = {\url{https://github.com/jobovy/streamgap-pepper}},
  commit = {176c8f67abf322a6a924a9e7757a004733a7413a}
}

@Article{su151411120,
AUTHOR = {Weligampola, Harshana and Ramanayake, Lakshitha and Ranasinghe, Yasiru and Ilangarathna, Gayanthi and Senarath, Neranjan and Samarakoon, Bhagya and Godaliyadda, Roshan and Herath, Vijitha and Ekanayake, Parakrama and Ekanayake, Janaka and Maheswaran, Muthucumaru and Theminimulle, Sandya and Rathnayake, Anuruddhika and Dharmaratne, Samath and Pinnawala, Mallika and Yatigammana, Sakunthala and Tilakaratne, Ganga},
TITLE = {Pandemic Simulator: An Agent-Based Framework with Human Behavior Modeling for Pandemic-Impact Assessment to Build Sustainable Communities},
JOURNAL = {Sustainability},
VOLUME = {15},
YEAR = {2023},
NUMBER = {14},
ARTICLE-NUMBER = {11120},
URL = {https://www.mdpi.com/2071-1050/15/14/11120},
ISSN = {2071-1050},
DOI = {10.3390/su151411120}
}

@article{alsing2019fast,
  title={Fast likelihood-free cosmology with neural density estimators and active learning},
  author={Alsing, Justin and Charnock, Tom and Feeney, Stephen and Wandelt, Benjamin},
  journal={Monthly Notices of the Royal Astronomical Society},
  volume={488},
  number={3},
  pages={4440--4458},
  year={2019},
  publisher={Oxford University Press}
}

@article{cobb2019ensemble,
  title={An ensemble of bayesian neural networks for exoplanetary atmospheric retrieval},
  author={Cobb, Adam D and Himes, Michael D and Soboczenski, Frank and Zorzan, Simone and O Beirne, Molly D and Baydin, Atilim Gunes and Gal, Yarin and Domagal-Goldman, Shawn D and Arney, Giada N and Angerhausen, Daniel and others},
  journal={The astronomical journal},
  volume={158},
  number={1},
  pages={33},
  year={2019},
  publisher={IOP Publishing}
}

@article{walmsley2020galaxy,
  title={Galaxy Zoo: probabilistic morphology through Bayesian CNNs and active learning},
  author={Walmsley, Mike and Smith, Lewis and Lintott, Chris and Gal, Yarin and Bamford, Steven and Dickinson, Hugh and Fortson, Lucy and Kruk, Sandor and Masters, Karen and Scarlata, Claudia and others},
  journal={Monthly Notices of the Royal Astronomical Society},
  volume={491},
  number={2},
  pages={1554--1574},
  year={2020},
  publisher={Oxford University Press}
}

@article{Hermans:2020skz,
  title={Towards constraining warm dark matter with stellar streams through neural simulation-based inference},
  author={Hermans, Joeri and Banik, Nilanjan and Weniger, Christoph and Bertone, Gianfranco and Louppe, Gilles},
  journal={Monthly Notices of the Royal Astronomical Society},
  volume={507},
  number={2},
  pages={1999--2011},
  year={2021},
  publisher={Oxford University Press}
}

@article{gonccalves2020training,
  title={Training deep neural density estimators to identify mechanistic models of neural dynamics},
  author={Gon{\c{c}}alves, Pedro J and Lueckmann, Jan-Matthis and Deistler, Michael and Nonnenmacher, Marcel and {\"O}cal, Kaan and Bassetto, Giacomo and Chintaluri, Chaitanya and Podlaski, William F and Haddad, Sara A and Vogels, Tim P and others},
  journal={Elife},
  volume={9},
  pages={e56261},
  year={2020},
  publisher={eLife Sciences Publications Limited}
}

@article{delaunoy2020lightning,
  title={Lightning-fast gravitational wave parameter inference through neural amortization},
  author={Delaunoy, Arnaud and Wehenkel, Antoine and Hinderer, Tanja and Nissanke, Samaya and Weniger, Christoph and Williamson, Andrew R and Louppe, Gilles},
  journal={arXiv preprint arXiv:2010.12931},
  year={2020}
}

@article{zhang2023sensitivity,
  title={Sensitivity-guided iterative parameter identification and data generation with BayesFlow and PELS-VAE for model calibration},
  author={Zhang, Yi and Mikelsons, Lars},
  journal={Advanced Modeling and Simulation in Engineering Sciences},
  volume={10},
  number={1},
  pages={9},
  year={2023},
  publisher={Springer}
}

@article{zeng2023probabilistic,
  title={Probabilistic damage detection using a new likelihood-free Bayesian inference method},
  author={Zeng, Jice and Todd, Michael D and Hu, Zhen},
  journal={Journal of Civil Structural Health Monitoring},
  volume={13},
  number={2},
  pages={319--341},
  year={2023},
  publisher={Springer}
}

@article{lemos2023robust,
  title={Robust simulation-based inference in cosmology with Bayesian neural networks},
  author={Lemos, Pablo and Cranmer, Miles and Abidi, Muntazir and Hahn, ChangHoon and Eickenberg, Michael and Massara, Elena and Yallup, David and Ho, Shirley},
  journal={Machine Learning: Science and Technology},
  volume={4},
  number={1},
  pages={01LT01},
  year={2023},
  publisher={IOP Publishing}
}

@article{wagner2024strong,
  title={A Strong Gravitational Lens Is Worth a Thousand Dark Matter Halos: Inference on Small-Scale Structure Using Sequential Methods},
  author={Wagner-Carena, Sebastian and Lee, Jaehoon and Pennington, Jeffrey and Aalbers, Jelle and Birrer, Simon and Wechsler, Risa H},
  journal={arXiv preprint arXiv:2404.14487},
  year={2024}
}

@article{reza2024constraining,
  title={Constraining Cosmology with Simulation-based inference and Optical Galaxy Cluster Abundance},
  author={Reza, Moonzarin and Zhang, Yuanyuan and Avestruz, Camille and Strigari, Louis E and Shevchuk, Simone and Villaescusa-Navarro, Francisco},
  journal={arXiv preprint arXiv:2409.20507},
  year={2024}
}

@article{swierc2024domain,
  title={Domain-Adaptive Neural Posterior Estimation for Strong Gravitational Lens Analysis},
  author={Swierc, Paxson and Tamargo-Arizmendi, Marcos and {\'C}iprijanovi{\'c}, Aleksandra and Nord, Brian D},
  journal={arXiv preprint arXiv:2410.16347},
  year={2024}
}

@article{morandini2024reconstructing,
  title={Reconstructing axion-like particles from beam dumps with simulation-based inference},
  author={Morandini, Alessandro and Ferber, Torben and Kahlhoefer, Felix},
  journal={The European Physical Journal C},
  volume={84},
  number={2},
  pages={1--26},
  year={2024},
  publisher={Springer}
}

@article{berteaud2024simulation,
  title={Simulation-based inference of radio millisecond pulsars in globular clusters},
  author={Berteaud, Joanna and Eckner, Christopher and Calore, Francesca and Clavel, Ma{\"\i}ca and Haggard, Daryl},
  journal={arXiv preprint arXiv:2405.15691},
  year={2024}
}

@article{dimitriou2023fast,
  title={Fast likelihood-free reconstruction of gravitational wave backgrounds},
  author={Dimitriou, Androniki and Figueroa, Daniel G and Zaldivar, Bryan},
  journal={arXiv preprint arXiv:2309.08430},
  year={2023}
}

@article{vasist2023neural,
  title={Neural posterior estimation for exoplanetary atmospheric retrieval},
  author={Vasist, Malavika and Rozet, Fran{\c{c}}ois and Absil, Olivier and Molli{\`e}re, Paul and Nasedkin, Evert and Louppe, Gilles},
  journal={Astronomy \& Astrophysics},
  volume={672},
  pages={A147},
  year={2023},
  publisher={EDP Sciences}
}

@article{lotka,
  title={Analytical note on certain rhythmic relations in organic systems},
  author={Lotka, Alfred J},
  journal={Proceedings of the National Academy of Sciences},
  volume={6},
  number={7},
  pages={410--415},
  year={1920},
  publisher={National Acad Sciences}
}

@article{volterra1926fluctuations,
  title={Fluctuations in the abundance of a species considered mathematically},
  author={Volterra, Vito},
  journal={Nature},
  volume={118},
  number={2972},
  pages={558--560},
  year={1926}
}

@article{shestopaloff2014bayesian,
  title={On Bayesian inference for the M/G/1 queue with efficient MCMC sampling},
  author={Shestopaloff, Alexander Y and Neal, Radford M},
  journal={arXiv preprint arXiv:1401.5548},
  year={2014}
}

@misc{weinberg,
  author = {Cranmer, Kyle and Heinrich, Lukas and Head, Tim and Louppe, Gilles},
  title = {{``Active Sciencing'' with Reusable Workflows}},
  year = {2017},
  publisher = {GitHub},
  journal = {GitHub repository},
  howpublished = {\url{https://github.com/cranmer/active_sciencing}}
}

@InProceedings{lueckmann2021benchmarking, 
  title     = {Benchmarking Simulation-Based Inference},
  author    = {Lueckmann, Jan-Matthis and Boelts, Jan and Greenberg, David and Goncalves, Pedro and Macke, Jakob}, 
  booktitle = {Proceedings of The 24th International Conference on Artificial Intelligence and Statistics}, 
  pages     = {343--351}, 
  year      = {2021}, 
  editor    = {Banerjee, Arindam and Fukumizu, Kenji}, 
  volume    = {130}, 
  series    = {Proceedings of Machine Learning Research}, 
  publisher = {PMLR}
}

@article{goodfellow2014generative,
  title={Generative adversarial nets},
  author={Goodfellow, Ian and Pouget-Abadie, Jean and Mirza, Mehdi and Xu, Bing and Warde-Farley, David and Ozair, Sherjil and Courville, Aaron and Bengio, Yoshua},
  journal={Advances in neural information processing systems},
  volume={27},
  year={2014}
}

@inproceedings{rezende2015variational,
  title={Variational inference with normalizing flows},
  author={Rezende, Danilo and Mohamed, Shakir},
  booktitle={International conference on machine learning},
  pages={1530--1538},
  year={2015},
  organization={PMLR}
}

@inproceedings{dinh2015nice,
  author    = {Laurent Dinh and
               David Krueger and
               Yoshua Bengio},
  title     = {{NICE:} Non-linear Independent Components Estimation},
  booktitle = {3rd International Conference on Learning Representations, {ICLR} 2015,
               San Diego, CA, USA, May 7-9, 2015, Workshop Track Proceedings},
  year      = {2015},
}

@inproceedings{dinh2017density,
  author    = {Laurent Dinh and
               Jascha Sohl{-}Dickstein and
               Samy Bengio},
  title     = {Density estimation using Real {NVP}},
  booktitle = {5th International Conference on Learning Representations, {ICLR} 2017,
               Toulon, France, April 24-26, 2017, Conference Track Proceedings},
  year      = {2017}
}

@article{durkan2019neural,
  title={Neural spline flows},
  author={Durkan, Conor and Bekasov, Artur and Murray, Iain and Papamakarios, George},
  journal={Advances in neural information processing systems},
  volume={32},
  year={2019}
}

@article{papamakarios2019normalizing,
  title={Normalizing flows for probabilistic modeling and inference},
  author={Papamakarios, George and Nalisnick, Eric and Rezende, Danilo Jimenez and Mohamed, Shakir and Lakshminarayanan, Balaji},
  journal={arXiv preprint arXiv:1912.02762},
  year={2019}
}

@article{song2020score,
  title={Score-based generative modeling through stochastic differential equations},
  author={Song, Yang and Sohl-Dickstein, Jascha and Kingma, Diederik P and Kumar, Abhishek and Ermon, Stefano and Poole, Ben},
  journal={arXiv preprint arXiv:2011.13456},
  year={2020}
}

@article{Neyman1937Outline,
  title={Outline of a Theory of Statistical Estimation Based on the Classical Theory of Probability},
  author={Jerzy Neyman},
  journal={Philosophical Transactions of the Royal Society A},
  year={1937},
  volume={236},
  pages={333-380},
  url={https://api.semanticscholar.org/CorpusID:19584450}
}

@article{metropolis1953equation,
  title={Equation of state calculations by fast computing machines},
  author={Metropolis, Nicholas and Rosenbluth, Arianna W and Rosenbluth, Marshall N and Teller, Augusta H and Teller, Edward},
  journal={The journal of chemical physics},
  volume={21},
  number={6},
  pages={1087--1092},
  year={1953},
  publisher={American Institute of Physics}
}

@book{popper1959logic,
  address = {London},
  author = {Popper, Karl R.},
  description = {Wujastyk's main bibtex file, April 30, 2010},
  howpublished = {Paperback},
  isbn = {0415278449},
  publisher = {Routledge},
  title = {The Logic of Scientific Discovery},
  year = 1959
}

@article{hastings1970monte,
    author = {Hastings, W. K.},
    title = "{Monte Carlo sampling methods using Markov chains and their applications}",
    journal = {Biometrika},
    volume = {57},
    number = {1},
    pages = {97-109},
    year = {1970},
    month = {04},
    issn = {0006-3444},
    doi = {10.1093/biomet/57.1.97},
    url = {https://doi.org/10.1093/biomet/57.1.97},
    eprint = {https://academic.oup.com/biomet/article-pdf/57/1/97/23940249/57-1-97.pdf},
}

@article{box1976science,
  title={Science and statistics},
  author={Box, George EP},
  journal={Journal of the American Statistical Association},
  volume={71},
  number={356},
  pages={791--799},
  year={1976},
  publisher={Taylor \& Francis}
}

@techreport{raftery1991many,
  title={How many iterations in the Gibbs sampler?},
  author={Raftery, Adrian E and Lewis, Steven},
  year={1991},
}

@techreport{geweke1991evaluating,
  title={Evaluating the accuracy of sampling-based approaches to the calculation of posterior moments},
  author={Geweke, John F and others},
  year={1991},
  institution={Federal Reserve Bank of Minneapolis}
}

@article{gelman1992inference,
  title={Inference from iterative simulation using multiple sequences},
  author={Gelman, Andrew and Rubin, Donald B},
  journal={Statistical science},
  volume={7},
  number={4},
  pages={457--472},
  year={1992},
  publisher={Institute of Mathematical Statistics}
}

@article{hyndman1996computing,
  title={Computing and graphing highest density regions},
  author={Hyndman, Rob J},
  journal={The American Statistician},
  volume={50},
  number={2},
  pages={120--126},
  year={1996},
  publisher={Taylor \& Francis}
}

@article{bernardo2005intrinsic,
  title={Intrinsic credible regions: An objective Bayesian approach to interval estimation},
  author={Bernardo, Jos{\'e} M},
  journal={Test},
  volume={14},
  pages={317--384},
  year={2005},
  publisher={Springer}
}

@article{Goodman2010ensemble,
author = {Jonathan Goodman and Jonathan Weare},
title = {{Ensemble samplers with affine invariance}},
volume = {5},
journal = {Communications in Applied Mathematics and Computational Science},
number = {1},
publisher = {MSP},
pages = {65 -- 80},
year = {2010},
doi = {10.2140/camcos.2010.5.65},
URL = {https://doi.org/10.2140/camcos.2010.5.65}
}

@book{box2011bayesian,
  title={Bayesian inference in statistical analysis},
  author={Box, George EP and Tiao, George C},
  volume={40},
  year={1973},
  publisher={John Wiley \& Sons}
}

@article{sugiyama2012density,
  title={Density-ratio matching under the Bregman divergence: a unified framework of density-ratio estimation},
  author={Sugiyama, Masashi and Suzuki, Taiji and Kanamori, Takafumi},
  journal={Annals of the Institute of Statistical Mathematics},
  volume={64},
  number={5},
  pages={1009--1044},
  year={2012},
  publisher={Springer}
}

@article{neal2012mcmc,
  title={MCMC using Hamiltonian dynamics},
  author={Neal, Radford M},
  journal={arXiv preprint arXiv:1206.1901},
  year={2012}
}

@inproceedings{vovk2012conditional,
  title={Conditional validity of inductive conformal predictors},
  author={Vovk, Vladimir},
  booktitle={Asian conference on machine learning},
  pages={475--490},
  year={2012},
  organization={PMLR}
}

@article{schall2012empirical,
  title={The empirical coverage of confidence intervals: Point estimates and confidence intervals for confidence levels},
  author={Schall, Robert},
  journal={Biometrical journal},
  volume={54},
  number={4},
  pages={537--551},
  year={2012},
  publisher={Wiley Online Library}
}

@article{nelson2013run,
  title={Run dmc: an efficient, parallel code for analyzing radial velocity observations using n-body integrations and differential evolution markov chain monte carlo},
  author={Nelson, Benjamin and Ford, Eric B and Payne, Matthew J},
  journal={The Astrophysical Journal Supplement Series},
  volume={210},
  number={1},
  pages={11},
  year={2013},
  publisher={IOP Publishing}
}

@article{lin2014integrated,
  title={An integrated procedure for bayesian reliability inference using MCMC},
  author={Lin, Jing},
  journal={Journal of Quality and Reliability Engineering},
  volume={2014},
  year={2014},
  publisher={Hindawi}
}

@article{blei2014build,
  title={Build, compute, critique, repeat: Data analysis with latent variable models},
  author={Blei, David M},
  journal={Annual Review of Statistics and Its Application},
  volume={1},
  number={1},
  pages={203--232},
  year={2014},
  publisher={Annual Reviews}
}

@article{bissiri2016general,
  title={A general framework for updating belief distributions},
  author={Bissiri, Pier Giovanni and Holmes, Chris C and Walker, Stephen G},
  journal={Journal of the Royal Statistical Society: Series B (Statistical Methodology)},
  volume={78},
  number={5},
  pages={1103--1130},
  year={2016},
  publisher={Wiley Online Library}
}

@article{dixit2017mcmc,
  title={MCMC diagnostics for higher dimensions using Kullback Leibler divergence},
  author={Dixit, Anand and Roy, Vivekananda},
  journal={Journal of Statistical Computation and Simulation},
  volume={87},
  number={13},
  pages={2622--2638},
  year={2017},
  publisher={Taylor \& Francis}
}

@article{holmes2017assigning,
  title={Assigning a value to a power likelihood in a general Bayesian model},
  author={Holmes, Chris C and Walker, Stephen G},
  journal={Biometrika},
  volume={104},
  number={2},
  pages={497--503},
  year={2017},
  publisher={Oxford University Press}
}

@article{grunwald2017inconsistency,
  title={Inconsistency of Bayesian inference for misspecified linear models, and a proposal for repairing it},
  author={Gr{\"u}nwald, Peter and Van Ommen, Thijs},
  journal={Bayesian Analysis},
  volume={12},
  number={4},
  pages={1069--1103},
  year={2017},
  publisher={International Society for Bayesian Analysis}
}

@phdthesis{dixit2018developments,
  title={Developments in MCMC diagnostics and sparse Bayesian learning models},
  author={Dixit, Anand Ulhas},
  year={2018},
  school={Iowa State University}
}

@article{hogg2018data,
  title={Data analysis recipes: Using markov chain monte carlo},
  author={Hogg, David W and Foreman-Mackey, Daniel},
  journal={The Astrophysical Journal Supplement Series},
  volume={236},
  number={1},
  pages={11},
  year={2018},
  publisher={IOP Publishing}
}

@article{miller2018robust,
  title={Robust Bayesian inference via coarsening},
  author={Miller, Jeffrey W and Dunson, David B},
  journal={Journal of the American Statistical Association},
  year={2018},
  publisher={Taylor \& Francis}
}

@article{roy2020convergence,
  title={Convergence diagnostics for markov chain monte carlo},
  author={Roy, Vivekananda},
  journal={Annual Review of Statistics and Its Application},
  volume={7},
  pages={387--412},
  year={2020},
  publisher={Annual Reviews}
}

@article{wenger2022posterior,
  title={Posterior and computational uncertainty in Gaussian processes},
  author={Wenger, Jonathan and Pleiss, Geoff and Pf{\"o}rtner, Marvin and Hennig, Philipp and Cunningham, John P},
  journal={Advances in Neural Information Processing Systems},
  volume={35},
  pages={10876--10890},
  year={2022}
}

@article{mackay1992bayesian,
  title={Bayesian interpolation},
  author={MacKay, David JC},
  journal={Neural computation},
  volume={4},
  number={3},
  pages={415--447},
  year={1992},
  publisher={MIT Press One Rogers Street, Cambridge, MA 02142-1209, USA journals-info}
}

@inproceedings{welling2011bayesian,
  title={Bayesian learning via stochastic gradient Langevin dynamics},
  author={Welling, Max and Teh, Yee W},
  booktitle={Proceedings of the 28th international conference on machine learning (ICML-11)},
  pages={681--688},
  year={2011},
  organization={Citeseer}
}

@inproceedings{chen2014stochastic,
  title={Stochastic gradient hamiltonian monte carlo},
  author={Chen, Tianqi and Fox, Emily and Guestrin, Carlos},
  booktitle={International conference on machine learning},
  pages={1683--1691},
  year={2014},
  organization={PMLR}
}

@inproceedings{blundell2015weight,
  title={Weight uncertainty in neural network},
  author={Blundell, Charles and Cornebise, Julien and Kavukcuoglu, Koray and Wierstra, Daan},
  booktitle={International conference on machine learning},
  pages={1613--1622},
  year={2015},
  organization={PMLR}
}

@article{gal2016uncertainty,
  title={Uncertainty in deep learning},
  author={Gal, Yarin and others},
  year={2016},
  publisher={phd thesis, University of Cambridge}
}

@inproceedings{flam2017mapping,
  title={Mapping Gaussian process priors to Bayesian neural networks},
  author={Flam-Shepherd, Daniel and Requeima, James and Duvenaud, David},
  booktitle={NIPS Bayesian deep learning workshop},
  volume={3},
  year={2017}
}

@inproceedings{sun2018functional,
  title={FUNCTIONAL VARIATIONAL BAYESIAN NEURAL NETWORKS},
  author={Sun, Shengyang and Zhang, Guodong and Shi, Jiaxin and Grosse, Roger},
  booktitle={International Conference on Learning Representations},
  year={2018}
}

@inproceedings{shi2018spectral,
  title={A spectral approach to gradient estimation for implicit distributions},
  author={Shi, Jiaxin and Sun, Shengyang and Zhu, Jun},
  booktitle={International Conference on Machine Learning},
  pages={4644--4653},
  year={2018},
  organization={PMLR}
}

@inproceedings{depeweg2018decomposition,
  title={Decomposition of uncertainty in Bayesian deep learning for efficient and risk-sensitive learning},
  author={Depeweg, Stefan and Hernandez-Lobato, Jose-Miguel and Doshi-Velez, Finale and Udluft, Steffen},
  booktitle={International conference on machine learning},
  pages={1184--1193},
  year={2018},
  organization={PMLR}
}

@article{maddox2019simple,
  title={A simple baseline for bayesian uncertainty in deep learning},
  author={Maddox, Wesley J and Izmailov, Pavel and Garipov, Timur and Vetrov, Dmitry P and Wilson, Andrew Gordon},
  journal={Advances in neural information processing systems},
  volume={32},
  year={2019}
}

@inproceedings{pearce2020uncertainty,
  title={Uncertainty in neural networks: Approximately bayesian ensembling},
  author={Pearce, Tim and Leibfried, Felix and Brintrup, Alexandra},
  booktitle={International conference on artificial intelligence and statistics},
  pages={234--244},
  year={2020},
  organization={PMLR}
}

@inproceedings{wenzel2020good,
  title={How Good is the Bayes Posterior in Deep Neural Networks Really?},
  author={Wenzel, Florian and Roth, Kevin and Veeling, Bastiaan and Swiatkowski, Jakub and Tran, Linh and Mandt, Stephan and Snoek, Jasper and Salimans, Tim and Jenatton, Rodolphe and Nowozin, Sebastian},
  booktitle={International Conference on Machine Learning},
  pages={10248--10259},
  year={2020},
  organization={PMLR}
}

@article{he2020bayesian,
  title={Bayesian deep ensembles via the neural tangent kernel},
  author={He, Bobby and Lakshminarayanan, Balaji and Teh, Yee Whye},
  journal={Advances in neural information processing systems},
  volume={33},
  pages={1010--1022},
  year={2020}
}

@article{ma2021functional,
  title={Functional variational inference based on stochastic process generators},
  author={Ma, Chao and Hern{\'a}ndez-Lobato, Jos{\'e} Miguel},
  journal={Advances in Neural Information Processing Systems},
  volume={34},
  pages={21795--21807},
  year={2021}
}

@article{fortuin2022priors,
  title={Priors in bayesian deep learning: A review},
  author={Fortuin, Vincent},
  journal={International Statistical Review},
  volume={90},
  number={3},
  pages={563--591},
  year={2022},
  publisher={Wiley Online Library}
}

@article{rudner2022tractable,
  title={Tractable function-space variational inference in Bayesian neural networks},
  author={Rudner, Tim GJ and Chen, Zonghao and Teh, Yee Whye and Gal, Yarin},
  journal={Advances in Neural Information Processing Systems},
  volume={35},
  pages={22686--22698},
  year={2022}
}

@article{tran2022all,
  title={All you need is a good functional prior for Bayesian deep learning},
  author={Tran, Ba-Hien and Rossi, Simone and Milios, Dimitrios and Filippone, Maurizio},
  journal={The Journal of Machine Learning Research},
  volume={23},
  number={1},
  pages={3210--3265},
  year={2022},
  publisher={JMLRORG}
}

@inproceedings{kozyrskiy2023imposing,
  title={Imposing Functional Priors on Bayesian Neural Networks},
  author={Kozyrskiy, Bogdan and Milios, Dimitrios and Filippone, Maurizio},
  booktitle={ICPRAM 2023, 12th International Conference on Pattern Recognition Applications and Methods},
  year={2023}
}

@article{papamarkou2024position,
  title={Position Paper: Bayesian Deep Learning in the Age of Large-Scale AI},
  author={Papamarkou, Theodore and Skoularidou, Maria and Palla, Konstantina and Aitchison, Laurence and Arbel, Julyan and Dunson, David and Filippone, Maurizio and Fortuin, Vincent and Hennig, Philipp and Hubin, Aliaksandr and others},
  journal={arXiv preprint arXiv:2402.00809},
  year={2024}
}

@misc{lalsuite,
   author         = "{LIGO Scientific Collaboration}",
   title          = "{LIGO} {A}lgorithm {L}ibrary - {LALS}uite",
   howpublished   = "free software (GPL)",
   doi            = "10.7935/GT1W-FZ16",
   year           = "2018"
}

@article{Biwer:2018osg,
    author = "Biwer, C. M. and Capano, Collin D. and De, Soumi and Cabero, Miriam and Brown, Duncan A. and Nitz, Alexander H. and Raymond, V.",
    title = "{PyCBC Inference: A Python-based parameter estimation toolkit for compact binary coalescence signals}",
    eprint = "1807.10312",
    archivePrefix = "arXiv",
    primaryClass = "astro-ph.IM",
    doi = "10.1088/1538-3873/aaef0b",
    journal = "Publ. Astron. Soc. Pac.",
    volume = "131",
    number = "996",
    pages = "024503",
    year = "2019"
}

@article{sbi,
  title={sbi: A toolkit for simulation-based inference},
  author={Tejero-Cantero, Alvaro and Boelts, Jan and Deistler, Michael and Lueckmann, Jan-Matthis and Durkan, Conor and Gon{\c{c}}alves, Pedro and Greenberg, David and Macke, Jakob},
  journal={The Journal of Open Source Software},
  volume={5},
  number={52},
  pages={2505},
  year={2020}
}

@misc{Rozet_Zuko_2022,
        author = {Rozet, Fran{\c{c}}ois},
        doi = {10.5281/zenodo.7625672},
        license = {MIT},
        title = {{Zuko}},
        url = {https://pypi.org/project/zuko},
        year = {2022}
}

@software{rozet2023lampe,
  author       = {{Rozet, Fran{\c{c}}ois}},
  title        = {LAMPE},
  year         = 2023,
  doi          = {10.5281/zenodo.8405783},
  url          = {https://pypi.org/project/lampe}
}

@article{breiman1996bagging,
  title={Bagging predictors},
  author={Breiman, Leo},
  journal={Machine learning},
  volume={24},
  number={2},
  pages={123--140},
  year={1996},
  publisher={Springer}
}

@article{pritchard1999population,
  title={Population growth of human Y chromosomes: a study of Y chromosome microsatellites.},
  author={Pritchard, Jonathan K and Seielstad, Mark T and Perez-Lezaun, Anna and Feldman, Marcus W},
  journal={Molecular biology and evolution},
  volume={16},
  number={12},
  pages={1791--1798},
  year={1999},
  publisher={Oxford University Press}
}

@article{bishop1994mixture,
  title={Mixture density networks.},
  author={Bishop, Christopher M},
  journal={Technical Report},
  year={1994},
  publisher={Aston University}
}

@article{aad2012observation,
  title={Observation of a new particle in the search for the Standard Model Higgs boson with the ATLAS detector at the LHC},
  author={Aad, Georges and Abajyan, Tatevik and Abbott, B and Abdallah, J and Khalek, S Abdel and Abdelalim, Ahmed Ali and Aben, R and Abi, B and Abolins, M and AbouZeid, OS and others},
  journal={Physics Letters B},
  volume={716},
  number={1},
  pages={1--29},
  year={2012},
  publisher={Elsevier}
}

@article{abbott2016gw151226,
  title={GW151226: observation of gravitational waves from a 22-solar-mass binary black hole coalescence},
  author={Abbott, Benjamin P and Abbott, R and Abbott, TD and Abernathy, MR and Acernese, F and Ackley, K and Adams, C and Adams, T and Addesso, P and Adhikari, RX and others},
  journal={Physical review letters},
  volume={116},
  number={24},
  pages={241103},
  year={2016},
  publisher={APS}
}

@article{sason2016f,
  title={$ f $-divergence Inequalities},
  author={Sason, Igal and Verd{\'u}, Sergio},
  journal={IEEE Transactions on Information Theory},
  volume={62},
  number={11},
  pages={5973--6006},
  year={2016},
  publisher={IEEE}
}

@article{lakshminarayanan2017simple,
  title={Simple and scalable predictive uncertainty estimation using deep ensembles},
  author={Lakshminarayanan, Balaji and Pritzel, Alexander and Blundell, Charles},
  journal={Advances in neural information processing systems},
  volume={30},
  year={2017}
}

@article{gilman2018probing,
  title={Probing the nature of dark matter by forward modelling flux ratios in strong gravitational lenses},
  author={Gilman, Daniel and Birrer, Simon and Treu, Tommaso and Keeton, Charles R and Nierenberg, Anna},
  journal={Monthly Notices of the Royal Astronomical Society},
  volume={481},
  number={1},
  pages={819--834},
  year={2018},
  publisher={Oxford University Press}
}

@article{banik2018probing,
  title={Probing the nature of dark matter particles with stellar streams},
  author={Banik, Nilanjan and Bertone, Gianfranco and Bovy, Jo and Bozorgnia, Nassim},
  journal={Journal of Cosmology and Astroparticle Physics},
  volume={2018},
  number={07},
  pages={061},
  year={2018},
  publisher={IOP Publishing}
}

@article{Planck:2018vyg,
    author = "Aghanim, N. and others",
    collaboration = "Planck",
    title = "{Planck 2018 results. VI. Cosmological parameters}",
    eprint = "1807.06209",
    archivePrefix = "arXiv",
    primaryClass = "astro-ph.CO",
    doi = "10.1051/0004-6361/201833910",
    journal = "Astron. Astrophys.",
    volume = "641",
    pages = "A6",
    year = "2020",
    note = "[Erratum: Astron.Astrophys. 652, C4 (2021)]"
}

@article{villaescusa2020quijote,
  title={The quijote simulations},
  author={Villaescusa-Navarro, Francisco and Hahn, ChangHoon and Massara, Elena and Banerjee, Arka and Delgado, Ana Maria and Ramanah, Doogesh Kodi and Charnock, Tom and Giusarma, Elena and Li, Yin and Allys, Erwan and others},
  journal={The Astrophysical Journal Supplement Series},
  volume={250},
  number={1},
  pages={2},
  year={2020},
  publisher={IOP Publishing}
}

@article{cuesta2023point,
  title={A point cloud approach to generative modeling for galaxy surveys at the field level},
  author={Cuesta-Lazaro, Carolina and Mishra-Sharma, Siddharth},
  journal={arXiv preprint arXiv:2311.17141},
  year={2023}
}
  \cleardoublepage

\end{document}